\documentclass[10pt]{article}
\usepackage{flexrsi}   %

\showplansfalse

\newcommand{\name}{Flex-RSI}

\title{\name: Benchmarking Recursive\\Self-Improvement of Frontier Agents}
\shorttitle{\name: Benchmarking Recursive Self-Improvement of Frontier Agents}
\author{Flex-RSI Team}
\affiliation{Texas A\&M University\enspace\textperiodcentered\enspace University of Cambridge\enspace\textperiodcentered\enspace National University of Singapore\enspace\textperiodcentered\enspace
NVIDIA\enspace\textperiodcentered\enspace Massachusetts Institute of Technology\enspace\textperiodcentered\enspace
MIT-IBM Watson AI Lab\enspace\textperiodcentered\enspace The University of Texas at Austin\enspace\textperiodcentered\enspace
Carnegie Mellon University\enspace\textperiodcentered\enspace University of Pennsylvania\enspace\textperiodcentered\enspace University of Southern California\enspace\textperiodcentered\enspace
Purdue University\enspace\textperiodcentered\enspace Adobe\enspace\textperiodcentered\enspace University of Toronto}
\abstract{Frontier language models are increasingly deployed as agents, and how well an agent performs depends not only
on its model but also on its harness: the prompts, tools, memory, and control loop through which the model acts.
When an agent is applied to a new problem, a natural question is whether it can use the feedback of the task to
improve the way it works, and not merely its answer. Most existing benchmarks keep the agent's harness fixed and score
the solution it produces, or target large-scale model training. We introduce \name{}, a benchmark for the recursive
self-improvement of agents on vision and robotics tasks under a fixed budget. In each round, an executor improves
the task solution, which is kept only if it scores higher on held-out validation data, and an improver then revises
the harness, including the executor and itself. Every run starts from the same general-purpose harness, and the
final solution is scored on hidden test data. Evaluating closed-source and open-weight models on tasks spanning
perception, reasoning, robot action, 3D reconstruction, visual creation, and infrastructure, we find that
self-improvement helps most when the agent builds its own solution, that feedback recording only success or failure
holds back open-weight models, and that open-weight models fall furthest behind where outputs must pass strict checks
of physical stability or exact correctness. Paired runs further show that editing the harness itself contributes to
the gains.}
\metadata{Project page}{\url{https://flex-rsi.github.io}}
\metadata{Authors}{The full list of authors and their contributions is in \cref{sec:contributors}}

\hypersetup{pdftitle={Flex-RSI: Benchmarking Recursive Self-Improvement of Frontier Agents},
            pdfauthor={Flex-RSI Team: Jingdong Zhang, Jionghao Wang, Zhengming Yu, Yancheng Cai, Wei Chow, Chris Choy, Zhiyang Dou, Hang Hua, Jiachen Li, Zhixuan Liu, Chen Wang, Songlin Wei, Siyuan Yang, Yu Yuan, Xiaohang Zhan, Yucheng Zhang, Shuhong Zheng, Xin Li, Lingjie Liu, Wojciech Matusik, Zhengzhong Tu, Wenping Wang}}

\begin{document}
\maketitle

\section{Introduction}
\label{sec:intro}

Frontier large language models now routinely act as agents that write code, conduct experiments, and read the results
over multiple steps~\citep{anthropic2026opus55,openai2026gpt6,google2026gemini38,qwen2026qwen38,deepseek2026v4,glm2026glm5}.
An agent's performance depends not only on the underlying model but also on its \emph{harness}: the prompts, tools,
memory, and control loop through which the model acts. This raises a practical question for anyone who deploys an
agent on a new problem: given only the task, can the agent use the task's feedback to improve the way it works under a fixed budget, and not merely its answer?

Most existing benchmarks run each model inside a fixed harness, often the vendor's own, and score only the solution
or artifact it produces, mostly on machine-learning research tasks with a continuous
metric~\citep{chan2024mlebench,wijk2024rebench}; those that also let the agent revise its
harness~\citep{rsiexam2026} or that center on upstream model training at scale~\citep{openrsi2026} remain within
machine-learning research. We introduce \name{} to measure how well a frontier agent improves both its solution and
its own harness on
general vision and robotics tasks, a setting in which agents are increasingly deployed, each attempt consumes real
compute, and the goal is a better solution within hours to days.

\begin{figure}[t]
  \centering
  \includegraphics[width=\linewidth]{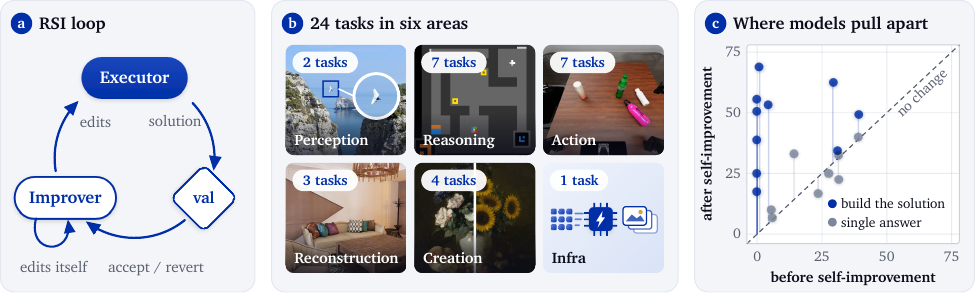}
  \caption{\textbf{\name{} at a glance.} \textbf{(a)}~The RSI loop. In each round, the executor edits the task
  solution, which is kept or reverted according to its score on held-out validation data; the improver then edits
  the harness, including the executor and itself. \textbf{(b)}~The 24 tasks fall
  into six areas. Each tile shows an example task from one area, and its badge gives the number of tasks in that
  area. \textbf{(c)}~Each dot is a task, placed by the score difference between the best and the worst model before
  (horizontal) and after (vertical) self-improvement; its height above the diagonal shows how much further apart the
  models end up. Blue: tasks in which the agent builds the solution. Gray: tasks whose solution is a single answer.}
  \label{fig:overview}
\end{figure}

In \name{}, the harness consists of two components that alternate over rounds (\cref{fig:overview}a). The
\emph{executor} improves the task solution: it modifies the solution, runs experiments, and evaluates them on the labeled
development data, and its candidate is kept only if it also scores higher on held-out validation data. The
\emph{improver} then reviews the round and modifies the harness itself, including the executor's prompts, tools, and
workflow as well as its own prompts, settings, and lessons. Every run starts from the same general-purpose harness
and, on each task, the same starter solution.

The tasks differ in what the agent must produce (a single answer or a solution it builds), how the evaluator scores
it (on a scale or only as success or failure), and how strictly the output is checked.
From experiments on 24 tasks spanning perception, reasoning, robot action, 3D reconstruction, visual creation, and
infrastructure (\cref{fig:overview}b), with 18 closed-source and open-weight models, we find:
\begin{enumerate}[leftmargin=*,itemsep=2pt]
\item \textbf{Tasks in which the agent builds the solution leave more room for self-improvement:} tasks whose
solution is a single answer to each question leave little room to improve, while tasks in which the agent builds its own solution leave the most, and there models
pull apart (\cref{fig:overview}c).
\item \textbf{The granularity of feedback matters:} feedback that scores each episode on a scale helps all models,
whereas feedback that records only success or failure leaves open-weight models barely improving on the solutions
they build.
\item \textbf{Strict checks of physical stability and exact correctness pose the greatest challenge:} outputs that
look plausible often fail checks of physical stability or exact correctness, and open-weight models fall furthest behind on them.
\end{enumerate}
Paired runs further show that letting the improver edit the harness raises the test score in six of seven pairs
of otherwise identical runs (\cref{sec:harness}).

\section{Related work}
\label{sec:related}

\subsection{Benchmarks for self-improving agents}

Benchmarks for machine-learning engineering and scientific research measure whether agents can turn a task
description into a working result. MLE-bench evaluates machine-learning engineering on competition
tasks~\citep{chan2024mlebench}; RE-Bench compares agents with human experts on research-engineering
problems~\citep{wijk2024rebench}; and PaperBench assesses the replication of machine-learning
research~\citep{starace2025paperbench}. Related evaluations examine the design and execution of machine-learning
experiments~\citep{huang2023mlagentbench} and the generation of executable programs for data-driven scientific
discovery~\citep{chen2025scienceagentbench}. Their primary evaluation targets are the models, code,
and research results that an agent produces.

Evaluations framed around recursive self-improvement cover several different objects of improvement.
RSI-Exam supports executable research on both task methods and harnesses for frozen models, followed by hidden-test
evaluation~\citep{rsiexam2026}. OpenRSI Index spans foundation-model development, including large-scale training
and lighter-weight inference, agent, and systems tasks~\citep{openrsi2026}. Autonomous post-training is evaluated
through controlled training runs and assessments of the resulting
models~\citep{rank2026posttrainbench,huang2026rsiarenarecursiveselfimprovement},
while algorithm-design tasks ask agents to improve training procedures that are subsequently evaluated by
retraining~\citep{chi2026ai4ai}.
Sustained machine-learning research with repeated validation feedback and a separate hidden test is evaluated
in AutoResearchExam~\citep{ramesh2026autoresearchexam}. Harness engineering also appears alongside model-training
tasks in Vals RSI Index~\citep{valsai2026rsiindex}. TraceTrial-Bench evaluates the reliability of agent harnesses
themselves rather than their improvement~\citep{cao2026by}.
These settings share an interest in autonomous experimentation, but differ in whether the optimized artifact
is data, a training algorithm, model weights, or the harness around a model.

Complementary studies examine how experience becomes reusable improvement. Data-centric research under a fixed
post-training stack is studied in RSIBench-Data~\citep{meng2026rsibench}. Procedural-ability transfer and the
conversion of episodic experience into reusable skills are evaluated in EvoAgentBench~\citep{gao2026evoagentbench}
and SkillEvolBench~\citep{lei2026skillevolbench}, respectively.
EvolveTool-Bench checks whether evolved tools satisfy independent conformance tests beyond completing the
current task~\citep{kaliyev2026taskcompletionverificationvsconformancegap}.
The contribution of persistent experience to later personal-agent behavior is isolated through matched
persistence-on/off comparisons in PAST-Bench~\citep{xue2026past}.
Whether self-evolving language-model agents learn new tasks without forgetting earlier ones is evaluated in a
related study~\citep{chen2026selfb}. EmbodiedCLBench evaluates whether embodied agents can reuse the experience gained
on basic tasks to solve unseen compositions of them~\citep{huang2026embodiedclbench},
and Embodied Skill Bench measures how packaged embodied skills help agents on physical tasks~\citep{han2026embodied}. Together, these studies motivate evaluating both
immediate task gains and improvements that
remain useful on subsequent tasks.

\name{} studies task-solution and harness improvement together on vision, robotics, and related downstream tasks.
For each task, every model starts from the same general-purpose harness and starter solution, then alternates
execution and harness revision under a fixed budget. The improver can edit the executor's prompts, tools, and
workflow as well as its own prompts, settings, and lessons. This shared protocol separates three questions: how much the task solution
improves, what additional gains come from editing the harness, and whether a harness built by one model helps another
model on the same task. A review of the evidence behind recursive self-improvement claims finds that few
studies compare improvement procedures from identical starting artifacts at a matched budget~\citep{chen2026product};
the paired runs with and without the improver (\cref{sec:harness}) make this comparison.

\subsection{Frontier agents}

Language-model agents connect reasoning to actions and environmental feedback through an execution loop.
ReAct interleaves reasoning and tool use~\citep{yao2022react}, Reflexion carries verbal feedback into subsequent
trials through episodic memory~\citep{shinn2023reflexion}, and Voyager accumulates executable skills through
open-ended interaction in Minecraft~\citep{wang2023voyager}. Coding interfaces such as Claude Code and Codex CLI
expose file editing and command execution~\citep{anthropic_claudecode,openai_codexcli}. OpenHands provides
infrastructure for software-development agents~\citep{wang2025openhands}, while SWE-agent studies how an
agent--computer interface supports repository navigation, editing, and testing~\citep{yang2024swe}. Research
agents extend these capabilities to experimentation: AIDE searches over candidate
solution code~\citep{jiang2025aide}, and The AI Scientist combines research ideation, experiments, paper writing,
and review in an automated workflow~\citep{lu2024ai}. Frontier models have also been evaluated as robot agents that act in the control
loop~\citep{huang2026llms,zhou2026multimodal,yao2026agent} or write reusable robot tools~\citep{ge2026code}, across
many embodiments~\citep{yang2026robotworld} and against learned policies~\citep{zhang2026bigym}, and coding agents
have been asked to reconstruct dynamic scenes as executable graphics programs~\citep{shen20264dcodebench}.

Automated agent design makes parts of this execution machinery optimization variables. Promptbreeder evolves
both task prompts and the mutation prompts that revise them~\citep{fernando2023promptbreeder}, while GEPA uses
reflection on execution traces to propose and select prompt updates~\citep{agrawal2026gepa}. ADAS searches over
agent programs using an archive of prior designs~\citep{hu2025automated}, and AFlow searches over code-represented
workflows~\citep{zhang2025aflow}. A builder model can also learn principles for constructing the execution
environment of a fixed target model~\citep{qian2026meta}. AlphaEvolve applies
evolutionary program search to scientific and algorithmic discovery, with executable evaluators selecting candidate
task programs~\citep{novikov2025alphaevolve}. These approaches optimize objects ranging from prompts to complete
programs, with different boundaries around the optimization procedure itself.

Self-modifying agents extend this search to their own implementations. STOP applies its code improver to the
improver itself~\citep{zelikman2023self}; G\"odel Agent and SICA likewise revise agent logic and orchestration
code~\citep{yin2025godel,robeyns2025self}. The Darwin G\"odel Machine retains an archive of empirically evaluated,
self-modified agents~\citep{zhang2026darwingodelmachineopenended}, while the Huxley--G\"odel Machine also considers
a candidate's potential to produce better descendants~\citep{wang2026huxley}. Meta-Harness searches harness code
using previous candidates, scores, and execution traces~\citep{lee2026meta}. RHI represents the harness as a
prompt-level specification of the agent loop and refines it using pairwise feedback over its own revision
history~\citep{lee2026recursive}. Self-Harness uses execution failures
to guide revisions to the model's own harness~\citep{zhang2026self}, and AIDE$^2$ recursively revises research-agent
code~\citep{srikanth2026recursive}.

Several systems explicitly distinguish task improvement from improvement of the process that produces it.
Hyperagents combine a task agent and a meta agent in a shared editable program, allowing the self-modification
mechanism itself to evolve and studying transfer of meta-level improvements~\citep{zhang2026hyperagents}. HSI
separates a task harness, an evolver, and a meta-evolver under a fixed outer anchor~\citep{zhou2026hierarchical}.
In robotics, RegenHarness revises harness configuration from execution records, with regression checks and release
gates governing acceptance~\citep{wang2026regenharness}. Harness evolution around frozen models now spans video
understanding~\citep{luo2026video,liang2026vidharness,lin2026videoharness}, image generation with a frozen
generator~\citep{han2026imagersi}, 3D reconstruction~\citep{yao2026code}, and embodied agents, including robot manipulation, where the
harness is revised from diagnosed execution failures~\citep{tian2026physevo,feng2026enactopt,yu2026robodoctor,du2026vlm},
experience is organized into knowledge or executable stacks shared across tasks~\citep{bao2026roboharn,chen2026compose,wen2026roborsi},
the system decides where to spend robot trials~\citep{song2026embodiedrsi},
and changes are tested in simulation before they reach the robot~\citep{hu2026simex}.
Studies of these loops find that the number of rollouts the optimizer sees in each round and the selection rule
determine whether an evolved robot harness generalizes~\citep{wei2026and}, that accepting any edit that raises an
aggregate validation score admits regressions and favors noise~\citep{wang2026sage}, that proxy improvements need not
hold where the update is deployed~\citep{wang2026better}, and that an exact validation score after every attempt
opens the validation split to adaptive selection~\citep{yao2026explicit}. RRSI addresses the risk that harness
evolution overfits its training tasks, with gains that shrink or vanish on out-of-distribution benchmarks, by
constraining how edits are proposed and selected~\citep{xia2026rrsi}. Other studies ask which failures harness
evolution can repair and when weight training should take over~\citep{tian2026evolution,wang2026adarsi}, argue that
rewriting the scaffold around frozen weights avoids diminishing returns but not an eventual
ceiling~\citep{bobadillasuarez2026scaffold},
and report that weak evaluators and early failures in long skill chains stall improvement over long robot
runs~\citep{wang2026stops}. Surveys organize these developments by the components that evolve and the feedback
that drives their updates~\citep{gao2025survey,fang2025comprehensive,chen2026self}, and recent surveys and position papers cover
recursive self-improvement in physical AI~\citep{han2026self,zhou2026improves}, agent harnesses for video
generation~\citep{fan2026video}, and the use of frontier models for 3D modeling, computational design, and
robotics~\citep{dou2026frontier3drobotics}.

These systems propose particular ways to accumulate experience, search for agent designs, or revise an improvement
procedure. \name{} evaluates the ability to make such revisions through a shared executor--improver harness
(\cref{sec:method-harness}). Each frontier model operates both components from the same initial harness and, on
each task, the same starter solution and budget. The executor changes the task solution; the improver changes the
executor and its own instructions and lessons. The evaluator, data splits, and budget accounting remain fixed.
This setting makes the model's capacity to improve its operating machinery the object of comparison across vision,
robotics, and related tasks, while keeping changes to the solution distinct from changes to the harness.

\subsection{\name{} tasks}
\label{sec:task-sources}

\name{} draws its 24 tasks from six areas, combining established benchmarks, source datasets, and our own
curated creation tasks. \Cref{tab:tasks} summarizes their sources and metrics. Each task provides a starter
solution, development data, hidden validation and test splits, and a fixed improvement budget, so that these
resources support the evaluation of changes to both task solutions and agent harnesses.

\paragraph{Action}
The manipulation tasks use simulated environments from RoboDojo~\citep{chen2026robodojo} and RoboCasa365~\citep{nasiriany2026robocasa365}, which
extends RoboCasa's household simulation platform~\citep{nasiriany2024robocasa}. Active Search uses H*Bench's panoramic environments
for visual search~\citep{yu2025thinking360deghumanoidvisual}, while RoboLab uses the RoboLab-120 suite to evaluate a generalist tabletop manipulation
policy~\citep{yang2026robolab}. The LIBERO-based tasks~\citep{liu2023libero} examine robustness to changes in initial conditions and task
specifications through LIBERO-PRO~\citep{zhou2025libero}, and adaptation to changes during execution through LIBERO-MAX~\citep{zhang2026libero}. Embodied-SWE draws on EmbodiedSWE-Bench, where agents write robot-control programs and receive
scores based on physical task progress~\citep{you2026embodiedswe}. Related simulation benchmarks extend such
evaluation to whole-body humanoid loco-manipulation, as in SIMPLE~\citep{wei2026simple}.

\paragraph{Reasoning}
Ego-Exo Memory uses the Ego-Exo4D portion of EgoExoMem for memory reasoning across synchronized first- and
third-person videos~\citep{liu2026egoexomem,grauman2024ego}. Kinematics uses QuantiPhy to infer physical quantities such as sizes,
velocities, and accelerations from video~\citep{puyin2026quantiphy}; Mass adapts product photographs and metadata from Amazon
Berkeley Objects for weight estimation~\citep{collins2022abo}. Abstract reasoning spans grid transformations
in ARC-AGI-1 and ARC-AGI-2~\citep{chollet2019measure,chollet2025arc} and interactive exploration in ARC-AGI-3~\citep{foundation2026arc}. Spatial-VSI
uses VSI-Bench's spatial questions about indoor videos~\citep{yang2025thinking}.

\paragraph{Creation}
Image Editing uses our own curated images and editing instructions, and Text-to-Image uses original Chinese and
English prompts. Multi-turn Editing follows the five-turn sessions of MSE-Bench, introduced with VINCIE, in which
each instruction acts on the previous result~\citep{qu2026vincie}. Video Generation uses the compositional prompts and
evaluation criteria of T2V-CompBench~\citep{sun2025t2v}.

\paragraph{Reconstruction}
3D Tracking adapts IGGT's instance-tracking setting~\citep{li2025iggt} on ScanNet and ScanNet++~\citep{dai2017scannet,yeshwanth2023scannet++}. Scene from
Photo uses HARMONY300, introduced with HARMONY for reconstructing furnished rooms from single images~\citep{sun2026harmony};
its synthetic scenes derive from 3D-FRONT~\citep{fu20213d}. Scene from Video uses the indoor scenes released with HoloScene~\citep{xia2026holoscene}, drawn from Replica and
ScanNet++~\citep{straub2019replica,yeshwanth2023scannet++}, together with CoDimRecon's geometry, rendering,
and physical-stability evaluation setting~\citep{xie2026codimrecon}.

\paragraph{Perception}
Visual Search combines high-resolution questions from V*Bench and HR-Bench 8K~\citep{wu2024v,wang2025divide}. Video Quality
combines subjective quality scores for YouTube-UGC videos~\citep{wang2019youtube,yim2020subjective} with human preferences
on generated-video pairs from Rapidata's Veo~3.1 comparison dataset~\citep{rapidata_veo31_preferences}.

\paragraph{Infrastructure}
VAE Decode uses the variational autoencoder released with MiniMax H3~\citep{minimax_h3_modelcard} and asks the agent to accelerate its
decoder while preserving the reference outputs on unseen inputs.

\section{\name}
\label{sec:benchmark}

\name{} evaluates whether an agent, given a task and a fixed budget, can improve not only its solution to the task
but also the harness through which it operates. We first define the task setting (\cref{sec:task-formulation}), then the
self-improving harness (\cref{sec:method-harness}), and finally how runs are evaluated (\cref{sec:protocol}).

\subsection{Problem setup}
\label{sec:task-formulation}

Each task provides the agent with a starter solution, compute resources, a budget, and data partitioned into
three disjoint splits. The \emph{development split}, including its labels, is available to the agent for
experimentation and evaluation. The \emph{validation split} is withheld from the agent and determines which candidate
solutions are retained. The \emph{test split} is likewise withheld and is used only once, after the run terminates,
to evaluate the final solution. The objective of the agent is to maximize the test score of its final solution within
the budget.

On a given task, every run starts from the same starter solution and is subject to the same budget, which limits
wall-clock time, GPU hours, model usage, the number of validation evaluations, and the number of rounds. The 24 tasks are listed in \cref{tab:tasks} and described in \cref{app:tasks}.

\begin{table}[t]
  \centering\footnotesize
  \caption{\textbf{The task suite by area.} Source benchmark or dataset, primary metric, and size of the hidden test
  split of each task. SPL: success weighted by path length; MRA: mean relative accuracy; T-mIoU: temporal mean IoU
  of the instance tracks; RHAE: the official ARC-AGI-3 score of action efficiency relative to humans; SRCC: Spearman
  rank correlation; VLM: vision-language model. Ours: data collected for
  \name{} (\cref{app:tasks}).}
  \label{tab:tasks}
  \begin{fittabular}{@{}lllll@{}}
    \toprule
    \colhead{Area} & \colhead{Task} & \colhead{Source} & \colhead{Metric} & \colhead{Test split} \\
    \midrule
    \multirow{7}{*}{Action}
      & RoboDojo & RoboDojo & Task score & 10 tasks $\times$ 5 layouts \\
      & RoboCasa & RoboCasa365 & Success rate & 8 tasks $\times$ 5 episodes \\
      & Active Search & H*Bench & SPL & 47 episodes \\
      & RoboLab & RoboLab-120 & RoboLab score & 24 tasks $\times$ 2 episodes \\
      & LIBERO-MAX & LIBERO-MAX & Dynamic success rate & 24 episode pairs \\
      & LIBERO-PRO & LIBERO-PRO & Success rate & 30 episodes \\
      & Embodied-SWE & EmbodiedSWE-Bench & Rubric score & 4 seeds $\times$ 8 environments \\
    \midrule
    \multirow{7}{*}{Reasoning}
      & Ego-Exo Memory & EgoExoMem (Ego-Exo4D) & Accuracy & 120 questions \\
      & Kinematics & QuantiPhy & MRA & 67 questions \\
      & Mass & Amazon Berkeley Objects & MRA & 600 items \\
      & ARC-AGI-1 & ARC-AGI-1 evaluation set & pass@2 & 400 puzzles \\
      & ARC-AGI-2 & ARC-AGI-2 evaluation set & pass@2 & 120 puzzles \\
      & ARC-AGI-3 & ARC-AGI-3 public games & RHAE & 10 games \\
      & Spatial-VSI & VSI-Bench & VSI-Bench score & 320 questions \\
    \midrule
    \multirow{4}{*}{Creation}
      & Image Editing & Ours & VLM judge + checks & 80 images \\
      & Multi-turn Editing & MSE-Bench & VLM judge & 40 sessions \\
      & Video Generation & T2V-CompBench & Composition score & 20 prompts \\
      & Text-to-Image & Ours & VLM judge & 80 prompts \\
    \midrule
    \multirow{3}{*}{Reconstruction}
      & 3D Tracking & IGGT (ScanNet, ScanNet++) & T-mIoU & 10 clips \\
      & Scene from Photo & HARMONY300 & Composite score & 60 photos \\
      & Scene from Video & HoloScene & Composite score & 6 scenes \\
    \midrule
    \multirow{2}{*}{Perception}
      & Visual Search & V*Bench, HR-Bench 8K & Accuracy & 140 questions \\
      & Video Quality & YouTube-UGC, Veo~3.1 pairs & (SRCC + pairwise acc.)\,/\,2 & 40 videos, 20 pairs \\
    \midrule
    Infrastructure & VAE Decode & MiniMax H3 VAE decoder & $\log_2$ speedup & 12 scenes \\
    \bottomrule
  \end{fittabular}
\end{table}

\subsection{The self-improving harness}
\label{sec:method-harness}

A harness is the program that mediates between a language model and a task. Through its prompts, tools, control
loop, and memory, it determines the context presented to the model and the actions available to it. Since the harness
consists of code and text, the agent can modify it in the same manner as any other program, while the model weights
remain fixed. In \name{}, the harness $H=(E,I)$ comprises two components that query the same model. The
\emph{executor} $E$ operates on the task solution: it modifies the solution, conducts experiments, and evaluates the
results on the development split. The \emph{improver} $I$ operates on the harness: it analyses the executor's
trajectories and results, diagnoses failures, and revises the prompts, tools, and workflow of the executor as well as
its own instructions and accumulated lessons.

Let $s_0$ denote the starter solution, $\mathcal{D}_{\mathrm{dev}}$, $\mathcal{D}_{\mathrm{val}}$, and
$\mathcal{D}_{\mathrm{test}}$ the development, validation, and test splits, and $S(s;\mathcal{D})$ the score that the
task evaluator assigns to a solution $s$ on a split $\mathcal{D}$. A run starts from the harness $H_0=(E_0,I_0)$ and
proceeds over rounds $t=1,2,\dots$. In round $t$, the executor of the current harness modifies the accepted solution
$s_{t-1}$, using only the development split, and submits a candidate
\begin{equation}
  \tilde{s}_t = E_{t-1}\bigl(s_{t-1};\,\mathcal{D}_{\mathrm{dev}}\bigr).
  \label{eq:executor}
\end{equation}
The candidate is then evaluated on the validation split and retained only if it improves on the accepted solution by
more than a noise margin $\delta$, which the starting harness sets to twice the standard deviation of the starter's validation
score and which the improver may revise (\cref{app:protocol}):
\begin{equation}
  s_t =
  \begin{cases}
    \tilde{s}_t & \text{if } S(\tilde{s}_t;\mathcal{D}_{\mathrm{val}}) > S(s_{t-1};\mathcal{D}_{\mathrm{val}}) + \delta,\\
    s_{t-1} & \text{otherwise}.
  \end{cases}
  \label{eq:accept}
\end{equation}
Finally,
the improver examines the records $\tau_{1:t}$ of all rounds so far, which comprise the executor's trajectories,
experiments, and scores, and produces the harness for the next round:
\begin{equation}
  (E_t,\,I_t) = I_{t-1}\bigl(E_{t-1},\,I_{t-1};\,\tau_{1:t}\bigr).
  \label{eq:improver}
\end{equation}
The run terminates at the round $T$ in which the budget is exhausted or progress stops (\cref{app:protocol}), and its
result is the final solution $s_T$.

\Cref{eq:improver} makes the recursion explicit: the improver is both the operator that produces the next harness
and one of its arguments, so that a modification of $I$ changes how all subsequent modifications are made. The
decomposition also separates the two kinds of improvement. A change to the solution takes effect only after it has
been confirmed on data unseen by the agent (\cref{eq:accept}), whereas a change to the harness is reflected in the
candidates of subsequent rounds (\cref{eq:executor}).

All runs start from the same general-purpose harness $H_0$, which contains no task- or model-specific components.
The evaluator $S$, the data splits, and the accounting of the budget are fixed and cannot be modified by the agent.
Differences between runs can therefore be attributed to the agents themselves, and runs of different models on the
same task are directly comparable.

\subsection{Evaluation}
\label{sec:protocol}

We report the test score of the final solution, $S(s_T;\mathcal{D}_{\mathrm{test}})$, and its improvement over the
starter solution, $S(s_T;\mathcal{D}_{\mathrm{test}})-S(s_0;\mathcal{D}_{\mathrm{test}})$. Agent code is executed
in a sandbox without access to the validation or test data, so no run can read the held-out data.

\section{Experiments}
\label{sec:experiments}

This section reports what the agents achieve, in the order in which a reader would ask about it: how well each model
does over all tasks and within each area (\cref{sec:main-results}), how much of that is due to self-improvement
(\cref{sec:selfimprove}), in which rounds the gains arrive (\cref{sec:rounds}), and what they cost
(\cref{sec:cost}). It reports what happens; \cref{sec:analysis} examines why.

\subsection{Models and setup}
\label{sec:setup}

We evaluate 18 models, comprising 11 closed-source models, namely Claude Opus~5.5~\citep{anthropic2026opus55}, Claude
Sonnet~5.5~\citep{anthropic2026sonnet55}, GPT-6 Astra~\citep{openai2026gpt6}, GPT-6.1 Sol~\citep{openai2026gpt61sol},
GPT-5.6 Sol~\citep{openai2026gpt56}, Gemini~4 Argon~\citep{google2026gemini4argon}, Gemini~3.8
Flash~\citep{google2026gemini38}, Gemini~3.7 Flash~\citep{google2026gemini37}, Grok~4.6~\citep{xai2026grok46}, Muse
Spark~1.3~\citep{meta2026musespark}, and Muse Wonder~1.0~\citep{meta2026musewonder}, and 7 open-weight models, namely
Qwen3.8-27B and Qwen3.8-Flash-Next~\citep{qwen2026qwen38}, Qwen3.6-35B-A3B~\citep{qwen2026qwen36},
Gemma-4-31B-it~\citep{gemma2026gemma4}, MiniMax-M3~\citep{minimax2026m3}, GLM-5.3 Flash~\citep{glm2026glm5}, and
DeepSeek-V4.1 Flash~\citep{deepseek2026v4}.\footnote{We evaluate the 1017 version of DeepSeek-V4.1 Flash; it is referred to as DeepSeek-V4.1 Flash throughout.} Closed-source models are accessed through their providers' interfaces,
and open-weight models are served on our own GPUs. Every model operates within the same starting harness
(\cref{sec:method-harness}) and, on a given task, under the same budget.

\paragraph{How results are reported}
Each model is run once on each task it is evaluated on; where a run was repeated, the latest run is reported.
RoboDojo and RoboCasa are suites of 10 and 8 robot tasks, each run separately for 10 rounds and aggregated into one
suite score. A run's result is the test score of its final solution on the task's primary metric, on a 0--100 scale.
Because the metrics differ across tasks, models are compared through ranks: on each task, a model scores the share of
the other models evaluated on the task that it beats, from 0 (last) to 100 (first). Its score in an area is the mean
over the tasks of the area it was run on, and its overall score is the mean of its five area scores. To measure
self-improvement itself, we also report the validation score of the starter solution and the best validation score
reached during the run. The main results cover the 20 tasks evaluated with at least six models; the four tasks
evaluated with two models so far are reported in \cref{tab:res-small}.

\subsection{Main results}
\label{sec:main-results}

\begin{figure}[t]
  \centering
  \begin{subfigure}[b]{0.54\linewidth}
    \centering
    \includegraphics[width=\linewidth]{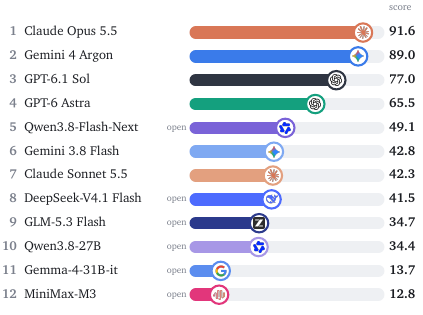}
    \caption{Score of every model run on at least five tasks.}
    \label{fig:leaderboard}
  \end{subfigure}\hfill
  \begin{subfigure}[b]{0.44\linewidth}
    \centering
    \includegraphics[width=\linewidth]{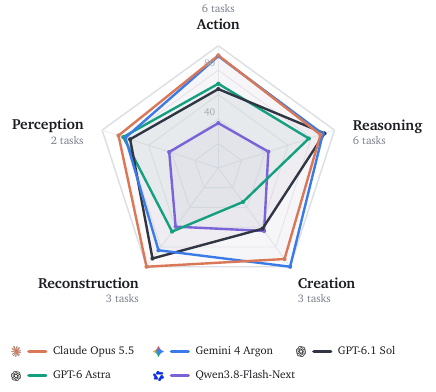}
    \caption{Per-area score of the five leading models.}
    \label{fig:radar}
  \end{subfigure}
  \caption{\textbf{Leaderboard.} On each task a model scores the share of the other models it beats (100 = best).
  (b)~gives a model's mean over the tasks of each area it was run on, further out is stronger; (a)~gives the mean of
  these five area scores.}
  \label{fig:board}
\end{figure}

\Cref{fig:leaderboard} gives one score per model and \cref{fig:radar} breaks it down by area. Claude Opus~5.5 has
the highest score (91.6): it is one of two models run on all 20 tasks and has the best test score on 11 of them.
Gemini~4 Argon follows closely (89.0) over the nine tasks it was run on, then GPT-6.1 Sol (77.0) and GPT-6 Astra
(65.5). The best open-weight model, Qwen3.8-Flash-Next (49.1), places fifth, ahead of Gemini~3.8 Flash and Claude
Sonnet~5.5. The leaders differ by area: Sol matches Opus and Argon on reasoning but trails them on action and
creation, and Astra is weakest on creation. The rest of this section takes the areas in turn; each table gives the
test score of every model on every task of the area, and \cref{app:results} adds the starter's score, the
validation scores and the rounds of every run (and the average rank of each model, \cref{tab:leaderboard}).

\begin{table}[!htb]
  \centering\footnotesize
  \caption{\textbf{Action.} Test score on each task (0--100); models ordered by average rank over these tasks.}
  \label{tab:lb-action}
  \begin{fittabular}{@{}l c cccccc}
    \toprule
    \colhead{Model} & \phead{Avg.\ rank} & \colhead{Active Search} & \colhead{RoboLab} & \colhead{LIBERO-MAX} & \colhead{LIBERO-PRO} & \colhead{RoboDojo} & \colhead{RoboCasa} \\
    \midrule
    \mlogo{claude}Claude Opus 5.5 & \pbest{1.5} & 74.2 & \best{71.9} & \best{45.8} & \best{63.3} & \best{55.1} & \best{57.5}\\\rowrule
    \mlogo{gemini}Gemini 4 Argon & \pcol{1.7} & \best{79.7} & \weak{--} & \weak{--} & \weak{--} & 46.8 & 40.0\\\rowrule
    \mlogo{gpt}GPT-6.1 Sol & \pcol{2.8} & 74.4 & 62.5 & 41.7 & 53.3 & 41.5 & 32.5\\\rowrule
    \mlogo{gpt}GPT-6 Astra & \pcol{3.5} & 78.2 & 33.0 & \weak{--} & \weak{--} & 28.6 & 37.5\\\rowrule
    \mlogo{qwen}Qwen3.6-35B-A3B & \pcol{3.5} & \weak{--} & \weak{--} & 25.0 & 53.3 & \weak{--} & \weak{--}\\\rowrule
    \mlogo{gpt}GPT-5.6 Sol & \pcol{4.0} & \weak{--} & 37.8 & \weak{--} & \weak{--} & \weak{--} & \weak{--}\\\rowrule
    \mlogo{qwen}Qwen3.8-Flash-Next & \pcol{5.2} & 59.3 & 17.4 & \best{45.8} & 53.3 & 8.4 & 5.0\\\rowrule
    \mlogo{qwen}Qwen3.8-27B & \pcol{6.0} & 69.0 & 31.6 & 25.0 & 50.0 & \weak{--} & 0.0\\\rowrule
    \mlogo{claude}Claude Sonnet 5.5 & \pcol{6.0} & 55.8 & 41.0 & \weak{--} & \weak{--} & \weak{--} & \weak{--}\\\rowrule
    \mlogo{glm}GLM-5.3 Flash & \pcol{6.0} & 60.5 & \weak{--} & \weak{--} & \weak{--} & \weak{--} & 7.5\\\rowrule
    \mlogo{gemini}Gemini 3.8 Flash & \pcol{6.0} & 65.5 & \weak{--} & \weak{--} & \weak{--} & \weak{--} & \weak{--}\\\rowrule
    \mlogo{meta}Muse Wonder 1.0 & \pcol{6.0} & \weak{--} & \weak{--} & \weak{--} & \weak{--} & 10.6 & \weak{--}\\\rowrule
    \mlogo{gemma}Gemma-4-31B-it & \pcol{6.4} & 52.2 & 5.2 & 29.2 & 53.3 & \weak{--} & 2.5\\\rowrule
    \mlogo{deepseek}DeepSeek-V4.1 Flash & \pcol{8.0} & 47.1 & \weak{--} & \weak{--} & \weak{--} & 17.8 & 0.0\\\rowrule
    \mlogo{minimax}MiniMax-M3 & \pcol{9.0} & 32.5 & 8.3 & \weak{--} & \weak{--} & 0.0 & 0.0 \\
    \bottomrule
  \end{fittabular}
\end{table}

\paragraph{Action}
The model's solution drives an agent: an active searcher in 360\textdegree{} scenes (Active Search, SPL), one
generalist tabletop policy (RoboLab), a Franka arm in two LIBERO variants, a dual-arm ARX~X5 robot (RoboDojo, 10
tasks), and a mobile manipulator in a simulated kitchen (RoboCasa, 8 tasks). Opus has the best test score on five of
the six tasks, and Argon on Active Search (79.7). The gap between closed-source and open-weight models is widest here: on
RoboDojo and RoboCasa the best open-weight models reach 17.8 and 7.5, against 55.1 and 57.5 for Opus. These two
suites start from nothing, the starter scoring 0 on validation in 4 of 8 RoboDojo runs and 7 of 10 RoboCasa runs, so
a model must first write a working policy before a round can improve the score. \Cref{tab:res-robodojo,tab:res-robocasa} give the results of the two
suites task by task.

\begin{table}[!htb]
  \centering\footnotesize
  \caption{\textbf{Reasoning.} Test score on each task (0--100); models ordered by average rank over these tasks.}
  \label{tab:lb-reasoning}
  \begin{fittabular}{@{}l c cccccc}
    \toprule
    \colhead{Model} & \phead{Avg.\ rank} & \colhead{Ego-Exo} & \colhead{Kinematics} & \colhead{Mass} & \colhead{ARC-1} & \colhead{ARC-2} & \colhead{ARC-3} \\
    \midrule
    \mlogo{gpt}GPT-6.1 Sol & \pbest{1.8} & 67.5 & \weak{--} & \weak{--} & 98.8 & \best{95.8} & \best{44.2}\\\rowrule
    \mlogo{claude}Claude Opus 5.5 & \pcol{1.8} & \best{72.5} & 54.6 & \best{63.5} & 98.0 & 95.0 & 40.4\\\rowrule
    \mlogo{gemini}Gemini 4 Argon & \pcol{2.0} & 70.0 & \weak{--} & \weak{--} & \weak{--} & \weak{--} & \weak{--}\\\rowrule
    \mlogo{claude}Claude Sonnet 5.5 & \pcol{2.5} & \weak{--} & \best{57.2} & 62.5 & \weak{--} & \weak{--} & \weak{--}\\\rowrule
    \mlogo{gpt}GPT-5.6 Sol & \pcol{2.5} & \weak{--} & 50.0 & 63.0 & \weak{--} & \weak{--} & \weak{--}\\\rowrule
    \mlogo{gpt}GPT-6 Astra & \pcol{2.8} & 65.8 & \weak{--} & \weak{--} & \best{99.8} & 93.3 & 36.5\\\rowrule
    \mlogo{qwen}Qwen3.8-Flash-Next & \pcol{5.2} & 59.2 & 49.2 & 60.4 & 80.8 & 22.5 & 0.9\\\rowrule
    \mlogo{gemini}Gemini 3.7 Flash & \pcol{5.5} & \weak{--} & 46.3 & 60.6 & \weak{--} & \weak{--} & \weak{--}\\\rowrule
    \mlogo{qwen}Qwen3.8-27B & \pcol{5.7} & 60.8 & 43.4 & 62.7 & 77.2 & 19.2 & 0.3\\\rowrule
    \mlogo{deepseek}DeepSeek-V4.1 Flash & \pcol{6.0} & 50.8 & \weak{--} & \weak{--} & 91.0 & 46.7 & 0.6\\\rowrule
    \mlogo{gemini}Gemini 3.8 Flash & \pcol{6.0} & 63.3 & 48.2 & 57.5 & \weak{--} & \weak{--} & \weak{--}\\\rowrule
    \mlogo{grok}Grok 4.6 & \pcol{7.5} & \weak{--} & 37.2 & 59.3 & \weak{--} & \weak{--} & \weak{--}\\\rowrule
    \mlogo{glm}GLM-5.3 Flash & \pcol{7.7} & 56.7 & \weak{--} & \weak{--} & 63.5 & 14.2 & \weak{--}\\\rowrule
    \mlogo{gemma}Gemma-4-31B-it & \pcol{7.8} & 59.2 & \weak{--} & \weak{--} & 54.0 & 1.7 & 0.0\\\rowrule
    \mlogo{minimax}MiniMax-M3 & \pcol{9.0} & 51.7 & \weak{--} & \weak{--} & \weak{--} & 4.2 & \weak{--} \\
    \bottomrule
  \end{fittabular}
\end{table}

\paragraph{Reasoning}
The model answers memory questions over synchronized egocentric and exocentric video (Ego-Exo Memory), estimates
physical quantities from video and images (Kinematics and Mass, mean relative accuracy), and solves ARC-AGI-1 and -2
grid puzzles and the interactive games of ARC-AGI-3. Closed models lead every task: Opus on Ego-Exo Memory (72.5) and
Mass (63.5), Sonnet on Kinematics (57.2), Astra on ARC-AGI-1 (99.8), and Sol on ARC-AGI-2 (95.8) and ARC-AGI-3
(44.2). Open-weight models are close on the physical quantities (Qwen3.8-27B scores 62.7 on Mass) but fall away on
ARC: the best of them reaches 46.7 on ARC-AGI-2 and 0.9 on ARC-AGI-3. ARC-AGI-3 is also one of the tasks where self-improvement does
least, with a median validation score of 3.8 for the starter and 4.1 for the best solution.

\begin{table}[!htb]
  \begin{minipage}[t]{0.495\linewidth}
  \centering\footnotesize
  \caption{\textbf{Creation.} Test score on each task (0--100); models ordered by average rank over these tasks. Image, Multi-turn and Video: Image Editing, Multi-turn Editing and Video Generation.}
  \label{tab:lb-creation}
  \setlength{\tabcolsep}{3.5pt}
  \begin{tabular}{@{}lcccc}
    \toprule
    \colhead{Model} & \phead{Avg.\ rank} & \colhead{Image} & \colhead{Multi-turn} & \colhead{Video} \\
    \midrule
    \mlogo{gemini}Gemini 4 Argon & \pbest{1.0} & \best{74.2} & \best{71.1} & \weak{--}\\\rowrule
    \mlogo{claude}Claude Opus 5.5 & \pcol{1.7} & 73.8 & 70.3 & \best{93.9}\\\rowrule
    \mlogo{qwen}Qwen3.8-Flash-Next & \pcol{3.7} & 70.0 & 69.2 & 91.4\\\rowrule
    \mlogo{gpt}GPT-6.1 Sol & \pcol{4.0} & 69.9 & 68.3 & 91.5\\\rowrule
    \mlogo{qwen}Qwen3.8-27B & \pcol{4.7} & 69.7 & 70.2 & 81.8\\\rowrule
    \mlogo{gpt}GPT-6 Astra & \pcol{6.5} & 69.7 & 67.8 & \weak{--}\\\rowrule
    \mlogo{gemma}Gemma-4-31B-it & \pcol{6.7} & 68.4 & 66.3 & 87.1\\\rowrule
    \mlogo{claude}Claude Sonnet 5.5 & \pcol{7.0} & 66.0 & 69.8 & \weak{--}\\\rowrule
    \mlogo{minimax}MiniMax-M3 & \pcol{7.5} & 64.3 & \weak{--} & 87.3\\\rowrule
    \mlogo{deepseek}DeepSeek-V4.1 Flash & \pcol{8.0} & 67.6 & \weak{--} & \weak{--}\\\rowrule
    \mlogo{gemini}Gemini 3.8 Flash & \pcol{9.0} & 67.2 & \weak{--} & \weak{--}\\\rowrule
    \mlogo{glm}GLM-5.3 Flash & \pcol{9.5} & 60.6 & \weak{--} & 79.6 \\
    \bottomrule
  \end{tabular}
  \end{minipage}\hfill
  \begin{minipage}[t]{0.495\linewidth}
  \centering\footnotesize
  \caption{\textbf{Reconstruction.} Test score on each task (0--100); models ordered by average rank over these tasks. Photo and Video: Scene from Photo and Scene from Video.\newline\mbox{}}
  \label{tab:lb-reconstruction}
  \setlength{\tabcolsep}{3.5pt}
  \begin{tabular}{@{}lcccc}
    \toprule
    \colhead{Model} & \phead{Avg.\ rank} & \colhead{3D Track.} & \colhead{Photo} & \colhead{Video} \\
    \midrule
    \mlogo{claude}Claude Opus 5.5 & \pbest{1.0} & \best{79.6} & \best{66.5} & \best{70.9}\\\rowrule
    \mlogo{gpt}GPT-6.1 Sol & \pcol{2.0} & 77.3 & \weak{--} & \weak{--}\\\rowrule
    \mlogo{gemini}Gemini 4 Argon & \pcol{3.0} & 73.8 & \weak{--} & \weak{--}\\\rowrule
    \mlogo{glm}GLM-5.3 Flash & \pcol{4.0} & 61.2 & \weak{--} & \weak{--}\\\rowrule
    \mlogo{qwen}Qwen3.8-Flash-Next & \pcol{4.0} & 58.9 & 40.0 & 34.5\\\rowrule
    \mlogo{gpt}GPT-6 Astra & \pcol{4.3} & 52.8 & 53.2 & 53.4\\\rowrule
    \mlogo{deepseek}DeepSeek-V4.1 Flash & \pcol{5.0} & 60.2 & \weak{--} & \weak{--}\\\rowrule
    \mlogo{qwen}Qwen3.8-27B & \pcol{6.3} & 52.4 & 29.0 & 26.7\\\rowrule
    \mlogo{meta}Muse Wonder 1.0 & \pcol{7.0} & 57.9 & \weak{--} & \weak{--}\\\rowrule
    \mlogo{minimax}MiniMax-M3 & \pcol{7.0} & 27.9 & 33.3 & 26.7\\\rowrule
    \mlogo{gemini}Gemini 3.8 Flash & \pcol{8.0} & 53.7 & \weak{--} & \weak{--}\\\rowrule
    \mlogo{gemma}Gemma-4-31B-it & \pcol{8.0} & 32.0 & 15.4 & 17.5\\\rowrule
    \mlogo{claude}Claude Sonnet 5.5 & \pcol{11.0} & 47.9 & \weak{--} & \weak{--} \\
    \bottomrule
  \end{tabular}
  \end{minipage}
\end{table}

\paragraph{Creation}
The model edits images from instructions (Image Editing), edits one photo over five turns (Multi-turn Editing), and
builds text-to-video workflows around a frozen generator (Video Generation, composition score). Scores are
compressed: the 12 models span 60.6 to 74.2 on Image Editing, and the top six on Multi-turn Editing lie within 2.8
points. Open-weight models are most competitive here: Qwen3.8-Flash-Next ranks third in the area and scores 91.4 on
Video Generation, against 93.9 for Opus.

\paragraph{Reconstruction}
The model tracks object instances in 3D (3D Tracking, temporal mIoU) and rebuilds a furnished room from a photo or an
indoor scene from a video (scene score). Opus leads all three tasks (79.6, 66.5 and 70.9), 13 to 18 points ahead of
the next model, GPT-6 Astra, on the two scene tasks. Reconstruction gains the most from self-improvement: the median
validation score rises from 19.4 to 62.2 on 3D Tracking and from 7.7 to 38.0 on Scene from Photo.

\begin{table}[!htb]
  \centering\footnotesize
  \caption{\textbf{Perception.} Test score on each task (0--100); models ordered by average rank over these tasks.}
  \label{tab:lb-perception}
  \begin{tabular*}{\linewidth}{@{\extracolsep{\fill}}lccc@{}}
    \toprule
    \colhead{Model} & \phead{Avg.\ rank} & \colhead{Visual Search} & \colhead{Video Quality} \\
    \midrule
    \mlogo{claude}Claude Opus 5.5 & \pbest{2.0} & 96.4 & 82.0\\\rowrule
    \mlogo{gpt}GPT-6 Astra & \pcol{3.0} & 93.6 & \best{82.1}\\\rowrule
    \mlogo{gpt}GPT-6.1 Sol & \pcol{3.0} & 96.4 & 80.8\\\rowrule
    \mlogo{gemini}Gemini 4 Argon & \pcol{3.0} & \best{98.6} & 80.6\\\rowrule
    \mlogo{gemini}Gemini 3.8 Flash & \pcol{5.5} & 95.0 & 77.5\\\rowrule
    \mlogo{deepseek}DeepSeek-V4.1 Flash & \pcol{6.5} & 77.1 & 82.0\\\rowrule
    \mlogo{qwen}Qwen3.8-27B & \pcol{6.5} & 85.7 & 78.0\\\rowrule
    \mlogo{qwen}Qwen3.8-Flash-Next & \pcol{7.0} & 88.6 & 76.6\\\rowrule
    \mlogo{glm}GLM-5.3 Flash & \pcol{8.0} & 85.7 & 76.2\\\rowrule
    \mlogo{claude}Claude Sonnet 5.5 & \pcol{9.0} & 80.7 & \weak{--}\\\rowrule
    \mlogo{gemma}Gemma-4-31B-it & \pcol{10.5} & 72.9 & 74.3\\\rowrule
    \mlogo{minimax}MiniMax-M3 & \pcol{11.5} & 70.7 & 69.8 \\
    \bottomrule
  \end{tabular*}
\end{table}

\paragraph{Perception}
The model finds small targets in high-resolution images (Visual Search, accuracy) and rates the perceptual quality of
real user videos (Video Quality). The starters already score high, with a median validation score of 80.0 and 78.2,
and the leaders are a few points apart: Gemini~4 Argon scores 98.6 on Visual Search, and on Video Quality GPT-6 Astra
(82.1), Claude Opus~5.5 and DeepSeek-V4.1 Flash (82.0) are within 0.1 of each other. With little headroom left,
the best open-weight model is 10 points behind on Visual Search and level on
Video Quality.

\subsection{How much self-improvement adds}
\label{sec:selfimprove}

\begin{figure}[t]
  \centering
  \begin{subfigure}[b]{0.514\linewidth}
    \centering
    \includegraphics[width=\linewidth]{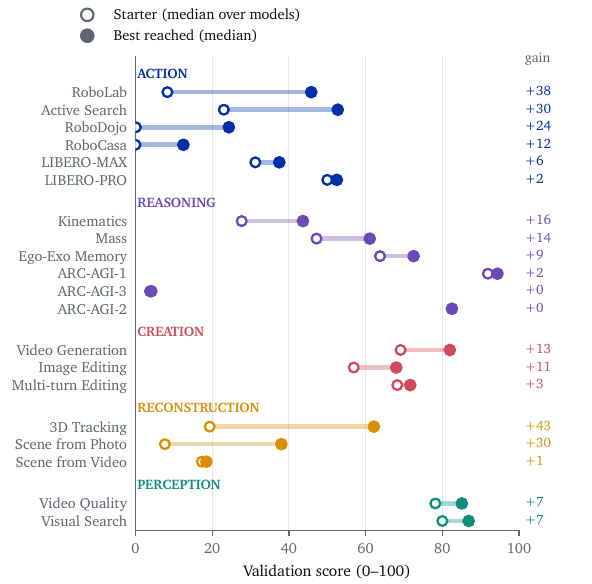}
    \caption{By task: median over its models.}
    \label{fig:gains}
  \end{subfigure}\hfill
  \begin{subfigure}[b]{0.466\linewidth}
    \centering
    \includegraphics[width=\linewidth]{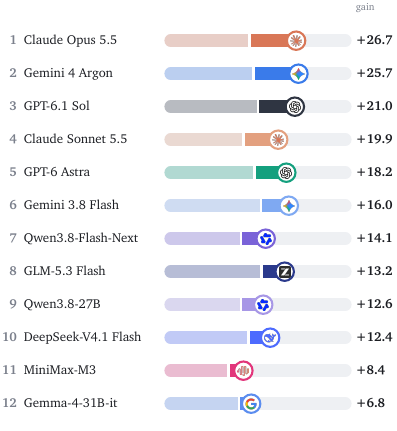}
    \caption{By model: mean over its runs.}
    \label{fig:improve}
  \end{subfigure}
  \caption{\textbf{Where self-improvement helps.} Validation score of the starter (open circle; light bar) and best
  validation score reached during the run (filled circle; dark bar), on a 0--100 scale. (a)~One row per task, grouped
  by area and sorted by gain; the number on the right is the difference between the two medians. (b)~One row per model run on at least five tasks, sorted by the mean gain.}
  \label{fig:selfimprove}
\end{figure}

The scores above combine what a model can do from the start with what it gains by improving. \Cref{fig:selfimprove}
separates the two by comparing the starter's validation score with the best validation score reached during the run.
Self-improvement raises the validation score in 150 of the 177 runs (85\%), by a median of 10.4 points.

The gain depends on how much the task leaves to build. It is largest where the agent builds most of its solution
from a weak starter: on 3D Tracking, RoboLab and Active Search the median validation score rises from 19.4, 8.3 and
23.0 to 62.2, 45.8 and 52.7 (\cref{fig:gains}). It is smallest where the starter already solves most of the task, as
on ARC-AGI-1 and ARC-AGI-2 (median starter 91.9 and 82.5, a median gain of 0 points on both). Two tasks do not fit this
pattern: ARC-AGI-3 and Scene from Video start low (3.8 and 17.3) and still barely improve (to 4.1 and 18.5).

The gain also depends on the model. Closed-source models improve more often and by more: they raise the validation
score in 94\% of their 85 runs, by a median of 14.9 points, against 76\% of 92 runs and 8.5 points for open-weight
models. Per model (\cref{fig:improve}), the mean gain ranges from 26.7 points for Claude Opus~5.5 and 25.7 for
Gemini~4 Argon to 6.8 for Gemma-4-31B-it, and the four models with the highest scores in \cref{fig:leaderboard}
have four of the five largest gains.

\subsection{Improvement over rounds}
\label{sec:rounds}

\begin{figure}[!tb]
  \centering
  \begin{subfigure}[b]{0.49\linewidth}
    \centering
    \includegraphics[width=\linewidth]{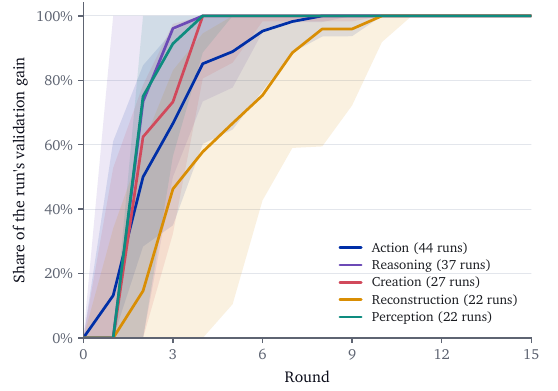}
    \caption{Share of the gain reached by each round.}
    \label{fig:rounds}
  \end{subfigure}\hfill
  \begin{subfigure}[b]{0.49\linewidth}
    \centering
    \includegraphics[width=\linewidth]{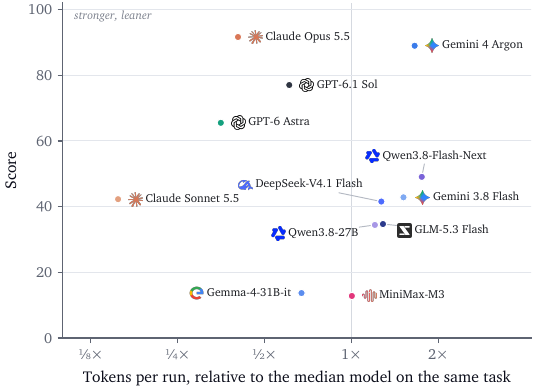}
    \caption{Token use against score.}
    \label{fig:tokens}
  \end{subfigure}
  \caption{\textbf{When the gains arrive and what they cost.} (a)~For each run with a validation gain, the best
  validation score after each round as a share of the run's whole gain; the line is the median over the runs of an
  area and the band the middle half. Runs that stop earlier keep their final value, and runs longer than 15 rounds
  are shown up to round~15. (b)~Tokens per run relative to the median model on the same task (geometric mean over the
  tasks with at least three models with usage records), against the score of \cref{fig:leaderboard}.}
  \label{fig:cost}
\end{figure}

\Cref{fig:rounds} shows in which rounds the gains arrive. For each run that improves, we express its best validation
score after every round as a share of the run's whole gain, and take the median over the runs of each area. Most of
the gain comes early: the median run reaches 90\% of its gain by round~3 in Perception and Reasoning and by round~4
in Creation. Action and Reconstruction take longer, reaching 90\% only by rounds~6 and~8. These are the two areas
with the weakest starters (median validation score 16.7 and 19.4, against 60.0 to 78.8 in the other three), where
the agent likely has to build a working solution before it can refine it.

\subsection{Cost}
\label{sec:cost}

Runs differ widely in what they spend. A run on one task, counting each robot task of RoboDojo and RoboCasa
separately, uses a median of 49.7 million tokens, and the middle 80\% of runs use between 8.6 and 93.4 million. We
compare tokens rather than prices, since the open-weight models are served on our own GPUs. To compare models on the
same footing, \cref{fig:tokens} expresses each model's token use relative to the median model on the same task.
Spending more does not buy a higher score: across the 12 models, the rank correlation between relative token use and
score is $-0.05$. Claude Opus~5.5, GPT-6.1 Sol and GPT-6 Astra, three of the four highest-scoring models, use 0.35 to 0.61 times
the median, whereas Gemini~4 Argon reaches a score close to that of Opus with 1.66 times. Five of the six open-weight
models use at least the median (1.0 to 1.75 times), and all six score below 50.

\providecommand{\qframe}[1]{\includegraphics[width=2.2cm,height=1.65cm]{figures/qual/robo/#1}}
\providecommand{\qhead}{\noindent\hspace*{2.45cm}\parbox{6.66cm}{\centering\scriptsize\textbf{Before}}%
  \hspace{2mm}\parbox{6.66cm}{\centering\scriptsize\textbf{After}}\par\vspace{1mm}}
\providecommand{\qrobo}[8]{\noindent\parbox[t]{2.35cm}{\vspace{0pt}\raggedright\scriptsize\textbf{#1}\\#2\\\textcolor{muted}{#3}}%
  \hspace{1mm}\parbox[t]{6.66cm}{\vspace{0pt}\qframe{#4-before-0}\hspace{0.3mm}\qframe{#4-before-1}\hspace{0.3mm}\qframe{#4-before-2}\\[0.6mm]%
    \scriptsize\textbf{#5.} #6}%
  \hspace{2mm}\parbox[t]{6.66cm}{\vspace{0pt}\qframe{#4-after-0}\hspace{0.3mm}\qframe{#4-after-1}\hspace{0.3mm}\qframe{#4-after-2}\\[0.6mm]%
    \scriptsize\textbf{#7.} #8}\par\vspace{2.2mm}}
\providecommand{\qbox}[3]{\parbox[c][#2][c]{#1}{\centering\includegraphics[width=#1,height=#2,keepaspectratio]{#3}}}
\providecommand{\qedit}[6]{\parbox[t]{8.1cm}{\vspace{0pt}\scriptsize\parbox[t][3\baselineskip][t]{8.1cm}{\textbf{#1.} \textit{#2}}\\[0.4mm]%
  \qbox{2.65cm}{2.2cm}{figures/qual/edit/#3-in}\hfill\qbox{2.65cm}{2.2cm}{figures/qual/edit/#3-before}\hfill\qbox{2.65cm}{2.2cm}{figures/qual/edit/#3-after}\\[0.3mm]%
  \makebox[2.65cm]{Input}\hfill\makebox[2.65cm]{Starter: #4}\hfill\makebox[2.65cm]{Final: #5}\ifx&#6&\else\\[0.3mm]\textcolor{muted}{#6}\fi}}
\providecommand{\qsearch}[4]{\noindent\parbox{\linewidth}{\scriptsize\textbf{#1.} #2}\\[0.5mm]%
  \includegraphics[width=0.497\linewidth]{figures/qual/hstar/#3-before}\hfill\includegraphics[width=0.497\linewidth]{figures/qual/hstar/#3-after}\par\vspace{2mm}}

\subsection{Qualitative examples}
\label{sec:qualitative}

\begin{figure}[!tbp]
  \centering
\qhead
\qrobo{Fill Egg Holder}{RoboDojo}{Claude Opus 5.5}{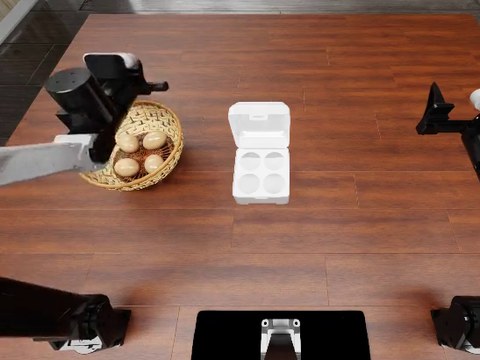}{Round 0 (starter), fail, score 0}{The gripper pokes around in the basket and never lifts an egg.}{Round 3, partial, score 90}{It moves all four eggs into the holder, standing upright; the lid stays open, which costs the last 10 points.}
\qrobo{Pick From Conveyor By Image}{RoboDojo}{GPT-6.1 Sol}{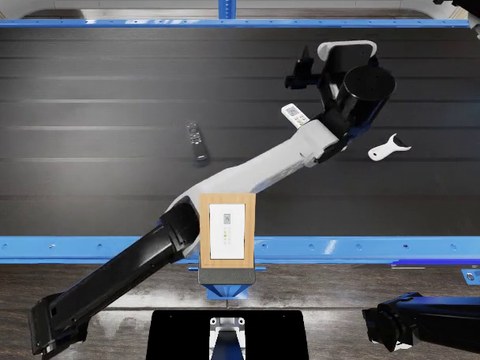}{Round 0 (starter), fail, score 0}{The right arm knocks the basket away, and the left gripper closes beside the objects without lifting any of them.}{Round 6, success}{One arm lifts the basket and holds it over the conveyor while the other picks the object shown on the board and drops it in.}
\qrobo{Play Tic-Tac-Toe}{RoboDojo}{GPT-6 Astra}{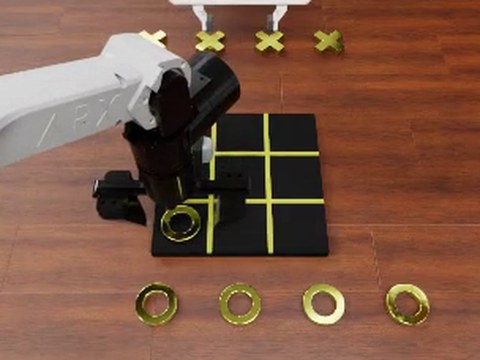}{Round 1, fail, score 10}{It sets one ring on the board, then its policy stops responding and the episode ends, so the opponent never gets a turn.}{Round 8, success}{It takes turns with the opponent arm and places all five of its rings until the board is full.}
\qrobo{Kettle Boiling}{RoboCasa}{Claude Opus 5.5}{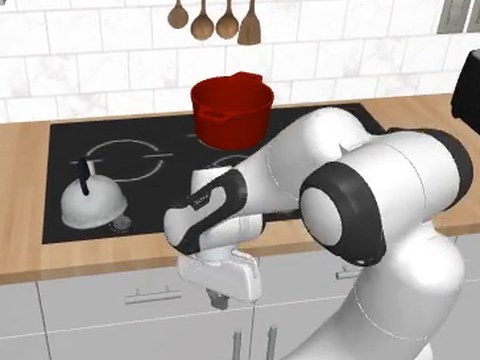}{Round 0 (starter), fail}{The kettle is already on a burner, but the arm wanders around the stove and the burner never comes on.}{Round 2, success}{It sets the kettle on the front burner, then turns that burner's knob; the ring under the kettle glows red.}
\qrobo{Coffee Setup Mug}{RoboCasa}{GPT-6.1 Sol}{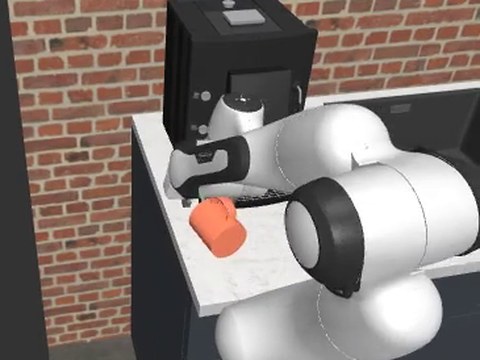}{Round 0 (starter), fail}{The gripper comes down on the mug from the side and knocks it over; it never gets a hold.}{Round 3, success}{It picks the mug up and sets it upright on the drip tray, under the spout.}
\qrobo{Stack the rubiks cubes in a tower}{RoboLab}{Claude Opus 5.5}{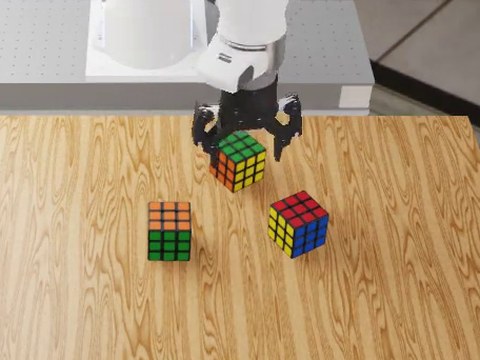}{Round 0 (starter), fail, score 50}{It puts one cube on another, then knocks the pair over going for the third; the cubes end up side by side. It stops when its 40 model calls run out.}{Round 7, success}{It stacks the cubes one by one and sets the third on top of the other two.}
  \caption{\textbf{Robot manipulation.} Each row shows one episode of a robot task played by the solution of an earlier round (Before) and by that of a later round (After) of the same run, three frames each, cropped to the part of the scene where the two differ. The rows give the benchmark, the model of the run, the round of each solution and the outcome of the episode.}
  \label{fig:qual-robot}
\end{figure}

\begin{figure}[!tbp]
  \centering
\qsearch{Object search}{Find the ad `BROADWAY SHOWS'.}{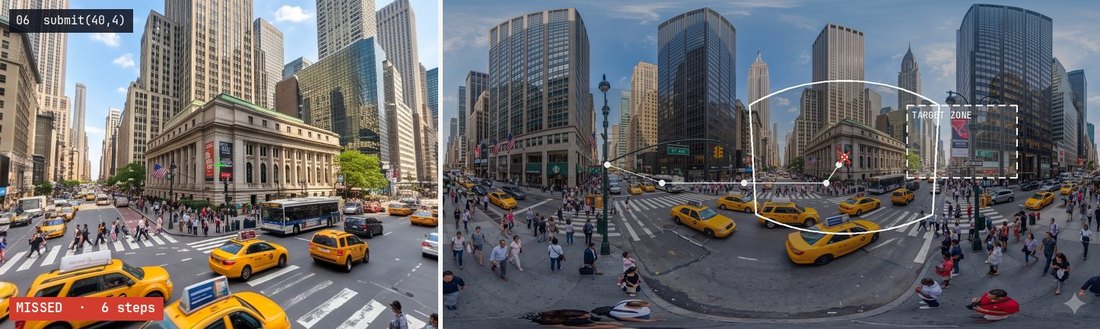}{}
\qsearch{Path search}{If you want to enter the red house, where should you move?}{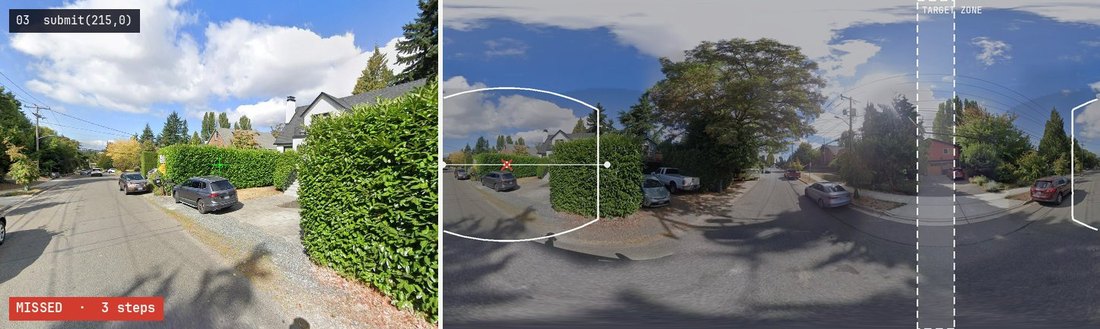}{}
\qsearch{Path search}{You want to travel to Frankfurt, your flight number is `LH496', where'd you go?}{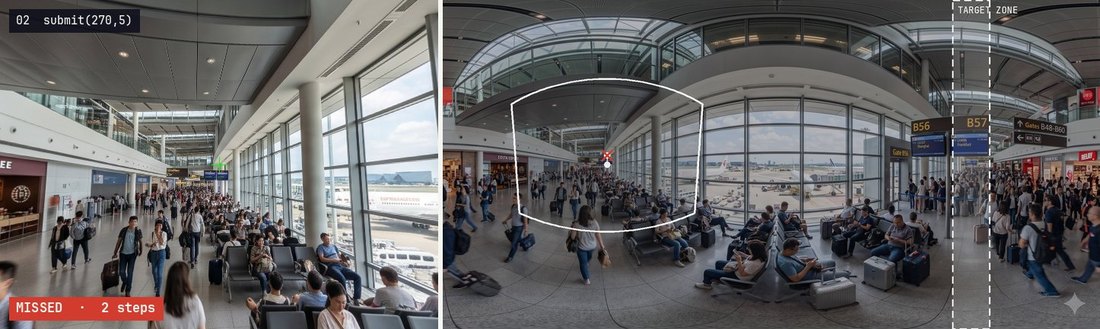}{}
  \caption{\textbf{Active Search.} Episodes of the starter solution (left of each row) and of the final solution of the run of Claude Opus~5.5 (right). Each image shows the last view of the episode (left part) and the panorama with the path of the view and the target zone (right part); the label gives the outcome and the number of actions, and the top line the last action. Object search asks for a named object, path search for the direction in which to move.}
  \label{fig:qual-search}
\end{figure}

\begin{figure}[!tbp]
  \centering
\noindent\parbox{\linewidth}{\scriptsize\textbf{Puzzle 689c358e}}\\[0.5mm]
\noindent\begin{tabular}[b]{@{}c@{}}\includegraphics{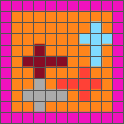}\,\raisebox{0.4\height}{\scriptsize$\to$}\,\includegraphics{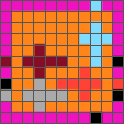}\\[0.5mm]\scriptsize Example 1\end{tabular}\hfill\begin{tabular}[b]{@{}c@{}}\includegraphics{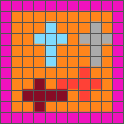}\,\raisebox{0.4\height}{\scriptsize$\to$}\,\includegraphics{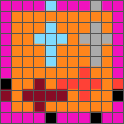}\\[0.5mm]\scriptsize Example 2\end{tabular}\hfill\begin{tabular}[b]{@{}c@{}}\includegraphics{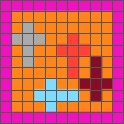}\\[0.5mm]\scriptsize Test input\end{tabular}\hfill\begin{tabular}[b]{@{}c@{}}\includegraphics{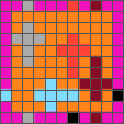}\\[0.5mm]\scriptsize Starter (wrong)\end{tabular}\hfill\begin{tabular}[b]{@{}c@{}}\includegraphics{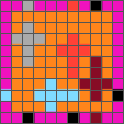}\\[0.5mm]\scriptsize Final (correct)\end{tabular}\par\vspace{3mm}
\noindent\parbox{\linewidth}{\scriptsize\textbf{Puzzle 25c199f5}}\\[0.5mm]
\noindent\begin{tabular}[b]{@{}c@{}}\includegraphics{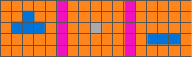}\,\raisebox{0.4\height}{\scriptsize$\to$}\,\includegraphics{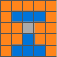}\\[0.5mm]\scriptsize Example 1\end{tabular}\hfill\begin{tabular}[b]{@{}c@{}}\includegraphics{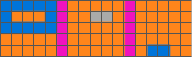}\,\raisebox{0.4\height}{\scriptsize$\to$}\,\includegraphics{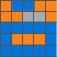}\\[0.5mm]\scriptsize Example 2\end{tabular}\hfill\begin{tabular}[b]{@{}c@{}}\includegraphics{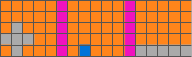}\\[0.5mm]\scriptsize Test input\end{tabular}\hfill\begin{tabular}[b]{@{}c@{}}\includegraphics{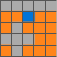}\\[0.5mm]\scriptsize Starter (wrong)\end{tabular}\hfill\begin{tabular}[b]{@{}c@{}}\includegraphics{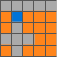}\\[0.5mm]\scriptsize Final (correct)\end{tabular}\par\vspace{3mm}
\noindent\parbox{\linewidth}{\scriptsize\textbf{Puzzle 8b28cd80}}\\[0.5mm]
\noindent\begin{tabular}[b]{@{}c@{}}\includegraphics{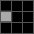}\,\raisebox{0.4\height}{\scriptsize$\to$}\,\includegraphics{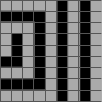}\\[0.5mm]\scriptsize Example 1\end{tabular}\hfill\begin{tabular}[b]{@{}c@{}}\includegraphics{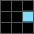}\,\raisebox{0.4\height}{\scriptsize$\to$}\,\includegraphics{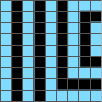}\\[0.5mm]\scriptsize Example 2\end{tabular}\hfill\begin{tabular}[b]{@{}c@{}}\includegraphics{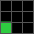}\\[0.5mm]\scriptsize Test input\end{tabular}\hfill\begin{tabular}[b]{@{}c@{}}\parbox[b]{1.6cm}{\centering\scriptsize\textcolor{muted}{no answer}\\[4mm]}\\[0.5mm]\scriptsize Starter\end{tabular}\hfill\begin{tabular}[b]{@{}c@{}}\includegraphics{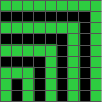}\\[0.5mm]\scriptsize Final (correct, 2nd attempt)\end{tabular}\par\vspace{3mm}
  \caption{\textbf{ARC-AGI-2.} Three validation puzzles: two of the training examples (input $\to$ output), the test input, and the answers of the starter solution and of the final solution of the run of Claude Opus~5.5. A puzzle counts as solved if one of two attempts is correct.}
  \label{fig:qual-arc}
\end{figure}

\begin{figure}[!tbp]
  \centering
\noindent\qedit{Add object, exact count}{Add one more bee-eater of the same species on the branch so there are three in total}{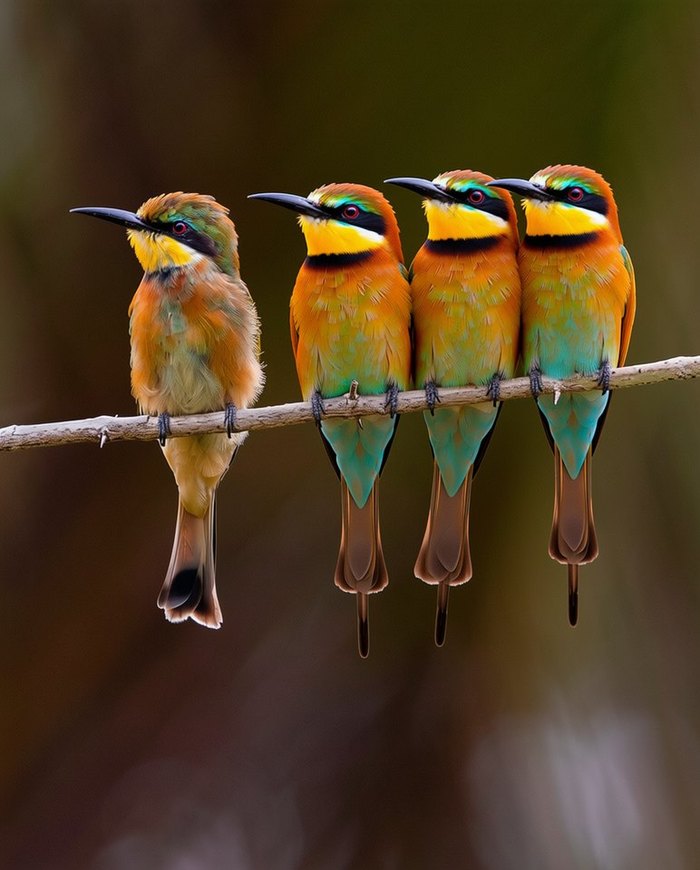}{0}{95}{}\hfill\qedit{Solve, then write}{Write the answer to this expression after the equals sign on the chalkboard}{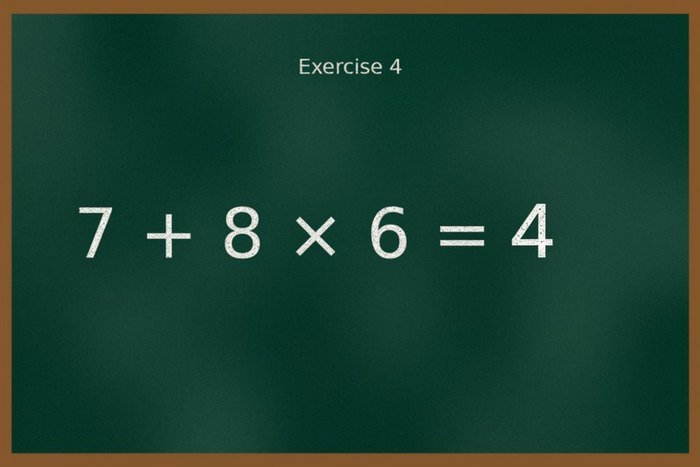}{0}{89}{}\par\vspace{2.5mm}
\noindent\qedit{What happens over time}{Show what these bananas look like a week later}{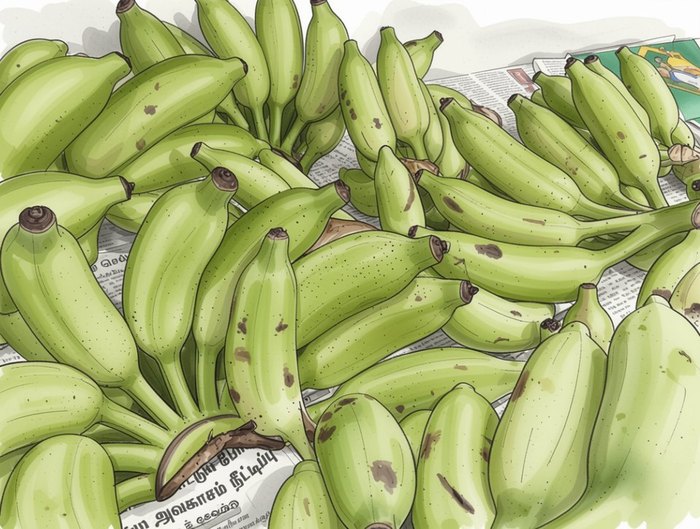}{0}{77}{}\hfill\qedit{Clean up and fill in}{Clean up the litter on the reedy shore and fill it in naturally}{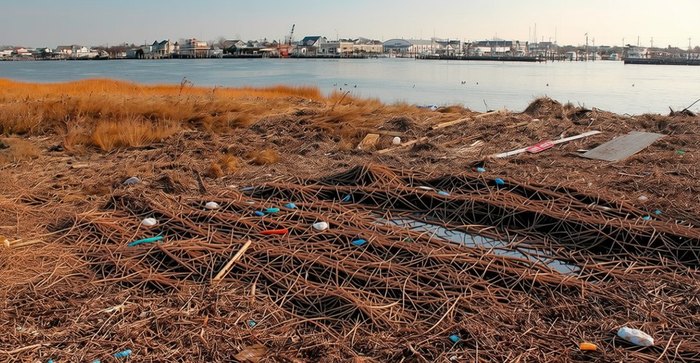}{0}{83}{}\par\vspace{2.5mm}
\noindent\qedit{Replace one thing, keep the rest}{Replace the flowers in the painting with sunflowers; keep the vase, the peach and the grapes unchanged}{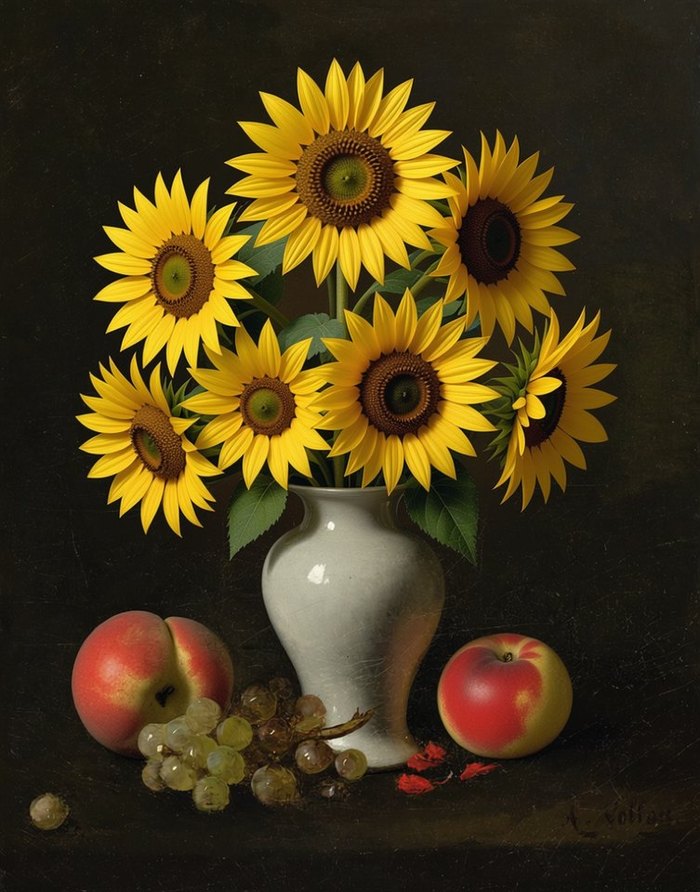}{63}{91}{}\hfill\qedit{Soft-effect removal, haze}{Remove the haze over the city so the sky is clear and the buildings look crisp with natural colors; keep the buildings unchanged}{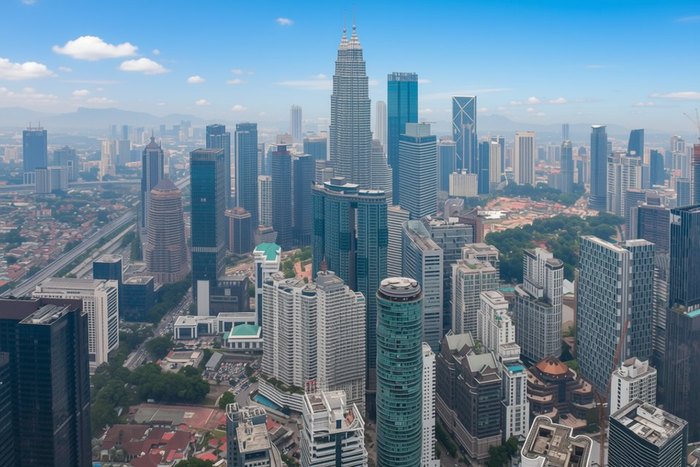}{51}{75}{}\par\vspace{2.5mm}
  \caption{\textbf{Image Editing.} For each edit instruction, the input image and the outputs of the starter solution and of the final solution of the run of Claude Opus~5.5, with the score that the task evaluator assigns to each output (multiplied by 100).}
  \label{fig:qual-edit}
\end{figure}

\begin{figure}[!tbp]
  \centering
\noindent\makebox[3.1cm]{\scriptsize\textbf{Photo}}\hfill\makebox[3.1cm]{\scriptsize\textbf{Starter}}\hfill\makebox[3.1cm]{\scriptsize\textbf{Round 1}}\hfill\makebox[3.1cm]{\scriptsize\textbf{Final (round 4)}}\hfill\makebox[3.1cm]{\scriptsize\textbf{Ground truth}}\par\vspace{0.8mm}
\noindent{\setlength\fboxsep{0pt}\fcolorbox{hairline}{white}{\includegraphics[width=3.1cm,height=2.32cm]{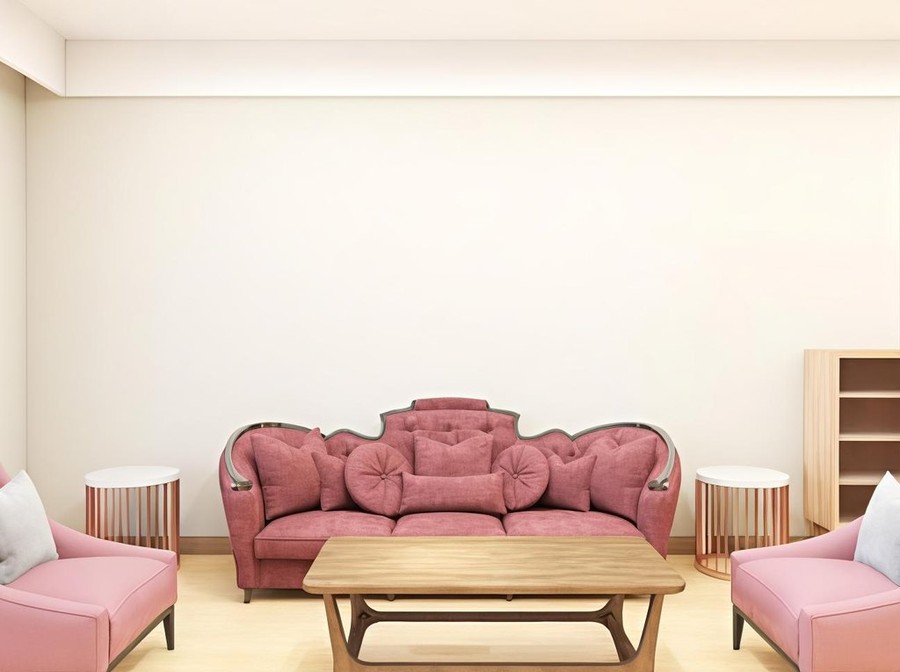}}}\hfill{\setlength\fboxsep{0pt}\fcolorbox{hairline}{white}{\parbox[b][2.32cm][c]{3.1cm}{\centering\scriptsize\textcolor{muted}{empty room,\\no furniture}}}}\hfill{\setlength\fboxsep{0pt}\fcolorbox{hairline}{white}{\includegraphics[width=3.1cm,height=2.32cm]{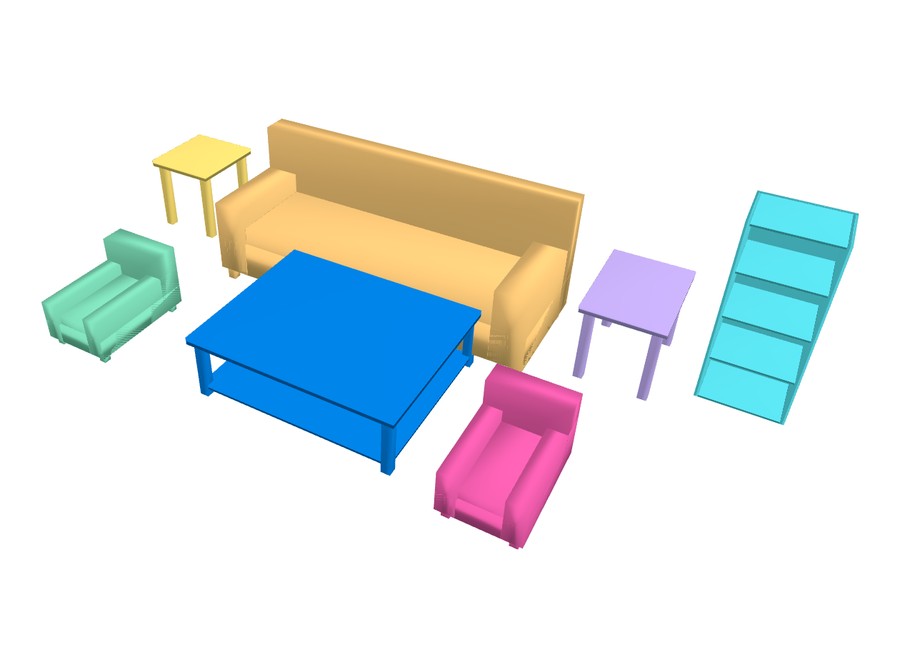}}}\hfill{\setlength\fboxsep{0pt}\fcolorbox{hairline}{white}{\includegraphics[width=3.1cm,height=2.32cm]{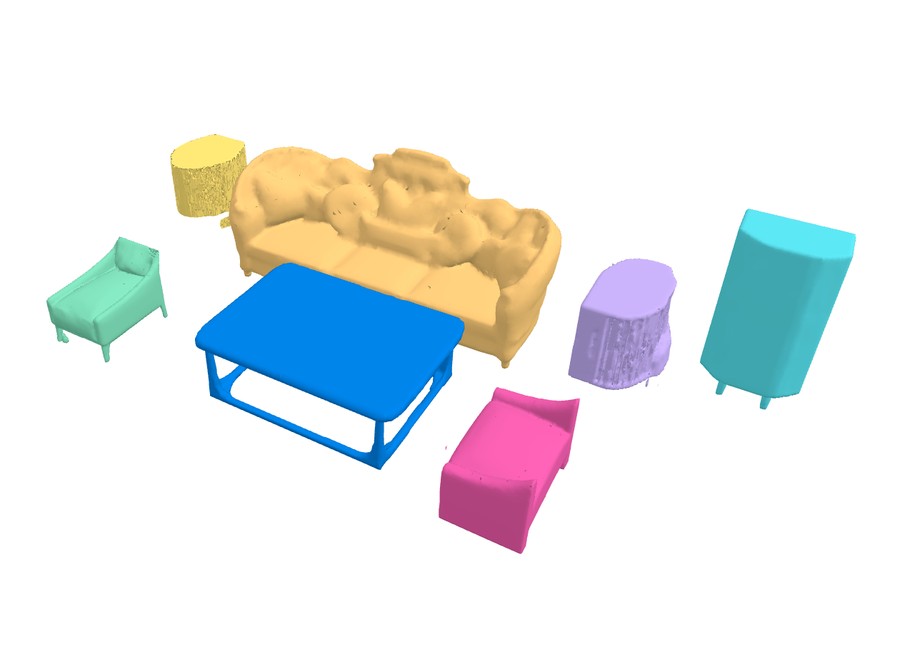}}}\hfill{\setlength\fboxsep{0pt}\fcolorbox{hairline}{white}{\includegraphics[width=3.1cm,height=2.32cm]{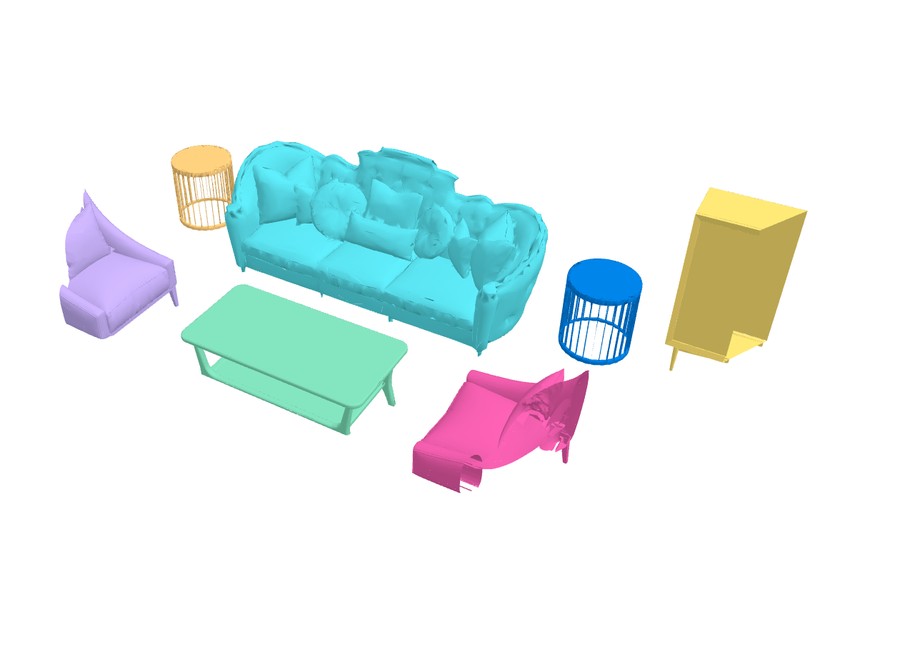}}}\par\vspace{1.2mm}
\noindent{\setlength\fboxsep{0pt}\fcolorbox{hairline}{white}{\includegraphics[width=3.1cm,height=2.32cm]{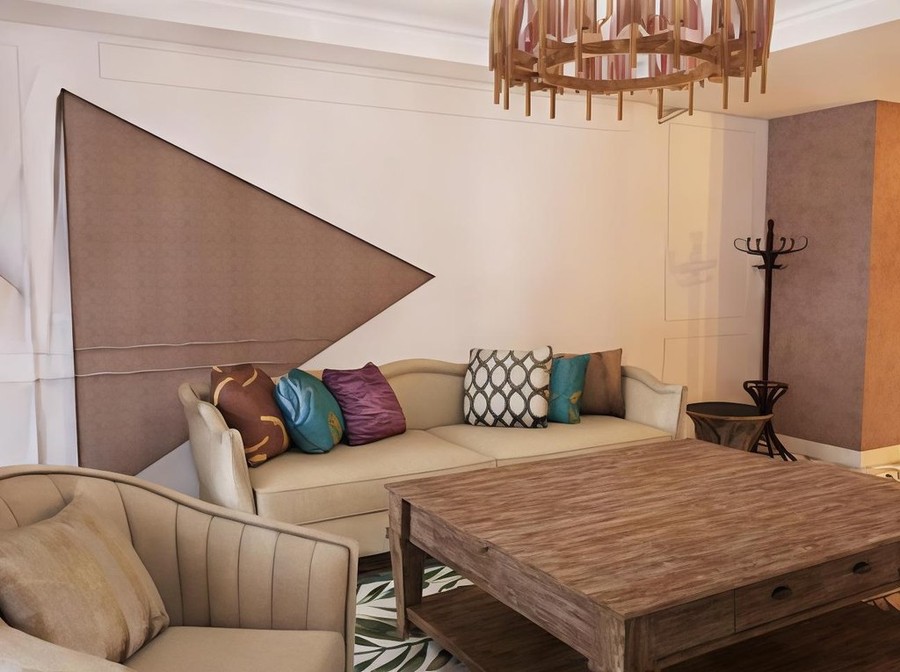}}}\hfill{\setlength\fboxsep{0pt}\fcolorbox{hairline}{white}{\parbox[b][2.32cm][c]{3.1cm}{\centering\scriptsize\textcolor{muted}{empty room,\\no furniture}}}}\hfill{\setlength\fboxsep{0pt}\fcolorbox{hairline}{white}{\includegraphics[width=3.1cm,height=2.32cm]{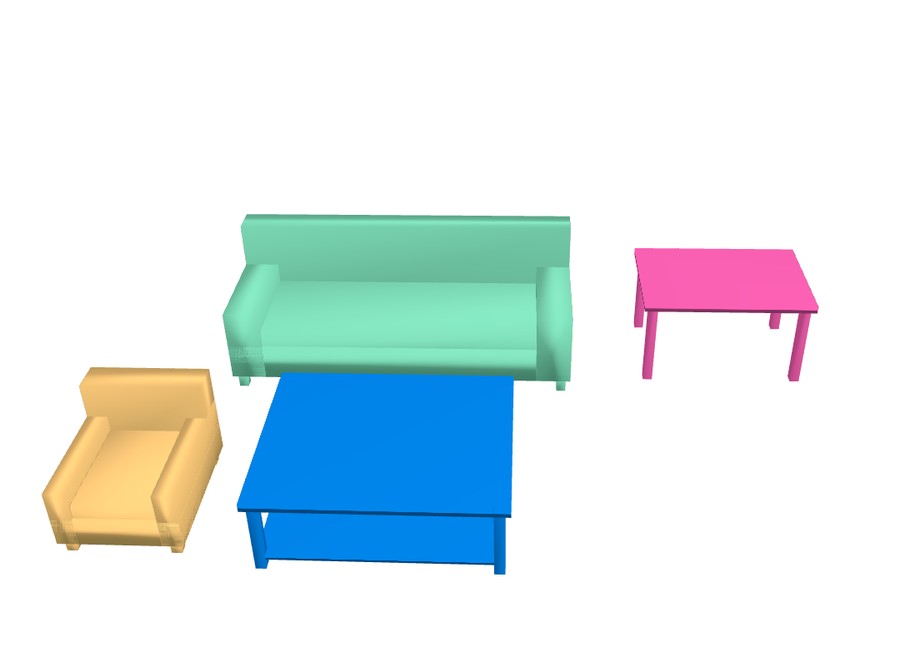}}}\hfill{\setlength\fboxsep{0pt}\fcolorbox{hairline}{white}{\includegraphics[width=3.1cm,height=2.32cm]{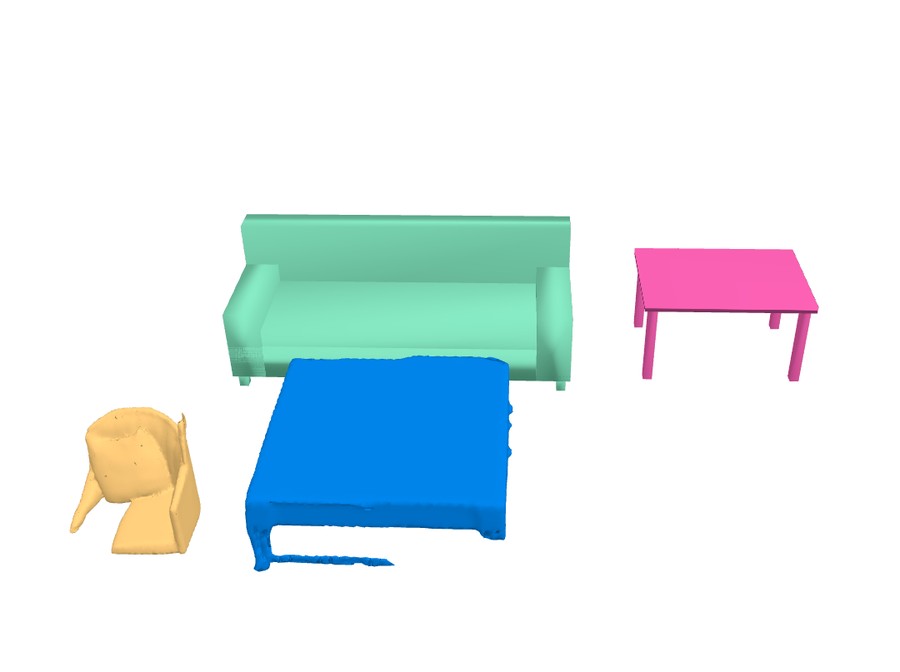}}}\hfill{\setlength\fboxsep{0pt}\fcolorbox{hairline}{white}{\includegraphics[width=3.1cm,height=2.32cm]{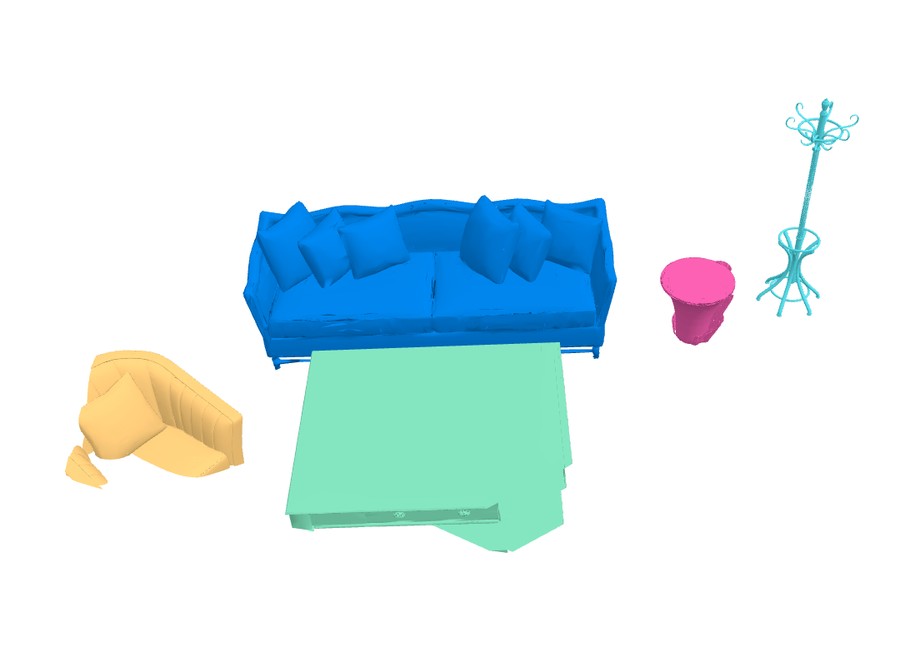}}}\par\vspace{1.2mm}
  \caption{\textbf{Scene from Photo.} Development photos and the furniture reconstructed by the starter solution, by the solution after round~1, and by the final solution of the run of Claude Opus~5.5, next to the annotated furniture, seen from above the photo's camera; each color is one piece, and the room box is not drawn. The starter returns an empty room box; the solution of round~1 places a box template for each piece; the final solution generates the shape of each piece.}
  \label{fig:qual-scene}
\end{figure}

\Cref{fig:qual-robot,fig:qual-search,fig:qual-arc,fig:qual-edit,fig:qual-scene} compare the behavior of solutions before and after self-improvement on robot manipulation, active search, abstract reasoning, image editing, and 3D reconstruction from a photo. Further examples on these and the remaining tasks are given in \cref{app:qualitative}.

\FloatBarrier

\section{Analysis and takeaways}
\label{sec:analysis}

\Cref{tab:takeaways} lists, for every task evaluated with at least six models, the quantities on which the
findings of \cref{sec:intro} rest. The tasks fall into two groups according to what the agent has to produce. On
seven tasks, the solution gives a single answer to each question, and the starter solution obtains it by passing the
question to a model: the model being evaluated, or on Kinematics and Mass a fixed open-weight model. On the other thirteen, the agent has to build the solution itself, namely a program, a robot
policy, or a processing pipeline that produces the output. These thirteen tasks are divided further according to how
the task evaluator scores an episode: on nine it assigns a score on a scale, and on four it records only whether the
episode or level succeeded.
\Cref{sec:open-ended,sec:feedback,sec:strict} examine the three findings in turn, and \cref{sec:harness} examines
whether editing the harness contributes to the improvement.

\begin{table}[!htb]
  \centering\footnotesize
  \caption{\textbf{Evidence for the findings.} Starter solution gives the median validation score of the starter
  solution over the runs of a task, and Largest gain the largest increase of the validation score over the starter
  solution in any run. Best minus worst model gives the difference between the highest and the lowest validation
  score among the models, at the start of the run and at the best validation score reached. Median gain gives the
  median increase of the validation score over the runs of closed-source and of open-weight models, and Best test
  score the highest test score of a closed-source and of an open-weight model. The four tasks evaluated with two models are omitted. All scores are on the 0--100 scale of
  the result tables.}
  \label{tab:takeaways}
  \begin{fittabular}{@{}lrrrrrrrr@{}}
    \toprule
    & & & \multicolumn{2}{c}{\colhead{Best minus worst model}} & \multicolumn{2}{c}{\colhead{Median gain}} & \multicolumn{2}{c}{\colhead{Best test score}} \\
    \cmidrule(lr){4-5}\cmidrule(lr){6-7}\cmidrule(l){8-9}
    \colhead{Task} & \stackhead{Starter}{solution} & \stackhead{Largest}{gain} & \colhead{Start} & \colhead{End} & \stackhead{Closed-}{source} & \stackhead{Open-}{weight} & \stackhead{Closed-}{source} & \stackhead{Open-}{weight} \\
    \midrule
    \multicolumn{9}{@{}l}{\textit{The solution gives a single answer to each question}} \\
    Visual Search & 80.0 & 20.0 & 31.3 & 22.5 & 6.2 & 11.9 & 98.6 & 88.6 \\
    Video Quality & 78.2 & 14.8 & 23.5 & 16.7 & 6.7 & 12.7 & 82.1 & 82.0 \\
    Ego-Exo Memory & 63.7 & 13.8 & 27.5 & 25.0 & 7.5 & 5.6 & 72.5 & 60.8 \\
    Kinematics & 27.7 & 33.5 & 14.2 & 33.1 & 8.8 & 28.5 & 57.2 & 49.2 \\
    Mass & 47.2 & 16.6 & 6.0 & 5.8 & 13.9 & 11.2 & 63.5 & 62.7 \\
    ARC-AGI-1 & 91.9 & 11.2 & 31.2 & 32.5 & 1.2 & 0.0 & 99.8 & 91.0 \\
    ARC-AGI-2 & 82.5 & 3.8 & 38.8 & 40.0 & 1.2 & 0.0 & 95.8 & 46.7 \\
    \midrule
    \multicolumn{9}{@{}l}{\textit{The agent builds the solution; the evaluator scores each episode on a scale}} \\
    RoboDojo & 0.1 & 68.4 & 0.9 & 68.9 & 46.8 & 9.9 & 55.1 & 17.8 \\
    Active Search & 23.0 & 35.4 & 30.8 & 34.4 & 28.4 & 29.3 & 79.7 & 69.0 \\
    RoboLab & 8.3 & 54.2 & 29.2 & 62.5 & 39.6 & 18.7 & 71.9 & 31.6 \\
    Image Editing & 56.9 & 22.3 & 4.6 & 13.9 & 15.1 & 9.3 & 74.2 & 70.0 \\
    Multi-turn Editing & 68.3 & 4.2 & 0.4 & 2.3 & 3.7 & 2.1 & 71.1 & 70.2 \\
    Video Generation & 69.1 & 23.9 & 1.0 & 20.2 & 21.6 & 8.7 & 93.9 & 91.4 \\
    3D Tracking & 19.4 & 65.4 & 0.0 & 55.6 & 42.8 & 39.8 & 79.6 & 61.2 \\
    Scene from Photo & 7.7 & 60.1 & 0.0 & 50.6 & 53.6 & 23.0 & 66.5 & 40.0 \\
    Scene from Video & 17.3 & 38.8 & 0.0 & 38.8 & 24.7 & 0.0 & 70.9 & 34.5 \\
    \midrule
    \multicolumn{9}{@{}l}{\textit{The agent builds the solution; the evaluator records only success or failure}} \\
    RoboCasa & 0.0 & 55.0 & 5.0 & 58.8 & 51.2 & 5.6 & 57.5 & 7.5 \\
    LIBERO-MAX & 31.2 & 18.8 & 0.0 & 18.8 & 12.5 & 3.1 & 45.8 & 45.8 \\
    LIBERO-PRO & 50.0 & 25.0 & 0.0 & 25.0 & 17.5 & 0.0 & 63.3 & 53.3 \\
    ARC-AGI-3 & 3.8 & 18.4 & 39.0 & 49.3 & 16.4 & 0.2 & 44.2 & 0.9 \\
    \bottomrule
  \end{fittabular}
\end{table}

\subsection{Tasks in which the agent builds the solution leave more room for self-improvement}
\label{sec:open-ended}
\begin{finding}When the agent builds the solution, self-improvement raises the score the most and separates the
models; when the solution is a single answer, the starter solution already scores high and the models end about as
far apart as they began. On 3D Tracking, every run starts from 19.4 and the best run gains 65.4 points.\end{finding}

On tasks whose solution is a single answer, the starter solution already passes each question to the model, so it
scores high from the start and its score depends on the model: its median validation score over these seven tasks is 78.2, and the
models are 27.5 points apart before the first round (medians over tasks). The largest gain that any run achieves is
14.8 points at the median, and at the end of the runs the models are 25.0 points apart, about as far as at the
start. On the tasks in which the agent builds the solution, every model starts from the same program, although
its score can depend on the model that it calls. On the nine of them
whose evaluator scores each episode on a scale, this program scores 19.4 at the median, the largest gain is 38.8
points at the median, and the models, 0.9 points apart at the start, end 38.8 points apart; on the four whose
evaluator records only success or failure, they move from 2.5 to 37.2 points apart. The same pattern appears in \cref{fig:overview}c. The
exceptions are Kinematics, whose solution is a single answer but whose starter solution scores only 27.7 and on which
a run gains 33.5 points, and Multi-turn Editing, on which the agent builds the solution but no run gains more than
4.2 points.

\subsection{The granularity of feedback matters}
\label{sec:feedback}
\begin{finding}When the evaluator scores each episode on a scale, open-weight models also improve the solutions they
build; when it records only success or failure, they barely improve, while closed-source models still do. On RoboCasa, the
median open-weight run gains 5.6 points and the median closed-source run 51.2.\end{finding}

On the nine tasks in which the agent builds the solution and the evaluator scores each episode on a scale, the
median gain over the open-weight runs of a task is 9.9
points at the median over tasks, against 28.4 for the closed-source runs, and on 3D Tracking and Active Search it is
close to that of the closed-source runs (39.8 against 42.8, and 29.3 against 28.4). On the four tasks whose evaluator records only whether an episode or a
level succeeded, namely RoboCasa, LIBERO-MAX, LIBERO-PRO, and ARC-AGI-3, the median open-weight gain falls to 1.6
points, while the closed-source runs still gain 16.9. The median open-weight run does not improve at all on
LIBERO-PRO and gains 0.2 points on ARC-AGI-3. Since a candidate is retained only if it raises the validation score
(\cref{eq:accept}), a score on a scale rewards a change that improves part of an episode, whereas a score that
records only success or failure rewards a change only once it turns failed episodes into successful ones.

\subsection{Strict checks of physical stability and exact correctness pose the greatest challenge}
\label{sec:strict}
\begin{finding}Outputs that look correct often fail checks of physical stability or exact correctness, and tasks
scored by such checks are where open-weight models fall furthest behind. On Scene from Video, no object placed by an
open-weight model remains stable in simulation, against 70.7\% for Claude Opus~5.5.\end{finding}

The difference between the best closed-source and the best open-weight model is largest on tasks whose evaluator
applies such a check: whether a robot task succeeds in simulation (RoboCasa, 50.0 points; RoboLab, 40.3; RoboDojo,
37.3), whether the output grid is exactly right (ARC-AGI-2, 49.2) or a level is cleared (ARC-AGI-3, 43.3; on ARC-AGI-1,
whose puzzles the starter solutions already nearly solve, it is 8.8), and
whether a reconstructed scene is physically and geometrically consistent (Scene from Video, 36.4; Scene from Photo,
26.5). It is smallest on tasks scored by a judge or with a tolerance: Video Quality (0.1), Mass (0.8), Multi-turn
Editing (1.0), Video Generation (2.5), and Image Editing (4.2). The component scores of the two scene tasks show how
plausible outputs fail these checks (\cref{tab:res-photo,tab:res-video}). On Scene from Photo, the renderings of the
open-weight scenes resemble the photo, with an Appearance score of up to 81.8, but the furniture is not recovered as
complete and correctly placed objects: their Objects score is at most 16.5, against 62.5 for both closed-source
models. On Scene from Video, the Physics score, the fraction of objects that remain stable in simulation, is 0.0
for all four open-weight models, against 70.7 for Claude Opus~5.5 and 37.7 for GPT-6 Astra. The two LIBERO
tasks are the exceptions among the robot tasks: the best open-weight model matches the best closed-source model on
LIBERO-MAX (45.8) and trails it by 10.0 points on LIBERO-PRO.

\subsection{Does editing the harness help?}
\label{sec:harness}
\begin{finding}Letting the improver edit the harness raises the test score in six of seven pairs of otherwise
identical runs, and most where the unchanged starting harness makes little progress. On Active Search,
Qwen3.8-Flash-Next reaches 21.0 without the improver and 59.3 with it.\end{finding}

\begin{table}[!htb]
  \centering\footnotesize
  \caption{\textbf{Frozen and self-improving harness.} Each row compares two runs of the same model on the same task
  whose settings (task version, budget, protocol, and seed) are identical except for the improver. With the frozen
  harness, the improver does not run and the starting harness is never modified, so only the executor works on the
  solution; with the self-improving harness, the improver edits the executor and itself after every round, as in all
  other runs of this report. Validation is the best validation score reached during the run, Test the test score of
  the final solution, and $\Delta$ the test score with the self-improving harness minus that with the frozen one.
  The robot tasks are single RoboCasa tasks with 5 episodes per split. Claude Opus~5.5 with the frozen harness
  stopped after round~8, when its validation score reached the maximum; a second self-improving run of Claude
  Opus~5.5 on Turn On Sink Faucet with the same settings reached a test score of 40.0. In the RoboCasa runs of
  DeepSeek-V4.1 Flash and Qwen3.8-Flash-Next, the model also reasons while acting as the policy, so the Open Drawer
  run of DeepSeek-V4.1 Flash differs from the one in \cref{tab:res-robocasa}. All scores are multiplied by 100.}
  \label{tab:ablation-harness}
  \setlength{\tabcolsep}{3pt}
  \begin{fittabular}{@{}llrrrrrrr@{}}
    \toprule
    & & \multicolumn{3}{c}{\colhead{Frozen harness}} & \multicolumn{3}{c}{\colhead{Self-improving harness}} & \\
    \cmidrule(lr){3-5}\cmidrule(lr){6-8}
    \colhead{Task} & \colhead{Model} & \colhead{Validation} & \colhead{Test} & \colhead{Rounds} & \colhead{Validation} & \phead{Test} & \colhead{Rounds} & \colhead{$\Delta$ Test} \\
    \midrule
    \multirow{1}{*}{RoboCasa: Open Drawer} & \mlogo{deepseek}DeepSeek-V4.1 Flash & 0.0 & 0.0 & 5 & 40.0 & \pcol{40.0} & 10 & +40.0 \\
    \midrule
    \multirow{2}{*}{\shortstack[l]{RoboCasa: Turn On\\Sink Faucet}} & \mlogo{qwen}Qwen3.8-Flash-Next & 20.0 & 0.0 & 9 & 40.0 & \pcol{20.0} & 10 & +20.0 \\\rowrulefrom{2-9}
     & \mlogo{claude}Claude Opus 5.5 & 100.0 & 80.0 & 8 & 80.0 & \pcol{80.0} & 10 & \phantom{+}0.0 \\
    \midrule
    \multirow{2}{*}{Active Search} & \mlogo{qwen}Qwen3.8-27B & 41.2 & 50.7 & 10 & 51.7 & \pcol{69.0} & 15 & +18.3 \\\rowrulefrom{2-9}
     & \mlogo{qwen}Qwen3.8-Flash-Next & 16.5 & 21.0 & 5 & 50.3 & \pcol{59.3} & 12 & +38.3 \\
    \midrule
    \multirow{2}{*}{3D Tracking} & \mlogo{qwen}Qwen3.8-27B & 48.9 & 45.9 & 11 & 53.9 & \pcol{52.4} & 9 & +6.5 \\\rowrulefrom{2-9}
     & \mlogo{qwen}Qwen3.8-Flash-Next & 63.4 & 56.9 & 9 & 65.7 & \pcol{58.9} & 12 & +2.0 \\
    \bottomrule
  \end{fittabular}
\end{table}

\begin{table}[!htb]
  \centering\footnotesize
  \caption{\textbf{Starting a weaker model from a stronger model's harness.} On two RoboCasa tasks, Active Search, and
  3D Tracking, a weaker model is run from the harness that a stronger model had at the end of its own run on the task (the source run, last row
  of each block). Executor gives the executor the run starts from: the starting harness of every run, or the final
  executor of the source run. Improver gives the improver that runs between rounds: the model's own, which edits the
  executor and itself; none, in which case the executor stays frozen; the final improver of the source run, kept fixed,
  which edits only the executor; or the model's own improver starting from the final improver of the source run, which
  edits the executor and itself. The first row of each block is the self-improving run of
  \cref{tab:ablation-harness}. Validation is the best validation score reached during the run and Test the test
  score of the final solution. Runs from a source run's harness use the code and protocol of the source run; the RoboCasa runs use 5
  episodes per split. All
  scores are multiplied by 100.}
  \label{tab:harness-transfer}
  \setlength{\tabcolsep}{3pt}
  \begin{fittabular}{@{}llllrrr@{}}
    \toprule
    \colhead{Task} & \colhead{Model} & \colhead{Executor} & \colhead{Improver} & \colhead{Validation} & \phead{Test} & \colhead{Rounds} \\
    \midrule
    \multirow{4}{*}{\shortstack[l]{RoboCasa:\\Open Drawer}} & \mlogo{deepseek}DeepSeek-V4.1 Flash & Starting harness & Own & 40.0 & 40.0 & 10 \\\rowrulefrom{2-7}
     & \mlogo{deepseek}DeepSeek-V4.1 Flash & From Gemini 4 Argon & None (executor frozen) & 40.0 & 20.0 & 10 \\\rowrulefrom{2-7}
     & \mlogo{deepseek}DeepSeek-V4.1 Flash & From Gemini 4 Argon & Fixed, from Gemini 4 Argon & 80.0 & 60.0 & 10 \\\rowrulefrom{2-7}
     & \mlogo{gemini}Gemini 4 Argon (source run) & Starting harness & Own & 60.0 & 80.0 & 10 \\
    \midrule
    \multirow{4}{*}{\shortstack[l]{RoboCasa: Turn On\\Sink Faucet}} & \mlogo{qwen}Qwen3.8-Flash-Next & Starting harness & Own & 40.0 & 20.0 & 10 \\\rowrulefrom{2-7}
     & \mlogo{qwen}Qwen3.8-Flash-Next & From Claude Opus 5.5 & None (executor frozen) & 20.0 & 0.0 & 10 \\\rowrulefrom{2-7}
     & \mlogo{qwen}Qwen3.8-Flash-Next & From Claude Opus 5.5 & Fixed, from Claude Opus 5.5 & 0.0 & 0.0 & 10 \\\rowrulefrom{2-7}
     & \mlogo{claude}Claude Opus 5.5 (source run) & Starting harness & Own & 80.0 & 80.0 & 10 \\
    \midrule
    \multirow{3}{*}{Active Search} & \mlogo{qwen}Qwen3.8-27B & Starting harness & Own & 51.7 & 69.0 & 15 \\\rowrulefrom{2-7}
     & \mlogo{qwen}Qwen3.8-27B & From Gemini 4 Argon & Own, from Gemini 4 Argon & 39.5 & 49.2 & 9 \\\rowrulefrom{2-7}
     & \mlogo{gemini}Gemini 4 Argon (source run) & Starting harness & Own & 67.3 & 79.7 & 12 \\
    \midrule
    \multirow{3}{*}{3D Tracking} & \mlogo{qwen}Qwen3.8-27B & Starting harness & Own & 53.9 & 52.4 & 9 \\\rowrulefrom{2-7}
     & \mlogo{qwen}Qwen3.8-27B & From Claude Opus 5.5 & Own, from Claude Opus 5.5 & 59.0 & 53.8 & 15 \\\rowrulefrom{2-7}
     & \mlogo{claude}Claude Opus 5.5 (source run) & Starting harness & Own & 84.7 & 79.6 & 15 \\
    \bottomrule
  \end{fittabular}
\end{table}

\Cref{tab:ablation-harness} compares pairs of runs that differ only in whether the improver runs. In the run
without the improver, labeled frozen harness in the table, the starting harness is never modified; in the run with
it, labeled self-improving harness, the improver edits the executor and itself after every round, as in all other
runs of this report. The test score with the improver is higher in six of the seven pairs, by 2.0 to 40.0 points; in
the seventh, Claude Opus~5.5 on Turn On Sink Faucet reaches 80.0 in both runs, and its run without the improver
reached the maximum validation score in round~8. The difference is largest where the unchanged starting harness makes
little progress: DeepSeek-V4.1 Flash remains at 0.0 on Open Drawer without the improver and reaches 40.0 with it,
and Qwen3.8-Flash-Next gains 20.0 on Turn On Sink Faucet and 38.3 on Active Search. On 3D Tracking, the executor
alone already raises the two Qwen models from the 19.4 of the starter solution to validation scores of 48.9 and
63.4, and the improver adds 6.5 and 2.0 points on the test split. The runs without the improver also end after fewer
rounds in six of the seven pairs.

\Cref{tab:harness-transfer} asks whether a harness built by a stronger model helps a weaker one. On Open Drawer,
DeepSeek-V4.1 Flash started from the final executor of Gemini~4 Argon reaches 20.0 when the executor is not modified
further and 60.0 when Argon's final improver continues to edit it, against 40.0 when it improves its own harness from
the start and 80.0 for Argon. On Turn On Sink Faucet, Qwen3.8-Flash-Next started from the harness of Claude Opus~5.5 reaches 0.0 in both
settings, below the 20.0 it reaches when it improves its own harness and the 80.0 of Opus. On Active Search and
3D Tracking, Qwen3.8-27B started from the final harness of the source run and improved it further with its own
improver. On Active Search, it reaches 49.2 from the harness of Gemini~4 Argon, against 69.0 when it improves its own
harness from the start and 79.7 for Argon. On 3D Tracking, it reaches 53.8 from the harness of Claude Opus~5.5,
against 52.4 from the starting harness and 79.6 for Opus. A harness built by a stronger model thus raised the score
of the weaker model on two of the four tasks, by 20.0 points on Open Drawer and 1.4 points on 3D Tracking, and in
neither case did the weaker model approach the score of the stronger one.

\section{Conclusion}
\label{sec:conclusion}

We introduced \name{}, a benchmark that measures whether a frontier agent can improve both its task solution and its
own harness within a fixed budget, with every run starting from the same general-purpose harness and every final
solution scored on a hidden test split. Across 24 tasks and 18 models, three findings emerge. Self-improvement
raises the score most on tasks in which the agent builds its own solution, and on these tasks models that start from
the same program end far apart, whereas on tasks whose solution is a single answer the starter solution already
scores high and the models end about as far apart as they began. Open-weight models improve when the evaluator scores each episode on a scale, but
barely improve when it records only success or failure. The largest gaps between closed-source and open-weight
models arise where the evaluator checks physical stability or exact correctness, even when the outputs look
plausible. Paired runs further show that editing the harness contributes to the improvement, most where
the unchanged starting harness makes little progress, whereas a harness built by a stronger model raised the score
of a weaker model on only two of four tasks.

The improvement measured here is mostly at the level of the harness. The evaluated model's weights never change, and on
most tasks no model is trained; on some, the solution may train or fine-tune helper models on the development split.
\name{} therefore does not measure a model that modifies its own parameters. Extending the benchmark to weight-level
self-improvement, and studying how it interacts with changes to the harness, is left to future work. The leaderboard is
maintained on the project page, \url{https://flex-rsi.github.io}.

\section{Contributors}
\label{sec:contributors}

\name{} is a community effort. Organizers lead the project, contributors bring its tasks, code and experiments, and
advisors guide its direction. The organizers are listed first;
the other contributors and the advisors follow in alphabetical order of surname, and the contributions of each
person are given below.

\newcommand{\contrib}[2]{\noindent\hspace*{0.1\linewidth}\makebox[0.3\linewidth][l]{#1}#2\par}
\newcommand{\contribgroup}[1]{\par\addvspace{6pt}\noindent\hspace*{0.1\linewidth}\colhead{#1}\par\nopagebreak\vspace{1pt}}

{\small\setlength{\parskip}{0pt}
  \contribgroup{Organizers}
  \contrib{Jingdong Zhang\textsuperscript{*}}{Texas A\&M University}
  \contrib{Jionghao Wang\textsuperscript{*}}{Texas A\&M University}
  \contrib{Zhengming Yu}{Texas A\&M University}
  \contribgroup{Contributors (alphabetical order)}
  \contrib{Yancheng Cai}{University of Cambridge}
  \contrib{Wei Chow}{National University of Singapore}
  \contrib{Chris Choy}{NVIDIA}
  \contrib{Zhiyang Dou}{Massachusetts Institute of Technology}
  \contrib{Hang Hua}{MIT-IBM Watson AI Lab}
  \contrib{Jiachen Li}{The University of Texas at Austin}
  \contrib{Zhixuan Liu}{Carnegie Mellon University}
  \contrib{Chen Wang}{University of Pennsylvania}
  \contrib{Songlin Wei}{University of Southern California}
  \contrib{Siyuan Yang}{Texas A\&M University}
  \contrib{Yu Yuan}{Purdue University}
  \contrib{Xiaohang Zhan}{Adobe}
  \contrib{Yucheng Zhang}{Purdue University}
  \contrib{Shuhong Zheng}{University of Toronto}
  \contribgroup{Advisors (alphabetical order)}
  \contrib{Xin Li}{Texas A\&M University}
  \contrib{Lingjie Liu}{University of Pennsylvania}
  \contrib{Wojciech Matusik}{Massachusetts Institute of Technology}
  \contrib{Zhengzhong Tu}{Texas A\&M University}
  \contrib{Wenping Wang}{Texas A\&M University}
\par}\medskip
{\small\textsuperscript{*}Organizers, equal contribution.}

\paragraph{Contributions}
\begin{itemize}[leftmargin=*,itemsep=1pt]
\item \textbf{Project leads:} Jingdong Zhang, Jionghao Wang, Zhengming Yu (benchmark protocol, organization, project
  page, and paper).
\item \textbf{Action tasks:} Jionghao Wang (RoboDojo, RoboCasa), Zhiyang Dou (RoboDojo, RoboCasa), Yucheng Zhang (RoboDojo,
  RoboCasa), Songlin Wei (RoboDojo, RoboCasa), Jingdong Zhang (Active Search, LIBERO-MAX, LIBERO-PRO), Xiaohang Zhan
  (Active Search), Chris Choy (RoboLab), Jiachen Li (LIBERO-MAX, LIBERO-PRO), Hang Hua (Embodied-SWE).
\item \textbf{Reasoning tasks:} Jingdong Zhang (Ego-Exo Memory), Yu Yuan (Kinematics, Mass), Zhengming Yu (ARC-AGI-1,
  ARC-AGI-2, ARC-AGI-3), Hang Hua (Spatial-VSI).
\item \textbf{Creation tasks:} Xiaohang Zhan (Image Editing), Zhixuan Liu (Multi-turn Editing), Shuhong Zheng (Video
  Generation), Jingdong Zhang (Video Generation), Wei Chow (Text-to-Image).
\item \textbf{Reconstruction tasks:} Jingdong Zhang (3D Tracking), Xiaohang Zhan (3D Tracking), Chen Wang (Scene from
  Photo, Scene from Video).
\item \textbf{Perception tasks:} Jingdong Zhang (Visual Search, Video Quality), Xiaohang Zhan (Visual Search, Video
  Quality), Yancheng Cai (Video Quality).
\item \textbf{Infrastructure task:} Siyuan Yang (VAE Decode).
\item \textbf{Harness and codebase:} Jingdong Zhang, Jionghao Wang, Zhengming Yu, Chris Choy, Jiachen Li, Shuhong Zheng,
  Wei Chow, Zhixuan Liu, Zhiyang Dou, Hang Hua, Xiaohang Zhan, Chen Wang.
\item \textbf{Experiments and ablations:} Jingdong Zhang, Jionghao Wang, Zhengming Yu, Zhiyang Dou, Yucheng Zhang.
\end{itemize}

\bibliographystyle{plainnat}
\bibliography{references}

\begin{thebibliography}{188}
\providecommand{\natexlab}[1]{#1}
\providecommand{\url}[1]{\texttt{#1}}
\expandafter\ifx\csname urlstyle\endcsname\relax
  \providecommand{\doi}[1]{doi: #1}\else
  \providecommand{\doi}{doi: \begingroup \urlstyle{rm}\Url}\fi

\bibitem[Agrawal et~al.(2026)Agrawal, Tan, Soylu, Ziems, Khare, Opsahl-Ong,
  Singhvi, Shandilya, Ryan, Jiang, et~al.]{agrawal2026gepa}
Lakshya~A Agrawal, Shangyin Tan, Dilara Soylu, Noah Ziems, Rishi Khare, Krista
  Opsahl-Ong, Arnav Singhvi, Herumb Shandilya, Michael~J Ryan, Meng Jiang,
  et~al.
\newblock {GEPA}: Reflective prompt evolution can outperform reinforcement
  learning.
\newblock In \emph{International Conference on Learning Representations},
  volume 2026, pages 8479--8565, 2026.

\bibitem[Aky{\"u}rek et~al.(2024)Aky{\"u}rek, Damani, Zweiger, Qiu, Guo, Pari,
  Kim, and Andreas]{akyurek2024surprising}
Ekin Aky{\"u}rek, Mehul Damani, Adam Zweiger, Linlu Qiu, Han Guo, Jyothish
  Pari, Yoon Kim, and Jacob Andreas.
\newblock The surprising effectiveness of test-time training for few-shot
  learning.
\newblock \emph{arXiv preprint arXiv:2411.07279}, 2024.

\bibitem[Ancuti et~al.(2020)Ancuti, Ancuti, and Timofte]{ancuti2020nh}
Codruta~O. Ancuti, Cosmin Ancuti, and Radu Timofte.
\newblock {NH-HAZE}: An image dehazing benchmark with non-homogeneous hazy and
  haze-free images.
\newblock In \emph{IEEE/CVF Conference on Computer Vision and Pattern
  Recognition Workshops (CVPRW)}, pages 444--445, 2020.

\bibitem[Anderson et~al.(2018)Anderson, Chang, Chaplot, Dosovitskiy, Gupta,
  Koltun, Kosecka, Malik, Mottaghi, Savva, and Zamir]{anderson2018on}
Peter Anderson, Angel Chang, Devendra~Singh Chaplot, Alexey Dosovitskiy,
  Saurabh Gupta, Vladlen Koltun, Jana Kosecka, Jitendra Malik, Roozbeh
  Mottaghi, Manolis Savva, and Amir~R. Zamir.
\newblock On evaluation of embodied navigation agents.
\newblock \emph{arXiv preprint arXiv:1807.06757}, 2018.

\bibitem[{Anthropic}(2026{\natexlab{a}})]{anthropic2026opus55}
{Anthropic}.
\newblock System card: {Claude Opus 5.5}.
\newblock
  \url{https://www-cdn.anthropic.com/fc1b44717c85dc068bc6ba5024219938094694bd/Claude\%20Opus\%205.5\%20System\%20Card.pdf},
  September 2026{\natexlab{a}}.

\bibitem[{Anthropic}(2026{\natexlab{b}})]{anthropic2026sonnet55}
{Anthropic}.
\newblock System card: {Claude Sonnet 5.5}.
\newblock \url{https://www.anthropic.com/claude-sonnet-5-5-system-card},
  September 2026{\natexlab{b}}.

\bibitem[{Anthropic}(2026{\natexlab{c}})]{anthropic_claudecode}
{Anthropic}.
\newblock {Claude Code}: Overview.
\newblock Official documentation, 2026{\natexlab{c}}.
\newblock URL \url{https://code.claude.com/docs/en/overview}.
\newblock Accessed October 29, 2026.

\bibitem[{ARC Prize Foundation}(2026)]{foundation2026arc}
{ARC Prize Foundation}.
\newblock {ARC-AGI-3}: A new challenge for frontier agentic intelligence.
\newblock \emph{arXiv preprint arXiv:2603.24621}, 2026.

\bibitem[Ardelean et~al.(2025)Ardelean, {\"O}zer, and
  Egger]{ardelean2025gen3dsr}
Andreea Ardelean, Mert {\"O}zer, and Bernhard Egger.
\newblock {Gen3DSR}: Generalizable {3D} scene reconstruction via divide and
  conquer from a single view.
\newblock In \emph{International Conference on 3D Vision (3DV)}, pages
  616--626, 2025.

\bibitem[Bai et~al.(2025{\natexlab{a}})Bai, Cai, et~al.]{bai2025qwen3vl}
Shuai Bai, Yuxuan Cai, et~al.
\newblock {Qwen3-VL} technical report.
\newblock \emph{arXiv preprint arXiv:2511.21631}, 2025{\natexlab{a}}.

\bibitem[Bai et~al.(2025{\natexlab{b}})]{bai2025qwen}
Shuai Bai et~al.
\newblock {Qwen2.5-VL} technical report.
\newblock \emph{arXiv preprint arXiv:2502.13923}, 2025{\natexlab{b}}.

\bibitem[Bao et~al.(2026)Bao, Huang, Lin, Feng, Li, Zhao, Li, He, Chi, and
  Zhang]{bao2026roboharn}
Shifeng Bao, Fanding Huang, Yihan Lin, Youhe Feng, Guanlin Li, Chen Zhao, Yang
  Li, Jiawei He, Cheng Chi, and Jing Zhang.
\newblock {RoboHarn-Evo}: Evolving hierarchical physical knowledge for
  self-improving robotic manipulation.
\newblock \emph{arXiv preprint arXiv:2609.37583}, 2026.

\bibitem[Baruch et~al.(2021)Baruch, Chen, Dehghan, Dimry, Feigin, Fu, Gebauer,
  Joffe, Kurz, Schwartz, and Shulman]{baruch2021arkitscenes}
Gilad Baruch, Zhuoyuan Chen, Afshin Dehghan, Tal Dimry, Yuri Feigin, Peter Fu,
  Thomas Gebauer, Brandon Joffe, Daniel Kurz, Arik Schwartz, and Elad Shulman.
\newblock {ARKitScenes}: A diverse real-world dataset for {3D} indoor scene
  understanding using mobile {RGB-D} data.
\newblock \emph{arXiv preprint arXiv:2111.08897}, 2021.

\bibitem[Bobadilla-Suarez et~al.(2026)Bobadilla-Suarez, Suh, and
  Fortin]{bobadillasuarez2026scaffold}
Sebastian Bobadilla-Suarez, Bob Suh, and Ryan Fortin.
\newblock Audit the scaffold, not the checkpoint: A stationarity dichotomy for
  recursive self-improvement in agentic coding.
\newblock \emph{arXiv preprint arXiv:2609.34924}, 2026.

\bibitem[Brown et~al.(2025)Brown, Yang, Yang, Fergus, and
  Xie]{brown2025benchmark}
Ellis Brown, Jihan Yang, Shusheng Yang, Rob Fergus, and Saining Xie.
\newblock Benchmark designers should ``train on the test set'' to expose
  exploitable non-visual shortcuts.
\newblock \emph{arXiv preprint arXiv:2511.04655}, 2025.

\bibitem[Cao et~al.(2026)Cao, Zhu, Wang, Li, Li, Liu, Liu, Wang, Zhang, Lin,
  Yu, Liu, et~al.]{cao2026by}
Zhiyong Cao, Hengjie Zhu, Yongxin Wang, Xuchen Li, Jintao Li, Shuxia Liu,
  Dunqiang Liu, Bo~Wang, Kairan Zhang, Cong Lin, JiangDi Yu, Siyuan Liu, et~al.
\newblock Trial by trace: Benchmarking {AI} agent harnesses for reliability and
  failure attribution.
\newblock \emph{arXiv preprint arXiv:2610.32285}, 2026.

\bibitem[Chan et~al.(2025)Chan, Chowdhury, Jaffe, et~al.]{chan2024mlebench}
Jun~Shern Chan, Neil Chowdhury, Oliver Jaffe, et~al.
\newblock {MLE-bench}: Evaluating machine learning agents on machine learning
  engineering.
\newblock In \emph{International Conference on Learning Representations}, 2025.

\bibitem[Chen et~al.(2026{\natexlab{a}})Chen, Wang, and Qu]{chen2026self}
Mingguang Chen, Licheng Wang, and Bo~Qu.
\newblock Recursive self-improvement in {AI}: From bounded self-refinement to
  autonomous research loops.
\newblock \emph{arXiv preprint arXiv:2607.07663}, 2026{\natexlab{a}}.

\bibitem[Chen et~al.(2026{\natexlab{b}})Chen, Jaff, Nashold, and
  Krishnan]{valsai2026rsiindex}
Oliver Chen, Geby Jaff, Langston Nashold, and Rayan Krishnan.
\newblock {Vals RSI Index}.
\newblock Vals AI, August 2026{\natexlab{b}}.
\newblock URL \url{https://vals.ai/benchmarks/rsi_index}.

\bibitem[Chen et~al.(2026{\natexlab{c}})Chen, Hadfield, Zook, Uy, Song,
  Coumans, Yang, Ladhak, Qu, Birchfield, Tremblay, and Blukis]{chen2026volo}
Siyi Chen, Hugo Hadfield, Alex Zook, Mikaela~Angelina Uy, Chan~Hee Song, Erwin
  Coumans, Xuning Yang, Faisal Ladhak, Qing Qu, Stan Birchfield, Jonathan
  Tremblay, and Valts Blukis.
\newblock {VoLo}: A physical orchestrator for open-vocabulary long-horizon
  manipulation.
\newblock \emph{arXiv preprint arXiv:2606.07723}, 2026{\natexlab{c}}.

\bibitem[Chen et~al.(2026{\natexlab{d}})Chen, Chen, Li, Tang, Su, Lu, Wan,
  Chen, Liu, Yan, et~al.]{chen2026robodojo}
Tianxing Chen, Yue Chen, Zixuan Li, Junyuan Tang, Kailun Su, Haoran Lu, Weijie
  Wan, Baijun Chen, Songling Liu, Haowen Yan, et~al.
\newblock {RoboDojo}: A unified sim-and-real benchmark for comprehensive
  evaluation of generalist robot manipulation policies.
\newblock \emph{arXiv preprint arXiv:2607.04434}, 2026{\natexlab{d}}.

\bibitem[Chen et~al.(2026{\natexlab{e}})Chen, Huang, Xu, Zhao, McQuinn, Liu,
  and Yao]{chen2026selfb}
Xiaoyin Chen, Yizheng Huang, Yinglun Xu, Ziyun Zhao, Emmett McQuinn, Julie Liu,
  and Hope Yao.
\newblock Are self-evolving agents continual learners?
\newblock \emph{arXiv preprint arXiv:2610.25858}, 2026{\natexlab{e}}.

\bibitem[Chen et~al.(2026{\natexlab{f}})Chen, Liu, Wang, Jin, Zhu, Ruan, and
  Han]{chen2026compose}
Yangzhi Chen, Feiran Liu, Zishuo Wang, Zijian Jin, Xin Zhu, Lecheng Ruan, and
  Wenjuan Han.
\newblock Learning to compose: Recursive self-improvement of shared executable
  stacks for zero-shot long-horizon robot manipulation.
\newblock \emph{arXiv preprint arXiv:2610.24464}, 2026{\natexlab{f}}.

\bibitem[Chen et~al.(2026{\natexlab{g}})Chen, Zhu, and Zheng]{chen2026product}
Zihan Chen, Di~Zhu, and Lei~Nico Zheng.
\newblock Auditing the product, not the recursion: The evaluation evidence
  behind recursive self-improvement claims.
\newblock \emph{arXiv preprint arXiv:2610.27610}, 2026{\natexlab{g}}.

\bibitem[Chen et~al.(2025)Chen, Chen, Ning, Zhang, Wang, Yu, Li, Liao, Wei, Lu,
  et~al.]{chen2025scienceagentbench}
Ziru Chen, Shijie Chen, Yuting Ning, Qianheng Zhang, Boshi Wang, Botao Yu,
  Yifei Li, Zeyi Liao, Chen Wei, Zitong Lu, et~al.
\newblock {ScienceAgentBench}: Toward rigorous assessment of language agents
  for data-driven scientific discovery.
\newblock In \emph{International Conference on Learning Representations},
  volume 2025, pages 96934--96990, 2025.

\bibitem[Cheng et~al.(2022)Cheng, Misra, Schwing, Kirillov, and
  Girdhar]{cheng2022masked}
Bowen Cheng, Ishan Misra, Alexander~G. Schwing, Alexander Kirillov, and Rohit
  Girdhar.
\newblock Masked-attention mask transformer for universal image segmentation.
\newblock In \emph{Proceedings of the IEEE/CVF Conference on Computer Vision
  and Pattern Recognition (CVPR)}, pages 1290--1299, 2022.

\bibitem[Chi et~al.(2026)Chi, Li, Hong, Wang, Gao, Yang, He, Zheng, Xiao, and
  Na]{chi2026ai4ai}
Yizhe Chi, Wenyi Li, Deyao Hong, Xiaoqiu Wang, Mingju Gao, Kaisen Yang,
  Bingxiang He, Youjie Zheng, Calvin Xiao, and Qinhuai Na.
\newblock {AI4AI-Bench}: Benchmarking {LLM} agents in algorithmic design for
  recursive self-improvement.
\newblock \emph{arXiv preprint arXiv:2608.20318}, 2026.

\bibitem[Chollet(2019)]{chollet2019measure}
Fran{\c{c}}ois Chollet.
\newblock On the measure of intelligence.
\newblock \emph{arXiv preprint arXiv:1911.01547}, 2019.

\bibitem[Chollet et~al.(2024)Chollet, Knoop, Kamradt, and
  Landers]{chollet2024arc}
Francois Chollet, Mike Knoop, Gregory Kamradt, and Bryan Landers.
\newblock {ARC} prize 2024: Technical report.
\newblock \emph{arXiv preprint arXiv:2412.04604}, 2024.

\bibitem[Chollet et~al.(2025)Chollet, Knoop, Kamradt, Landers, and
  Pinkard]{chollet2025arc}
Francois Chollet, Mike Knoop, Gregory Kamradt, Bryan Landers, and Henry
  Pinkard.
\newblock {ARC-AGI-2}: A new challenge for frontier {AI} reasoning systems.
\newblock \emph{arXiv preprint arXiv:2505.11831}, 2025.

\bibitem[Collins et~al.(2022)Collins, Goel, Deng, Luthra, Xu, Gundogdu, Zhang,
  Vicente, Dideriksen, Arora, et~al.]{collins2022abo}
Jasmine Collins, Shubham Goel, Kenan Deng, Achleshwar Luthra, Leon Xu, Erhan
  Gundogdu, Xi~Zhang, Tomas F~Yago Vicente, Thomas Dideriksen, Himanshu Arora,
  et~al.
\newblock {ABO}: Dataset and benchmarks for real-world {3D} object
  understanding.
\newblock In \emph{2022 IEEE/CVF Conference on Computer Vision and Pattern
  Recognition (CVPR)}, pages 21094--21104. IEEE, 2022.

\bibitem[Dai et~al.(2017)Dai, Chang, Savva, Halber, Funkhouser, and
  Nie{\ss}ner]{dai2017scannet}
Angela Dai, Angel~X Chang, Manolis Savva, Maciej Halber, Thomas Funkhouser, and
  Matthias Nie{\ss}ner.
\newblock {ScanNet}: Richly-annotated {3D} reconstructions of indoor scenes.
\newblock In \emph{2017 IEEE conference on computer vision and pattern
  recognition (CVPR)}, pages 2432--2443. IEEE, 2017.

\bibitem[{DeepSeek-AI}(2026)]{deepseek2026v4}
{DeepSeek-AI}.
\newblock {DeepSeek-V4}: Towards highly efficient million-token context
  intelligence.
\newblock \emph{arXiv preprint arXiv:2606.19348}, 2026.

\bibitem[Dou et~al.(2026)Dou, Meindl, Watanabe, Deng, Huang, Sadalski, Liang,
  Guo, Jones, and Matusik]{dou2026frontier3drobotics}
Zhiyang Dou, Jamison Meindl, Akihisa Watanabe, Anna Deng, Tianyu Huang, Igor
  Sadalski, Harrison Liang, Minghao Guo, Benjamin~Tod Jones, and Wojciech
  Matusik.
\newblock On the opportunities and risks of frontier models for {3D} modeling,
  computational design and robotics.
\newblock Computational Design and Fabrication Group (CDFG), MIT CSAIL,
  September 2026.
\newblock URL
  \url{https://mit-cdfg.github.io/Survey-AI-for-3D-modeling-Robotics/}.
\newblock Living survey.

\bibitem[Du et~al.(2026)Du, Guo, Zhang, Wang, Dong, Gao, Han, Lu, Yu, Shen,
  Chang, and Li]{du2026vlm}
Tingwen Du, Haoyu Guo, Hanxue Zhang, Xuanxu Wang, Junting Dong, Chao Gao,
  Mingfei Han, Tao Lu, Mulin Yu, Chunhua Shen, Xiaojun Chang, and Zhihui Li.
\newblock From {VLM} planners to robotic agents: Self-evolving harnesses for
  history-dependent manipulation.
\newblock \emph{arXiv preprint arXiv:2610.29560}, 2026.

\bibitem[Fan et~al.(2026)Fan, Zheng, Mao, Ran, Hu, Mao, Zhao, Wu, Fang, Chen,
  Zhu, Jiang, et~al.]{fan2026video}
Liaoyuan Fan, Yuhao Zheng, Dongxing Mao, Hangyu Ran, Siyuan Hu, Weijia Mao, Rui
  Zhao, Weijia Wu, Guian Fang, Yanzhe Chen, Zeyu Zhu, Yuxin Jiang, et~al.
\newblock Video generation meets agent harness: Agentic video generation,
  agentic world modeling and recursive self-improvement.
\newblock \emph{arXiv preprint arXiv:2610.40191}, 2026.

\bibitem[Fang et~al.(2025)Fang, Peng, Zhang, Wang, Yi, Zhang, Xu, Wu, Liu, Li,
  et~al.]{fang2025comprehensive}
Jinyuan Fang, Yanwen Peng, Xi~Zhang, Yingxu Wang, Xinhao Yi, Guibin Zhang,
  Yi~Xu, Bin Wu, Siwei Liu, Zihao Li, et~al.
\newblock A comprehensive survey of self-evolving ai agents: A new paradigm
  bridging foundation models and lifelong agentic systems.
\newblock \emph{arXiv preprint arXiv:2508.07407}, 2025.

\bibitem[Fei et~al.(2025)Fei, Wang, Shi, Dai, Cai, Qian, Ji, He, Zhang, Fei,
  Fu, Gong, and Qiu]{fei2025liberoplus}
Senyu Fei, Siyin Wang, Junhao Shi, Zihao Dai, Jikun Cai, Pengfang Qian, Li~Ji,
  Xinzhe He, Shiduo Zhang, Zhaoye Fei, Jinlan Fu, Jingjing Gong, and Xipeng
  Qiu.
\newblock {LIBERO-Plus}: In-depth robustness analysis of vision-language-action
  models.
\newblock \emph{arXiv preprint arXiv:2510.13626}, 2025.

\bibitem[Feng et~al.(2026)Feng, Pan, Kang, Chen, Yang, Ye, Wang, and
  Li]{feng2026enactopt}
ZhiYuan Feng, Zhuoshi Pan, Zhaolu Kang, Zhongxia Chen, Shuo Yang, Sen Ye, Heng
  Wang, and Han Li.
\newblock {EnactOpt}: Diagnosis-guided harness evolution for embodied agents.
\newblock \emph{arXiv preprint arXiv:2610.14977}, 2026.

\bibitem[Fernando et~al.(2023)Fernando, Banarse, Michalewski, Osindero, and
  Rockt{\"a}schel]{fernando2023promptbreeder}
Chrisantha Fernando, Dylan Banarse, Henryk Michalewski, Simon Osindero, and Tim
  Rockt{\"a}schel.
\newblock Promptbreeder: Self-referential self-improvement via prompt
  evolution.
\newblock \emph{arXiv preprint arXiv:2309.16797}, 2023.

\bibitem[Fischler and Bolles(1981)]{fischler1981random}
Martin~A. Fischler and Robert~C. Bolles.
\newblock Random sample consensus: A paradigm for model fitting with
  applications to image analysis and automated cartography.
\newblock \emph{Communications of the ACM}, 24\penalty0 (6):\penalty0 381--395,
  1981.

\bibitem[Fu et~al.(2021)Fu, Cai, Gao, Zhang, Wang, Li, Zeng, Sun, Jia, Zhao,
  et~al.]{fu20213d}
Huan Fu, Bowen Cai, Lin Gao, Ling-Xiao Zhang, Jiaming Wang, Cao Li, Qixun Zeng,
  Chengyue Sun, Rongfei Jia, Binqiang Zhao, et~al.
\newblock {3D-FRONT}: {3D} furnished rooms with layouts and semantics.
\newblock In \emph{2021 IEEE/CVF International Conference on Computer Vision
  (ICCV)}, pages 10913--10922. IEEE, 2021.

\bibitem[{Galbot Team} et~al.(2026){Galbot Team}, Chen, Cheng, Deng, Ding,
  Dong, Gao, Jia, Li, Li, et~al.]{galbot2026systematically}
{Galbot Team}, Xuchuan Chen, Xiaoqian Cheng, Yu~Deng, Lihe Ding, Shaocong Dong,
  Xiangjun Gao, Haozhe Jia, Zekai Li, Zhoujian Li, et~al.
\newblock Systematically exploring the capabilities of {GPT-6 Astra} as
  embodied policies.
\newblock \emph{arXiv preprint arXiv:2609.38537}, 2026.

\bibitem[Gao et~al.(2026{\natexlab{a}})Gao, Geng, Hua, Hu, Juan, Liu, Liu, Qiu,
  Qi, Wu, et~al.]{gao2025survey}
Huan-ang Gao, Jiayi Geng, Wenyue Hua, Mengkang Hu, Xinzhe Juan, Hongzhang Liu,
  Shilong Liu, Jiahao Qiu, Xuan Qi, Yiran Wu, et~al.
\newblock A survey of self-evolving agents: What, when, how, and where to
  evolve on the path to artificial super intelligence.
\newblock \emph{Transactions on Machine Learning Research}, 2026{\natexlab{a}}.

\bibitem[Gao et~al.(2026{\natexlab{b}})Gao, Hu, Chen, Yao, Wang, Bai, Wu, Han,
  Cong, Gui, Deng, and Li]{gao2026evoagentbench}
Xingze Gao, Chuanrui Hu, Hongda Chen, Pengfei Yao, Zhao Wang, Yi~Bai, Zhengwei
  Wu, Yunyun Han, Xiaofeng Cong, Jie Gui, Yafeng Deng, and Teng Li.
\newblock {EvoAgentBench}: Benchmarking agent self-evolution via ability
  transfer.
\newblock In \emph{Findings of the Association for Computational Linguistics:
  EMNLP 2026}, 2026{\natexlab{b}}.

\bibitem[Ge et~al.(2026)Ge, Zhou, Zeng, Wan, Wu, Zheng, Wang, Jia, Shangguan,
  Zhu, Liu, He, et~al.]{ge2026code}
Shijia Ge, Alex Zhou, Jianshu Zeng, Yexing Wan, Di~Wu, Zelin Zheng, Yazhe Wang,
  Zhiqi Jia, Xuan Shangguan, Jay Zhu, Yijun Liu, Lingyu He, et~al.
\newblock Make code as policy great again: Frontier agents write, call, and
  evolve robot tools.
\newblock \emph{arXiv preprint arXiv:2609.39018}, 2026.

\bibitem[{Gemma Team}(2026)]{gemma2026gemma4}
{Gemma Team}.
\newblock {Gemma 4} technical report.
\newblock \emph{arXiv preprint arXiv:2607.02770}, 2026.

\bibitem[Ghosh et~al.(2023)Ghosh, Hajishirzi, and Schmidt]{ghosh2023geneval}
Dhruba Ghosh, Hannaneh Hajishirzi, and Ludwig Schmidt.
\newblock {GenEval}: An object-focused framework for evaluating text-to-image
  alignment.
\newblock In \emph{Advances in Neural Information Processing Systems (NeurIPS),
  Datasets and Benchmarks Track}, 2023.

\bibitem[{GLM-5 Team}(2026)]{glm2026glm5}
{GLM-5 Team}.
\newblock {GLM-5}: from vibe coding to agentic engineering.
\newblock \emph{arXiv preprint arXiv:2602.15763}, 2026.

\bibitem[{GLM-V Team}(2025)]{hong2025glm}
{GLM-V Team}.
\newblock {GLM-4.5V} and {GLM-4.1V-Thinking}: Towards versatile multimodal
  reasoning with scalable reinforcement learning.
\newblock \emph{arXiv preprint arXiv:2507.01006}, 2025.

\bibitem[{Google}(2026{\natexlab{a}})]{google2026gemini37}
{Google}.
\newblock {Gemini 3.7 Flash}.
\newblock Gemini API documentation,
  \url{https://ai.google.dev/gemini-api/docs/models/gemini-3.7-flash},
  2026{\natexlab{a}}.

\bibitem[{Google}(2026{\natexlab{b}})]{google2026gemini4argon}
{Google}.
\newblock {Gemini 4 Argon}.
\newblock Gemini API documentation,
  \url{https://ai.google.dev/gemini-api/docs/models/gemini-4-argon},
  2026{\natexlab{b}}.

\bibitem[{Google DeepMind}(2026)]{google2026gemini38}
{Google DeepMind}.
\newblock {Gemini 3.8 Flash} model card.
\newblock \url{https://deepmind.google/models/model-cards/gemini-3-8-flash/},
  September 2026.

\bibitem[Grauman et~al.(2024)Grauman, Westbury, Torresani, Kitani, Malik,
  Afouras, Ashutosh, Baiyya, Bansal, Boote, et~al.]{grauman2024ego}
Kristen Grauman, Andrew Westbury, Lorenzo Torresani, Kris Kitani, Jitendra
  Malik, Triantafyllos Afouras, Kumar Ashutosh, Vijay Baiyya, Siddhant Bansal,
  Bikram Boote, et~al.
\newblock {Ego-Exo4D}: Understanding skilled human activity from first- and
  third-person perspectives.
\newblock In \emph{Proceedings of the IEEE/CVF conference on computer vision
  and pattern recognition}, pages 19383--19400, 2024.

\bibitem[Han et~al.(2026{\natexlab{a}})Han, Wang, Ling, Gong, Liu, Jiang, Cao,
  and Xie]{han2026imagersi}
Shuo Han, Zhilin Wang, Peng Ling, Wenjia Gong, Yang Liu, Guanjun Jiang, Yukun
  Cao, and Xike Xie.
\newblock {ImageRSI}: Verification-driven recursive self-improvement for
  agentic image generation.
\newblock \emph{arXiv preprint arXiv:2610.32120}, 2026{\natexlab{a}}.

\bibitem[Han et~al.(2026{\natexlab{b}})Han, Luo, Fu, Zhu, Feng, Wang, Xu, Meng,
  and Zhang]{han2026self}
Xiaofeng Han, Kaize Luo, Zenghuang Fu, Zichong Zhu, Zhe Feng, Changwei Wang,
  Rongtao Xu, Weiliang Meng, and Xiaopeng Zhang.
\newblock Recursive self-improvement in physical {AI}: Mechanisms, evidence,
  and evaluation.
\newblock \emph{arXiv preprint arXiv:2610.38266}, 2026{\natexlab{b}}.

\bibitem[Han et~al.(2026{\natexlab{c}})Han, Guo, Zeng, Cui, Xin, Zhou, Zhou,
  Zhang, and Li]{han2026embodied}
Zhi Han, Zihan Guo, Chenxi Zeng, Zicai Cui, Yue Xin, Ming Zhou, Dongzhan Zhou,
  Weinan Zhang, and Yang Li.
\newblock {Embodied Skill Bench}: Benchmarking how well agents with embodied
  skills perform on physical tasks.
\newblock \emph{arXiv preprint arXiv:2610.16238}, 2026{\natexlab{c}}.

\bibitem[Hu et~al.(2026)Hu, Martin-Martin, Stone, Duan, Jiang, and
  Shi]{hu2026simex}
Jiaheng Hu, Roberto Martin-Martin, Peter Stone, Rocky Duan, Zhenyu Jiang, and
  Guanya Shi.
\newblock {SimEX}: Simulation-integrated robotics {AutoResearch}.
\newblock \emph{arXiv preprint arXiv:2609.38982}, 2026.

\bibitem[Hu et~al.(2025)Hu, Lu, and Clune]{hu2025automated}
Shengran Hu, Cong Lu, and Jeff Clune.
\newblock Automated design of agentic systems.
\newblock In \emph{International Conference on Learning Representations},
  volume 2025, pages 21344--21377, 2025.

\bibitem[Huang et~al.(2026{\natexlab{a}})Huang, Wang, Wu, Ran, Guo, Zhang, Men,
  Wang, Zhou, Yao, and Wu]{huang2026llms}
Da~Huang, Jichao Wang, Junhui Wu, Tong Ran, Linting Guo, Siying Zhang, Chuang
  Men, Sukai Wang, Yue Zhou, Maoqing Yao, and Mo~Wu.
\newblock Frontier {LLMs} enter the world of robotics: A cross-model evaluation
  on {RoboColiseum}.
\newblock \emph{arXiv preprint arXiv:2610.27865}, 2026{\natexlab{a}}.

\bibitem[Huang et~al.(2026{\natexlab{b}})Huang, Chen, Liu, He, He, Huang,
  Zheng, Zhang, and Zhong]{huang2026embodiedclbench}
Jie Huang, Yanan Chen, Ruixun Liu, Kaichen He, Xinyi He, Rui Huang, Zilong
  Zheng, Zhenliang Zhang, and Yiwu Zhong.
\newblock {EmbodiedCLBench}: Evaluating continual learning for self-evolving
  embodied agents.
\newblock \emph{arXiv preprint arXiv:2610.13021}, 2026{\natexlab{b}}.

\bibitem[Huang et~al.(2023{\natexlab{a}})Huang, Sun, Xie, Li, and
  Liu]{huang2023t2i}
Kaiyi Huang, Kaiyue Sun, Enze Xie, Zhenguo Li, and Xihui Liu.
\newblock {T2I-CompBench}: A comprehensive benchmark for open-world
  compositional text-to-image generation.
\newblock In \emph{Advances in Neural Information Processing Systems (NeurIPS),
  Datasets and Benchmarks Track}, 2023{\natexlab{a}}.

\bibitem[Huang et~al.(2023{\natexlab{b}})Huang, Vora, Liang, and
  Leskovec]{huang2023mlagentbench}
Qian Huang, Jian Vora, Percy Liang, and Jure Leskovec.
\newblock {MLAgentBench}: Evaluating language agents on machine learning
  experimentation.
\newblock \emph{arXiv preprint arXiv:2310.03302}, 2023{\natexlab{b}}.

\bibitem[Huang et~al.(2026{\natexlab{c}})Huang, Wang, Li, Chen, Haupt, Luo,
  Liu, Choi, Juefei-Xu, Zhang, Poovendran, Sra, Liu, Burtenshaw, Wolf,
  Pentland, Chen, and Xu]{huang2026rsiarenarecursiveselfimprovement}
Yue Huang, Yanbo Wang, Lichi Li, Yuanyuan Chen, Andreas Haupt, Xiaonan Luo,
  Zheyuan Liu, Jihye Choi, Felix Juefei-Xu, Xiangliang Zhang, Radha Poovendran,
  Misha Sra, Ying Liu, Ben Burtenshaw, Thomas Wolf, Alex Pentland, Zichen Chen,
  and Zhangchen Xu.
\newblock {RSI Arena}: Towards recursive self-improvement with human feedback,
  2026{\natexlab{c}}.
\newblock URL \url{https://arxiv.org/abs/2610.33624}.

\bibitem[Jiang et~al.(2025)Jiang, Schmidt, Srikanth, Xu, Kaplan, Jacenko, and
  Wu]{jiang2025aide}
Zhengyao Jiang, Dominik Schmidt, Dhruv Srikanth, Dixing Xu, Ian Kaplan, Deniss
  Jacenko, and Yuxiang Wu.
\newblock {AIDE}: {AI}-driven exploration in the space of code.
\newblock \emph{arXiv preprint arXiv:2502.13138}, 2025.

\bibitem[Kaliyev and
  Maryanskyy(2026)]{kaliyev2026taskcompletionverificationvsconformancegap}
Alibek Kaliyev and Artem Maryanskyy.
\newblock Beyond task completion: A verification-vs.-conformance gap in
  tool-evolving agents, 2026.
\newblock URL \url{https://arxiv.org/abs/2604.00392}.
\newblock KDD 2026 Workshop on Agentic AI Evaluation and Trustworthiness.

\bibitem[Kang et~al.(2026)Kang, Wang, Fan, and Liu]{kang2026encore}
Yifan Kang, Zihan Wang, Zhiwen Fan, and Bangya Liu.
\newblock Encore: Few-shot agentic discovery of manipulation strategies.
\newblock \emph{arXiv preprint arXiv:2609.37359}, 2026.

\bibitem[Ku et~al.(2024)Ku, Jiang, Wei, Yue, and Chen]{ku2024viescore}
Max Ku, Dongfu Jiang, Cong Wei, Xiang Yue, and Wenhu Chen.
\newblock {VIEScore}: Towards explainable metrics for conditional image
  synthesis evaluation.
\newblock In \emph{Proceedings of the 62nd Annual Meeting of the Association
  for Computational Linguistics (Volume 1: Long Papers)}, pages 12268--12290,
  2024.

\bibitem[Kuroki et~al.(2026)Kuroki, Kawaharazuka, and Tang]{kuroki2026learning}
So~Kuroki, Kento Kawaharazuka, and Yujin Tang.
\newblock Learning and transferring closed-loop robot software.
\newblock \emph{arXiv preprint arXiv:2609.19906}, 2026.

\bibitem[Lee et~al.(2026{\natexlab{a}})Lee, Xu, Seely, Lee, Zaharia, and
  Tang]{lee2026recursive}
Hyunin Lee, Jinglue Xu, Jeffrey Seely, Donghyun Lee, Matei Zaharia, and Yujin
  Tang.
\newblock Recursive harness self-improvement.
\newblock \emph{arXiv preprint arXiv:2607.15524}, 2026{\natexlab{a}}.

\bibitem[Lee et~al.(2026{\natexlab{b}})Lee, Nair, Zhang, Lee, Khattab, and
  Finn]{lee2026meta}
Yoonho Lee, Roshen Nair, Qizheng Zhang, Kangwook Lee, Omar Khattab, and Chelsea
  Finn.
\newblock Meta-harness: End-to-end optimization of model harnesses.
\newblock \emph{arXiv preprint arXiv:2603.28052}, 2026{\natexlab{b}}.

\bibitem[Lei et~al.(2026)Lei, Wan, Zhang, Alam, Zhong, Huang, Wang, Zhang,
  Zhou, Hsieh, et~al.]{lei2026skillevolbench}
Yingtie Lei, Zhongwei Wan, Jiankun Zhang, Samiul Alam, Zixuan Zhong, Peizhou
  Huang, Xin Wang, Jingxuan Zhang, Donghao Zhou, Yunta Hsieh, et~al.
\newblock {SkillEvolBench}: Benchmarking the evolution from episodic experience
  to procedural skills.
\newblock \emph{arXiv preprint arXiv:2605.24117}, 2026.

\bibitem[Li et~al.(2026)Li, Zou, Liu, Zhang, Hong, Cao, Lan, Zhang, Yu, Zhang,
  et~al.]{li2025iggt}
Hao Li, Zhengyu Zou, Fangfu Liu, Xuanyang Zhang, Fangzhou Hong, Yukang Cao,
  Yushi Lan, Manyuan Zhang, Gang Yu, Dingwen Zhang, et~al.
\newblock {IGGT}: Instance-grounded geometry transformer for semantic {3D}
  reconstruction.
\newblock In \emph{International Conference on Learning Representations}, 2026.

\bibitem[Liang et~al.(2026)Liang, Wu, and Dong]{liang2026vidharness}
Susan Liang, Jianmin Wu, and Daxiang Dong.
\newblock {VidHarness}: Evolving agent harnesses for cost-efficient long video
  understanding.
\newblock \emph{arXiv preprint arXiv:2609.38413}, 2026.

\bibitem[Lin et~al.(2026)Lin, Dong, Xu, Chang, Ning, Xia, Yang, Zhao, Song,
  Liu, Huang, Song, et~al.]{lin2026videoharness}
Zihao Lin, Haoyu Dong, Haobo Xu, Chi-Chih Chang, Xuying Ning, Yinglong Xia,
  Shuang Yang, Zhuokai Zhao, Xinying Song, Dongfang Liu, Lifu Huang, Yang Song,
  et~al.
\newblock {VideoHarness-SRSI}: Structured recursive self-improvement for video
  understanding harness.
\newblock \emph{arXiv preprint arXiv:2610.19424}, 2026.

\bibitem[Liu et~al.(2023)Liu, Zhu, Gao, Feng, Liu, Zhu, and
  Stone]{liu2023libero}
Bo~Liu, Yifeng Zhu, Chongkai Gao, Yihao Feng, Qiang Liu, Yuke Zhu, and Peter
  Stone.
\newblock {LIBERO}: Benchmarking knowledge transfer for lifelong robot
  learning.
\newblock \emph{Advances in Neural Information Processing Systems},
  36:\penalty0 44776--44791, 2023.

\bibitem[Liu et~al.(2026{\natexlab{a}})Liu, Shen, Cheng, Mi, Tang, Liu, Xiang,
  Li, Li, and Tang]{liu2026roboicl}
Fangcheng Liu, Yeqing Shen, Anda Cheng, Weishi Mi, Chao Tang, Chenyuan Liu,
  Yushun Xiang, Tingguang Li, Yong-Lu Li, and Yehui Tang.
\newblock {RoboICL}: Embodied in-context learning with {GPT-6 Astra}.
\newblock \emph{arXiv preprint arXiv:2609.34261}, 2026{\natexlab{a}}.

\bibitem[Liu et~al.(2024{\natexlab{a}})Liu, Li, Li, Li, Zhang, Shen, and
  Lee]{liu2024llava}
Haotian Liu, Chunyuan Li, Yuheng Li, Bo~Li, Yuanhan Zhang, Sheng Shen, and
  Yong~Jae Lee.
\newblock {LLaVA-NeXT}: Improved reasoning, {OCR}, and world knowledge.
\newblock Blog post, 2024{\natexlab{a}}.
\newblock URL \url{https://llava-vl.github.io/blog/2024-01-30-llava-next/}.

\bibitem[Liu et~al.(2026{\natexlab{b}})Liu, Zheng, Chen, Wen, Quan, Wu, Zhang,
  Yang, Peng, and Stiefelhagen]{liu2026egoexomem}
Ruiping Liu, Junwei Zheng, Yufan Chen, Di~Wen, Shaofang Quan, Chengzhi Wu,
  Jiaming Zhang, Kailun Yang, Kunyu Peng, and Rainer Stiefelhagen.
\newblock {EgoExoMem}: Cross-view memory reasoning over synchronized egocentric
  and exocentric videos.
\newblock \emph{arXiv preprint arXiv:2605.18734}, 2026{\natexlab{b}}.

\bibitem[Liu et~al.(2024{\natexlab{b}})]{liu2024grounding}
Shilong Liu et~al.
\newblock Grounding {DINO}: Marrying {DINO} with grounded pre-training for
  open-set object detection.
\newblock In \emph{European Conference on Computer Vision (ECCV)},
  2024{\natexlab{b}}.

\bibitem[Liu et~al.(2025)Liu, Han, et~al.]{liu2025step1x}
Shiyu Liu, Yucheng Han, et~al.
\newblock {Step1X-Edit}: A practical framework for general image editing.
\newblock \emph{arXiv preprint arXiv:2504.17761}, 2025.

\bibitem[Lowe(2004)]{lowe2004distinctive}
David~G. Lowe.
\newblock Distinctive image features from scale-invariant keypoints.
\newblock \emph{International Journal of Computer Vision}, 60\penalty0
  (2):\penalty0 91--110, 2004.

\bibitem[Lu et~al.(2024)Lu, Lu, Lange, Foerster, Clune, and Ha]{lu2024ai}
Chris Lu, Cong Lu, Robert~Tjarko Lange, Jakob Foerster, Jeff Clune, and David
  Ha.
\newblock The {AI} {Scientist}: Towards fully automated open-ended scientific
  discovery.
\newblock \emph{arXiv preprint arXiv:2408.06292}, 2024.

\bibitem[Luo et~al.(2026)Luo, Guo, and Li]{luo2026video}
Bingjun Luo, Jialin Guo, and Siqi Li.
\newblock Video-{RSI}: Recursive self-improvement of video understanding agents
  via harness evolution.
\newblock \emph{arXiv preprint arXiv:2609.37950}, 2026.

\bibitem[Meng et~al.(2026)Meng, Du, Chen, Zhao, Lu, Hu, and
  Shieh]{meng2026rsibench}
Fanqing Meng, Lingxiao Du, Qiguang Chen, Ziqi Zhao, Haocheng Lu, Mengkang Hu,
  and Michael~Qizhe Shieh.
\newblock {RSIBench-Data}: Benchmarking data-centric research for recursive
  self-improvement.
\newblock \emph{arXiv preprint arXiv:2607.25886}, 2026.

\bibitem[{Meta}(2026{\natexlab{a}})]{meta2026musespark}
{Meta}.
\newblock Introducing {Muse Spark 1.3}.
\newblock \url{https://research.meta.ai/blog/introducing-muse-spark-1-3},
  September 2026{\natexlab{a}}.

\bibitem[{Meta}(2026{\natexlab{b}})]{meta2026musewonder}
{Meta}.
\newblock Introducing {Muse Wonder} and {Muse Wonder Full}: Frontier models for
  agents.
\newblock
  \url{https://about.fb.com/news/2026/10/introducing-muse-wonder-models/},
  October 2026{\natexlab{b}}.

\bibitem[{MiniMax}(2026{\natexlab{a}})]{minimax2026m3}
{MiniMax}.
\newblock {MiniMax-M3}.
\newblock Model card, \url{https://huggingface.co/MiniMaxAI/MiniMax-M3},
  2026{\natexlab{a}}.

\bibitem[{MiniMax}(2026{\natexlab{b}})]{minimax_h3_modelcard}
{MiniMax}.
\newblock {MiniMax H3}.
\newblock Hugging Face model card, 2026{\natexlab{b}}.
\newblock URL \url{https://huggingface.co/MiniMaxAI/MiniMax-H3}.

\bibitem[Nasiriany et~al.(2024)Nasiriany, Maddukuri, Zhang, Parikh, Lo, Joshi,
  Mandlekar, and Zhu]{nasiriany2024robocasa}
Soroush Nasiriany, Abhiram Maddukuri, Lance Zhang, Adeet Parikh, Aaron Lo,
  Abhishek Joshi, Ajay Mandlekar, and Yuke Zhu.
\newblock {RoboCasa}: Large-scale simulation of everyday tasks for generalist
  robots.
\newblock In \emph{Robotics: Science and Systems}, 2024.

\bibitem[Nasiriany et~al.(2026)Nasiriany, Nasiriany, Maddukuri, and
  Zhu]{nasiriany2026robocasa365}
Soroush Nasiriany, Sepehr Nasiriany, Abhiram Maddukuri, and Yuke Zhu.
\newblock {RoboCasa365}: A large-scale simulation framework for training and
  benchmarking generalist robots.
\newblock In \emph{International Conference on Learning Representations},
  volume 2026, pages 98643--98667, 2026.

\bibitem[Novikov et~al.(2025)Novikov, V{\~u}, Eisenberger, Dupont, Huang,
  Wagner, Shirobokov, Kozlovskii, Ruiz, Mehrabian,
  et~al.]{novikov2025alphaevolve}
Alexander Novikov, Ng{\^a}n V{\~u}, Marvin Eisenberger, Emilien Dupont, Po-Sen
  Huang, Adam~Zsolt Wagner, Sergey Shirobokov, Borislav Kozlovskii,
  Francisco~JR Ruiz, Abbas Mehrabian, et~al.
\newblock {AlphaEvolve}: A coding agent for scientific and algorithmic
  discovery.
\newblock \emph{arXiv preprint arXiv:2506.13131}, 2025.

\bibitem[{OpenAI}(2026{\natexlab{a}})]{openai2026gpt56}
{OpenAI}.
\newblock {GPT-5.6} system card.
\newblock \url{https://deploymentsafety.openai.com/gpt-5-6}, July
  2026{\natexlab{a}}.

\bibitem[{OpenAI}(2026{\natexlab{b}})]{openai2026gpt6}
{OpenAI}.
\newblock {GPT-6 Astra} system card.
\newblock \url{https://deploymentsafety.openai.com/gpt-6-astra}, September
  2026{\natexlab{b}}.

\bibitem[{OpenAI}(2026{\natexlab{c}})]{openai2026gpt61sol}
{OpenAI}.
\newblock Addendum: {GPT-6.1 Sol} system card.
\newblock
  \url{https://cdn.openai.com/pdf/38e3efcf-545e-44cd-99ec-2b7eb395f4cc/oai_GPT_6_1_Sol.pdf},
  September 2026{\natexlab{c}}.

\bibitem[{OpenAI}(2026{\natexlab{d}})]{openai_codexcli}
{OpenAI}.
\newblock {Codex CLI}.
\newblock Official documentation, 2026{\natexlab{d}}.
\newblock URL \url{https://developers.openai.com/codex/cli/}.
\newblock Accessed October 29, 2026.

\bibitem[{OpenRSI Foundation}(2026)]{openrsi2026}
{OpenRSI Foundation}.
\newblock {OpenRSI Index}.
\newblock \url{https://index.openrsi.foundation}, 2026.

\bibitem[Perazzi et~al.(2016)Perazzi, Pont-Tuset, McWilliams, Van~Gool, Gross,
  and Sorkine-Hornung]{perazzi2016benchmark}
Federico Perazzi, Jordi Pont-Tuset, Brian McWilliams, Luc Van~Gool, Markus
  Gross, and Alexander Sorkine-Hornung.
\newblock A benchmark dataset and evaluation methodology for video object
  segmentation.
\newblock In \emph{Proceedings of the IEEE Conference on Computer Vision and
  Pattern Recognition (CVPR)}, pages 724--732, 2016.

\bibitem[Puyin et~al.(2026)Puyin, Xiang, Mao, Wei, Chen, Masood, Fei-Fei, and
  Adeli]{puyin2026quantiphy}
Li~Puyin, Tiange Xiang, Ella Mao, Shirley Wei, Xinye Chen, Adnan Masood,
  Li~Fei-Fei, and Ehsan Adeli.
\newblock {QuantiPhy}: A quantitative benchmark evaluating physical reasoning
  abilities of vision-language models.
\newblock In \emph{Proceedings of the IEEE/CVF Conference on Computer Vision
  and Pattern Recognition}, pages 33174--33184, 2026.

\bibitem[Qian et~al.(2026)Qian, Zhu, Li, Wang, and Ji]{qian2026meta}
Cheng Qian, Kunlun Zhu, Beibin Li, Zhenhailong Wang, and Heng Ji.
\newblock Learning meta-skills for agent harness design in test-time {AI4AI}.
\newblock \emph{arXiv preprint arXiv:2609.38143}, 2026.

\bibitem[Qu et~al.(2026)Qu, Cheng, Yang, Zhao, Lin, Shi, Li, Wang, Chua, and
  Jiang]{qu2026vincie}
Leigang Qu, Feng Cheng, Ziyan Yang, Qi~Zhao, Shanchuan Lin, Yichun Shi, Yicong
  Li, Wenjie Wang, Tat-Seng Chua, and Lu~Jiang.
\newblock {VINCIE}: Unlocking in-context image editing from video.
\newblock In \emph{International Conference on Learning Representations},
  volume 2026, pages 114873--114918, 2026.

\bibitem[{Qwen Team}(2026{\natexlab{a}})]{qwen2026qwen36}
{Qwen Team}.
\newblock {Qwen3.6-35B-A3B}: Agentic coding power, now open to all.
\newblock \url{https://qwen.ai/blog?id=qwen3.6-35b-a3b}, 2026{\natexlab{a}}.

\bibitem[{Qwen Team}(2026{\natexlab{b}})]{qwen2026qwen38}
{Qwen Team}.
\newblock {Qwen3.8-Max}: A new bar for coding and cowork.
\newblock \url{https://qwen.ai/blog?id=qwen3.8}, August 2026{\natexlab{b}}.

\bibitem[Radford et~al.(2021)Radford, Kim, Hallacy, Ramesh, Goh, Agarwal,
  Sastry, Askell, Mishkin, Clark, Krueger, and Sutskever]{radford2021learning}
Alec Radford, Jong~Wook Kim, Chris Hallacy, Aditya Ramesh, Gabriel Goh,
  Sandhini Agarwal, Girish Sastry, Amanda Askell, Pamela Mishkin, Jack Clark,
  Gretchen Krueger, and Ilya Sutskever.
\newblock Learning transferable visual models from natural language
  supervision.
\newblock In \emph{Proceedings of the 38th International Conference on Machine
  Learning (ICML)}, volume 139 of \emph{Proceedings of Machine Learning
  Research}, pages 8748--8763, 2021.

\bibitem[Ramesh et~al.(2026)Ramesh, Devic, Garg, Parulekar, Mahnot,
  Pimpalgaonkar, Suresh, Dimakis, and
  Sathiamoorthy]{ramesh2026autoresearchexam}
Anirudha Ramesh, Siddartha Devic, Shivank Garg, Advait Parulekar, Drish Mahnot,
  Shreyas Pimpalgaonkar, Vishnu Suresh, Alex Dimakis, and Maheswaran
  Sathiamoorthy.
\newblock {AutoResearchExam}, September 2026.
\newblock URL \url{https://benchmarks.bespokelabs.ai/autoresearchexam/}.
\newblock Published September 9, 2026.

\bibitem[Rank et~al.(2026)Rank, Bhatnagar, Prabhu, Eisenberg, Nguyen, Bethge,
  and Andriushchenko]{rank2026posttrainbench}
Ben Rank, Hardik Bhatnagar, Ameya Prabhu, Shira Eisenberg, Karina Nguyen,
  Matthias Bethge, and Maksym Andriushchenko.
\newblock {PostTrainBench}: Can {LLM} agents automate {LLM} post-training?
\newblock \emph{arXiv preprint arXiv:2603.08640}, 2026.

\bibitem[{Rapidata}(2025)]{rapidata_veo31_preferences}
{Rapidata}.
\newblock {Rapidata Video Generation Veo 3.1 Human Preference}.
\newblock Hugging Face dataset card, 2025.
\newblock URL
  \url{https://huggingface.co/datasets/Rapidata/text-2-video-human-preferences-veo3.1}.

\bibitem[Ravi et~al.(2024)Ravi, Gabeur, Hu, Hu, Ryali, Ma, Khedr,
  et~al.]{ravi2024sam2}
Nikhila Ravi, Valentin Gabeur, Yuan-Ting Hu, Ronghang Hu, Chaitanya Ryali,
  Tengyu Ma, Haitham Khedr, et~al.
\newblock {SAM} 2: Segment anything in images and videos.
\newblock \emph{arXiv preprint arXiv:2408.00714}, 2024.

\bibitem[Robeyns et~al.(2025)Robeyns, Szummer, and Aitchison]{robeyns2025self}
Maxime Robeyns, Martin Szummer, and Laurence Aitchison.
\newblock A self-improving coding agent.
\newblock \emph{arXiv preprint arXiv:2504.15228}, 2025.

\bibitem[{RSI-Exam Team}(2026)]{rsiexam2026}
{RSI-Exam Team}.
\newblock {RSI-Exam}: Benchmarking recursive self-improvement through
  executable research.
\newblock \url{https://github.com/aiming-lab/RSI-Exam}, 2026.

\bibitem[{SAM 3D Team} et~al.(2025){SAM 3D Team}, Chen, Chu, Gleize, Liang,
  Sax, et~al.]{chen2025sam}
{SAM 3D Team}, Xingyu Chen, Fu-Jen Chu, Pierre Gleize, Kevin~J. Liang,
  Alexander Sax, et~al.
\newblock {SAM 3D}: {3D}fy anything in images.
\newblock \emph{arXiv preprint arXiv:2511.16624}, 2025.

\bibitem[Shen et~al.(2021)Shen, Xia, Li, Mart{\'i}n-Mart{\'i}n, Fan, Wang,
  P{\'e}rez-D'Arpino, Buch, Srivastava, Tchapmi, Tchapmi, Vainio, Wong,
  Fei-Fei, and Savarese]{shen2021igibson}
Bokui Shen, Fei Xia, Chengshu Li, Roberto Mart{\'i}n-Mart{\'i}n, Linxi Fan,
  Guanzhi Wang, Claudia P{\'e}rez-D'Arpino, Shyamal Buch, Sanjana Srivastava,
  Lyne~P. Tchapmi, Micael~E. Tchapmi, Kent Vainio, Josiah Wong, Li~Fei-Fei, and
  Silvio Savarese.
\newblock {iGibson} 1.0: A simulation environment for interactive tasks in
  large realistic scenes.
\newblock In \emph{IEEE/RSJ International Conference on Intelligent Robots and
  Systems (IROS)}, 2021.

\bibitem[Shen et~al.(2025)Shen, Zhao, Zhao, Xu, Zhang, Zhu, and
  Yin]{shen2024zoomeye}
Haozhan Shen, Kangjia Zhao, Tiancheng Zhao, Ruochen Xu, Zilun Zhang, Mingwei
  Zhu, and Jianwei Yin.
\newblock {ZoomEye}: Enhancing multimodal {LLMs} with human-like zooming
  capabilities through tree-based image exploration.
\newblock In \emph{Proceedings of the 2025 Conference on Empirical Methods in
  Natural Language Processing}, 2025.

\bibitem[Shen et~al.(2026{\natexlab{a}})Shen, Kova\v{c}i\v{c}, Kulits, Wang,
  Li, Tenenbaum, Yuille, Chen, and Wu]{shen20264dcodebench}
Ruihong Shen, \v{Z}iga Kova\v{c}i\v{c}, Peter Kulits, Xingrui Wang, Zizhang Li,
  Joshua~B. Tenenbaum, Alan Yuille, Jieneng Chen, and Jiajun Wu.
\newblock {4DCodeBench}: Benchmarking agents on inverse graphics of dynamic
  scenes.
\newblock \emph{arXiv preprint arXiv:2610.03715}, 2026{\natexlab{a}}.

\bibitem[Shen et~al.(2026{\natexlab{b}})Shen, You, Liu, Zheng, Zha, Yamazaki,
  Zhang, Huang, Sun, Chen, et~al.]{you2026embodiedswe}
Zeyu Shen, Haoxiang You, Yilang Liu, Zhicheng Zheng, Lihan Zha, Kashu Yamazaki,
  Mingtong Zhang, Suning Huang, Jiankai Sun, Qianzhong Chen, et~al.
\newblock {EmbodiedSWE}: Coding agents for long horizon dexterous robotics.
\newblock \emph{arXiv preprint arXiv:2609.27308}, 2026{\natexlab{b}}.

\bibitem[Shinn et~al.(2023)Shinn, Cassano, Gopinath, Narasimhan, and
  Yao]{shinn2023reflexion}
Noah Shinn, Federico Cassano, Ashwin Gopinath, Karthik Narasimhan, and Shunyu
  Yao.
\newblock Reflexion: Language agents with verbal reinforcement learning.
\newblock \emph{Advances in neural information processing systems},
  36:\penalty0 8634--8652, 2023.

\bibitem[Song et~al.(2026)Song, Liang, Fu, Wang, Yang, and
  Liu]{song2026embodiedrsi}
Python Song, Zhixuan Liang, Kelsey Fu, Mengdi Wang, Junfeng Yang, and Shilong
  Liu.
\newblock {EmbodiedRSI}: Active continual robot learning through
  hypothesis-guided co-evolution.
\newblock \emph{arXiv preprint arXiv:2610.10498}, 2026.

\bibitem[Spearman(1904)]{spearman1904proof}
Charles Spearman.
\newblock The proof and measurement of association between two things.
\newblock \emph{The American Journal of Psychology}, 15\penalty0 (1):\penalty0
  72--101, 1904.

\bibitem[Srikanth et~al.(2026)Srikanth, Zhao, Xu, Wu, and
  Jiang]{srikanth2026recursive}
Dhruv Srikanth, Bingchen Zhao, Dixing Xu, Yuxiang Wu, and Zhengyao Jiang.
\newblock Recursive self-improvement of {AI} research agents.
\newblock \emph{arXiv preprint arXiv:2609.26457}, 2026.

\bibitem[Standley et~al.(2017)Standley, Sener, Chen, and
  Savarese]{standley2017image2mass}
Trevor Standley, Ozan Sener, Dawn Chen, and Silvio Savarese.
\newblock image2mass: Estimating the mass of an object from its image.
\newblock In \emph{Proceedings of the 1st Annual Conference on Robot Learning},
  volume~78 of \emph{Proceedings of Machine Learning Research}, pages 324--333,
  2017.

\bibitem[Starace et~al.(2025)Starace, Jaffe, Sherburn,
  et~al.]{starace2025paperbench}
Giulio Starace, Oliver Jaffe, Dane Sherburn, et~al.
\newblock {PaperBench}: Evaluating {AI}'s ability to replicate {AI} research.
\newblock \emph{arXiv preprint arXiv:2504.01848}, 2025.

\bibitem[Straub et~al.(2019)Straub, Whelan, Ma, Chen, Wijmans, Green, Engel,
  Mur-Artal, Ren, Verma, et~al.]{straub2019replica}
Julian Straub, Thomas Whelan, Lingni Ma, Yufan Chen, Erik Wijmans, Simon Green,
  Jakob~J Engel, Raul Mur-Artal, Carl Ren, Shobhit Verma, et~al.
\newblock The replica dataset: A digital replica of indoor spaces.
\newblock \emph{arXiv preprint arXiv:1906.05797}, 2019.

\bibitem[Sun et~al.(2025)Sun, Huang, Liu, Wu, Xu, Li, and Liu]{sun2025t2v}
Kaiyue Sun, Kaiyi Huang, Xian Liu, Yue Wu, Zihan Xu, Zhenguo Li, and Xihui Liu.
\newblock {T2V-CompBench}: A comprehensive benchmark for compositional
  text-to-video generation.
\newblock In \emph{2025 IEEE/CVF Conference on Computer Vision and Pattern
  Recognition (CVPR)}, pages 8406--8416. IEEE, 2025.

\bibitem[Sun et~al.(2026)Sun, Wang, Song, Gu, and Liu]{sun2026harmony}
Shufan Sun, Chen Wang, Enxin Song, Jiatao Gu, and Lingjie Liu.
\newblock {HARMONY}: Hierarchical agentic reasoning for {MONocular}
  image-to-scene synthesis.
\newblock \emph{arXiv preprint arXiv:2609.26793}, 2026.

\bibitem[{Tencent Hunyuan Foundation Model Team}(2025)]{cao2025hunyuanimage}
{Tencent Hunyuan Foundation Model Team}.
\newblock {HunyuanImage 3.0} technical report.
\newblock \emph{arXiv preprint arXiv:2509.23951}, 2025.

\bibitem[Tian et~al.(2026{\natexlab{a}})Tian, Zhang, Liu, Liu, Liu, and
  Wang]{tian2026physevo}
Wenqing Tian, Zeyu Zhang, Zhaocheng Liu, Fengwei Liu, Qiang Liu, and Liang
  Wang.
\newblock {PhysEvo}: Astra can act, let it.
\newblock \emph{arXiv preprint arXiv:2610.08995}, 2026{\natexlab{a}}.

\bibitem[Tian et~al.(2026{\natexlab{b}})Tian, Hu, Wang, Lu, and
  Wu]{tian2026evolution}
Yuan Tian, Bing Hu, Hao Wang, Binghang Lu, and Fang Wu.
\newblock Harness evolution hits a ceiling: When weight training should begin.
\newblock \emph{arXiv preprint arXiv:2610.11655}, 2026{\natexlab{b}}.

\bibitem[Todorov et~al.(2012)Todorov, Erez, and Tassa]{todorov2012mujoco}
Emanuel Todorov, Tom Erez, and Yuval Tassa.
\newblock {MuJoCo}: A physics engine for model-based control.
\newblock In \emph{IEEE/RSJ International Conference on Intelligent Robots and
  Systems (IROS)}, pages 5026--5033, 2012.

\bibitem[Tu et~al.(2021)Tu, Wang, Birkbeck, Adsumilli, and Bovik]{tu2021ugc}
Zhengzhong Tu, Yilin Wang, Neil Birkbeck, Balu Adsumilli, and Alan~C. Bovik.
\newblock {UGC-VQA}: Benchmarking blind video quality assessment for user
  generated content.
\newblock \emph{IEEE Transactions on Image Processing}, 30:\penalty0
  4449--4464, 2021.

\bibitem[Vasluianu et~al.(2023)Vasluianu, Seizinger, and
  Timofte]{vasluianu2023wsrd}
Florin-Alexandru Vasluianu, Tim Seizinger, and Radu Timofte.
\newblock {WSRD}: A novel benchmark for high resolution image shadow removal.
\newblock In \emph{IEEE/CVF Conference on Computer Vision and Pattern
  Recognition Workshops (CVPRW)}, pages 1826--1835, 2023.

\bibitem[Wang et~al.(2023)Wang, Xie, Jiang, Mandlekar, Xiao, Zhu, Fan, and
  Anandkumar]{wang2023voyager}
Guanzhi Wang, Yuqi Xie, Yunfan Jiang, Ajay Mandlekar, Chaowei Xiao, Yuke Zhu,
  Linxi Fan, and Anima Anandkumar.
\newblock Voyager: An open-ended embodied agent with large language models.
\newblock \emph{arXiv preprint arXiv:2305.16291}, 2023.

\bibitem[Wang(2026)]{wang2026stops}
Jiaming Wang.
\newblock What stops recursive self-improvement in robotics? lessons from 123
  rounds of agentic skill discovery.
\newblock \emph{arXiv preprint arXiv:2609.31760}, 2026.

\bibitem[Wang et~al.(2026{\natexlab{a}})Wang, Jie, Yan, Heng, and
  Li]{wang2026regenharness}
Kailin Wang, Haoxiang Jie, Yaoyuan Yan, Zhiyou Heng, and Zhaosong Li.
\newblock {RegenHarness}: A robot agent harness with evidence-gated recursive
  self-improvement.
\newblock \emph{arXiv preprint arXiv:2609.27612}, 2026{\natexlab{a}}.

\bibitem[Wang et~al.(2026{\natexlab{b}})Wang, Zhao, Yuan, and
  Li]{wang2026better}
Ke~Wang, Zijie Zhao, Zhiyi Yuan, and Changlun Li.
\newblock When better gets worse: Improvement fidelity for self-improving
  agents in adaptive worlds.
\newblock \emph{arXiv preprint arXiv:2609.32677}, 2026{\natexlab{b}}.

\bibitem[Wang et~al.(2026{\natexlab{c}})Wang, Chang, Ma, Yang, Feng, Liu, Yang,
  Cao, Wang, Su, and Wang]{wang2026adarsi}
Peidong Wang, Ying Chang, Zhiming Ma, Xiaocui Yang, Shi Feng, Daqi Liu, Yuzhe
  Yang, Xu~Cao, Chuang Wang, Jian Su, and Daling Wang.
\newblock {AdaRSI}: On-demand recursive self-improvement for embodied agents.
\newblock \emph{arXiv preprint arXiv:2610.20474}, 2026{\natexlab{c}}.

\bibitem[Wang et~al.(2025{\natexlab{a}})Wang, Ding, Zeng, Zhou, Shen, Luo, Yu,
  and Tao]{wang2025divide}
Wenbin Wang, Liang Ding, Minyan Zeng, Xiabin Zhou, Li~Shen, Yong Luo, Wei Yu,
  and Dacheng Tao.
\newblock Divide, conquer and combine: A training-free framework for
  high-resolution image perception in multimodal large language models.
\newblock In \emph{Proceedings of the AAAI Conference on Artificial
  Intelligence}, volume~39, pages 7907--7915, 2025{\natexlab{a}}.

\bibitem[Wang et~al.(2026{\natexlab{d}})Wang, Pi{\k{e}}kos, Nanbo, Laakom,
  Chen, Ostaszewski, Zhuge, and Schmidhuber]{wang2026huxley}
Wenyi Wang, Piotr Pi{\k{e}}kos, Li~Nanbo, Firas Laakom, Yimeng Chen, Mateusz
  Ostaszewski, Mingchen Zhuge, and J{\"u}rgen Schmidhuber.
\newblock {Huxley-G{\"o}del Machine}: Human-level coding agent development by
  an approximation of the optimal self-improving machine.
\newblock In \emph{International Conference on Learning Representations},
  volume 2026, pages 80356--80386, 2026{\natexlab{d}}.

\bibitem[Wang et~al.(2025{\natexlab{b}})Wang, Li, Song, Xu, Tang, Zhuge, Pan,
  Song, Li, Singh, et~al.]{wang2025openhands}
Xingyao Wang, Boxuan Li, Yufan Song, Frank~F Xu, Xiangru Tang, Mingchen Zhuge,
  Jiayi Pan, Yueqi Song, Bowen Li, Jaskirat Singh, et~al.
\newblock {OpenHands}: An open platform for {AI} software developers as
  generalist agents.
\newblock In \emph{International Conference on Learning Representations},
  volume 2025, pages 65882--65919, 2025{\natexlab{b}}.

\bibitem[Wang et~al.(2026{\natexlab{e}})Wang, Xia, Liu, Zhang, Wu, Yang, He,
  Guo, and Lei]{wang2026sage}
Yihao Wang, Linhan Xia, Rui Liu, Zhaofeng Zhang, Hongyu Wu, Yang Yang, Jinglu
  He, Yu~Guo, and Kai Lei.
\newblock {SAGE}: A statistical acceptance gate for self-evolving agents.
\newblock \emph{arXiv preprint arXiv:2609.36043}, 2026{\natexlab{e}}.

\bibitem[Wang et~al.(2019)Wang, Inguva, and Adsumilli]{wang2019youtube}
Yilin Wang, Sasi Inguva, and Balu Adsumilli.
\newblock {YouTube} {UGC} dataset for video compression research.
\newblock In \emph{2019 IEEE 21st international workshop on multimedia signal
  processing (MMSP)}, pages 1--5. IEEE, 2019.

\bibitem[Wang et~al.(2004)Wang, Bovik, Sheikh, and Simoncelli]{wang2004image}
Zhou Wang, Alan~C. Bovik, Hamid~R. Sheikh, and Eero~P. Simoncelli.
\newblock Image quality assessment: From error visibility to structural
  similarity.
\newblock \emph{IEEE Transactions on Image Processing}, 13\penalty0
  (4):\penalty0 600--612, 2004.

\bibitem[Wei et~al.(2026{\natexlab{a}})Wei, Ni, Liu, Zhao, Ye, Jing, Xia, Liu,
  Leong, Heng, Huang, and Wang]{wei2026simple}
Songlin Wei, Zhenhao Ni, Jie Liu, Zhenyu Zhao, Junjie Ye, Hongyi Jing, Junkai
  Xia, Xiawei Liu, Michael Leong, Liang Heng, Di~Huang, and Yue Wang.
\newblock {SIMPLE}: Simulation-based policy learning and evaluation for
  humanoid loco-manipulation.
\newblock \emph{arXiv preprint arXiv:2606.08278}, 2026{\natexlab{a}}.

\bibitem[Wei et~al.(2022)Wei, Liu, Ling, and Su]{wei2022approximate}
Xinyue Wei, Minghua Liu, Zhan Ling, and Hao Su.
\newblock Approximate convex decomposition for {3D} meshes with collision-aware
  concavity and tree search.
\newblock \emph{ACM Transactions on Graphics}, 41\penalty0 (4), 2022.
\newblock Article 42.

\bibitem[Wei et~al.(2026{\natexlab{b}})Wei, Tan, Wang, and Zhou]{wei2026and}
Zhijie Wei, Ferris Tan, Jinghui Wang, and Hao Zhou.
\newblock Scale and selection: What makes automatic harness evolution work for
  visual-interface robot agents.
\newblock \emph{arXiv preprint arXiv:2609.39304}, 2026{\natexlab{b}}.

\bibitem[Wen et~al.(2026)Wen, Chen, Cao, Chen, Zou, Yu, Kuang, Xue, Lv, Wen,
  and Lu]{wen2026roborsi}
Zimo Wen, Yijin Chen, Yuxuan Cao, Wendi Chen, Yanwen Zou, Wenye Yu, Fuhang
  Kuang, Han Xue, Jun Lv, Chuan Wen, and Cewu Lu.
\newblock {RoboRSI}: Stable, efficient, and reusable robot self-evolution in
  complex real-world environments.
\newblock \emph{arXiv preprint arXiv:2610.12424}, 2026.

\bibitem[Wijk et~al.(2024)Wijk, Lin, Becker, et~al.]{wijk2024rebench}
Hjalmar Wijk, Tao Lin, Joel Becker, et~al.
\newblock {RE-Bench}: Evaluating frontier {AI} {R\&D} capabilities of language
  model agents against human experts.
\newblock \emph{arXiv preprint arXiv:2411.15114}, 2024.

\bibitem[Wu et~al.(2025{\natexlab{a}})]{wu2025qwen}
Chenfei Wu et~al.
\newblock {Qwen-Image} technical report.
\newblock \emph{arXiv preprint arXiv:2508.02324}, 2025{\natexlab{a}}.

\bibitem[Wu et~al.(2025{\natexlab{b}})Wu, Liu, Hung, and Duan]{wu2025spatial}
Diankun Wu, Fangfu Liu, Yi-Hsin Hung, and Yueqi Duan.
\newblock {Spatial-MLLM}: Boosting {MLLM} capabilities in visual-based spatial
  intelligence.
\newblock In \emph{Advances in Neural Information Processing Systems},
  2025{\natexlab{b}}.

\bibitem[Wu et~al.(2023)Wu, Zhang, Liao, Chen, Hou, Wang, Sun, Yan, and
  Lin]{wu2023exploring}
Haoning Wu, Erli Zhang, Liang Liao, Chaofeng Chen, Jingwen Hou, Annan Wang,
  Wenxiu Sun, Qiong Yan, and Weisi Lin.
\newblock Exploring video quality assessment on user generated contents from
  aesthetic and technical perspectives.
\newblock In \emph{Proceedings of the IEEE/CVF International Conference on
  Computer Vision (ICCV)}, pages 20144--20154, 2023.

\bibitem[Wu and Xie(2024)]{wu2024v}
Penghao Wu and Saining Xie.
\newblock {V*}: Guided visual search as a core mechanism in multimodal {LLMs}.
\newblock In \emph{2024 IEEE/CVF Conference on Computer Vision and Pattern
  Recognition (CVPR)}, pages 13084--13094. IEEE, 2024.

\bibitem[{xAI}(2026)]{xai2026grok46}
{xAI}.
\newblock Model card: {Grok 4.6}.
\newblock \url{https://media.x.ai/v1/website/card-4p6-4cd2dc57.pdf}, August
  2026.

\bibitem[Xia et~al.(2025)Xia, Lin, Hsu, Leboutet, Gao, Paulitsch, Ummenhofer,
  and Wang]{xia2026holoscene}
Hongchi Xia, Chih-Hao Lin, Hao-Yu Hsu, Quentin Leboutet, Katelyn Gao, Michael
  Paulitsch, Benjamin Ummenhofer, and Shenlong Wang.
\newblock {HoloScene}: Simulation-ready interactive {3D} worlds from a single
  video.
\newblock \emph{Advances in Neural Information Processing Systems},
  38:\penalty0 32501--32524, 2025.

\bibitem[Xia et~al.(2026)Xia, Han, Wang, Chen, Zhuang, Lee, Huang, Yu, CuiZhu,
  Ming, Yao, Gokturk, Pfister, and Lee]{xia2026rrsi}
Peng Xia, Rujun Han, Zifeng Wang, Yanfei Chen, Yufan Zhuang, Yoonho Lee,
  Chengsong Huang, Han Yu, Zhongying CuiZhu, Yifei Ming, Huaxiu Yao, Burak
  Gokturk, Tomas Pfister, and Chen-Yu Lee.
\newblock {RRSI}: Regularized recursive self-improvement of agent harnesses.
\newblock \emph{arXiv preprint arXiv:2609.24972}, 2026.

\bibitem[Xie et~al.(2026)Xie, Wang, Lin, Wang, and Li]{xie2026codimrecon}
Shuzhao Xie, Lelin Wang, Guying Lin, Zhi Wang, and Minchen Li.
\newblock {CoDimRecon}: Agentic reconstruction of sim-ready {3D} scenes with
  deformable curves, surfaces, and volumes.
\newblock \emph{arXiv preprint arXiv:2609.36024}, 2026.

\bibitem[Xue et~al.(2026)Xue, Ding, Shen, Wang, Yin, Wu, Chen, Wang, and
  Yang]{xue2026past}
Shuhan Xue, Zixin Ding, Yichen Shen, Yinjie Wang, Zhenfei Yin, Yingcheng Wu,
  Yuxin Chen, Mengdi Wang, and Ling Yang.
\newblock {PAST-Bench}: Benchmarking the foundations of recursive
  self-improvement in personal agents.
\newblock \emph{arXiv preprint arXiv:2608.04003}, 2026.

\bibitem[Yang et~al.(2025)Yang, Yang, Gupta, Han, Fei-Fei, and
  Xie]{yang2025thinking}
Jihan Yang, Shusheng Yang, Anjali~W Gupta, Rilyn Han, Li~Fei-Fei, and Saining
  Xie.
\newblock Thinking in space: How multimodal large language models see,
  remember, and recall spaces.
\newblock In \emph{2025 IEEE/CVF Conference on Computer Vision and Pattern
  Recognition (CVPR)}, pages 10632--10643. IEEE, 2025.

\bibitem[Yang et~al.(2024)Yang, Jimenez, Wettig, Lieret, Yao, Narasimhan, and
  Press]{yang2024swe}
John Yang, Carlos Jimenez, Alexander Wettig, Kilian Lieret, Shunyu Yao, Karthik
  Narasimhan, and Ofir Press.
\newblock {SWE-agent}: Agent-computer interfaces enable automated software
  engineering.
\newblock \emph{Advances in Neural Information Processing Systems},
  37:\penalty0 50528--50652, 2024.

\bibitem[Yang et~al.(2026{\natexlab{a}})Yang, Dagli, Zook, Hadfield, Goyal,
  Birchfield, Ramos, and Tremblay]{yang2026robolab}
Xuning Yang, Rishit Dagli, Alex Zook, Hugo Hadfield, Ankit Goyal, Stan
  Birchfield, Fabio Ramos, and Jonathan Tremblay.
\newblock {RoboLab}: A high-fidelity simulation benchmark for analysis of task
  generalist policies.
\newblock In \emph{Robotics: Science and Systems}, 2026{\natexlab{a}}.

\bibitem[Yang et~al.(2026{\natexlab{b}})Yang, Li, Hu, Liu, Huang, Sun, Li, Wu,
  Huang, Huang, Sun, Liu, et~al.]{yang2026robotworld}
Zhiqin Yang, Chenxin Li, Xiaomeng Hu, Yibin Liu, Weidong Huang, Jiankai Sun,
  Haitao Li, Zijian Wu, Yuzhi Huang, Fanding Huang, Hanwen Sun, Jiashun Liu,
  et~al.
\newblock {RobotWorld}: Benchmarking multimodal agents for robot use across
  diverse tasks and embodiments.
\newblock \emph{arXiv preprint arXiv:2610.10409}, 2026{\natexlab{b}}.

\bibitem[Yao et~al.(2026{\natexlab{a}})Yao, Lin, Du, and Mu]{yao2026code}
Haoyu Yao, Chenguo Lin, Ruize Du, and Yadong Mu.
\newblock Learning to code the world: Harness evolution for compositional {3D}
  reconstruction.
\newblock \emph{arXiv preprint arXiv:2610.37046}, 2026{\natexlab{a}}.

\bibitem[Yao et~al.(2025)Yao, Zhang, Yan, Zeng, Zhang, Xu, Yang, Gu, and
  Yu]{yao2025cast}
Kaixin Yao, Longwen Zhang, Xinhao Yan, Yan Zeng, Qixuan Zhang, Lan Xu, Wei
  Yang, Jiayuan Gu, and Jingyi Yu.
\newblock {CAST}: Component-aligned {3D} scene reconstruction from an {RGB}
  image.
\newblock \emph{ACM Transactions on Graphics}, 44\penalty0 (4), 2025.
\newblock \doi{10.1145/3730841}.

\bibitem[Yao et~al.(2026{\natexlab{b}})Yao, Hu, Chen, Giahi, Maragheh,
  Morishetti, Hong, Su, Xu, Korpeoglu, and Achan]{yao2026explicit}
Kehui Yao, Chuanbo Hu, Jiao Chen, Ramin Giahi, Reza~Yousefi Maragheh, Lalitesh
  Morishetti, Ziwei Hong, Yiyun Su, Jianpeng Xu, Evren Korpeoglu, and Kannan
  Achan.
\newblock An explicit feedback contract for iterative {ML-Agent} harnesses.
\newblock \emph{arXiv preprint arXiv:2610.28798}, 2026{\natexlab{b}}.

\bibitem[Yao et~al.(2023)Yao, Zhao, Yu, Du, Shafran, Narasimhan, and
  Cao]{yao2022react}
Shunyu Yao, Jeffrey Zhao, Dian Yu, Nan Du, Izhak Shafran, Karthik Narasimhan,
  and Yuan Cao.
\newblock {ReAct}: Synergizing reasoning and acting in language models.
\newblock In \emph{International Conference on Learning Representations}, 2023.

\bibitem[Yao et~al.(2026{\natexlab{c}})Yao, Zou, Jiao, Gao, Gao, Wang, Chen,
  Zhu, Li, Yan, Wang, Ma, et~al.]{yao2026agent}
Yifei Yao, Yude Zou, Guanbo Jiao, Xing Gao, Ning Gao, Yukai Wang, Jiantong
  Chen, Zihou Zhu, Xinyu Li, Ningyu Yan, Shuai Wang, Jianming Ma, et~al.
\newblock Profiling a frontier agent with {EBench}: From direct control to
  harness with policy.
\newblock \emph{arXiv preprint arXiv:2610.35832}, 2026{\natexlab{c}}.

\bibitem[Yeshwanth et~al.(2023)Yeshwanth, Liu, Nie{\ss}ner, and
  Dai]{yeshwanth2023scannet++}
Chandan Yeshwanth, Yueh-Cheng Liu, Matthias Nie{\ss}ner, and Angela Dai.
\newblock {ScanNet++}: A high-fidelity dataset of {3D} indoor scenes.
\newblock In \emph{2023 IEEE/CVF International Conference on Computer Vision
  (ICCV)}, pages 12--22. IEEE, 2023.

\bibitem[Yim et~al.(2020)Yim, Wang, Birkbeck, and Adsumilli]{yim2020subjective}
Joong~Gon Yim, Yilin Wang, Neil Birkbeck, and Balu Adsumilli.
\newblock Subjective quality assessment for youtube ugc dataset.
\newblock In \emph{2020 IEEE International Conference on Image Processing
  (ICIP)}, pages 131--135. IEEE, 2020.

\bibitem[Yin et~al.(2025)Yin, Wang, Pan, Lin, Wan, and Wang]{yin2025godel}
Xunjian Yin, Xinyi Wang, Liangming Pan, Li~Lin, Xiaojun Wan, and William~Yang
  Wang.
\newblock {G{\"o}del Agent}: A self-referential agent framework for recursive
  self-improvement.
\newblock In \emph{Proceedings of the 63rd Annual Meeting of the Association
  for Computational Linguistics (Volume 1: Long Papers)}, pages 27890--27913,
  2025.

\bibitem[Yu et~al.(2026{\natexlab{a}})Yu, Han, Zhang, Yin, Chang, Han, Liu,
  Zhang, Pavone, Feng, Xie, and Li]{yu2025thinking360deghumanoidvisual}
Heyang Yu, Yinan Han, Xiangyu Zhang, Baiqiao Yin, Bowen Chang, Xiangyu Han,
  Xinhao Liu, Jing Zhang, Marco Pavone, Chen Feng, Saining Xie, and Yiming Li.
\newblock Thinking in 360$^{\circ}$: Humanoid visual search in the wild.
\newblock In \emph{IEEE/CVF Conference on Computer Vision and Pattern
  Recognition}, 2026{\natexlab{a}}.
\newblock URL \url{https://arxiv.org/abs/2511.20351}.

\bibitem[Yu et~al.(2026{\natexlab{b}})Yu, Li, Li, Yan, Peng, Zhong, Xin, Yao,
  Tian, and Li]{yu2026robodoctor}
Heyang Yu, Haomin Li, Sihang Li, Jixin Yan, Shiyun Peng, Wanqi Zhong, Tianyu
  Xin, Yuan Yao, Xiaoyu Tian, and Yiming Li.
\newblock {RoboDoctor}: Training-free recursive self-improvement for robot
  policies.
\newblock \emph{arXiv preprint arXiv:2610.35604}, 2026{\natexlab{b}}.

\bibitem[Zelikman et~al.(2024)Zelikman, Lorch, Mackey, and
  Kalai]{zelikman2023self}
Eric Zelikman, Eliana Lorch, Lester Mackey, and Adam~Tauman Kalai.
\newblock Self-taught optimizer ({STOP}): Recursively self-improving code
  generation.
\newblock In \emph{Conference on Language Modeling}, 2024.

\bibitem[Zhang et~al.(2026{\natexlab{a}})Zhang, Zhang, Li, Zhang, Chen, Zhang,
  Bai, and Hu]{zhang2026self}
Hangfan Zhang, Shao Zhang, Kangcong Li, Chen Zhang, Yang Chen, Yiqun Zhang, Lei
  Bai, and Shuyue Hu.
\newblock Self-harness: Harnesses that improve themselves.
\newblock \emph{arXiv preprint arXiv:2606.09498}, 2026{\natexlab{a}}.

\bibitem[Zhang et~al.(2025{\natexlab{a}})Zhang, Hu, Lu, Lange, and
  Clune]{zhang2026darwingodelmachineopenended}
Jenny Zhang, Shengran Hu, Cong Lu, Robert Lange, and Jeff Clune.
\newblock {Darwin G{\"o}del Machine}: Open-ended evolution of self-improving
  agents, 2025{\natexlab{a}}.
\newblock URL \url{https://arxiv.org/abs/2505.22954}.

\bibitem[Zhang et~al.(2026{\natexlab{b}})Zhang, Zhao, Yang, Foerster, Clune,
  Jiang, Devlin, and Shavrina]{zhang2026hyperagents}
Jenny Zhang, Bingchen Zhao, Wannan Yang, Jakob Foerster, Jeff Clune, Minqi
  Jiang, Sam Devlin, and Tatiana Shavrina.
\newblock Hyperagents.
\newblock \emph{arXiv preprint arXiv:2603.19461}, 2026{\natexlab{b}}.

\bibitem[Zhang et~al.(2025{\natexlab{b}})Zhang, Xiang, Yu, Teng, Chen, Chen,
  Zhuge, Cheng, Hong, Wang, et~al.]{zhang2025aflow}
Jiayi Zhang, Jinyu Xiang, Zhaoyang Yu, Fengwei Teng, Xionghui Chen, Jiaqi Chen,
  Mingchen Zhuge, Xin Cheng, Sirui Hong, Jinlin Wang, et~al.
\newblock {AFlow}: Automating agentic workflow generation.
\newblock In \emph{International Conference on Learning Representations},
  volume 2025, pages 34040--34077, 2025{\natexlab{b}}.

\bibitem[Zhang et~al.(2026{\natexlab{c}})Zhang, Li, Kautz, Wang, and
  Choy]{zhang2026s4vy}
Jingdong Zhang, Xin Li, Jan Kautz, Wenping Wang, and Chris Choy.
\newblock {S4VY}: Segment anything in feed-forward {4D} visual geometry.
\newblock \emph{arXiv preprint arXiv:2609.36875}, 2026{\natexlab{c}}.

\bibitem[Zhang et~al.(2026{\natexlab{d}})Zhang, Wang, Tu, Li, Wang, and
  Zhan]{zhang2026beyond}
Jingdong Zhang, Yizhou Wang, Zhengzhong Tu, Xin Li, Wenping Wang, and Xiaohang
  Zhan.
\newblock Beyond thinking: Imagining in 360$^{\circ}$ for humanoid visual
  search.
\newblock \emph{arXiv preprint arXiv:2605.09146}, 2026{\natexlab{d}}.

\bibitem[Zhang et~al.(2025{\natexlab{c}})]{zhang2025uniser}
Jingdong Zhang et~al.
\newblock {UniSER}: A foundation model for unified soft effects removal.
\newblock \emph{arXiv preprint arXiv:2511.14183}, 2025{\natexlab{c}}.

\bibitem[Zhang et~al.(2018)Zhang, Isola, Efros, Shechtman, and
  Wang]{zhang2018unreasonable}
Richard Zhang, Phillip Isola, Alexei~A. Efros, Eli Shechtman, and Oliver Wang.
\newblock The unreasonable effectiveness of deep features as a perceptual
  metric.
\newblock In \emph{Proceedings of the IEEE Conference on Computer Vision and
  Pattern Recognition (CVPR)}, pages 586--595, 2018.

\bibitem[Zhang et~al.(2026{\natexlab{e}})Zhang, Jin, Liu, Wang, Zhang, Zhang,
  Zhang, Zhang, Niu, Li, et~al.]{zhang2026libero}
Yunbei Zhang, Zijian Jin, Yuanzhe Liu, Janet Wang, Xilun Zhang, Yuyou Zhang,
  Zhenyu Zhang, Daoan Zhang, Shuaicheng Niu, Gen Li, et~al.
\newblock {LIBERO-MAX}: Do robot policies adapt when the world changes?
\newblock \emph{arXiv preprint arXiv:2609.36518}, 2026{\natexlab{e}}.

\bibitem[Zhang et~al.(2026{\natexlab{f}})Zhang, Zhu, Chen, Tuya, and
  James]{zhang2026bigym}
Zexi Zhang, Zecheng Zhu, Zidong Chen, Zulkhuu Tuya, and Stephen James.
\newblock {BiGym} 2.0: Benchmarking learned and agent-developed policies for
  humanoid household manipulation.
\newblock \emph{arXiv preprint arXiv:2610.07594}, 2026{\natexlab{f}}.

\bibitem[Zheng et~al.(2025)Zheng, Li, Wang, Liu, Kang, Feng, Zheng, Zou, Chen,
  Zeng, Zhang, Pang, Liu, Wang, and Zhan]{zheng2025xvla}
Jinliang Zheng, Jianxiong Li, Zhihao Wang, Dongxiu Liu, Xirui Kang, Yuchun
  Feng, Yinan Zheng, Jiayin Zou, Yilun Chen, Jia Zeng, Ya-Qin Zhang, Jiangmiao
  Pang, Jingjing Liu, Tai Wang, and Xianyuan Zhan.
\newblock {X-VLA}: Soft-prompted transformer as scalable cross-embodiment
  vision-language-action model.
\newblock \emph{arXiv preprint arXiv:2510.10274}, 2025.

\bibitem[Zheng et~al.(2024)]{zheng2024sglang}
Lianmin Zheng et~al.
\newblock {SGLang}: Efficient execution of structured language model programs.
\newblock In \emph{Advances in Neural Information Processing Systems
  (NeurIPS)}, 2024.

\bibitem[Zheng et~al.(2026)Zheng, Yang, Hong, Zhao, Xu, Yang, Shen, and
  Yu]{zheng2025deepeyes}
Ziwei Zheng, Michael Yang, Jack Hong, Chenxiao Zhao, Guohai Xu, Le~Yang, Chao
  Shen, and Xing Yu.
\newblock {DeepEyes}: Incentivizing ``thinking with images'' via reinforcement
  learning.
\newblock In \emph{International Conference on Learning Representations
  (ICLR)}, 2026.

\bibitem[Zhou et~al.(2026{\natexlab{a}})Zhou, Yang, Chen, Zhang, Fu, Li, Zhu,
  Liu, Yang, Meng, Wang, Han, et~al.]{zhou2026improves}
Hai Zhou, Fan Yang, Zeyi Chen, Lejun Zhang, Chenyang Fu, Jianping Li, Jiayi
  Zhu, Xianyang Liu, Yingxuan Yang, Lili Meng, Lu~Wang, Xinyang Han, et~al.
\newblock Who improves the robot? toward physical recursive self-improvement.
\newblock \emph{arXiv preprint arXiv:2610.18842}, 2026{\natexlab{a}}.

\bibitem[Zhou(2026)]{zhou2026hierarchical}
Tailin Zhou.
\newblock Hierarchical self-improvement: A framework for task-specific
  evolvable agent harnesses.
\newblock \emph{arXiv preprint arXiv:2608.08466}, 2026.

\bibitem[Zhou et~al.(2025)Zhou, Xu, Tie, Chen, Zhang, Chu, Zhou, and
  Sun]{zhou2025libero}
Xueyang Zhou, Yangming Xu, Guiyao Tie, Yongchao Chen, Guowen Zhang, Duanfeng
  Chu, Pan Zhou, and Lichao Sun.
\newblock {LIBERO-PRO}: Towards robust and fair evaluation of
  vision-language-action models beyond memorization.
\newblock \emph{arXiv preprint arXiv:2510.03827}, 2025.

\bibitem[Zhou et~al.(2026{\natexlab{b}})Zhou, Hou, Fu, Lyu, Ye, Shan, Guo, Liu,
  He, Shan, Zhang, Yang, et~al.]{zhou2026multimodal}
Youwei Zhou, Bohan Hou, Yankai Fu, Yaoxu Lyu, Yifan Ye, Shilin Shan, Chen Guo,
  Jiaming Liu, Ran He, Caifeng Shan, Shanghang Zhang, Jianfei Yang, et~al.
\newblock Large multimodal models for direct robot control: {GPT-6} as a
  beginning.
\newblock \emph{arXiv preprint arXiv:2610.28259}, 2026{\natexlab{b}}.

\bibitem[Zou et~al.(2026)Zou, Li, Jiao, Liu, Xiao, Zhou, Hong, Su, Zhang, and
  Liu]{zou2026iggt4d}
Zhengyu Zou, Hao Li, Kuixuan Jiao, Liu Liu, Tingyang Xiao, Xiaolin Zhou,
  Fangzhou Hong, Zhizhong Su, Dingwen Zhang, and Ziwei Liu.
\newblock {IGGT4D}: Streaming {4D} instance-grounded geometry transformer.
\newblock \emph{arXiv preprint arXiv:2607.19228}, 2026.

\end{thebibliography}

\clearpage
\appendix
\section{Task cards}
\label{app:tasks}

This appendix describes each of the 24 tasks of \cref{tab:tasks}: what it measures, how its data were built from
its source and divided into splits, what the agent has to produce and with which resources, the starter solution,
the metric and where it comes from, and the prior work on the source benchmark.
Every task follows the setting of \cref{sec:task-formulation}: the development split is available to the agent with its
labels, while the validation and test splits are withheld. Unless stated otherwise, the solution may call the
model being evaluated within the stated limits, and it may not use the annotations of the source benchmark.

\newcommand{\cardlabel}[1]{\textit{#1.}\enspace}
\newenvironment{taskcard}[1]{\par\medskip\Needspace{5\baselineskip}\noindent\textbf{#1}\par\nopagebreak\smallskip\begin{list}{}{%
  \setlength{\leftmargin}{0pt}\setlength{\itemsep}{1pt}\setlength{\parsep}{0pt}\setlength{\topsep}{0pt}}}{\end{list}}

\subsection{Action}

\begin{taskcard}{RoboDojo}
\item \cardlabel{Purpose}Whether the agent can turn a general multimodal model into a closed-loop policy for a
dual-arm robot. No policy is trained: the model decides the motions at evaluation time, and the agent improves the
program that queries it.
\item \cardlabel{Source and construction}Ten simulation tasks of RoboDojo~\citep{chen2026robodojo}. Full credit
(success) requires: four bottles in a dustbin, using a handover between the hands; a garment folded with sleeves in
and hem aligned to the shoulders; digit tiles placed in the order forming the largest number; three tubes inserted
upright into a rack; all five own pieces placed in tic-tac-toe against a randomly playing opponent arm; a T-shaped
block pushed onto its pad without lifting; eight xylophone keys struck left to right with a mallet; the conveyor
object shown on a board placed in a lifted basket; four eggs in a holder with its lid closed; and five nesting dolls
lined up by size. Several tasks also require a return to the start pose. All layouts come from RoboDojo's evaluation
seed 0: per task, validation takes the first five, test the next five, and development the rest, of which a fixed
panel of five is evaluated. Validation layouts of three tasks are those of \citet{galbot2026systematically}. Tasks
with a randomized variant (fold clothes, largest number, Push-T, nesting dolls) use two standard and three
randomized layouts per split. The agent sees full development results, the validation score only, and no test
results.
\item \cardlabel{Setting}An ARX X5 dual-arm robot in Isaac Sim at 25\,Hz, run through the pinned single-layout RoboDojo client of RoboICL~\citep{liu2026roboicl}, 500 to 1100 steps per episode, starting
in an elbow-up pose slightly in front of RoboDojo's. The policy receives three RGB cameras, proprioception, and the
head-camera calibration, never object poses or rewards, and returns chunks of at most 64 end-effector or joint
targets, limited to 3\,cm and 0.15\,rad per step. It may call the run's model at most 80 times per episode and no
other model; scripted motions and its own inverse kinematics are allowed.
\item \cardlabel{Starter solution}One model call per chunk of up to ten end-effector motions, given the head-camera
image at $320\times240$, the robot state, and its notes from earlier calls.
\item \cardlabel{Metric}Each episode receives the progress score $s\in[0,100]$ of its RoboDojo rubric (for example
10, 25, 40, 100 for one to four bottles), as defined in the RoboDojo task documentation~\citep{chen2026robodojo}, 100 only on success, scored as the scene stands when it ends. The task score
is $S_t=\frac{1}{5}\sum_e s_{t,e}$ over the test layouts and the suite score $\frac{1}{10}\sum_t S_t$, a task without a
run counting as zero. The success rate is also reported.
\item \cardlabel{Prior work}RoboDojo has 42 simulation tasks and evaluates 30 trained manipulation
policies~\citep{chen2026robodojo}. Frontier models have also been used as the policy:
\citet{galbot2026systematically} reach 48\% success on a RoboDojo subset with GPT-6 Astra combined with $\pi_{0.5}$;
RoboICL~\citep{liu2026roboicl} conditions the model on demonstrations and interaction memory (overall score 50.64 on
30 tasks against 33.68 for the strongest baseline); and Encore~\citep{kang2026encore} writes policy programs from a
few demonstrations.
\end{taskcard}

\begin{taskcard}{RoboCasa}
\item \cardlabel{Purpose}Whether the agent can obtain a model-driven policy for a mobile manipulator that carries
over from the kitchens and objects used in development to kitchens and objects it has never seen.
\item \cardlabel{Source and construction}Eight tasks of RoboCasa365~\citep{nasiriany2026robocasa365}, an extension of
RoboCasa~\citep{nasiriany2024robocasa}, with RoboCasa's own success checks. Six are atomic tasks from RoboCasa365's seen-task target set: turn on the microwave
(on, gripper away from the button), turn on the sink faucet (handle turned over 0.4\,rad), open the named cabinet
(every door at least 90\% open), open the named drawer (at least 95\% open), place a mug under the coffee dispenser,
and turn off the named burner. Two are composite: portion hot dogs (from its unseen composite set; one bun and one sausage on each of two plates) and
kettle boiling (kettle within 15\,cm of a burner that is on). An episode is fixed by a kitchen (layout and style),
object instances, and a seed from which RoboCasa places the robot and objects and words the instruction. Development
has five fixed episodes in RoboCasa's pretraining kitchens with pretraining objects; validation and test have five
episodes each in the same five target kitchens (the first five of RoboCasa's ten evaluation kitchens) with target objects, differing by seed; validation and test were reduced from ten to five episodes each, and every episode was checked to build and reset. The agent sees full
development results, the validation success rate only, and no test results; neither it nor the policy may load target
kitchens or objects.
\item \cardlabel{Setting}The PandaOmron robot (a Franka Panda arm on a mobile base with a torso lift) in MuJoCo at
20\,Hz, with the per-task horizons of RoboCasa's dataset registry (450 to 2250 steps). The policy receives three $512\times512$ RGB cameras,
proprioception, and calibration, never object states, and returns chunks of at most 64 rows of RoboCasa's
12-dimensional action. It may call the run's model at most 80 times per episode and no other model.
\item \cardlabel{Starter solution}One model call per chunk of up to ten actions, given one base-camera image at
$256\times256$, its calibration, the robot state, and its notes from earlier calls.
\item \cardlabel{Metric}An episode succeeds, and ends, as soon as the task's success check of RoboCasa365~\citep{nasiriany2026robocasa365} holds after a step; otherwise, including
when the policy stops early, it counts as a failure. The task score is the test success rate
$\mathrm{SR}_t=\frac{1}{5}\sum_e \mathbb{1}[\text{success}_{t,e}]$ and the suite score $\frac{1}{8}\sum_t \mathrm{SR}_t$,
a task without a run counting as zero.
\item \cardlabel{Prior work}RoboCasa365 trains and benchmarks generalist policies, evaluating in target kitchens
separate from the pretraining ones~\citep{nasiriany2026robocasa365}. \citet{galbot2026systematically} evaluate GPT-6
Astra as a direct and hybrid policy on it, and \citet{kuroki2026learning} optimize closed-loop control code, raising
success on four RoboCasa source tasks from 28.3\% to 64.2\% and reaching 57.0\% on nine transfer tasks.
\end{taskcard}

\begin{taskcard}{Active Search}
\item \cardlabel{Purpose}Active visual search: a humanoid robot that sees part of a 360-degree scene turns its head to
find a named object (object search) or the direction to move in (path search), with as few movements as possible;
most targets start out of view.
\item \cardlabel{Source and construction}H*Bench~\citep{yu2025thinking360deghumanoidvisual} annotates search tasks on
real panoramas of transport hubs, shops and streets: an instruction, a target yaw range (and, for objects, a pitch
range), and starting directions and difficulty levels; its public benchmark split has 2{,}400 object-search and
1{,}600 path-search episodes. Development and validation come from it: each task becomes one episode, with one start
drawn by a fixed seed from its annotated starts; each panorama yields at most one episode, so the splits share no
scene; both are stratified by search kind and difficulty in proportion to H*Bench. Development has 80 episodes (40
object, 40 path) and validation 100 (50 and 50). The test split uses the non-public in-the-wild panoramas of
Imagining in 360$^\circ$~\citep{zhang2026beyond}, eight from Google Street View and seven generated by image models.
A target was annotated by clicking its center (objects, widened by $\pm30^\circ$ in yaw and $\pm20^\circ$ in pitch)
or the direction to move in ($\pm10^\circ$), with starts at multiples of $90^\circ$; one start per task gives 47
episodes (33 object, 14 path) in 15 scenes. The agent sees the development
episodes with their annotations and panoramas.
\item \cardlabel{Setting}The solution receives the instruction and a 100-degree view ($960\times720$ pixels), never
the panorama. With at most ten actions it rotates the head by a relative angle or submits an absolute direction,
which ends the episode; no submission fails. As in H*Bench, success requires the submitted direction to lie in the
target ranges widened to at least the tolerances above. Up to 20 model calls per episode on average and open
models on one GPU are allowed; other H*Bench data are not.
\item \cardlabel{Starter solution}The H*Bench baseline with its prompts: one model call per step on the whole
conversation (every view with the environment's feedback), answered with one rotation or submission.
\item \cardlabel{Metric}Success weighted by path length (SPL)~\citep{anderson2018on}, counting actions: episode $e$
scores $s_e\,\ell_e/\max(\ell_e,p_e)$, with $s_e\in\{0,1\}$ success, $p_e$ the actions taken (including the
submission), and $\ell_e$ the fewest that can succeed (1 if the target center lies within $50^\circ$ of the starting
yaw, else 2). Scores are averaged within each search kind, and the two averages are averaged.
\item \cardlabel{Prior work}H*Bench reports about 30\% success for the best proprietary models; post-training
Qwen2.5-VL-3B (HVS-3B) with supervised fine-tuning and multi-turn reinforcement learning raises success from 14.83\%
to 47.38\% (objects) and from 6.44\% to 24.94\% (paths)~\citep{yu2025thinking360deghumanoidvisual}. Imagining in
360$^\circ$ adds an Imaginator that predicts the layout of seen and unseen regions for a frozen actor, raising HVS-3B
from 48.04\% to 62.75\% and from 24.12\% to 39.38\% on H*Bench~\citep{zhang2026beyond}.
\end{taskcard}

\begin{taskcard}{RoboLab}
\item \cardlabel{Purpose}Whether the agent can obtain one tabletop manipulation policy that follows instructions for
tasks it has never seen; the splits contain different tasks, so the score measures generalization across tasks.
\item \cardlabel{Source and construction}The pool is every RoboLab-120~\citep{yang2026robolab} task of difficulty
simple or moderate with a horizon of at most 60\,s (73 tasks). Each task is given one group of RoboLab's taxonomy:
procedural (affordance, reorientation, sorting, stacking), else relational (conjunction, counting, spatial), else
visual (color, semantics, size). Tasks are shuffled within each (group, difficulty) stratum with a fixed seed and
dealt by systematic sampling to development (16 tasks), validation (12), test (24), and an unused remainder, so every
split has each stratum in proportion; test has 4 procedural, 9 relational, 10 visual, and 1 unattributed task.
Development tasks include stacking bowls, packing cans into a bin, standing a jug upright, and selecting a hammer by
color. Each task keeps its RoboLab scene, default instruction, horizon, success predicate, and subtask tracker unchanged. Start states are deterministic, so a task's episodes differ
only through the policy; development runs one episode per task, validation and test two (16, 24, and 48 episodes).
The agent sees development tasks and results, validation aggregates only, and no held-out task names.
\item \cardlabel{Setting}The DROID setup, a Franka Panda arm with a Robotiq 2F-85 gripper, in Isaac Lab at 15\,Hz.
The policy receives the instruction, over-the-shoulder and wrist RGB images, over-the-shoulder depth, calibration, and
proprioception, never the task name or object poses, and returns chunks of absolute end-effector poses and gripper
commands, limited to 5\,cm and 0.25\,rad per step. It may call the run's model at most 40 times within 1200\,s of
thinking per episode, no other model, and may not key behavior on task identity.
\item \cardlabel{Starter solution}One model call per gripper waypoint, given both images, the pose, and calibration,
followed by a straight-line move and an open or close.
\item \cardlabel{Metric}RoboLab's total score~\citep{yang2026robolab},
$\mathrm{Score}=\frac{100}{N}\sum_{e=1}^{N}\bigl(\mathbb{1}[\text{success}_e]+(1-\mathbb{1}[\text{success}_e])\,q_e\bigr)$
over the $N=48$ test episodes, where $q_e\in[0,1]$ is the fraction of subtasks completed when the episode ends. The
success rate is also reported.
\item \cardlabel{Prior work}RoboLab analyses task-generalist policies along visual, procedural, and relational
competencies~\citep{yang2026robolab}. \citet{chen2026volo} evaluate agents without a trained policy on its 120 tasks
with vague instructions: code-as-policy (CaP-X) reaches 14.4\% success, VLM-based task and motion planning (TiPToP)
18.9\%, and their orchestrator without a vision--language--action policy 17.8\%.
\end{taskcard}

\begin{taskcard}{LIBERO-MAX}
\item \cardlabel{Purpose}Whether the agent can build a robot controller that adapts when the world changes during
execution, not only before it starts.
\item \cardlabel{Source and construction}LIBERO-MAX~\citep{zhang2026libero} has 8{,}000 paired cases on
LIBERO~\citep{liu2023libero} scenes, with starting configurations from LIBERO-Plus~\citep{fei2025liberoplus} and
LIBERO-PRO~\citep{zhou2025libero}. A case pairs a Base episode with a Dynamic episode that has the same start and policy
seed and receives one online event: target or receptacle relocation, camera shift, sensor-noise onset, illumination or
visual-theme switch, a burst of distractors, or an inserted obstacle. In the selected cases the event fires when the
robot comes within 0.18\,m of a designated object, before the grasp. Cases are selected from official metadata,
never from outcomes. All splits use three LIBERO tasks (a bowl from a drawer onto a plate, a sauce bottle into a
basket, a plate pushed to the front of a stove) and the same number of cases per event type: 8, 16 and 24 pairs for
development, validation and test. No initial physical state appears in two splits, and states used in earlier
experiments are excluded from validation and test. The agent sees the development case definitions.
\item \cardlabel{Setting}A simulated Franka Panda (MuJoCo). Each step, the solution receives RGB and metric depth from
an external and a wrist camera, their calibration, the robot state and the instruction, and returns an absolute
end-effector pose and gripper command. It may query a frozen X-VLA policy~\citep{zheng2025xvla} for 30-step action
chunks, possibly on transformed images, and call the model up to 16 times per episode; no other model, no training,
no simulator state. The Dynamic episode replays the Base episode exactly until the event.
\item \cardlabel{Starter solution}Executes the X-VLA chunks unchanged, without calling the model.
\item \cardlabel{Metric}Dynamic success rate $\frac{1}{P}\sum_{p=1}^{P}s^{\mathrm{dyn}}_p$ over all $P$ planned pairs,
with $s^{\mathrm{dyn}}_p\in\{0,1\}$ the LIBERO success check of the Dynamic episode; an episode that reaches the horizon
or its 20-minute limit unsolved fails. The Base success rate and the paired difference, as in LIBERO-MAX, are also
reported.
\item \cardlabel{Prior work}Across fourteen vision--language--action, hybrid and world-action policies, LIBERO-MAX
events lower success by 11.0 to 25.7 percentage points, mostly through geometry and observation
changes~\citep{zhang2026libero}.
\end{taskcard}

\begin{taskcard}{LIBERO-PRO}
\item \cardlabel{Purpose}Whether the agent can build a robot controller that stays successful when a familiar task is
perturbed before the episode starts, rather than relying on the conditions the underlying policy was trained under.
\item \cardlabel{Source and construction}LIBERO-PRO~\citep{zhou2025libero} perturbs LIBERO~\citep{liu2023libero}
tasks in their objects, initial states, instructions and environments, and releases the perturbed task definitions
and initial states. Ten of its categories are used: object appearance, object shape, object texture, object
positions, paraphrased instruction, changed task goal, camera view angle, initial robot pose, visual noise or glare,
and view occlusion. Each case is one perturbed initial state of one of the three LIBERO tasks of LIBERO-MAX. Cases are
selected from official metadata, with equal numbers per category in each split, spread over the three tasks: 10, 20
and 30 episodes for development, validation and test (ten per task in test). No initial physical state appears in
two splits or in LIBERO-MAX, and states used in earlier experiments are excluded from validation and test. The agent
sees the development case definitions.
\item \cardlabel{Setting}As in LIBERO-MAX: a controller that may query the frozen X-VLA policy~\citep{zheng2025xvla}
and call the model up to 16 times per episode, with a 20-minute limit per episode and no other model.
\item \cardlabel{Starter solution}Executes the X-VLA chunks unchanged, without calling the model.
\item \cardlabel{Metric}Success rate $\frac{1}{N}\sum_{i=1}^{N}s_i$ over all $N$ planned episodes, with $s_i$ the
native LIBERO success check; an episode that reaches the horizon or the time limit unsolved fails.
\item \cardlabel{Prior work}LIBERO-PRO reports that models above 90\% success under the standard LIBERO evaluation
fall to 0.0\% under its generalized setting, a sign of memorized training conditions~\citep{zhou2025libero}.
LIBERO-Plus~\citep{fei2025liberoplus} analyses robustness along seven dimensions: object layout, camera viewpoint,
initial robot state, language, lighting, background texture and sensor noise.
\end{taskcard}

\begin{taskcard}{Embodied-SWE}
\item \cardlabel{Purpose}Whether the agent can write, offline with a simulator, a control program for a long-horizon,
contact-rich tool-use task that then runs without any model.
\item \cardlabel{Source and construction}The spatula task of EmbodiedSWE-Bench~\citep{you2026embodiedswe}: a Franka
arm under operational-space control in Isaac Lab wedges a spatula under a slice of bread in a pan, flips it, reloads
it onto the blade and serves it flat onto a plate, touching the bread only with the tool. The environment is built
with the benchmark's tools in its setting without agent tools, and the grader is withheld. Each evaluation runs eight
parallel environments with random bread size and pose, plate position and spatula pose. Development uses the build
seed (8 trajectories); validation 2 and test 4 hidden random seeds (16 and 32 trajectories). The agent receives the
benchmark's task statement, its rule for batched solutions and its staged-planning guide.
\item \cardlabel{Setting}A program that commands the robot only through the environment's step function and may read
object poses. Writing simulator state, applying external forces, changing physics or the scene, or altering the
grader scores zero. A rollout has 30 minutes of wall time, and no model is called during evaluation. During
development the agent has full simulator access on one GPU.
\item \cardlabel{Starter solution}Holds the arm still for ten steps and scores zero.
\item \cardlabel{Metric}The EmbodiedSWE-Bench rubric score~\citep{you2026embodiedswe}: each trajectory receives the
credit of the furthest stage it reaches, namely spatula lifted (0.2), bread on the blade (0.4), bread flipped and flat
(0.6), bread carried on the blade (0.8) and bread on the plate (1.0). The score is the mean over all trajectories of
the split; a crashed rollout scores zero, and a timed-out one is scored as it stands. The final-state success rate is
also reported.
\item \cardlabel{Prior work}EmbodiedSWE-Bench~\citep{you2026embodiedswe} evaluates coding agents that write such
programs in one four-hour run per task, and uses their solutions to generate demonstrations for
vision--language--action models. On this task, single runs of frontier coding agents reach rubric scores from 0.2 to
1.0, with one run disqualified for reward hacking.
\end{taskcard}

\subsection{Reasoning}

\begin{taskcard}{Ego-Exo Memory}
\item \cardlabel{Purpose}Memory reasoning across two synchronized views of one activity. The evidence for an answer is
visible only at particular moments, often in one view, and the model sees few frames per call, so the solution must
choose which moments and views to show.
\item \cardlabel{Source and construction}The questions are the Ego-Exo4D portion of
EgoExoMem~\citep{liu2026egoexomem}: 764 human-verified four-option questions on 125 recordings of cooking and
bicycle repair from Ego-Exo4D~\citep{grauman2024ego}. Each recording has an egocentric view from the glasses of the
person performing the activity ($448\times448$ pixels) and an exocentric view from a fixed camera ($796\times448$
pixels), synchronized at 30 frames per second and lasting from 20 seconds to 23 minutes. The questions concern habitual
location, instantaneous position, resulting location, egocentric direction, object state, allocentric relation, and
temporal order. All recordings of one session (the same room and people) are assigned to one split, stratified by
activity, with about 35\%, 25\%, and 40\% of the questions in the development, validation, and test splits. Each split
then keeps a fixed-seed sample stratified by activity and question type and spread over its videos: 80 questions on
38 videos, 80 on 33, and 120 on 47. The answer options are shuffled once and the videos are anonymized. With each
development question the agent also receives the annotated moments at which the answer is visible and the view that
shows it.
\item \cardlabel{Setting}Given the question, both videos, and frames of both views at one frame per second, the
solution returns an option letter. It may call the model three times per question on average over a split, with at
most $16\times448\times448$ pixels per call; further calls are refused. It may run open-weight models, for example for
frame retrieval, on one GPU, has no network access during evaluation, and may not use the annotations of EgoExoMem or
Ego-Exo4D.
\item \cardlabel{Starter solution}Shows the model eight evenly spaced frames of each view, each labeled with its view
and time and downscaled so that all sixteen fit the pixel limit, followed by the question, in a single call.
\item \cardlabel{Metric}Accuracy over the questions of a split; a question without a valid letter counts as wrong.
Chance is 0.25.
\item \cardlabel{Prior work}EgoExoMem evaluates multimodal language models on 2.6K cross-view memory questions of eight
types. On the full benchmark, the best existing model reaches 55.3\%, and the training-free frame selection method
E$^2$-Select, which allocates the frame budget by relevance and samples frames per view with a $k$-DPP, reaches
58.2\%~\citep{liu2026egoexomem}.
\end{taskcard}

\begin{taskcard}{Kinematics}
\item \cardlabel{Purpose}Quantitative physical reasoning from video: building a measurement procedure around a fixed
vision--language model, so that quantities follow from pixels, time, and a known scale.
\item \cardlabel{Source and construction}QuantiPhy~\citep{puyin2026quantiphy} contains 3{,}355 questions on 569 videos
from physics simulations, laboratory recordings, internet clips, and segmented objects. The task uses its public
labeled validation set, 158 questions on 24 videos. Each question asks for the size, distance, velocity, or
acceleration of an object, possibly at a stated time, in stated units, and gives one prior in physical units: a known
length, speed, or acceleration, and for some videos depths. The videos are assigned to fixed lists, so that no video
appears in two splits and every split mixes the four video sources and the static and dynamic priors and targets:
39 questions on 6 videos for development, 52 on 8 for validation, and 67 on 10 for testing.
\item \cardlabel{Setting}The solution returns a number in the unit of the question. It does not call the model being
evaluated: the only language or vision--language model it may use is Qwen3.8-27B with open weights, served locally
with at most 32 images ($32\times640\times360$ pixels in total) per call and eight calls per question on average over a
split. It may add non-language vision models and classical code on one GPU, such as point trackers, optical flow,
detectors, segmenters, and depth or pose estimators, but no other language model, neither in the solution nor to label
data, and no network access during evaluation.
\item \cardlabel{Starter solution}Shows the fixed model four evenly spaced frames, the question, and the prior in a
single call with reasoning disabled, and reads the first number of the reply.
\item \cardlabel{Metric}The mean relative accuracy (MRA). For a prediction $\hat y_i$ of the true value $y_i$,
\[
\mathrm{MRA}=\frac{1}{N}\sum_{i=1}^{N}\frac{1}{|\mathcal{C}|}\sum_{\theta\in\mathcal{C}}
\mathbf{1}\!\left[\frac{|\hat y_i-y_i|}{|y_i|}<1-\theta\right],\qquad \mathcal{C}=\{0.1,0.2,\dots,0.9,0.95\},
\]
and a question without a numeric answer scores zero. MRA was introduced by VSI-Bench~\citep{yang2025thinking} with
thresholds from 0.5 to 0.95; the threshold set above is that of QuantiPhy~\citep{puyin2026quantiphy}.
\item \cardlabel{Prior work}QuantiPhy evaluates 21 vision--language models that answer directly from the video. The
best reaches an MRA of 53.1, below the average human score of 55.6; the models rely on pretrained world knowledge more
than on the given video and prior, and a four-step chain of thought (measure the source property in pixels, estimate
the scale, measure the target in pixels, convert) lowers the score of 19 of the 21 models~\citep{puyin2026quantiphy}.
\end{taskcard}

\begin{taskcard}{Mass}
\item \cardlabel{Purpose}Estimating a property that is not directly visible: weight depends on material, on
whether an object is solid or hollow, and on how its volume is filled.
\item \cardlabel{Source and construction}The items are product listings of Amazon Berkeley
Objects~\citep{collins2022abo} that state a weight, all three dimensions, and a main photograph. Consumables (food,
supplements, cleaning and medical supplies) and multipacks or sets, recognized by name, are removed, as are items
outside 10\,g to 150\,kg and items whose weight divided by the volume of their bounding box lies outside 5 to
5000\,kg/m$^3$, which removes unit and listing errors. Variants of one product (the same brand and the same first six
words of the name) form one group, and one item per group is drawn with a fixed seed. The groups are split into 300
development, 300 validation, and 600 test items, with at most 3\% of a split from one product type, giving about 130
product types per split. Each item keeps its main photograph and up to two others, with the longer side at most
1024 pixels.
\item \cardlabel{Setting}Given the photographs, a coarse product type, and the listed dimensions in centimetres, but
not the name, brand, or description, the solution states the weight in kilograms. As in Kinematics, the only language
or vision--language model is the fixed Qwen3.8-27B, with at most eight images ($8\times1024\times1024$ pixels) per call
and four calls per item on average. Non-language models such as segmenters, depth estimators, image encoders, and
regressors trained on the development split may run on one GPU; other language models, text encoders applied to other
data, and any external source of the weights are not allowed.
\item \cardlabel{Starter solution}Shows the fixed model the main photograph, the product type, and the dimensions in a
single call with reasoning disabled, and reads the first number of the reply.
\item \cardlabel{Metric}The mean relative accuracy with the thresholds of Kinematics, averaged over items; an item
without a numeric answer scores zero.
\item \cardlabel{Prior work}Image2mass~\citep{standley2017image2mass} collected about 150{,}000 online product listings
with images, sizes, and weights, and estimates mass from an image together with the object's dimensions, comparing
learned models with human estimates. Amazon Berkeley Objects~\citep{collins2022abo} provides catalogue images,
metadata, and 3D models of real products for object understanding.
\end{taskcard}

\begin{taskcard}{ARC-AGI-1 and ARC-AGI-2}
\item \cardlabel{Purpose}Abstract reasoning from few examples: inferring a grid transformation from a few
demonstration pairs and applying it to new inputs, on puzzles designed to require little prior knowledge.
\item \cardlabel{Source and construction}The puzzles come from the official public repositories at fixed versions:
400 training and 400 evaluation puzzles for ARC-AGI-1~\citep{chollet2019measure}, and 1000 training and 120
evaluation puzzles for ARC-AGI-2~\citep{chollet2025arc}. Each puzzle has demonstration pairs and one or more test
inputs; grids are at most $30\times30$ with ten colors. Training puzzles identical to an evaluation puzzle are removed
(one for ARC-AGI-1). The remaining training puzzles are ordered by a seeded hash: the first 80 form the validation
split, and the rest form the development pool, which the agent receives with answers (319 puzzles for ARC-AGI-1); a
fixed panel of 32 development puzzles is scored in development evaluations. The test split consists of all official
evaluation puzzles, 400 for ARC-AGI-1 and 120 for ARC-AGI-2, so validation and test puzzles are disjoint.
\item \cardlabel{Setting}Given the demonstration pairs and the test inputs, the solution returns at most two candidate
grids per test input. It may call the model up to 2000 times per evaluation, with four puzzles solved in parallel in
independent sessions, and runs in a sandbox without a GPU or network access. It may use program search, hand-written
rules, or models trained on the development puzzles, but may not obtain or store held-out answers. Invalid predictions
score zero.
\item \cardlabel{Starter solution}Asks the model for two candidate outputs as JSON in a single call per puzzle; an
unparseable reply counts as no answer.
\item \cardlabel{Metric}Pass@2, in percent: with $g_{ij}^{1},g_{ij}^{2}$ the two candidates for test input $j$ of
puzzle $i$,
\[
\text{pass@2}=\frac{100}{N}\sum_{i=1}^{N}\prod_{j}\mathbf{1}\!\left[y_{ij}\in\{g_{ij}^{1},g_{ij}^{2}\}\right],
\]
so a puzzle counts only when every test output is matched exactly. The two attempts per test input follow the ARC
Prize protocol~\citep{chollet2024arc}.
\item \cardlabel{Prior work}ARC~\citep{chollet2019measure} was proposed to measure the efficiency of skill acquisition
on novel problems. In ARC Prize 2024 the best private-set score rose from 33\% to 55.5\%, mainly through deep-learning-guided
program synthesis and test-time training~\citep{chollet2024arc}; test-time training brings an 8B language model to
53.0\% on the public validation set, and 61.9\% when ensembled with program synthesis~\citep{akyurek2024surprising}.
ARC-AGI-2 raises the difficulty: on its semi-private set, o3 (medium) scores 3.0\%, against 53.0\% on
ARC-AGI-1~\citep{chollet2025arc}.
\end{taskcard}

\begin{taskcard}{ARC-AGI-3}
\item \cardlabel{Purpose}Learning an interactive environment without instructions: finding out by acting what the
controls do and what completes a level, then completing levels with few actions.
\item \cardlabel{Source and construction}ARC-AGI-3~\citep{foundation2026arc} consists of turn-based games on grids
of up to $64\times64$ cells with 16 colors, each with several levels and a human baseline action count per level.
Its public set has 25 games, and official scores use 110 further private games. The task uses the 25 public games
at a fixed version of the official toolkit, split by game with a seeded hash into 10 development, 5 validation, and
10 test games. The agent sees only the identifiers of the development games and plays them through the interface;
the game code, the human baselines, and the validation and test games are not accessible to it.
\item \cardlabel{Setting}The solution observes the current frame, the available actions, and the number of completed
levels, and submits actions: directions, a game-specific action, a click at a grid position, and undo, whose effects
must be inferred; a reset is allowed only after the game is lost. Each game allows at most 1000 actions, 120 model
calls, and 1200 seconds, and each evaluation at most 1200 model calls. Games are played one at a time, each from a
fresh start with a fixed seed, and no memory or files carry over between games. The solution may not contain game
solutions or game code.
\item \cardlabel{Starter solution}Sends the model the current frame as an image and as rows of color values, a short
memory, and the recent actions, and executes up to eight actions returned per call along with an updated memory.
\item \cardlabel{Metric}Relative human action efficiency (RHAE), computed by the official toolkit. With $h_{\ell}$
the human baseline and $a_{\ell}$ the agent's actions on level $\ell$, a completed level scores
$S_{\ell}=\min(1.15,\,h_{\ell}/a_{\ell})^{2}$ and an uncompleted level zero; with weights $w_{\ell}=\ell$ over the
$n$ levels of a game of which $k$ are completed,
\[
E=\min\!\left(\frac{\sum_{\ell=1}^{k}w_{\ell}}{\sum_{\ell=1}^{n}w_{\ell}},\;
\frac{\sum_{\ell=1}^{n}w_{\ell}S_{\ell}}{\sum_{\ell=1}^{n}w_{\ell}}\right),
\]
and the score is the mean of $E$ over games, multiplied by 100~\citep{foundation2026arc}. A game stopped by a limit keeps
its partial score.
\item \cardlabel{Prior work}ARC-AGI-3~\citep{foundation2026arc} reports that humans solve all of its environments,
while frontier AI systems scored below 1\% as of March 2026; in its preview competition, the best entry, which trains
a convolutional network with reinforcement learning to predict which actions change the frame, scored 12.58\%.
\end{taskcard}

\begin{taskcard}{Spatial-VSI}
\item \cardlabel{Purpose}Spatial reasoning about indoor scenes from a moving camera: counts, metric sizes and
distances, relative directions, routes, and the order of appearance.
\item \cardlabel{Source and construction}VSI-Bench~\citep{yang2025thinking} contains over 5000 questions on 288 videos
of real rooms from ScanNet, ScanNet++, and ARKitScenes~\citep{dai2017scannet,yeshwanth2023scannet++,baruch2021arkitscenes}.
It has ten question types in eight groups: object counting, absolute distance, object size, and room size, answered
by numbers, and relative distance, relative direction (easy, medium, hard), route planning, and appearance order,
answered by option letters. The scenes of a fixed dataset version are shuffled with a fixed seed and dealt to the
test, validation, and development splits in proportions 40\%, 30\%, and 30\%; each split then keeps questions
stratified so that the eight groups are weighted as in the official score. The splits hold 80 questions from 46
scenes for development, 480 from 84 scenes for validation, and 320 from 100 scenes for testing.
\item \cardlabel{Setting}Given the video (640 by 480 pixels, one to three minutes), the question, and the options,
the solution returns an answer. It may call the model twice per question on average over a split and uses no other
model and no downloaded weights; its own code, such as geometry and classical vision with OpenCV, is allowed. It may
not read the annotations of the source datasets or access the network, has no GPU, and has four hours per evaluation.
\item \cardlabel{Starter solution}Shows the model 16 evenly spaced frames, 512 pixels wide, with the official
VSI-Bench prompt, in one call per question.
\item \cardlabel{Metric}The overall score of VSI-Bench~\citep{yang2025thinking}. A multiple-choice answer is correct
when its first word is the option letter; a numerical answer receives the mean relative accuracy with thresholds
$\{0.50,0.55,\dots,0.95\}$ and the condition $|\hat y-y|/y\le1-\theta$, and zero if it is not a number. Scores are
averaged within each group, the three direction levels are first averaged into one, and the eight groups are
averaged, giving a value between 0 and 1.
\item \cardlabel{Prior work}VSI-Bench finds that the best model it tests, Gemini-1.5 Pro, averages 45.4\%, against
about 79\% for humans on a subset; chain-of-thought, self-consistency, and tree-of-thoughts prompting do not help,
whereas asking the model to draw a cognitive map raises relative-distance accuracy from 46.0\% to
56.0\%~\citep{yang2025thinking}. VSI-Bench-Debiased removes questions that can be answered without the
video~\citep{brown2025benchmark}, and Spatial-MLLM adds a 3D spatial encoder initialized from a visual geometry model
to a multimodal language model~\citep{wu2025spatial}.
\end{taskcard}

\subsection{Creation}

\begin{taskcard}{Image Editing}
\item \cardlabel{Purpose}Whether the agent can build an instruction-based image editor that performs the edit
requested in English or Chinese while leaving the rest of the photograph intact, across local, global, text,
retouching and restoration, reasoning, and compound edits. The editing model is not prescribed: choosing, combining,
and adapting open models is part of the task.
\item \cardlabel{Source and construction}The evaluation set was built for this task in two versions. The first has 100
images; the second, used here, adds 86 images and 14 restoration photographs, for 200 images. The 186 photographs,
paintings, and illustrations are CC0 or public-domain works from Wikimedia Commons, each checked for its licence and
excluding identifiable people, minors, trademarks, and third-party works; two images of puzzles (an arithmetic problem on a
blackboard and a number sequence) were made for the task, and four images carry a reproducible synthetic degradation (blur, color cast, or low contrast) to be undone.
The 14 restoration photographs show haze, fog, smoke, lens flare, reflections on glass, or cast shadows; all come from the
data of UniSER~\citep{zhang2025uniser}, including ten of its 39 in-the-wild evaluation images and three of its paired
images from NH-HAZE~\citep{ancuti2020nh}, FlareReal600, and WSRD+~\citep{vasluianu2023wsrd}. Images are resized to at
most $1536\times1024$ pixels with sides that are multiples of 16, and span outdoor (85), product (34), indoor (32),
low-quality (21), illustration (13), text (11), and document (4) images. Each image has one instruction, written in
Chinese (106) or English (94) together with a translation into the other language, and is assigned to one of 23 kinds
of edit in six families: local semantic edits, namely adding, removing, or replacing an object, changing its
attributes, changing a pose, setting a count, or moving objects (71 images); global edits of style, background, or
lighting, weather, and season (30); text editing, namely replacing, writing, or editing Chinese text (22); retouching
and restoration, namely tone, old or degraded photographs and soft effects, removal of clutter, and outpainting to a new
aspect ratio (34); edits that require inference, such as candles after three hours of burning, the farthest of several
glasses, leaves in late autumn, or a completed sudoku (25); and compound or constrained instructions (18). Each sample
is labeled with a difficulty (44, 114, and 42 at three levels) and the size of the edited region (small, medium,
large, or global), and is annotated for scoring: a mask of the region allowed to change (92 images), objects that must
appear (43) or disappear (32), a target count (11), text that must be readable (24) or gone (14), a target aspect ratio
(5), content that must be kept (74), the parts of a compound instruction (13), and a description of the intended result
of an inference edit (31). The images are divided into 50 development, 70 validation, and 80 test images, stratified by
kind of edit. The development samples come with their annotations, and their evaluations return per-sample judge
rationales and side-by-side comparisons; validation and test return only the aggregate score.
\item \cardlabel{Setting}The solution is an editing program that receives the image, the instruction, and its
language, and writes an image at the input size. It may use any open-weight editing or generation model, several
models at once, vision models, classical image processing, and its own fine-tuning on the development split;
Qwen-Image-2.1, Qwen-Image-Edit-2511 with distilled few-step adapters~\citep{wu2025qwen}, and
HunyuanImage-3.0-Instruct~\citep{cao2025hunyuanimage} are provided. It may not use the judge's model family, remote
services, or the annotations. Evaluation runs offline on four 80\,GB GPUs with, on average, 150\,s per image per GPU;
the solution does not call the model being evaluated.
\item \cardlabel{Starter solution}Qwen-Image-Edit-2509 with an eight-step distilled adapter, given the instruction
unchanged, producing one output per image.
\item \cardlabel{Metric}A frozen judge, GLM-4.5V~\citep{hong2025glm} at temperature 0, sees the source and the output and
rates instruction compliance $\mathrm{SC}$ and preservation $\mathrm{PR}$, and, from the output alone, perceptual
quality $\mathrm{PQ}$, each from 0 to 10. The preservation rubric depends on the kind of edit (for example, only content
and composition for style transfer, only the kept subjects for a background change), and the judge is told the
annotated parts, the kept content, and, for inference edits, the intended result. Global deformation is measured rather
than judged: a SIFT~\citep{lowe2004distinctive} and RANSAC~\citep{fischler1981random} homography from input to output
gives the 90th-percentile displacement $p_{90}$ in pixels at a 512-pixel short side, $g=\exp(-\max(0,p_{90}-0.5)/6)$, and
$\mathrm{PR}'=\max(0,\mathrm{PR}-5(1-g))$. The judge score is
$J=\tfrac{1}{10}\sqrt{\mathrm{SC}\cdot\mathrm{PQ}}\,(0.5+0.05\,\mathrm{PR}')$. The programmatic score $P$ averages the
checks that apply: object presence and absence and the exact count, answered by the judge; for text, one minus the
lowest character error rate over two OCR readings and the judge's transcription, and the absence of text to be removed;
the aspect ratio of outpainted images, $\max(0,1-5|r/r^\ast-1|)$; preservation,
$\mathrm{clip}(1-(\mathrm{MAD}-0.02)/0.10)$ from the mean absolute color difference outside the dilated edit region after
alignment; and detail, $\mathrm{clip}((cs-0.55)/0.40)$ from the contrast--structure term of SSIM~\citep{wang2004image} at a
1024-pixel short side. Without a mask, the edit region is formed from the judge's boxes of the edited objects. A sample
scores $0.5J+0.5P$ ($J$ if no check applies), and 0 if $\mathrm{SC}<3$ or no image is produced. The score is the mean of
the six family means, in $[0,1]$.
\item \cardlabel{Prior work}GEdit-Bench, introduced with the Step1X-Edit editor~\citep{liu2025step1x}, evaluates
instruction-based editing on real user requests with a vision--language judge of semantic consistency and perceptual
quality, which VIEScore~\citep{ku2024viescore} combines as $\sqrt{\mathrm{SC}\cdot\mathrm{PQ}}$; the judge score above
takes this form and adds the preservation term and the programmatic checks. UniSER~\citep{zhang2025uniser} removes soft
effects such as haze, flare, shadows, and reflections with a single diffusion model and evaluates its in-the-wild images,
which lack ground truth, with reference-free metrics and a vision--language judge.
\end{taskcard}

\begin{taskcard}{Multi-turn Editing}
\item \cardlabel{Purpose}Whether the agent can improve a multi-turn editing system, in which each instruction is
applied to the previous result and becomes known only at its turn, so that errors in content and image quality
accumulate over the session.
\item \cardlabel{Source and construction}All 100 sessions of MSE-Bench, the multi-turn editing benchmark introduced with
VINCIE~\citep{qu2026vincie}, at a fixed revision. Each session is a real photograph, used unchanged at its original size
(many are small), and five instructions that add, remove, or replace objects, change attributes, backgrounds, posture,
expression, orientation, position, actions, or interactions, restyle the image, or move the camera (dolly, pan, tilt,
orbit); MSE-Bench labels 20 operation types. The sessions are split into 30 development, 30 validation, and 40 test
sessions by a seeded search that balances the per-turn counts of the 20 operation types, the sessions with a camera
operation, and the images under 512 pixels on the short side. For the development sessions the agent also sees the
MSE-Bench annotations (operation types and descriptions of the intended results); the solution never receives them.
\item \cardlabel{Setting}The solution is a session object that is reset with the source image and returns one image per
instruction; it may keep earlier images and instructions. The editing model is fixed to the original Qwen-Image-Edit
(August 2025)~\citep{wu2025qwen} with its four- and eight-step distilled adapters, since the later 2509 and 2511 releases
degrade into blocky noise when edited repeatedly. Prompt rewriting, sampler settings, candidate selection, masking and
compositing, and Qwen2.5-VL-7B~\citep{bai2025qwen} as a helper are allowed; training the editing model, network access,
and any copy of MSE-Bench or of VINCIE's outputs are not. Each of four GPUs runs one process with 300\,s of start-up and,
on average, 60\,s per turn. A failed turn, an aspect-ratio change above 5\%, or a timeout scores 0 for that turn and the
rest of the session. The solution does not call the model being evaluated.
\item \cardlabel{Starter solution}Edits the previous image with each turn's instruction, using the eight-step adapter
and a fixed seed.
\item \cardlabel{Metric}Each turn $k$ is judged by a frozen GLM-4.5V~\citep{hong2025glm} at about one megapixel. Given
the instructions of turns $1..k$ and the images up to turn $k$, MSE-Bench's own evaluation prompt rates prompt following
$\mathrm{PF}$ and consistency $\mathrm{CONS}$ with the previous image; a strict prompt rates the perceptual quality
$\mathrm{PQ}$ of the turn's image alone (all 0--10). The turn score is
$\tfrac{1}{10}\sqrt{\mathrm{PF}\cdot\mathrm{PQ}}\,(0.5+0.05\,\mathrm{CONS})$, or 0 if $\mathrm{PF}<3$; the score averages
over the five turns and then over sessions. MSE-Bench's cumulative success rate, the fraction of sessions in which every
turn up to $k$ has $\mathrm{PF}>6$ and $\mathrm{CONS}>6$, is reported for $k=1,3,5$.
\item \cardlabel{Prior work}VINCIE~\citep{qu2026vincie} learns in-context image editing from video and introduced
MSE-Bench to measure multi-turn editing by the success rate at each turn, judged by GPT-4o. Here the editing model is
fixed and the agent improves the system around it.
\end{taskcard}

\begin{taskcard}{Video Generation}
\item \cardlabel{Purpose}Whether the agent can build an executable text-to-video workflow around a frozen video
generator, deciding how to plan prompts, spend a fixed generation budget, inspect frames, and select candidates, so that
the videos follow compositional prompts.
\item \cardlabel{Source and construction}The prompts come from T2V-CompBench~\citep{sun2025t2v}, at a fixed revision of
its prompt lists, restricted to four of its seven categories: consistent attribute binding, action binding, generative
numeracy, and spatial relationships (two-dimensional only). Each split has 20 prompts, five per category. Near-duplicate
prompts were grouped into families (connected components of normalized token Jaccard similarity of at least 0.65); the
validation and test prompts were drawn in a salted hash order, excluding the families of development prompts and taking
at most one prompt per family. The agent sees the development prompts and their scores; validation and test prompts are
private.
\item \cardlabel{Setting}The generator is MiniMax H3~\citep{minimax_h3_modelcard} with frozen weights, producing 124
frames at $864\times480$ and 24 frames per second. For each prompt the workflow may make at most three generation
requests totalling 50 denoising steps, choosing prompts, seeds, step counts, and flow shift; sample frames; call the
model being evaluated at most twice (40 calls per evaluation), with images; and process videos locally. It returns one
video. It has no network, no direct GPU access, and no access to the evaluators, and may not add text or cues aimed at
the evaluators. Each evaluation runs on one 80\,GB H100 within 150 minutes; prompts unfinished when the workflow time
runs out score 0.
\item \cardlabel{Starter solution}A single generation request with the prompt unchanged and the default seed, 50
steps, and flow shift.
\item \cardlabel{Metric}The original T2V-CompBench evaluators score each video against the original prompt with their
default prompts, thresholds, and seeds: attribute binding and action binding with the multimodal-LLM evaluation using
LLaVA-v1.6-34B~\citep{liu2024llava} on sampled frames, and numeracy and spatial relationships with the detection-based
evaluation using Grounding DINO~\citep{liu2024grounding} (16 sampled frames, six of them in a grid for the language
model). Each video scores in $[0,1]$, a missing or corrupt video scores
0, and the score is the mean of the four category means. Because of the subset and the fixed resolution, the scores are
not comparable with the T2V-CompBench leaderboard.
\item \cardlabel{Prior work}T2V-CompBench~\citep{sun2025t2v} benchmarks text-to-video models on 1,400 prompts in seven
compositional categories, with multimodal-LLM-based, detection-based, and tracking-based metrics validated against human
judgments. Here the generator is fixed, and the agent improves the workflow around it.
\end{taskcard}

\begin{taskcard}{Text-to-Image}
\item \cardlabel{Purpose}Whether the agent can make a text-to-image program satisfy every constraint of new prompts in
Chinese and English while keeping image quality, either with a fixed generator or with any open model.
\item \cardlabel{Source and construction}The 200 prompts and their rubrics were written for this task: 100 in Chinese and
100 in English (not translations of each other), ten in each of 20 subcategories of six categories: subjects and
attributes (identity, color, material), counting and binding (single and multiple classes, attribute binding, excluded
content), spatial relations and interaction (position, depth and containment, actions and roles), text and layout
(English, Chinese, multiple text blocks), style and presentation (medium, lighting and season, camera and composition),
and reasoning and composition (world knowledge, causality and time, logical quantity, multiple constraints). Each
subcategory has two easy, five medium, and three hard prompts by the author's judgment. Each prompt has a fixed seed and
size ($1024\times1024$ for 145 prompts, $1280\times768$ or $768\times1280$ for 55) and a rubric of atomic criteria (954 in
total) of four types: semantic, count, absence, and exact text, the latter copied verbatim from the prompt; each
requirement is credited once. After authoring, every prompt and rubric was reviewed for physical consistency, count
logic, text literals, coverage, and duplicate credit, which revised 32 prompts, 51 rubrics, and 34 sizes; scripts check
quotas, uniqueness, field consistency, and split isolation, and flagged cross-split pairs of similar wording were
reviewed by hand. The prompts are split into 50 development, 70 validation, and 80 test prompts by semantic group (178
groups of prompts sharing a central scene), each split half Chinese and half English with 20/50/30\% easy, medium, and
hard prompts. Development prompts come with rubrics and per-image feedback; validation returns only the score.
\item \cardlabel{Setting}The solution receives a prompt, its language, size, and seed, and writes one image. With the
fixed generator, all images come from the frozen Qwen-Image-2.1~\citep{wu2025qwen}; prompt rewriting, layout planning,
sampling settings, several candidates with selection, Qwen2.5-VL-7B~\citep{bai2025qwen} as a helper, and ordinary image
tools are allowed, but no training or adapters. With any model, other open-weight generators, routing, and training on
the development split are also allowed. Neither may use the judge's model family, the rubrics, the network, or hosted
models at evaluation, which runs on four GPUs with at most 600\,s per image.
\item \cardlabel{Starter solution}One image from Qwen-Image-2.1 with the prompt unchanged and 40 sampling steps.
\item \cardlabel{Metric}A frozen GLM-4.5V~\citep{hong2025glm} judges each image: a semantic criterion scores 0, 0.25,
0.5, 0.75, or 1; a count, absence, or exact-text criterion scores 0 or 1; and the quality $Q\in[0,1]$ rates rendering
defects, excluding prompt compliance and intended style. With criterion scores $c_i$ and weights $w_i$ (all 1),
$C=\sum_i w_ic_i/\sum_i w_i$ and the image scores $C\,(0.5+0.5\,Q)$; a missing or wrongly sized image scores 0. Scores are
averaged within each category and then over the six categories. The reported score is the mean of the test scores of the
two settings.
\item \cardlabel{Prior work}GenEval~\citep{ghosh2023geneval} and T2I-CompBench~\citep{huang2023t2i} evaluate
compositional text-to-image generation with detectors and other vision models. The prompts here are new and these
benchmarks' scores are not reproduced.
\end{taskcard}

\subsection{Reconstruction}

\begin{taskcard}{3D Tracking}
\item \cardlabel{Purpose}Whether the agent can assemble a system that segments every object in a few widely separated
views of an indoor scene and gives each object the same identity in every view, from RGB images alone. Without
camera poses or depth, the views must be associated through appearance and implicit geometry: the cross-view
instance tracking that IGGT~\citep{li2025iggt} introduced as an evaluation of instance-grounded 3D reconstruction.
\item \cardlabel{Source and construction}The validation and test clips are the 3D tracking benchmark released with IGGT,
which samples a few views per scene from ScanNet~\citep{dai2017scannet} and ScanNet++~\citep{yeshwanth2023scannet++}
for spatial coverage with sufficient overlap and annotates a subset of objects by hand with identities consistent
across views. Each clip has six to nine RGB views, resized to $592\times448$ pixels, typically with four to eight
annotated objects such as furniture, lamps, and decorations; walls, floors, and ceilings are not annotated. The 8
ScanNet++ clips form the validation split and the 10 ScanNet clips the test split. The development split consists of 100 clips built
from ScanNet++ training scenes that share no scene with the validation clips: eight views are drawn from a random
stretch covering 30 to 80\% of a capture, and up to eight object-like instances are kept (visible in at least two
views, at most a quarter of any view, and at least 0.1\% of the clip's pixels); a clip with fewer than four is drawn
again. The agent sees the development views with their instance maps.
\item \cardlabel{Setting}The solution is a program that receives the views of one clip and returns one instance map
per view, with 0 for the background and identities consistent across views. It may use any publicly released
pretrained model or library and needs no training; it does not call the model being evaluated. It runs on one GPU
without network access, and a whole split must finish within 45 minutes.
\item \cardlabel{Starter solution}Clusters the colors of each view into six clusters with $k$-means, splits them
into connected regions covering 0.2 to 20\% of the image, and links each region to the existing track with the
nearest mean color, or starts a new track.
\item \cardlabel{Metric}Evaluation is class-agnostic. Predicted and annotated tracks are matched one to one, greedily in
decreasing order of their temporal IoU, which accumulates intersection and union over the views: for an annotated
object with masks $G_f$ and its matched prediction $P_f$ in views $f=1,\dots,S$,
\[
\mathrm{T\text{-}IoU}=\frac{\sum_{f}|P_f\cap G_f|}{\sum_{f}|P_f\cup G_f|}.
\]
T-mIoU is the mean of T-IoU over the annotated objects of a clip, averaged over clips (0 to 1). An annotated object
without a match scores 0, and predicted tracks without a match are not penalised. T-SR is the share of annotated
objects whose per-view IoU exceeds 0.5 in every view in which they appear. The region similarity $\mathcal{J}$ (mean
per-view IoU), the boundary measure $\mathcal{F}$ (tolerance 0.8\% of the image diagonal), and
$\mathcal{J}\&\mathcal{F}$ of DAVIS~\citep{perazzi2016benchmark} are also reported. IGGT named T-mIoU and T-SR but
defined them per view, averaging the IoU over views and counting an object as tracked when its prediction is
non-empty in every view; the accumulated IoU, the IoU threshold of T-SR, and the greedy matching follow the
evaluation of S4VY~\citep{zhang2026s4vy}.
\item \cardlabel{Prior work}IGGT reconstructs geometry and predicts 3D-consistent instance features with one
transformer, trained with a 3D-consistent contrastive objective; IGGT4D~\citep{zou2026iggt4d} extends it to streaming
4D input; S4VY~\citep{zhang2026s4vy} adds persistent object queries to a feed-forward 4D geometry model. On the ten
ScanNet clips under this metric, S4VY reports a T-mIoU of 0.803, against 0.781 for IGGT4D, 0.635 for SAM~2
\citep{ravi2024sam2} with incremental object discovery, and 0.517 for IGGT.
\end{taskcard}

\begin{taskcard}{Scene from Photo}
\item \cardlabel{Purpose}Whether the agent can build a single-image 3D scene reconstruction system: from one photograph
of a furnished room, recover the camera, the room, and every piece of furniture as a complete object with plausible
unseen sides, with physically based materials and lighting, so that a path tracer reproduces the photograph.
\item \cardlabel{Source and construction}The photographs are those of HARMONY300~\citep{sun2026harmony}: 100 renders of
3D-FRONT rooms~\citep{fu20213d}, which have ground-truth geometry, and 200 real photographs, half of moderate and half
of high complexity. Each difficulty is split separately with a fixed seed (the renders stratified by their number of
annotated objects) into 10, 10, and 20 photographs for development, validation, and test, giving 30, 30, and 60; the
remaining 179 photographs are held out of every split, and one render whose image differs from its ground truth is
excluded. On the renders the annotated furniture and its meshes come from the 3D-FRONT ground truth. On the real
photographs up to eight pieces of furniture per photograph are annotated by Mask2Former~\citep{cheng2022masked}
panoptic segments of furniture classes (score at least 0.5, at least 0.5\% of the image); on the renders this
procedure recovers 85\% of the annotated objects. The agent sees the development photographs with all labels.
\item \cardlabel{Setting}The solution writes a Blender scene per photograph within 120\,s per photograph; the evaluator
renders it with Cycles from the scene's own camera under fixed rules (only meshes, physically based shader nodes,
non-emissive materials, at most 64 lights). It may use any public pretrained model on one GPU, but not HARMONY300,
3D-FRONT, 3D-FUTURE, or HARMONY's code, and does not call the model being evaluated.
\item \cardlabel{Starter solution}An empty room box in flat colors matched to the photograph, one ceiling light, and
the same guessed camera with a 60\textdegree{} field of view for every photograph.
\item \cardlabel{Metric}$s=(A+O+G)/3$ in $[0,1]$, with $A$ and $O$ averaged over all photographs and $G$ over the
renders. Appearance
$A=\tfrac13[\max(0,1-\mathrm{NCLIP}/0.25)+\max(0,1-\mathrm{PL}/0.1)+1-\mathrm{LPIPS}]$, where N-CLIP
(one minus the cosine similarity of CLIP ViT-B/32 embeddings~\citep{radford2021learning}) and the photometric loss PL
are HARMONY's metrics and LPIPS is that of \citet{zhang2018unreasonable}. Objects
$O=\mathrm{mean}_k\, r_k\min(1,\mathrm{IoU}_k/0.5)$, where $\mathrm{IoU}_k$ compares the meshes assigned to object $k$
with its mask and the realness $r_k=\mathrm{clip}((\min_j P_j-0.5)/0.4,0,1)$; $P_j$ is the probability that
Qwen3-VL-8B~\citep{bai2025qwen3vl}, shown untextured views from four directions, judges the reconstruction a more
real object than shortcut $j$ (its visible surface as a sheet, that sheet extruded to a prism, or its bounding box).
Geometry $G=(F_{0.1}+F_{0.01}+F^{\mathrm{obj}}+F^{\mathrm{obj}}_{\mathrm{hidden}})/400$, with HARMONY's normalized
scene F-scores in the protocol of Gen3DSR~\citep{ardelean2025gen3dsr} (furniture only, both point sets normalized by
their bounding box and aligned by ICP, thresholds 0.1 and 0.01) and per-object F-scores at 5\,cm, in percent, after
scaling by the median ratio of true to rendered depth; $F^{\mathrm{obj}}_{\mathrm{hidden}}$ counts recall only on the
parts of the object the photograph does not show. A missing or unloadable scene, or one with more than 4 million
triangles, renders black and has no geometry. For calibration, the true meshes textured with the photograph score
0.879 on the development renders, and HARMONY's released reconstructions score 0.422 on the test split.
\item \cardlabel{Prior work}HARMONY reconstructs such scenes by hierarchical agentic reasoning with a vision-language
model and was compared with Gen3DSR, CAST~\citep{yao2025cast}, and SAM~3D~\citep{chen2025sam}, among others.
\end{taskcard}

\begin{taskcard}{Scene from Video}
\item \cardlabel{Purpose}Whether the agent can build a video-to-simulation pipeline: from the frames of an indoor
video alone, estimate the cameras and reconstruct a scene of separate textured objects on a fixed background that
has the true shape, reproduces the frames, and remains at rest under physical simulation.
\item \cardlabel{Source and construction}The scenes are the nine released with HoloScene~\citep{xia2026holoscene}, each
with frames, ground-truth cameras, per-object instance masks, a support graph, and a ground-truth surface. The test
split is the six scenes evaluated by CoDimRecon~\citep{xie2026codimrecon}: three Replica rooms~\citep{straub2019replica}
and three ScanNet++ captures~\citep{yeshwanth2023scannet++}. Validation is one iGibson house~\citep{shen2021igibson};
development is another iGibson house and a game room filmed with a phone (undistorted). The iGibson and phone
sequences keep every third frame. Each face of the ground-truth surface is labeled by the masks that see it, and each
object is marked dynamic if it rests on the floor or on another object and static if it hangs from a wall or the
ceiling. The agent sees the development frames with all labels and the scoring code.
\item \cardlabel{Setting}The input is only the RGB frames, without cameras, masks, or depth, as in CoDimRecon. The
solution writes a glTF scene (at most 4 million triangles) and a camera for every frame, within 1200\,s per scene, on
one GPU and 16 CPU cores. The scene is placed in the ground-truth frame by the similarity transform that aligns its
cameras with the true ones, rendered unlit from the true cameras, and each mesh node is assigned to the object or
background whose masks it covers most. Any public pretrained model may be used, but not the HoloScene release or its
source datasets; the model being evaluated is not called.
\item \cardlabel{Starter solution}An empty room box in the mean color of the frames, with every camera at one point.
\item \cardlabel{Metric}Per scene, $s=(G+A+P)/3$ with $G=(\mathrm{F1}_{5\,\mathrm{cm}}+\mathrm{NC})/200$ on 500{,}000
points of the whole scene, $A=(\mathrm{SSIM}+1-\mathrm{LPIPS})/2$ over all input frames~\citep{wang2004image,
zhang2018unreasonable}, and $P$ the share of dynamic objects that are reconstructed (mask IoU at least 0.5) and stay
in place; $s$ is averaged over scenes. Stability follows CoDimRecon: each dynamic object, decomposed into convex hulls
by CoACD~\citep{wei2022approximate}, is released alone in MuJoCo~\citep{todorov2012mujoco} for 3\,s with all else fixed,
and falls if it tilts more than 30\textdegree{} or moves more than $\max(5\,\mathrm{cm}, 0.2\,d)$, with $d$ its
bounding-box diagonal. A missing or unloadable scene scores as empty. The ground-truth surface colored from the frames
scores 0.936 on the test split.
\item \cardlabel{Prior work}HoloScene reconstructs simulation-ready interactive scenes from a video given its cameras
and instance masks. CoDimRecon reconstructs such scenes agentically from RGB views alone, representing objects as
curves, surfaces, and volumes, and compares with HoloScene and an agent that writes the scene in Blender.
\end{taskcard}

\subsection{Perception}

\begin{taskcard}{Visual Search}
\item \cardlabel{Purpose}Whether the agent can build a visual search procedure around a model with a limited field of
view: questions concern small objects in photographs of 2K to 8K resolution, while each model call may contain at
most one megapixel, so the solution must decide where to look.
\item \cardlabel{Source and construction}The questions come from V*Bench~\citep{wu2024v} (191 multiple-choice
questions on about 2K photographs: 115 on object attributes with four options, 76 on relative position with two) and
HR-Bench~8K~\citep{wang2025divide} (200 questions with four options on 151 photographs of about 8K resolution, half on
a single instance and half across instances; of its four rotated listings of each question one is drawn with a fixed
seed). V*Bench is split by category and HR-Bench by photograph, so no photograph appears in two splits. The test
split has 140 questions (70 from each benchmark). Development and validation are stratified subsets of 70 questions
(33 and 37) and 80 questions (40 and 40). The agent sees the development photographs, questions, answers, and, for
V*Bench, the boxes of the queried objects.
\item \cardlabel{Setting}The solution answers each question with an option letter. It may call the model being
evaluated on average at most five times per question, with at most $1024\times1024$ pixels per call in total; it may
crop, resize, and run open-weight models on one GPU, and may not use either benchmark beyond the development split.
\item \cardlabel{Starter solution}Shows the model the whole photograph reduced to one megapixel with the question, in a
single call.
\item \cardlabel{Metric}Accuracy over the questions of the split; a question without a valid letter counts as wrong.
Accuracies on V*Bench and HR-Bench~8K are reported separately. Chance accuracy on the validation split is 0.303.
\item \cardlabel{Prior work}V*Bench was introduced with SEAL, which guides a multimodal model to search the image for
the queried region, and HR-Bench with a training-free method that divides the image into patches and combines their
answers. Later methods explore zoomed regions as a tree (ZoomEye~\citep{shen2024zoomeye}) or learn to crop by
reinforcement learning (DeepEyes~\citep{zheng2025deepeyes}).
\end{taskcard}

\begin{taskcard}{Video Quality}
\item \cardlabel{Purpose}Whether the agent can turn a general multimodal model into a judge of perceptual video quality
that agrees with people, both for real user videos and for generated videos.
\item \cardlabel{Source and construction}The rated videos are 100 videos of YouTube-UGC~\citep{wang2019youtube}, chosen
from its 1{,}380 videos with mean opinion scores (MOS) excluding 4K: seven per content category for 14 categories
(eight for sports and gaming), evenly spaced in MOS within each category. Each lasts 16.6 to 20\,s at 360p to 1080p,
and its MOS (1 to 5) averages 105 to 154 crowdsourced ratings from the subjective study of
\citet{yim2020subjective}. The pairs are 50 pairs from a public human-preference study of Veo~3.1~\citep{rapidata_veo31_preferences},
each pitting Veo~3.1 against one of seven other text-to-video models on the same prompt (5 to 6\,s clips). Each pair
has 15 reliability-weighted crowd votes on three questions: prompt alignment, fewer visual artifacts, and overall
preference, the last two judged without the prompt. A clip with more than 55\% of the votes wins; otherwise the
question is a tie. The videos are dealt in order of category and MOS, and the pairs in order of opposing model, into
30, 30, and 40 videos and 15, 15, and 20 pairs; the clip order in each pair is randomized. The agent sees the
development videos, frames at two per second, and labels with vote shares and rating spreads.
\item \cardlabel{Setting}The solution scores each video and answers the three questions for each pair. It may call the
model being evaluated on average at most three times per item, with at most eight $1024\times576$ frames' worth of
pixels per call, and may run open-weight models on one GPU; it may not use the source datasets.
\item \cardlabel{Starter solution}Asks for a rating from 1 to 5 given eight evenly spaced frames, and answers all three
questions about a pair given six frames of each clip.
\item \cardlabel{Metric}$s=(\mathrm{SRCC}+\mathrm{Acc})/2$, where SRCC is Spearman's rank correlation~\citep{spearman1904proof}
between the solution's scores and the MOS over the split's videos (a video without a score receives the median of the
others), and Acc is the share of non-tied pair questions answered correctly, pooled over the three criteria (a missing
answer is wrong). PLCC and per-criterion accuracies are also reported. Chance is 0.25.
\item \cardlabel{Prior work}YouTube-UGC was collected for video compression research, and its MOS serve to evaluate
blind quality models such as VIDEVAL~\citep{tu2021ugc} and DOVER~\citep{wu2023exploring}.
\end{taskcard}

\subsection{Infrastructure}

\begin{taskcard}{VAE Decode}
\item \cardlabel{Purpose}Whether the agent can lower the latency of a fixed numerical computation, the video decoder of a
generative model, without changing what it computes.
\item \cardlabel{Source and construction}The computation is the decoder of the MiniMax H3 autoencoder~\citep{minimax_h3_modelcard},
a 36-layer vision transformer of about 2.4 billion parameters with FP32 weights run under FP16 autocast. It decodes a
window of seven latent frames of 24 channels into 28 RGB frames, in $256\times256$ tiles with at least 64 pixels of
overlap and linear blending; the reference implementation and weights are pinned by hash. Inputs are latent windows of
encoded videos at two sizes, $544\times960$ and $480\times832$ pixels, split by scene: 6 development, 6 validation, and
12 test scenes. Each request is either a real window with fresh noise ($+0.05\,\mathcal{N}(0,1)$) or a synthetic stress
window (Gaussian latents scaled per channel, clipped at $\pm5$). Per size and repetition there are four real and one
stress request on development (two repetitions) and validation (six), and eight and two on test (six), so 20, 60, and
120 requests; no input repeats. The development windows are visible.
\item \cardlabel{Setting}The solution is a decoder that receives latents in host memory and writes the full output into a
host buffer. Custom Triton or CUDA kernels built from source, fusions, CUDA graphs, compilation, batching of tiles,
and overlapping transfers are allowed. Lower precision, approximate or sparse attention, skipped tokens, tiles, or
frames, reuse across requests, special treatment of stress windows, more than one GPU, shipped binaries, and existing
optimized implementations of this decoder are not; accepted versions are reviewed for these rules. One H200, 90\,s of
setup, 30\,s per request; the solution does not call the model being evaluated.
\item \cardlabel{Starter solution}A direct eager PyTorch translation of the reference, bit-exact with it.
\item \cardlabel{Metric}Every frame must be finite with a per-frame RMSE below $6.3\times10^{-4}$ ($1.9\times10^{-3}$ for
stress windows) and a maximum absolute error below 0.12 relative to the reference; these bounds are twice the largest
error of legal reorderings, while FP16 accumulation, BF16, FP8, or dropped tokens exceed them. Fresh processes of the
solution and of the starter serve the same requests side by side, timed from input to copied output. With cost
$C=\bar t+t_{\mathrm{setup}}/1024$ ($\bar t$ the mean latency), a repetition scores
$0.75\log_2(C_{\mathrm{starter}}/C)_{544\times960}+0.25\log_2(C_{\mathrm{starter}}/C)_{480\times832}$; $s$ is the mean
over repetitions without the highest and lowest. One gate failure, crash, or timeout, or a median cost or 95th-percentile
latency above 1.5 times the starter's, gives $s=-1$. The reported speedup is $2^{s}$.
\item \cardlabel{Prior work}The serving frameworks vLLM-Omni and SGLang~\citep{zheng2024sglang} include optimized
implementations of the H3 decoder; these and code derived from them may not be used.
\end{taskcard}

\section{Protocol and harness}
\label{app:protocol}

This appendix specifies the parts of the protocol that \cref{sec:method-harness,sec:protocol} describe only in
outline: the structure of a round, the capabilities of the executor and the improver, the acceptance rule, the
budget and the conditions under which a run ends, and the measures that keep the evaluation intact. The rules are
identical for all models and tasks; the budget and two settings of the executor phase are fixed per task and
listed in \cref{tab:budgets}.

\subsection{Round structure}
\label{app:round}

A run begins by scoring the starter solution on the validation split, which gives the score against which the first
candidate is compared. Each round then proceeds in five steps: an executor phase that produces a candidate solution;
a check that reverts every change the executor made outside the files it may modify; the evaluation of the candidate
on the validation split; the acceptance decision of \cref{eq:accept}; and an improver phase that revises the
harness.

The executor phase normally ends when the executor declares its work finished, and the solution in its final state
becomes the candidate. It ends earlier when one of the executor's evaluations on the development split exceeds the
development score of the accepted solution by a fixed share of the distance between the starter's validation score
and the validation score of a strong known solution (\cref{tab:budgets}); the solution as it was in that evaluation
is then submitted. If the phase is interrupted before the executor finishes, because it reached its call limit, its
time limit, or the budget, the best state that the executor's development evaluations showed to exceed the accepted
solution is submitted, and nothing is submitted if there is none. A round that leaves the solution unchanged does
not use a validation evaluation.

\subsection{Executor}
\label{app:executor}

The executor works in a copy of the run's repository. Its tools allow it to run shell commands; to read, write, and
edit files; to view images; to run experiments on the compute allocated to the task, including GPUs where the task
provides them; to score the solution, or several variants of it at once, on the development split; and to query the
remaining budget. It may write only to the solution, to a notes file that persists across rounds and is shared with
the improver, and to a scratch directory that is emptied before each round. It observes development scores directly
and the validation scores of earlier rounds through the run records, but it has no access to the validation or test
data or to the evaluator. The starting harness allows 60 model calls per executor phase. Independently of the
harness, the protocol stops a phase after the number of model calls given in \cref{tab:budgets} or after 180
minutes (600 minutes on the ARC tasks), and an experiment after 240 minutes.

\subsection{Improver}
\label{app:improver}

After each round, the improver receives the executor's transcript, its experiments and notes, and the scores of all
rounds so far. It may modify the executor's prompts, settings, control loop, and tools, as well as its own prompts,
settings, acceptance rule, and a file of lessons that it maintains across rounds. A single improver phase may change
at most 400 lines of the harness; a larger change is reverted in full. Every four rounds, the improver additionally
reviews its own process: which kinds of harness changes were followed by accepted candidates and which by stalls,
whether the acceptance rule is too strict or too permissive, and which lessons to record. Changes to the harness take
effect from the following round. In the frozen-harness runs of \cref{tab:ablation-harness}, the improver is not run.

\subsection{Acceptance rule}
\label{app:accept}

The margin $\delta$ in \cref{eq:accept} is determined by the acceptance rule, which is part of the improver's
harness. In the starting harness, a candidate is accepted if its validation score exceeds that of the accepted
solution by more than $2\sigma$, where $\sigma$ is the standard deviation of the starter's validation score over
repeated evaluations (\cref{tab:budgets}). On tasks for which $\sigma$ was not measured, it is set to zero, and any
improvement on the validation split is accepted. Because the rule belongs to the harness, the improver may revise
it. If the rule fails to return a decision, and in every run with a frozen harness, the starting rule
$\delta=2\sigma$ is applied.

\subsection{Budget and termination}
\label{app:budget}

A run ends at the first of the following events: the wall-clock time, GPU hours, tokens (input, including cached
input, and output), or cost in dollars reach the task's limit; the number of validation evaluations reaches its
limit; the round cap is reached; or five consecutive rounds produce neither a candidate nor a change to the harness,
a rule that is not applied on LIBERO-MAX and LIBERO-PRO. The budget is checked before each model call, experiment,
and development evaluation and is charged when each of them ends, so that calls running concurrently can exceed a
limit by at most their combined cost. A submitted candidate is evaluated on the validation split even when the
budget is exhausted. Model calls that the evaluator makes on the validation and test splits, for tasks whose
solution calls the model, are recorded but not charged. Periods during which the model provider is unavailable are
not charged either, and a phase interrupted by such an outage is repeated. After the run, the final solution and
every accepted solution are scored on the test split (\cref{sec:protocol}).

\Cref{tab:budgets} gives the limits of each task. On a given task, all models have the same limits, with the
following exceptions. Gemini~3.8 Flash and Gemini~4 Argon are stopped after 80 model calls per executor phase on
Active Search, Ego-Exo Memory, 3D Tracking, Visual Search, and Video Quality. On Active Search, the dollar limit for
open-weight models is 250. On RoboLab, the runs of Claude Sonnet~5.5, GPT-5.6 Sol, and GPT-6.1 Sol have 240 GPU
hours. On Image Editing, the runs of Claude Opus~5.5, Claude Sonnet~5.5, GPT-6.1 Sol, and both Gemini models have 60
wall-clock hours and a dollar limit of 150. On Kinematics and Mass, the wall-clock limit of every model other than
Claude Opus~5.5 is the time used by the run of Claude Opus~5.5 (11.3 and 11.0 hours). On Embodied-SWE and
Spatial-VSI, Qwen3.8-Flash-Next is limited to the tokens used by the run of Claude Opus~5.5 (22M and 21M). On
3D Tracking, Claude Sonnet~5.5 has a dollar limit of 50. On RoboCasa, Muse Spark~1.3 has 48 GPU hours and 80M
tokens. On RoboDojo and RoboCasa, runs that continued past round 10 are scored with the solution they had after
round 10.

\begin{table}[!htb]
  \centering\footnotesize
  \caption{\textbf{Budget and run settings of each task.} The limits of one run, as recorded by the runs of
  \cref{sec:main-results} and \cref{tab:res-small}; for RoboDojo and RoboCasa they apply to each of their tasks.
  Wall: wall-clock hours. GPU: GPU hours. Tokens: input tokens, including cached input, and output tokens, in millions. Dollars: cost limit, which
  applies only to models with a price. Val.: validation evaluations. Rounds: the round cap. Calls: model calls after which an executor phase
  is stopped. Submit: the gain on the development split, as a share of the distance between the starter's
  validation score and that of a strong known solution, at which an executor phase ends and submits its candidate
  (--: only when the executor finishes). $\sigma$: the standard deviation of the starter's validation score, on the
  scale of the result tables (VAE Decode: $\log_2$ speedup; Text-to-Image: fixed-generator setting, then any-model setting); 0 where it was not measured. Settings that differ for
  some models are listed in the text.}
  \label{tab:budgets}
  \setlength{\tabcolsep}{3.2pt}
  \begin{fittabular}{@{}l rrrrrr rr r@{}}
    \toprule
    & \multicolumn{6}{c}{\colhead{Limits}} & \multicolumn{2}{c}{\colhead{Executor phase}} & \\
    \cmidrule(lr){2-7}\cmidrule(lr){8-9}
    \colhead{Task} & \colhead{Wall} & \colhead{GPU} & \colhead{Tokens} & \colhead{Dollars} & \colhead{Val.} & \colhead{Rounds}
      & \colhead{Calls} & \colhead{Submit} & \colhead{$\sigma$} \\
    \midrule
    RoboDojo           & 24   & 256 & 150 & 300 & 20 & 10  & 80 & 2\%   & 0 \\
    RoboCasa           & 24   & 256 & 150 & 300 & 20 & 10  & 80 & 2\%   & 0 \\
    Active Search      & 10   & 12  & 80  & 120 & 8  & 15  & 40 & 2\%   & 0 \\
    RoboLab            & 36   & 120 & 150 & 800 & 20 & 10  & 80 & 2\%   & 7.0 \\
    LIBERO-MAX         & 72   & 256 & 150 & 300 & 16 & 15  & 80 & 2\%   & 0 \\
    LIBERO-PRO         & 72   & 256 & 150 & 300 & 16 & 15  & 80 & 2\%   & 0 \\
    Embodied-SWE       & 12   & 12  & 120 & 150 & 10 & 10  & 60 & 5\%   & 2.0 \\
    \midrule
    Ego-Exo Memory     & 10   & 12  & 80  & 80  & 5  & 10  & 40 & 5\%   & 2.0 \\
    Kinematics         & 68   & 272 & 400 & 600 & 20 & 60  & 80 & 2\%   & 3.7 \\
    Mass               & 68   & 272 & 400 & 600 & 20 & 60  & 80 & 2\%   & 1.6 \\
    ARC-AGI-1, -2, -3  & 10   & 0   & 30  & 100 & 40 & 10  & 80 & --    & 0 \\
    Spatial-VSI        & 10   & 0   & 80  & 100 & 10 & 10  & 40 & 5\%   & 1.5 \\
    \midrule
    Image Editing      & 24   & 256 & 150 & 300 & 20 & 15  & 80 & 100\% & 0.48 \\
    Multi-turn Editing & 12   & 64  & 80  & 150 & 30 & 15  & 80 & 100\% & 0.55 \\
    Video Generation   & 12   & 64  & 80  & 10  & 30 & 200 & 80 & 2\%   & 0.013 \\
    Text-to-Image      & 24   & 256 & 150 & 300 & 20 & 25  & 80 & 2\%   & 0.20, 0.02 \\
    \midrule
    3D Tracking        & 10   & 12  & 80  & 100 & 15 & 15  & 40 & 2\%   & 0 \\
    Scene from Photo   & 12   & 36  & 80  & 100 & 10 & 15  & 40 & 2\%   & 0 \\
    Scene from Video   & 12   & 36  & 80  & 100 & 10 & 15  & 40 & 2\%   & 0 \\
    \midrule
    Visual Search      & 10   & 12  & 80  & 100 & 10 & 10  & 40 & 5\%   & 2.0 \\
    Video Quality      & 10   & 12  & 80  & 80  & 5  & 10  & 40 & 5\%   & 2.0 \\
    \midrule
    VAE Decode         & 12   & 16  & 200 & 400 & 60 & 200 & 80 & 5\%   & 0.019 \\
    \bottomrule
  \end{fittabular}
\end{table}

\subsection{Isolation and integrity}
\label{app:integrity}

After every phase, any change outside the paths that the phase may modify, including changes made through shell
commands, is reverted and recorded as a violation; the evaluator, the data splits, and the protocol therefore cannot
be altered by the agent. Files larger than 5\,MB are excluded from the run's repository. If a phase crashes in two
consecutive rounds, the harness is restored to the last version that completed a round without a crash. When a phase
ends, all processes it started are terminated, so that no process left running can modify the solution afterwards.
All model calls, experiments, and development evaluations are routed through the benchmark, which charges them to
the budget and records them.

\FloatBarrier
\section{Full results}
\label{app:results}
\Cref{tab:res-robodojo} to \cref{tab:res-vq} give the results of every run, one table per task in the order of
\cref{tab:tasks}: the test metrics of the final solution, the test score of the starter solution where it was
measured, the validation score at the start and the best validation score reached, and the number of rounds with
the condition that ended the run. \Cref{tab:res-small} reports, in the same layout, the four tasks on which only
two models have been evaluated so far. \Cref{tab:leaderboard} gives the average rank of every model over the tasks of the per-task tables, overall and per area.

\begin{table}[h]
  \centering\footnotesize
  \caption{\textbf{Leaderboard.} Average rank over the tasks of the per-task tables that a model was run on, overall and per area (lower is better); models run on at least five of these tasks. Tasks: the number of these tasks the model was run on. Wins: the number of them on which it has the highest test score.}
  \label{tab:leaderboard}
  \begin{fittabular}{@{}ll c ccccc cc}
    \toprule
    \multicolumn{2}{@{}l}{} & & \multicolumn{5}{c}{\colhead{Average rank by area}} & & \\
    \cmidrule(lr){4-8}
    \colhead{Model} & \colhead{Weights} & \phead{Overall} & \colhead{Action} & \colhead{Reasoning} & \colhead{Creation} & \colhead{Reconstruction} & \colhead{Perception} & \colhead{Tasks} & \colhead{Wins} \\
    \midrule
    \mlogo{claude}Claude Opus 5.5 & \wclosed & \pbest{1.6} & \best{1.5} & 1.8 & 1.7 & \best{1.0} & \best{2.0} & 20 & \best{11}\\\rowrule
    \mlogo{gemini}Gemini 4 Argon & \wclosed & \pcol{2.0} & 1.7 & 2.0 & \best{1.0} & 3.0 & 3.0 & 9 & 4\\\rowrule
    \mlogo{gpt}GPT-6.1 Sol & \wclosed & \pcol{2.8} & 2.8 & \best{1.8} & 4.0 & 2.0 & 3.0 & 16 & 2\\\rowrule
    \mlogo{gpt}GPT-6 Astra & \wclosed & \pcol{3.8} & 3.5 & 2.8 & 6.5 & 4.3 & 3.0 & 15 & 2\\\rowrule
    \mlogo{qwen}Qwen3.8-Flash-Next & \wopen & \pcol{5.0} & 5.2 & 5.2 & 3.7 & 4.0 & 7.0 & 20 & 1\\\rowrule
    \mlogo{qwen}Qwen3.8-27B & \wopen & \pcol{5.8} & 6.0 & 5.7 & 4.7 & 6.3 & 6.5 & 19 & 0\\\rowrule
    \mlogo{claude}Claude Sonnet 5.5 & \wclosed & \pcol{6.4} & 6.0 & 2.5 & 7.0 & 11.0 & 9.0 & 8 & 1\\\rowrule
    \mlogo{gemini}Gemini 3.8 Flash & \wclosed & \pcol{6.5} & 6.0 & 6.0 & 9.0 & 8.0 & 5.5 & 8 & 0\\\rowrule
    \mlogo{deepseek}DeepSeek-V4.1 Flash & \wopen & \pcol{6.7} & 8.0 & 6.0 & 8.0 & 5.0 & 6.5 & 11 & 0\\\rowrule
    \mlogo{glm}GLM-5.3 Flash & \wopen & \pcol{7.4} & 6.0 & 7.7 & 9.5 & 4.0 & 8.0 & 10 & 0\\\rowrule
    \mlogo{gemma}Gemma-4-31B-it & \wopen & \pcol{7.5} & 6.4 & 7.8 & 6.7 & 8.0 & 10.5 & 17 & 0\\\rowrule
    \mlogo{minimax}MiniMax-M3 & \wopen & \pcol{8.7} & 9.0 & 9.0 & 7.5 & 7.0 & 11.5 & 13 & 0 \\
    \bottomrule
  \end{fittabular}
\end{table}

\begin{table}[h]
  \centering\footnotesize
  \caption{\textbf{Tasks evaluated with two models.} Embodied-SWE, Spatial-VSI, Text-to-Image, and VAE Decode, in the layout of the per-task tables of this appendix. Score is the primary test metric of each task and Starter the primary test score of the starter solution; Validation gives the validation score of the starter and the best validation score reached during the run. Embodied-SWE: a robot arm in simulation must flip a slice of bread with a spatula and serve it onto a plate; the agent writes the robot program, and Score is the EmbodiedSWE-Bench rubric score, which credits each trajectory by the furthest stage of the task it reaches. Spatial-VSI: spatial questions about indoor videos, answered by a solver that calls the model; Score is the official VSI-Bench overall score. Text-to-Image: text-to-image generation on original Chinese and English prompts in two settings, one with the generator fixed to Qwen-Image-2.1 and one with any open-weight model or workflow; each image is scored by constraint completion $C$ times image quality $Q$ as rated by a frozen vision-language judge, $C(0.5+0.5Q)$, and Score, Starter, and Validation are the means over the two settings (test score of GPT-6 Astra 94.5 with the fixed generator and 93.4 with any model; of GPT-5.6 Sol 92.0 and 95.8), with Rounds counted per setting. VAE Decode: making the decoder of a video model faster without changing its output; Score is the speedup of the decoder over the starter on unseen inputs, and the starter has a speedup of 1.00$\times$. On Embodied-SWE and Spatial-VSI, Qwen3.8-Flash-Next is served on our own GPUs and has no price, so its run is limited to the number of tokens used by the run of Claude Opus~5.5 (22M on Embodied-SWE and 21M on Spatial-VSI). On Text-to-Image, a pre-release version of the task, both models run at medium reasoning effort with at most 25 rounds in each setting. Scores other than the speedup are multiplied by 100.}
  \label{tab:res-small}
  \setlength{\tabcolsep}{3pt}
  \begin{fittabular}{@{}lrllrrrrl@{}}
    \toprule
    & \multicolumn{3}{@{}l}{\colhead{Model}} & \multicolumn{2}{c}{\colhead{Hidden test}} & \colhead{Validation} & \multicolumn{2}{c}{\colhead{Run}} \\
    \cmidrule(r){2-4}\cmidrule(lr){5-6}\cmidrule(lr){7-7}\cmidrule(l){8-9}
    \colhead{Task} & \colhead{\#} & \colhead{Name} & \colhead{Weights} & \phead{Score} & \colhead{Starter} & \colhead{Start $\to$ best} & \colhead{Rounds} & \colhead{Ended by} \\
    \midrule
    \multirow{2}{*}{Embodied-SWE} & \weak{1} & \mlogo{claude}Claude Opus 5.5 & \wclosed & \pbest{44.4} & \weak{0.0} & \weak{0.0 $\to$} 41.2 & 7 & \weak{time budget} \\\rowrulefrom{2-9}
     & \weak{2} & \mlogo{qwen}Qwen3.8-Flash-Next & \wopen & \pcol{23.8} & \weak{0.0} & \weak{0.0 $\to$} 23.8 & 5 & \weak{token budget} \\
    \midrule
    \multirow{2}{*}{Spatial-VSI} & \weak{1} & \mlogo{claude}Claude Opus 5.5 & \wclosed & \pbest{65.9} & \weak{61.0} & \weak{63.6 $\to$} 70.2 & 10 & \weak{round limit} \\\rowrulefrom{2-9}
     & \weak{2} & \mlogo{qwen}Qwen3.8-Flash-Next & \wopen & \pcol{56.1} & \weak{53.4} & \weak{58.0 $\to$} 60.2 & 6 & \weak{token budget} \\
    \midrule
    \multirow{2}{*}{Text-to-Image} & \weak{1} & \mlogo{gpt}GPT-6 Astra & \wclosed & \pbest{94.0} & \weak{91.9} & \weak{94.1 $\to$} 95.5 & 25 & \weak{round limit} \\\rowrulefrom{2-9}
     & \weak{2} & \mlogo{gpt}GPT-5.6 Sol & \wclosed & \pcol{93.9} & \weak{91.9} & \weak{94.1 $\to$} 95.0 & 25 & \weak{round limit} \\
    \midrule
    \multirow{2}{*}{VAE Decode} & \weak{1} & \mlogo{claude}Claude Opus 5.5 & \wclosed & \pbest{2.13$\times$} & \weak{1.00$\times$} & \weak{1.00$\times$ $\to$} 2.14$\times$ & 41 & \weak{no further progress} \\\rowrulefrom{2-9}
     & \weak{2} & \mlogo{gpt}GPT-6 Astra & \wclosed & \pcol{1.96$\times$} & \weak{1.00$\times$} & \weak{1.00$\times$ $\to$} 1.95$\times$ & 16 & \weak{time budget} \\
    \bottomrule
  \end{fittabular}
\end{table}

\begin{table}[p]
  \centering\footnotesize
  \caption{\textbf{RoboDojo.} A dual-arm ARX X5 robot in simulation with the model as its policy, on 10 tasks (task score 0--100). Models are ranked by Suite on the hidden test split, the mean task score over all 10 tasks. The task columns give the test score of the final solution of each task, and Tasks~$>$\,0 the number of tasks with a test score above 0. Validation gives the suite mean of the starter's validation score and of the best validation score reached; Rounds is summed over the tasks. Every run counts up to round 10, and each task uses the model's latest run with a final test score. The starter's test score was not measured in every run and is not shown. Only models with a run on every task are listed.}
  \label{tab:res-robodojo}
  \setlength{\tabcolsep}{2.5pt}
  \begin{fittabular}{@{}rllrrrrrrrrrrrrrr@{}}
    \toprule
    \multicolumn{3}{@{}l}{\colhead{Model}} & \multicolumn{12}{c}{\colhead{Hidden test metrics}} & \colhead{Validation} & \colhead{Run} \\
    \cmidrule(r){1-3}\cmidrule(lr){4-15}\cmidrule(lr){16-16}\cmidrule(l){17-17}
    \colhead{\#} & \colhead{Name} & \colhead{Weights} & \phead{Suite} & \rothead{Tasks $>$\,0} & \rothead{Bottles} & \rothead{Number} & \rothead{Clothes} & \rothead{Tubes} & \rothead{Eggs} & \rothead{Conveyor} & \rothead{Tic-tac-toe} & \rothead{Xylophone} & \rothead{Push-T} & \rothead{Dolls} & \colhead{Start $\to$ best} & \colhead{Rounds} \\
    \midrule
    \weak{1} & \mlogo{claude}Claude Opus 5.5 & \wclosed & \pbest{55.1} & 9 & \best{70} & 27 & \best{68} & 72 & 19 & 0 & 95 & \best{100} & \best{80} & 20 & \weak{0.5 $\to$} 68.9 & 100 \\\rowrule
    \weak{2} & \mlogo{gemini}Gemini 4 Argon & \wclosed & \pcol{46.8} & 8 & \best{70} & \best{71} & 44 & \best{100} & \best{23} & 0 & \best{100} & 20 & 40 & 0 & \weak{0.9 $\to$} 47.7 & 90 \\\rowrule
    \weak{3} & \mlogo{gpt}GPT-6.1 Sol & \wclosed & \pcol{41.5} & 9 & 14 & 11 & 36 & 80 & 4 & \best{60} & 90 & 80 & 0 & \best{40} & \weak{0.2 $\to$} 50.5 & 100 \\\rowrule
    \weak{4} & \mlogo{gpt}GPT-6 Astra & \wclosed & \pcol{28.6} & 7 & 37 & 18 & 4 & 0 & 7 & 40 & \best{100} & 80 & 0 & 0 & \weak{0.7 $\to$} 30.0 & 100 \\\rowrule
    \weak{5} & \mlogo{deepseek}DeepSeek-V4.1 Flash & \wopen & \pcol{17.8} & 5 & 16 & 11 & 0 & 16 & 0 & 0 & 75 & 60 & 0 & 0 & \weak{0.0 $\to$} 18.7 & 100 \\\rowrule
    \weak{6} & \mlogo{meta}Muse Wonder 1.0 & \wclosed & \pcol{10.6} & 5 & 6 & 2 & 4 & 0 & 4 & 0 & 90 & 0 & 0 & 0 & \weak{0.0 $\to$} 15.4 & 100 \\\rowrule
    \weak{7} & \mlogo{qwen}Qwen3.8-Flash-Next & \wopen & \pcol{8.4} & 2 & 9 & 0 & 0 & 0 & 0 & 0 & 75 & 0 & 0 & 0 & \weak{0.0 $\to$} 9.9 & 100 \\\rowrule
    \weak{8} & \mlogo{minimax}MiniMax-M3 & \wopen & \pcol{0.0} & 0 & 0 & 0 & 0 & 0 & 0 & 0 & 0 & 0 & 0 & 0 & \weak{0.0 $\to$} 0.0 & 100 \\
    \bottomrule
  \end{fittabular}
\end{table}

\begin{table}[p]
  \centering\footnotesize
  \caption{\textbf{RoboCasa.} A mobile manipulator in simulated kitchens (RoboCasa365) with the model as its policy, on 8 tasks (success rate over 5 test episodes per task). Models are ranked by Suite on the hidden test split, the mean success rate over the 8 tasks. The task columns give the test success rate of the final solution of each task, and Tasks~$>$\,0 the number of tasks with a test success rate above 0. Validation gives the suite mean of the starter's validation score and of the best validation score reached; Rounds is summed over the tasks. Every run counts up to round 10, and each task uses the model's latest run with a final test score. The starter's test score was not measured in every run and is not shown. A ninth task, Heat kebab sandwich, is left out: no model reached a success on it in validation or test. $^\dagger$Muse Spark~1.3 is accessed through an internal endpoint; it is shown for comparison and not ranked. All scores are multiplied by 100.}
  \label{tab:res-robocasa}
  \setlength{\tabcolsep}{2.5pt}
  \begin{fittabular}{@{}rllrrrrrrrrrrrr@{}}
    \toprule
    \multicolumn{3}{@{}l}{\colhead{Model}} & \multicolumn{10}{c}{\colhead{Hidden test metrics}} & \colhead{Validation} & \colhead{Run} \\
    \cmidrule(r){1-3}\cmidrule(lr){4-13}\cmidrule(lr){14-14}\cmidrule(l){15-15}
    \colhead{\#} & \colhead{Name} & \colhead{Weights} & \phead{Suite} & \rothead{Tasks $>$\,0} & \rothead{Coffee} & \rothead{Kettle} & \rothead{Cabinet} & \rothead{Drawer} & \rothead{Hot dogs} & \rothead{Stove} & \rothead{Microwave} & \rothead{Faucet} & \colhead{Start $\to$ best} & \colhead{Rounds} \\
    \midrule
    \weak{1} & \mlogo{claude}Claude Opus 5.5 & \wclosed & \pbest{57.5} & 7 & 0 & \best{40} & \best{60} & \best{100} & \best{60} & 20 & \best{100} & \best{80} & \weak{5.0 $\to$} 60.0 & 80 \\\rowrule
    \weak{2} & \mlogo{gemini}Gemini 4 Argon & \wclosed & \pcol{40.0} & 7 & 0 & 20 & \best{60} & 80 & 20 & 20 & 80 & 40 & \weak{2.5 $\to$} 57.5 & 75 \\\rowrule
    \weak{3} & \mlogo{gpt}GPT-6 Astra & \wclosed & \pcol{37.5} & 6 & \best{60} & 20 & 0 & 80 & 20 & \best{40} & 80 & 0 & \weak{0.0 $\to$} 47.5 & 74 \\\rowrule
    \weak{4} & \mlogo{gpt}GPT-6.1 Sol & \wclosed & \pcol{32.5} & 6 & 40 & 0 & 20 & 80 & 0 & \best{40} & 60 & 20 & \weak{0.0 $\to$} 45.0 & 71 \\\rowrule
    \weak{5} & \mlogo{glm}GLM-5.3 Flash & \wopen & \pcol{7.5} & 1 & 0 & 0 & 0 & 0 & 0 & 0 & 0 & 60 & \weak{0.0 $\to$} 7.5 & 80 \\\rowrule
    \weak{6} & \mlogo{qwen}Qwen3.8-Flash-Next & \wopen & \pcol{5.0} & 2 & 0 & 0 & 0 & 0 & 0 & 0 & 20 & 20 & \weak{2.5 $\to$} 12.5 & 80 \\\rowrule
    \weak{7} & \mlogo{gemma}Gemma-4-31B-it & \wopen & \pcol{2.5} & 1 & 0 & 0 & 0 & 0 & 0 & 0 & 20 & 0 & \weak{0.0 $\to$} 1.3 & 78 \\\rowrule
    \weak{8} & \mlogo{minimax}MiniMax-M3 & \wopen & \pcol{0.0} & 0 & 0 & 0 & 0 & 0 & 0 & 0 & 0 & 0 & \weak{0.0 $\to$} 2.5 & 80 \\\rowrule
    \weak{8} & \mlogo{qwen}Qwen3.8-27B & \wopen & \pcol{0.0} & 0 & 0 & 0 & 0 & 0 & 0 & 0 & 0 & 0 & \weak{0.0 $\to$} 3.8 & 80 \\\rowrule
    \weak{8} & \mlogo{deepseek}DeepSeek-V4.1 Flash & \wopen & \pcol{0.0} & 0 & 0 & 0 & 0 & 0 & 0 & 0 & 0 & 0 & \weak{0.0 $\to$} 12.5 & 80 \\\rowrule
    \weak{--} & \mlogo{meta}Muse Spark 1.3$^\dagger$ & \wclosed & \pcol{0.0} & 0 & 0 & 0 & 0 & 0 & 0 & 0 & 0 & 0 & \weak{2.5 $\to$} 5.0 & 80 \\
    \bottomrule
  \end{fittabular}
\end{table}

\begin{table}[p]
  \centering\footnotesize
  \caption{\textbf{Active Search.} Active visual search in 360° scenes (H*Bench). Models are ranked by SPL on the hidden test split. Test metrics are those of the final solution, except Starter, which is the primary test score of the starter solution. Validation gives the validation score of the starter and the best validation score reached during the run. SPL is success weighted by path length; Success is the mean of Object SR and Path SR; Object SR and Path SR are the success rates of object-search and path-search episodes; Avg.\ steps is the average number of actions per episode. All scores except Avg.\ steps are multiplied by 100.}
  \label{tab:res-hstar}
  \setlength{\tabcolsep}{2pt}
  \begin{fittabular}{@{}rllrrrrrrrrl@{}}
    \toprule
    \multicolumn{3}{@{}l}{\colhead{Model}} & \multicolumn{6}{c}{\colhead{Hidden test metrics}} & \colhead{Validation} & \multicolumn{2}{c}{\colhead{Run}} \\
    \cmidrule(r){1-3}\cmidrule(lr){4-9}\cmidrule(lr){10-10}\cmidrule(l){11-12}
    \colhead{\#} & \colhead{Name} & \colhead{Weights} & \phead{SPL} & \colhead{Starter} & \colhead{Success} & \stackhead{Object}{SR} & \stackhead{Path}{SR} & \stackhead{Avg.}{steps} & \colhead{Start $\to$ best} & \colhead{Rounds} & \colhead{Ended by} \\
    \midrule
    \weak{1} & \mlogo{gemini}Gemini 4 Argon & \wclosed & \pbest{79.7} & \weak{45.9} & \best{100.0} & \best{100.0} & \best{100.0} & 2.53 & \weak{40.8 $\to$} 67.2 & 12 & \weak{no further progress} \\\rowrule
    \weak{2} & \mlogo{gpt}GPT-6 Astra & \wclosed & \pcol{78.2} & \weak{45.2} & 93.4 & 93.9 & 92.9 & 2.23 & \weak{35.0 $\to$} 64.8 & 15 & \weak{round limit} \\\rowrule
    \weak{3} & \mlogo{gpt}GPT-6.1 Sol & \wclosed & \pcol{74.4} & \weak{39.3} & 85.3 & 84.9 & 85.7 & 1.98 & \weak{26.6 $\to$} 59.7 & 15 & \weak{round limit} \\\rowrule
    \weak{4} & \mlogo{claude}Claude Opus 5.5 & \wclosed & \pcol{74.2} & \weak{31.1} & 83.2 & 87.9 & 78.6 & 2.21 & \weak{31.6 $\to$} 66.3 & 15 & \weak{round limit} \\\rowrule
    \weak{5} & \mlogo{qwen}Qwen3.8-27B & \wopen & \pcol{69.0} & \weak{18.1} & 78.1 & 84.9 & 71.4 & 2.23 & \weak{20.0 $\to$} 51.7 & 15 & \weak{round limit} \\\rowrule
    \weak{6} & \mlogo{gemini}Gemini 3.8 Flash & \wclosed & \pcol{65.5} & \weak{40.2} & 80.2 & 81.8 & 78.6 & 2.51 & \weak{32.7 $\to$} 59.7 & 11 & \weak{token budget} \\\rowrule
    \weak{7} & \mlogo{glm}GLM-5.3 Flash & \wopen & \pcol{60.5} & \weak{21.5} & 74.0 & 90.9 & 57.1 & 2.83 & \weak{17.1 $\to$} 52.5 & 15 & \weak{token budget} \\\rowrule
    \weak{8} & \mlogo{qwen}Qwen3.8-Flash-Next & \wopen & \pcol{59.3} & \weak{14.1} & 65.9 & 81.8 & 50.0 & 1.98 & \weak{16.7 $\to$} 50.3 & 12 & \weak{time budget} \\\rowrule
    \weak{9} & \mlogo{claude}Claude Sonnet 5.5 & \wclosed & \pcol{55.8} & \weak{26.0} & 62.3 & 81.8 & 42.9 & 2.23 & \weak{26.1 $\to$} 53.0 & 15 & \weak{round limit} \\\rowrule
    \weak{10} & \mlogo{gemma}Gemma-4-31B-it & \wopen & \pcol{52.2} & \weak{24.4} & 59.3 & 75.8 & 42.9 & 2.15 & \weak{17.6 $\to$} 44.4 & 12 & \weak{validation budget} \\\rowrule
    \weak{11} & \mlogo{deepseek}DeepSeek-V4.1 Flash & \wopen & \pcol{47.1} & \weak{26.6} & 59.3 & 75.8 & 42.9 & 2.38 & \weak{19.9 $\to$} 35.8 & 6 & \weak{time budget} \\\rowrule
    \weak{12} & \mlogo{minimax}MiniMax-M3 & \wopen & \pcol{32.5} & \weak{16.0} & 40.5 & 66.7 & 14.3 & 2.72 & \weak{10.0 $\to$} 32.8 & 15 & \weak{round limit} \\
    \bottomrule
  \end{fittabular}
\end{table}

\begin{table}[p]
  \centering\footnotesize
  \caption{\textbf{RoboLab.} One generalist tabletop manipulation policy for many tasks (pick and place, stacking, reorienting, choosing objects by color, size, kind or position), scored on tasks it never saw during development. Models are ranked by Score on the hidden test split. Test metrics are those of the final solution. Validation gives the validation score of the starter and the best validation score reached during the run. Score is the RoboLab score, which credits a successful episode fully and an unsuccessful one by the fraction of its subtasks completed; Success rate is the fraction of successful episodes. Score is on its own 0--100 scale, and Success rate is multiplied by 100.}
  \label{tab:res-robolab}
  \setlength{\tabcolsep}{3pt}
  \begin{fittabular}{@{}rllrrrrl@{}}
    \toprule
    \multicolumn{3}{@{}l}{\colhead{Model}} & \multicolumn{2}{c}{\colhead{Hidden test metrics}} & \colhead{Validation} & \multicolumn{2}{c}{\colhead{Run}} \\
    \cmidrule(r){1-3}\cmidrule(lr){4-5}\cmidrule(lr){6-6}\cmidrule(l){7-8}
    \colhead{\#} & \colhead{Name} & \colhead{Weights} & \phead{Score} & \colhead{Success rate} & \colhead{Start $\to$ best} & \colhead{Rounds} & \colhead{Ended by} \\
    \midrule
    \weak{1} & \mlogo{claude}Claude Opus 5.5 & \wclosed & \pbest{71.9} & \best{66.7} & \weak{16.7 $\to$} 70.8 & 10 & \weak{round limit} \\\rowrule
    \weak{2} & \mlogo{gpt}GPT-6.1 Sol & \wclosed & \pcol{62.5} & 56.2 & \weak{18.8 $\to$} 68.8 & 10 & \weak{round limit} \\\rowrule
    \weak{3} & \mlogo{claude}Claude Sonnet 5.5 & \wclosed & \pcol{41.0} & 31.2 & \weak{8.3 $\to$} 47.9 & 10 & \weak{round limit} \\\rowrule
    \weak{4} & \mlogo{gpt}GPT-5.6 Sol & \wclosed & \pcol{37.8} & 29.2 & \weak{2.1 $\to$} 29.2 & 10 & \weak{round limit} \\\rowrule
    \weak{5} & \mlogo{gpt}GPT-6 Astra & \wclosed & \pcol{33.0} & 20.8 & \weak{31.2 $\to$} 47.9 & 10 & \weak{round limit} \\\rowrule
    \weak{6} & \mlogo{qwen}Qwen3.8-27B & \wopen & \pcol{31.6} & 22.9 & \weak{4.2 $\to$} 37.5 & 10 & \weak{round limit} \\\rowrule
    \weak{7} & \mlogo{qwen}Qwen3.8-Flash-Next & \wopen & \pcol{17.4} & 6.2 & \weak{4.2 $\to$} 45.8 & 10 & \weak{round limit} \\\rowrule
    \weak{8} & \mlogo{minimax}MiniMax-M3 & \wopen & \pcol{8.3} & 4.2 & \weak{8.3 $\to$} 8.3 & 10 & \weak{round limit} \\\rowrule
    \weak{9} & \mlogo{gemma}Gemma-4-31B-it & \wopen & \pcol{5.2} & 0.0 & \weak{4.2 $\to$} 8.3 & 10 & \weak{GPU budget} \\
    \bottomrule
  \end{fittabular}
\end{table}

\begin{table}[p]
  \centering\footnotesize
  \caption{\textbf{LIBERO-MAX.} A Franka arm in LIBERO kitchen scenes where the world changes once the episode is under way; the model's controller drives it, with a frozen X-VLA policy, cameras and depth to call on (LIBERO-MAX, dynamic success rate over paired episodes). Models are ranked by Dynamic success on the hidden test split. Test metrics are those of the final solution, except Starter, which is the primary test score of the starter solution. Validation gives the validation score of the starter and the best validation score reached during the run. Dynamic success is the success rate on episodes in which the scene is perturbed once while the episode is under way; Base success is the success rate on the paired episodes without the perturbation. All scores are multiplied by 100.}
  \label{tab:res-liberomax}
  \setlength{\tabcolsep}{3pt}
  \begin{fittabular}{@{}rllrrrrrl@{}}
    \toprule
    \multicolumn{3}{@{}l}{\colhead{Model}} & \multicolumn{3}{c}{\colhead{Hidden test metrics}} & \colhead{Validation} & \multicolumn{2}{c}{\colhead{Run}} \\
    \cmidrule(r){1-3}\cmidrule(lr){4-6}\cmidrule(lr){7-7}\cmidrule(l){8-9}
    \colhead{\#} & \colhead{Name} & \colhead{Weights} & \phead{Dynamic success} & \colhead{Starter} & \colhead{Base success} & \colhead{Start $\to$ best} & \colhead{Rounds} & \colhead{Ended by} \\
    \midrule
    \weak{1} & \mlogo{claude}Claude Opus 5.5 & \wclosed & \pbest{45.8} & \weak{29.2} & 58.3 & \weak{31.2 $\to$} 50.0 & 15 & \weak{round limit} \\\rowrule
    \weak{1} & \mlogo{qwen}Qwen3.8-Flash-Next & \wopen & \pbest{45.8} & \weak{33.3} & \best{62.5} & \weak{31.2 $\to$} 37.5 & 15 & \weak{round limit} \\\rowrule
    \weak{3} & \mlogo{gpt}GPT-6.1 Sol & \wclosed & \pcol{41.7} & \weak{29.2} & 45.8 & \weak{31.2 $\to$} 37.5 & 15 & \weak{round limit} \\\rowrule
    \weak{4} & \mlogo{gemma}Gemma-4-31B-it & \wopen & \pcol{29.2} & \weak{25.0} & 54.2 & \weak{31.2 $\to$} 43.8 & 15 & \weak{round limit} \\\rowrule
    \weak{5} & \mlogo{qwen}Qwen3.6-35B-A3B & \wopen & \pcol{25.0} & \weak{25.0} & 54.2 & \weak{31.2 $\to$} 31.2 & 15 & \weak{round limit} \\\rowrule
    \weak{5} & \mlogo{qwen}Qwen3.8-27B & \wopen & \pcol{25.0} & \weak{25.0} & 54.2 & \weak{31.2 $\to$} 31.2 & 15 & \weak{round limit} \\
    \bottomrule
  \end{fittabular}
\end{table}

\begin{table}[p]
  \centering\footnotesize
  \caption{\textbf{LIBERO-PRO.} A Franka arm in LIBERO kitchen scenes with the task perturbed before it starts (objects, positions, wording, goal or camera); the model's controller drives it, with a frozen X-VLA policy, cameras and depth to call on (LIBERO-PRO, success rate). Models are ranked by Success rate on the hidden test split. Test metrics are those of the final solution, except Starter, which is the primary test score of the starter solution. Validation gives the validation score of the starter and the best validation score reached during the run. Success rate is the fraction of test episodes in which the task is completed. All scores are multiplied by 100.}
  \label{tab:res-liberopro}
  \setlength{\tabcolsep}{3pt}
  \begin{fittabular}{@{}rllrrrrl@{}}
    \toprule
    \multicolumn{3}{@{}l}{\colhead{Model}} & \multicolumn{2}{c}{\colhead{Hidden test metrics}} & \colhead{Validation} & \multicolumn{2}{c}{\colhead{Run}} \\
    \cmidrule(r){1-3}\cmidrule(lr){4-5}\cmidrule(lr){6-6}\cmidrule(l){7-8}
    \colhead{\#} & \colhead{Name} & \colhead{Weights} & \phead{Success rate} & \colhead{Starter} & \colhead{Start $\to$ best} & \colhead{Rounds} & \colhead{Ended by} \\
    \midrule
    \weak{1} & \mlogo{claude}Claude Opus 5.5 & \wclosed & \pbest{63.3} & \weak{53.3} & \weak{50.0 $\to$} 75.0 & 15 & \weak{round limit} \\\rowrule
    \weak{2} & \mlogo{gemma}Gemma-4-31B-it & \wopen & \pcol{53.3} & \weak{53.3} & \weak{50.0 $\to$} 50.0 & 15 & \weak{round limit} \\\rowrule
    \weak{2} & \mlogo{qwen}Qwen3.6-35B-A3B & \wopen & \pcol{53.3} & \weak{53.3} & \weak{50.0 $\to$} 50.0 & 15 & \weak{round limit} \\\rowrule
    \weak{2} & \mlogo{gpt}GPT-6.1 Sol & \wclosed & \pcol{53.3} & \weak{53.3} & \weak{50.0 $\to$} 60.0 & 15 & \weak{round limit} \\\rowrule
    \weak{2} & \mlogo{qwen}Qwen3.8-Flash-Next & \wopen & \pcol{53.3} & \weak{53.3} & \weak{50.0 $\to$} 50.0 & 15 & \weak{round limit} \\\rowrule
    \weak{6} & \mlogo{qwen}Qwen3.8-27B & \wopen & \pcol{50.0} & \weak{53.3} & \weak{50.0 $\to$} 55.0 & 15 & \weak{round limit} \\
    \bottomrule
  \end{fittabular}
\end{table}

\begin{table}[p]
  \centering\footnotesize
  \caption{\textbf{Ego-Exo Memory.} Memory reasoning over synchronized egocentric and exocentric videos (EgoExoMem, Ego-Exo4D part). Models are ranked by Accuracy on the hidden test split. Test metrics are those of the final solution, except Starter, which is the primary test score of the starter solution. Validation gives the validation score of the starter and the best validation score reached during the run. Accuracy is measured on all test questions; Cooking and Bike repair give the accuracy on the questions about cooking and bike-repair videos. All scores are multiplied by 100.}
  \label{tab:res-egoexo}
  \setlength{\tabcolsep}{3pt}
  \begin{fittabular}{@{}rllrrrrrrl@{}}
    \toprule
    \multicolumn{3}{@{}l}{\colhead{Model}} & \multicolumn{4}{c}{\colhead{Hidden test metrics}} & \colhead{Validation} & \multicolumn{2}{c}{\colhead{Run}} \\
    \cmidrule(r){1-3}\cmidrule(lr){4-7}\cmidrule(lr){8-8}\cmidrule(l){9-10}
    \colhead{\#} & \colhead{Name} & \colhead{Weights} & \phead{Accuracy} & \colhead{Starter} & \colhead{Cooking} & \colhead{Bike repair} & \colhead{Start $\to$ best} & \colhead{Rounds} & \colhead{Ended by} \\
    \midrule
    \weak{1} & \mlogo{claude}Claude Opus 5.5 & \wclosed & \pbest{72.5} & \weak{61.7} & \best{73.6} & \best{69.0} & \weak{63.7 $\to$} 73.8 & 10 & \weak{round limit} \\\rowrule
    \weak{2} & \mlogo{gemini}Gemini 4 Argon & \wclosed & \pcol{70.0} & \weak{65.0} & \best{73.6} & 58.6 & \weak{75.0 $\to$} 76.2 & 6 & \weak{validation budget} \\\rowrule
    \weak{3} & \mlogo{gpt}GPT-6.1 Sol & \wclosed & \pcol{67.5} & \weak{57.5} & 69.2 & 62.1 & \weak{67.5 $\to$} 75.0 & 10 & \weak{round limit} \\\rowrule
    \weak{4} & \mlogo{gpt}GPT-6 Astra & \wclosed & \pcol{65.8} & \weak{60.0} & 69.2 & 55.2 & \weak{75.0 $\to$} 81.2 & 10 & \weak{round limit} \\\rowrule
    \weak{5} & \mlogo{gemini}Gemini 3.8 Flash & \wclosed & \pcol{63.3} & \weak{59.2} & 64.8 & 58.6 & \weak{63.7 $\to$} 73.8 & 10 & \weak{round limit} \\\rowrule
    \weak{6} & \mlogo{qwen}Qwen3.8-27B & \wopen & \pcol{60.8} & \weak{55.0} & 62.6 & 55.2 & \weak{58.8 $\to$} 72.5 & 10 & \weak{round limit} \\\rowrule
    \weak{7} & \mlogo{qwen}Qwen3.8-Flash-Next & \wopen & \pcol{59.2} & \weak{54.2} & 59.3 & 58.6 & \weak{61.3 $\to$} 67.5 & 10 & \weak{round limit} \\\rowrule
    \weak{7} & \mlogo{gemma}Gemma-4-31B-it & \wopen & \pcol{59.2} & \weak{52.5} & 63.7 & 44.8 & \weak{61.3 $\to$} 66.2 & 5 & \weak{time budget} \\\rowrule
    \weak{9} & \mlogo{glm}GLM-5.3 Flash & \wopen & \pcol{56.7} & \weak{52.5} & 60.4 & 44.8 & \weak{60.0 $\to$} 65.0 & 8 & \weak{validation budget} \\\rowrule
    \weak{10} & \mlogo{minimax}MiniMax-M3 & \wopen & \pcol{51.7} & \weak{52.5} & 52.8 & 48.3 & \weak{47.5 $\to$} 56.2 & 10 & \weak{round limit} \\\rowrule
    \weak{11} & \mlogo{deepseek}DeepSeek-V4.1 Flash & \wopen & \pcol{50.8} & \weak{55.8} & 53.8 & 41.4 & \weak{66.2 $\to$} 66.2 & 7 & \weak{validation budget} \\
    \bottomrule
  \end{fittabular}
\end{table}

\begin{table}[p]
  \centering\footnotesize
  \caption{\textbf{Kinematics.} Sizes, distances, velocities and accelerations in physical units from a short video and one known quantity, with a fixed open VLM (QuantiPhy). Models are ranked by MRA on the hidden test split. Test metrics are those of the final solution, except Starter, which is the primary test score of the starter solution. Validation gives the validation score of the starter and the best validation score reached during the run. MRA is the mean relative accuracy, which averages over a range of thresholds whether the relative error of a predicted quantity falls below the threshold. All scores are multiplied by 100. $^\dagger$Muse Spark~1.3 is accessed through an internal endpoint; it is shown for comparison and not ranked.}
  \label{tab:res-kinematics}
  \setlength{\tabcolsep}{3pt}
  \begin{fittabular}{@{}rllrrrrl@{}}
    \toprule
    \multicolumn{3}{@{}l}{\colhead{Model}} & \multicolumn{2}{c}{\colhead{Hidden test metrics}} & \colhead{Validation} & \multicolumn{2}{c}{\colhead{Run}} \\
    \cmidrule(r){1-3}\cmidrule(lr){4-5}\cmidrule(lr){6-6}\cmidrule(l){7-8}
    \colhead{\#} & \colhead{Name} & \colhead{Weights} & \phead{MRA} & \colhead{Starter} & \colhead{Start $\to$ best} & \colhead{Rounds} & \colhead{Ended by} \\
    \midrule
    \weak{1} & \mlogo{claude}Claude Sonnet 5.5 & \wclosed & \pbest{57.2} & \weak{33.1} & \weak{26.4 $\to$} 56.5 & 30 & \weak{time budget} \\\rowrule
    \weak{2} & \mlogo{claude}Claude Opus 5.5 & \wclosed & \pcol{54.6} & \weak{43.6} & \weak{29.0 $\to$} 57.7 & 29 & \weak{no further progress} \\\rowrule
    \weak{3} & \mlogo{gpt}GPT-5.6 Sol & \wclosed & \pcol{50.0} & \weak{48.8} & \weak{34.2 $\to$} 41.7 & 18 & \weak{time budget} \\\rowrule
    \weak{4} & \mlogo{qwen}Qwen3.8-Flash-Next & \wopen & \pcol{49.2} & \weak{37.3} & \weak{22.1 $\to$} 45.6 & 14 & \weak{time budget} \\\rowrule
    \weak{5} & \mlogo{gemini}Gemini 3.8 Flash & \wclosed & \pcol{48.2} & \weak{42.5} & \weak{34.0 $\to$} 41.5 & 31 & \weak{time budget} \\\rowrule
    \weak{6} & \mlogo{gemini}Gemini 3.7 Flash & \wclosed & \pcol{46.3} & \weak{39.0} & \weak{29.8 $\to$} 39.8 & 24 & \weak{no further progress} \\\rowrule
    \weak{7} & \mlogo{qwen}Qwen3.8-27B & \wopen & \pcol{43.4} & \weak{42.1} & \weak{20.0 $\to$} 53.5 & 36 & \weak{time budget} \\\rowrule
    \weak{8} & \mlogo{grok}Grok 4.6 & \wclosed & \pcol{37.2} & \weak{47.2} & \weak{24.6 $\to$} 24.6 & 25 & \weak{time budget} \\\rowrule
    \weak{--} & \mlogo{meta}Muse Spark 1.3$^\dagger$ & \wclosed & \pcol{45.5} & \weak{36.4} & \weak{31.7 $\to$} 48.5 & 45 & \weak{no further progress} \\
    \bottomrule
  \end{fittabular}
\end{table}

\begin{table}[p]
  \centering\footnotesize
  \caption{\textbf{Mass.} A product's weight in kilograms from its catalog photos, type and dimensions, with a fixed open VLM (QuantiPhy-style, Amazon Berkeley Objects). Models are ranked by MRA on the hidden test split. Test metrics are those of the final solution, except Starter, which is the primary test score of the starter solution. Validation gives the validation score of the starter and the best validation score reached during the run. MRA is the mean relative accuracy, which averages over a range of thresholds whether the relative error of a predicted quantity falls below the threshold. All scores are multiplied by 100. $^\dagger$Muse Spark~1.3 is accessed through an internal endpoint; it is shown for comparison and not ranked.}
  \label{tab:res-mass}
  \setlength{\tabcolsep}{3pt}
  \begin{fittabular}{@{}rllrrrrl@{}}
    \toprule
    \multicolumn{3}{@{}l}{\colhead{Model}} & \multicolumn{2}{c}{\colhead{Hidden test metrics}} & \colhead{Validation} & \multicolumn{2}{c}{\colhead{Run}} \\
    \cmidrule(r){1-3}\cmidrule(lr){4-5}\cmidrule(lr){6-6}\cmidrule(l){7-8}
    \colhead{\#} & \colhead{Name} & \colhead{Weights} & \phead{MRA} & \colhead{Starter} & \colhead{Start $\to$ best} & \colhead{Rounds} & \colhead{Ended by} \\
    \midrule
    \weak{1} & \mlogo{claude}Claude Opus 5.5 & \wclosed & \pbest{63.5} & \weak{50.4} & \weak{46.4 $\to$} 63.0 & 29 & \weak{no further progress} \\\rowrule
    \weak{2} & \mlogo{gpt}GPT-5.6 Sol & \wclosed & \pcol{63.0} & \weak{49.8} & \weak{47.7 $\to$} 61.1 & 49 & \weak{no further progress} \\\rowrule
    \weak{3} & \mlogo{qwen}Qwen3.8-27B & \wopen & \pcol{62.7} & \weak{49.1} & \weak{50.3 $\to$} 60.9 & 13 & \weak{no further progress} \\\rowrule
    \weak{4} & \mlogo{claude}Claude Sonnet 5.5 & \wclosed & \pcol{62.5} & \weak{51.0} & \weak{50.2 $\to$} 61.7 & 60 & \weak{round limit} \\\rowrule
    \weak{5} & \mlogo{gemini}Gemini 3.7 Flash & \wclosed & \pcol{60.6} & \weak{51.8} & \weak{46.7 $\to$} 61.1 & 25 & \weak{no further progress} \\\rowrule
    \weak{6} & \mlogo{qwen}Qwen3.8-Flash-Next & \wopen & \pcol{60.4} & \weak{51.4} & \weak{49.4 $\to$} 61.1 & 24 & \weak{time budget} \\\rowrule
    \weak{7} & \mlogo{grok}Grok 4.6 & \wclosed & \pcol{59.3} & \weak{51.3} & \weak{46.2 $\to$} 57.1 & 21 & \weak{no further progress} \\\rowrule
    \weak{8} & \mlogo{gemini}Gemini 3.8 Flash & \wclosed & \pcol{57.5} & \weak{50.0} & \weak{44.3 $\to$} 59.1 & 32 & \weak{time budget} \\\rowrule
    \weak{--} & \mlogo{meta}Muse Spark 1.3$^\dagger$ & \wclosed & \pcol{56.9} & \weak{51.2} & \weak{45.3 $\to$} 56.3 & 38 & \weak{no further progress} \\
    \bottomrule
  \end{fittabular}
\end{table}

\begin{table}[p]
  \centering\footnotesize
  \caption{\textbf{ARC-AGI-1.} Abstract reasoning on grid puzzles from a few examples (ARC-AGI-1 public evaluation set, pass@2). Models are ranked by Accuracy on the hidden test split. Test metrics are those of the final solution, except Starter, which is the primary test score of the starter solution. Validation gives the validation score of the starter and the best validation score reached during the run. Accuracy is the fraction of test puzzles solved within two attempts (pass@2). All scores are multiplied by 100.}
  \label{tab:res-arc1}
  \setlength{\tabcolsep}{3pt}
  \begin{fittabular}{@{}rllrrrrl@{}}
    \toprule
    \multicolumn{3}{@{}l}{\colhead{Model}} & \multicolumn{2}{c}{\colhead{Hidden test metrics}} & \colhead{Validation} & \multicolumn{2}{c}{\colhead{Run}} \\
    \cmidrule(r){1-3}\cmidrule(lr){4-5}\cmidrule(lr){6-6}\cmidrule(l){7-8}
    \colhead{\#} & \colhead{Name} & \colhead{Weights} & \phead{Accuracy} & \colhead{Starter} & \colhead{Start $\to$ best} & \colhead{Rounds} & \colhead{Ended by} \\
    \midrule
    \weak{1} & \mlogo{gpt}GPT-6 Astra & \wclosed & \pbest{99.8} & \weak{99.2} & \weak{98.8 $\to$} 100.0 & 10 & \weak{round limit} \\\rowrule
    \weak{2} & \mlogo{gpt}GPT-6.1 Sol & \wclosed & \pcol{98.8} & \weak{98.8} & \weak{98.8 $\to$} 98.8 & 10 & \weak{round limit} \\\rowrule
    \weak{3} & \mlogo{claude}Claude Opus 5.5 & \wclosed & \pcol{98.0} & \weak{96.0} & \weak{98.8 $\to$} 100.0 & 10 & \weak{round limit} \\\rowrule
    \weak{4} & \mlogo{deepseek}DeepSeek-V4.1 Flash & \wopen & \pcol{91.0} & \weak{91.0} & \weak{93.8 $\to$} 93.8 & 2 & \weak{time budget} \\\rowrule
    \weak{5} & \mlogo{qwen}Qwen3.8-Flash-Next & \wopen & \pcol{80.8} & \weak{80.8} & \weak{90.0 $\to$} 90.0 & 1 & \weak{time budget} \\\rowrule
    \weak{6} & \mlogo{qwen}Qwen3.8-27B & \wopen & \pcol{77.2} & \weak{75.5} & \weak{83.8 $\to$} 95.0 & 5 & \weak{time budget} \\\rowrule
    \weak{7} & \mlogo{glm}GLM-5.3 Flash & \wopen & \pcol{63.5} & \weak{63.5} & \weak{76.2 $\to$} 76.2 & 6 & \weak{time budget} \\\rowrule
    \weak{8} & \mlogo{gemma}Gemma-4-31B-it & \wopen & \pcol{54.0} & \weak{54.0} & \weak{67.5 $\to$} 67.5 & 2 & \weak{time budget} \\
    \bottomrule
  \end{fittabular}
\end{table}

\begin{table}[p]
  \centering\footnotesize
  \caption{\textbf{ARC-AGI-2.} Abstract reasoning on grid puzzles from a few examples (ARC-AGI-2 public evaluation set, pass@2). Models are ranked by Accuracy on the hidden test split. Test metrics are those of the final solution, except Starter, which is the primary test score of the starter solution. Validation gives the validation score of the starter and the best validation score reached during the run. Accuracy is the fraction of test puzzles solved within two attempts (pass@2). All scores are multiplied by 100.}
  \label{tab:res-arc2}
  \setlength{\tabcolsep}{3pt}
  \begin{fittabular}{@{}rllrrrrl@{}}
    \toprule
    \multicolumn{3}{@{}l}{\colhead{Model}} & \multicolumn{2}{c}{\colhead{Hidden test metrics}} & \colhead{Validation} & \multicolumn{2}{c}{\colhead{Run}} \\
    \cmidrule(r){1-3}\cmidrule(lr){4-5}\cmidrule(lr){6-6}\cmidrule(l){7-8}
    \colhead{\#} & \colhead{Name} & \colhead{Weights} & \phead{Accuracy} & \colhead{Starter} & \colhead{Start $\to$ best} & \colhead{Rounds} & \colhead{Ended by} \\
    \midrule
    \weak{1} & \mlogo{gpt}GPT-6.1 Sol & \wclosed & \pbest{95.8} & \weak{92.5} & \weak{98.8 $\to$} 100.0 & 10 & \weak{round limit} \\\rowrule
    \weak{2} & \mlogo{claude}Claude Opus 5.5 & \wclosed & \pcol{95.0} & \weak{78.3} & \weak{96.2 $\to$} 100.0 & 10 & \weak{round limit} \\\rowrule
    \weak{3} & \mlogo{gpt}GPT-6 Astra & \wclosed & \pcol{93.3} & \weak{93.3} & \weak{98.8 $\to$} 98.8 & 10 & \weak{round limit} \\\rowrule
    \weak{4} & \mlogo{deepseek}DeepSeek-V4.1 Flash & \wopen & \pcol{46.7} & \weak{46.7} & \weak{92.5 $\to$} 92.5 & 4 & \weak{time budget} \\\rowrule
    \weak{5} & \mlogo{qwen}Qwen3.8-Flash-Next & \wopen & \pcol{22.5} & \weak{22.5} & \weak{82.5 $\to$} 82.5 & 4 & \weak{time budget} \\\rowrule
    \weak{6} & \mlogo{qwen}Qwen3.8-27B & \wopen & \pcol{19.2} & \weak{15.0} & \weak{77.5 $\to$} 80.0 & 3 & \weak{time budget} \\\rowrule
    \weak{7} & \mlogo{glm}GLM-5.3 Flash & \wopen & \pcol{14.2} & \weak{14.2} & \weak{68.8 $\to$} 68.8 & 2 & \weak{time budget} \\\rowrule
    \weak{8} & \mlogo{minimax}MiniMax-M3 & \wopen & \pcol{4.2} & \weak{4.2} & \weak{72.5 $\to$} 72.5 & 1 & \weak{time budget} \\\rowrule
    \weak{9} & \mlogo{gemma}Gemma-4-31B-it & \wopen & \pcol{1.7} & \weak{1.7} & \weak{60.0 $\to$} 60.0 & 2 & \weak{time budget} \\
    \bottomrule
  \end{fittabular}
\end{table}

\begin{table}[p]
  \centering\footnotesize
  \caption{\textbf{ARC-AGI-3.} Interactive puzzle games learned by playing, scored by action efficiency against humans (ARC-AGI-3 public games). Models are ranked by RHAE on the hidden test split. Test metrics are those of the final solution, except Starter, which is the primary test score of the starter solution. Validation gives the validation score of the starter and the best validation score reached during the run. RHAE is the official ARC-AGI-3 score, which rates each completed level by the agent's action efficiency relative to human players, weights the levels of a game by their number, and averages over games; it lies on a 0--100 scale and is not rescaled.}
  \label{tab:res-arc3}
  \setlength{\tabcolsep}{3pt}
  \begin{fittabular}{@{}rllrrrrl@{}}
    \toprule
    \multicolumn{3}{@{}l}{\colhead{Model}} & \multicolumn{2}{c}{\colhead{Hidden test metrics}} & \colhead{Validation} & \multicolumn{2}{c}{\colhead{Run}} \\
    \cmidrule(r){1-3}\cmidrule(lr){4-5}\cmidrule(lr){6-6}\cmidrule(l){7-8}
    \colhead{\#} & \colhead{Name} & \colhead{Weights} & \phead{RHAE} & \colhead{Starter} & \colhead{Start $\to$ best} & \colhead{Rounds} & \colhead{Ended by} \\
    \midrule
    \weak{1} & \mlogo{gpt}GPT-6.1 Sol & \wclosed & \pbest{44.2} & \weak{25.4} & \weak{29.8 $\to$} 48.2 & 10 & \weak{round limit} \\\rowrule
    \weak{2} & \mlogo{claude}Claude Opus 5.5 & \wclosed & \pcol{40.4} & \weak{21.6} & \weak{33.0 $\to$} 49.3 & 10 & \weak{round limit} \\\rowrule
    \weak{3} & \mlogo{gpt}GPT-6 Astra & \wclosed & \pcol{36.5} & \weak{36.5} & \weak{39.0 $\to$} 39.0 & 9 & \weak{no further progress} \\\rowrule
    \weak{4} & \mlogo{qwen}Qwen3.8-Flash-Next & \wopen & \pcol{0.9} & \weak{0.0} & \weak{0.0 $\to$} 4.1 & 7 & \weak{time budget} \\\rowrule
    \weak{5} & \mlogo{deepseek}DeepSeek-V4.1 Flash & \wopen & \pcol{0.6} & \weak{0.6} & \weak{3.8 $\to$} 3.8 & 3 & \weak{time budget} \\\rowrule
    \weak{6} & \mlogo{qwen}Qwen3.8-27B & \wopen & \pcol{0.3} & \weak{0.0} & \weak{0.0 $\to$} 0.3 & 7 & \weak{time budget} \\\rowrule
    \weak{7} & \mlogo{gemma}Gemma-4-31B-it & \wopen & \pcol{0.0} & \weak{0.0} & \weak{0.0 $\to$} 0.0 & 2 & \weak{time budget} \\
    \bottomrule
  \end{fittabular}
\end{table}

\begin{table}[p]
  \centering\footnotesize
  \caption{\textbf{Image Editing.} Instruction-based image editing with any open model or tool, covering local, global, text, retouching and restoration, reasoning, and compound edits. Models are ranked by Overall on the hidden test split. Test metrics are those of the final solution, except Starter, which is the primary test score of the starter solution. Validation gives the validation score of the starter and the best validation score reached during the run. Overall is the mean score over six families of edits, each edit scored by a vision-language judge and, where applicable, by programmatic checks: Local (a single object), Global (style and global changes), Text, Retouch (retouching and restoration), Reasoning (edits that require inference), and Compound (several edits in one instruction). All scores are multiplied by 100.}
  \label{tab:res-imgedit}
  \setlength{\tabcolsep}{2.5pt}
  \begin{fittabular}{@{}rllrrrrrrrrrrl@{}}
    \toprule
    \multicolumn{3}{@{}l}{\colhead{Model}} & \multicolumn{8}{c}{\colhead{Hidden test metrics}} & \colhead{Validation} & \multicolumn{2}{c}{\colhead{Run}} \\
    \cmidrule(r){1-3}\cmidrule(lr){4-11}\cmidrule(lr){12-12}\cmidrule(l){13-14}
    \colhead{\#} & \colhead{Name} & \colhead{Weights} & \phead{Overall} & \colhead{Starter} & \rothead{Local} & \rothead{Global} & \rothead{Text} & \rothead{Retouch} & \rothead{Reasoning} & \rothead{Compound} & \colhead{Start $\to$ best} & \colhead{Rounds} & \colhead{Ended by} \\
    \midrule
    \weak{1} & \mlogo{gemini}Gemini 4 Argon & \wclosed & \pbest{74.2} & \weak{55.5} & 80.2 & 72.2 & 87.8 & 62.4 & \best{73.7} & 68.7 & \weak{53.4 $\to$} 73.9 & 15 & \weak{round limit} \\\rowrule
    \weak{2} & \mlogo{claude}Claude Opus 5.5 & \wclosed & \pcol{73.8} & \weak{56.2} & 78.2 & 73.4 & 88.0 & \best{66.2} & 66.9 & 70.3 & \weak{54.8 $\to$} 77.1 & 15 & \weak{round limit} \\\rowrule
    \weak{3} & \mlogo{qwen}Qwen3.8-Flash-Next & \wopen & \pcol{70.0} & \weak{56.2} & \best{85.0} & 75.2 & 87.4 & 60.8 & 42.1 & 69.3 & \weak{57.6 $\to$} 67.9 & 15 & \weak{round limit} \\\rowrule
    \weak{4} & \mlogo{gpt}GPT-6.1 Sol & \wclosed & \pcol{69.9} & \weak{56.2} & 82.8 & 76.8 & \best{90.6} & 62.7 & 42.8 & 63.7 & \weak{55.6 $\to$} 71.1 & 15 & \weak{round limit} \\\rowrule
    \weak{5} & \mlogo{qwen}Qwen3.8-27B & \wopen & \pcol{69.7} & \weak{55.7} & 70.8 & 70.2 & 86.4 & 53.7 & 66.6 & 70.5 & \weak{56.4 $\to$} 64.8 & 15 & \weak{round limit} \\\rowrule
    \weak{6} & \mlogo{gpt}GPT-6 Astra & \wclosed & \pcol{69.7} & \weak{56.5} & 83.1 & 73.5 & 87.1 & 58.7 & 34.7 & \best{81.1} & \weak{58.0 $\to$} 68.9 & 15 & \weak{round limit} \\\rowrule
    \weak{7} & \mlogo{gemma}Gemma-4-31B-it & \wopen & \pcol{68.4} & \weak{56.1} & 76.0 & 73.3 & 85.7 & 62.9 & 44.3 & 68.4 & \weak{57.4 $\to$} 67.8 & 15 & \weak{round limit} \\\rowrule
    \weak{8} & \mlogo{deepseek}DeepSeek-V4.1 Flash & \wopen & \pcol{67.6} & \weak{56.2} & 78.2 & \best{79.3} & 77.4 & 53.9 & 40.2 & 76.6 & \weak{57.6 $\to$} 69.9 & 15 & \weak{round limit} \\\rowrule
    \weak{9} & \mlogo{gemini}Gemini 3.8 Flash & \wclosed & \pcol{67.2} & \weak{56.7} & 78.1 & 70.5 & 85.6 & 60.7 & 31.5 & 76.8 & \weak{53.4 $\to$} 68.1 & 15 & \weak{round limit} \\\rowrule
    \weak{10} & \mlogo{claude}Claude Sonnet 5.5 & \wclosed & \pcol{66.0} & \weak{57.0} & 77.3 & 71.0 & 86.0 & 61.4 & 41.1 & 59.0 & \weak{53.4 $\to$} 65.1 & 15 & \weak{round limit} \\\rowrule
    \weak{11} & \mlogo{minimax}MiniMax-M3 & \wopen & \pcol{64.3} & \weak{55.6} & 70.3 & 69.0 & 83.3 & 57.4 & 33.3 & 72.4 & \weak{57.6 $\to$} 63.2 & 15 & \weak{round limit} \\\rowrule
    \weak{12} & \mlogo{glm}GLM-5.3 Flash & \wopen & \pcol{60.6} & \weak{56.6} & 69.1 & 65.0 & 80.4 & 53.7 & 29.1 & 66.1 & \weak{57.6 $\to$} 63.9 & 15 & \weak{round limit} \\
    \bottomrule
  \end{fittabular}
\end{table}

\begin{table}[p]
  \centering\footnotesize
  \caption{\textbf{Multi-turn Editing.} Edit a photo over five turns, each instruction revealed only at its turn and applied to the previous result (MSE-Bench). Models are ranked by Score on the hidden test split. Test metrics are those of the final solution, except Starter, which is the primary test score of the starter solution. Validation gives the validation score of the starter and the best validation score reached during the run. Score is the mean per-turn score of a vision-language judge over the five turns of each session; Turn~1, Turn~3, and Turn~5 are the fractions of sessions in which every turn up to the first, third, and fifth is judged successful. All scores are multiplied by 100.}
  \label{tab:res-mtedit}
  \setlength{\tabcolsep}{3pt}
  \begin{fittabular}{@{}rllrrrrrrrl@{}}
    \toprule
    \multicolumn{3}{@{}l}{\colhead{Model}} & \multicolumn{5}{c}{\colhead{Hidden test metrics}} & \colhead{Validation} & \multicolumn{2}{c}{\colhead{Run}} \\
    \cmidrule(r){1-3}\cmidrule(lr){4-8}\cmidrule(lr){9-9}\cmidrule(l){10-11}
    \colhead{\#} & \colhead{Name} & \colhead{Weights} & \phead{Score} & \colhead{Starter} & \colhead{Turn 1} & \colhead{Turn 3} & \colhead{Turn 5} & \colhead{Start $\to$ best} & \colhead{Rounds} & \colhead{Ended by} \\
    \midrule
    \weak{1} & \mlogo{gemini}Gemini 4 Argon & \wclosed & \pbest{71.1} & \weak{64.8} & \best{95.0} & 62.5 & \best{50.0} & \weak{68.0 $\to$} 72.2 & 6 & \weak{time budget} \\\rowrule
    \weak{2} & \mlogo{claude}Claude Opus 5.5 & \wclosed & \pcol{70.3} & \weak{64.6} & 85.0 & 60.0 & 37.5 & \weak{68.1 $\to$} 72.3 & 15 & \weak{round limit} \\\rowrule
    \weak{3} & \mlogo{qwen}Qwen3.8-27B & \wopen & \pcol{70.2} & \weak{--} & 87.5 & 60.0 & 47.5 & \weak{68.2 $\to$} 70.8 & 8 & \weak{time budget} \\\rowrule
    \weak{4} & \mlogo{claude}Claude Sonnet 5.5 & \wclosed & \pcol{69.8} & \weak{65.1} & 87.5 & 57.5 & 42.5 & \weak{68.3 $\to$} 72.1 & 7 & \weak{time budget} \\\rowrule
    \weak{5} & \mlogo{qwen}Qwen3.8-Flash-Next & \wopen & \pcol{69.2} & \weak{64.8} & 90.0 & 60.0 & 42.5 & \weak{68.3 $\to$} 70.4 & 8 & \weak{time budget} \\\rowrule
    \weak{6} & \mlogo{gpt}GPT-6.1 Sol & \wclosed & \pcol{68.3} & \weak{65.0} & \best{95.0} & 57.5 & 42.5 & \weak{68.4 $\to$} 72.1 & 7 & \weak{time budget} \\\rowrule
    \weak{7} & \mlogo{gpt}GPT-6 Astra & \wclosed & \pcol{67.8} & \weak{65.1} & \best{95.0} & \best{67.5} & \best{50.0} & \weak{68.0 $\to$} 71.2 & 15 & \weak{round limit} \\\rowrule
    \weak{8} & \mlogo{gemma}Gemma-4-31B-it & \wopen & \pcol{66.3} & \weak{65.0} & \best{95.0} & \best{67.5} & 35.0 & \weak{68.3 $\to$} 69.9 & 6 & \weak{time budget} \\
    \bottomrule
  \end{fittabular}
\end{table}

\begin{table}[p]
  \centering\footnotesize
  \caption{\textbf{Video Generation.} Text-to-video workflows around a frozen video model (MiniMax H3, 480p), scored on compositional prompts: attribute and action binding, numeracy, spatial relations (T2V-CompBench). Models are ranked by Score on the hidden test split. Test metrics are those of the final solution, except Starter, which is the primary test score of the starter solution. Validation gives the validation score of the starter and the best validation score reached during the run. Score is the mean of the T2V-CompBench scores for attribute binding, action binding, numeracy, and spatial relations. All scores are multiplied by 100.}
  \label{tab:res-videogen}
  \setlength{\tabcolsep}{3pt}
  \begin{fittabular}{@{}rllrrrrl@{}}
    \toprule
    \multicolumn{3}{@{}l}{\colhead{Model}} & \multicolumn{2}{c}{\colhead{Hidden test metrics}} & \colhead{Validation} & \multicolumn{2}{c}{\colhead{Run}} \\
    \cmidrule(r){1-3}\cmidrule(lr){4-5}\cmidrule(lr){6-6}\cmidrule(l){7-8}
    \colhead{\#} & \colhead{Name} & \colhead{Weights} & \phead{Score} & \colhead{Starter} & \colhead{Start $\to$ best} & \colhead{Rounds} & \colhead{Ended by} \\
    \midrule
    \weak{1} & \mlogo{claude}Claude Opus 5.5 & \wclosed & \pbest{93.9} & \weak{81.7} & \weak{68.1 $\to$} 92.0 & 37 & \weak{no further progress} \\\rowrule
    \weak{2} & \mlogo{gpt}GPT-6.1 Sol & \wclosed & \pcol{91.5} & \weak{81.7} & \weak{68.1 $\to$} 87.4 & 13 & \weak{no further progress} \\\rowrule
    \weak{3} & \mlogo{qwen}Qwen3.8-Flash-Next & \wopen & \pcol{91.4} & \weak{80.6} & \weak{69.1 $\to$} 90.7 & 6 & \weak{time budget} \\\rowrule
    \weak{4} & \mlogo{minimax}MiniMax-M3 & \wopen & \pcol{87.3} & \weak{80.6} & \weak{69.1 $\to$} 76.5 & 5 & \weak{time budget} \\\rowrule
    \weak{5} & \mlogo{gemma}Gemma-4-31B-it & \wopen & \pcol{87.1} & \weak{80.6} & \weak{69.1 $\to$} 77.8 & 4 & \weak{time budget} \\\rowrule
    \weak{6} & \mlogo{qwen}Qwen3.8-27B & \wopen & \pcol{81.8} & \weak{80.6} & \weak{69.1 $\to$} 71.9 & 5 & \weak{time budget} \\\rowrule
    \weak{7} & \mlogo{glm}GLM-5.3 Flash & \wopen & \pcol{79.6} & \weak{80.6} & \weak{69.1 $\to$} 82.0 & 10 & \weak{time budget} \\
    \bottomrule
  \end{fittabular}
\end{table}

\begin{table}[p]
  \centering\footnotesize
  \caption{\textbf{3D Tracking.} 3D instance tracking (IGGT). Models are ranked by T-mIoU on the hidden test split. Test metrics are those of the final solution, except Starter, which is the primary test score of the starter solution. Validation gives the validation score of the starter and the best validation score reached during the run. T-mIoU is the mean intersection over union of the predicted instance tracks; T-SR is the fraction of tracks recovered successfully; $\mathcal{J}$, $\mathcal{F}$, and $\mathcal{J}\&\mathcal{F}$ are region similarity, contour accuracy, and their mean. All scores are multiplied by 100.}
  \label{tab:res-iggt}
  \setlength{\tabcolsep}{3pt}
  \begin{fittabular}{@{}rllrrrrrrrrl@{}}
    \toprule
    \multicolumn{3}{@{}l}{\colhead{Model}} & \multicolumn{6}{c}{\colhead{Hidden test metrics}} & \colhead{Validation} & \multicolumn{2}{c}{\colhead{Run}} \\
    \cmidrule(r){1-3}\cmidrule(lr){4-9}\cmidrule(lr){10-10}\cmidrule(l){11-12}
    \colhead{\#} & \colhead{Name} & \colhead{Weights} & \phead{T-mIoU} & \colhead{Starter} & \colhead{T-SR} & \colhead{$\mathcal{J}$} & \colhead{$\mathcal{F}$} & \colhead{$\mathcal{J}\&\mathcal{F}$} & \colhead{Start $\to$ best} & \colhead{Rounds} & \colhead{Ended by} \\
    \midrule
    \weak{1} & \mlogo{claude}Claude Opus 5.5 & \wclosed & \pbest{79.6} & \weak{21.2} & 65.0 & \best{76.1} & \best{73.2} & \best{74.7} & \weak{19.4 $\to$} 84.7 & 15 & \weak{round limit} \\\rowrule
    \weak{2} & \mlogo{gpt}GPT-6.1 Sol & \wclosed & \pcol{77.3} & \weak{21.2} & 60.4 & 74.5 & 71.1 & 72.8 & \weak{19.4 $\to$} 80.2 & 9 & \weak{time budget} \\\rowrule
    \weak{3} & \mlogo{gemini}Gemini 4 Argon & \wclosed & \pcol{73.8} & \weak{21.2} & \best{69.6} & 70.6 & 65.9 & 68.3 & \weak{19.4 $\to$} 78.6 & 8 & \weak{time budget} \\\rowrule
    \weak{4} & \mlogo{glm}GLM-5.3 Flash & \wopen & \pcol{61.2} & \weak{21.2} & 27.5 & 54.6 & 52.7 & 53.6 & \weak{19.4 $\to$} 64.5 & 12 & \weak{time budget} \\\rowrule
    \weak{5} & \mlogo{deepseek}DeepSeek-V4.1 Flash & \wopen & \pcol{60.2} & \weak{21.2} & 30.8 & 54.5 & 53.8 & 54.2 & \weak{19.4 $\to$} 70.1 & 9 & \weak{time budget} \\\rowrule
    \weak{6} & \mlogo{qwen}Qwen3.8-Flash-Next & \wopen & \pcol{58.9} & \weak{21.2} & 30.8 & 54.6 & 51.8 & 53.2 & \weak{19.4 $\to$} 65.7 & 12 & \weak{time budget} \\\rowrule
    \weak{7} & \mlogo{meta}Muse Wonder 1.0 & \wclosed & \pcol{57.9} & \weak{21.2} & 25.8 & 53.1 & 50.8 & 51.9 & \weak{19.4 $\to$} 62.2 & 6 & \weak{time budget} \\\rowrule
    \weak{8} & \mlogo{gemini}Gemini 3.8 Flash & \wclosed & \pcol{53.7} & \weak{21.2} & 22.1 & 46.4 & 46.2 & 46.3 & \weak{19.4 $\to$} 54.0 & 12 & \weak{time budget} \\\rowrule
    \weak{9} & \mlogo{gpt}GPT-6 Astra & \wclosed & \pcol{52.8} & \weak{21.2} & 20.4 & 48.5 & 46.5 & 47.5 & \weak{19.4 $\to$} 57.3 & 15 & \weak{round limit} \\\rowrule
    \weak{10} & \mlogo{qwen}Qwen3.8-27B & \wopen & \pcol{52.4} & \weak{21.2} & 12.5 & 40.8 & 38.3 & 39.6 & \weak{19.4 $\to$} 53.9 & 9 & \weak{time budget} \\\rowrule
    \weak{11} & \mlogo{claude}Claude Sonnet 5.5 & \wclosed & \pcol{47.9} & \weak{21.2} & 23.8 & 47.6 & 45.7 & 46.7 & \weak{19.4 $\to$} 50.1 & 13 & \weak{time budget} \\\rowrule
    \weak{12} & \mlogo{gemma}Gemma-4-31B-it & \wopen & \pcol{32.0} & \weak{21.2} & 5.4 & 26.2 & 21.4 & 23.8 & \weak{19.4 $\to$} 29.1 & 7 & \weak{time budget} \\\rowrule
    \weak{13} & \mlogo{minimax}MiniMax-M3 & \wopen & \pcol{27.9} & \weak{21.2} & 1.7 & 23.1 & 21.5 & 22.3 & \weak{19.4 $\to$} 29.4 & 15 & \weak{time budget} \\
    \bottomrule
  \end{fittabular}
\end{table}

\begin{table}[p]
  \centering\footnotesize
  \caption{\textbf{Scene from Photo.} Image-to-3D: rebuild a furnished room from one photo as a Blender scene, every piece of furniture a whole 3D object (HARMONY300). Models are ranked by Score on the hidden test split. Test metrics are those of the final solution, except Starter, which is the primary test score of the starter solution. Validation gives the validation score of the starter and the best validation score reached during the run. Score is the mean of three terms: Appearance compares renderings of the scene with the photo, Objects measures how well each annotated object is recovered as a complete and correctly placed object, and Geometry compares the scene geometry with the ground truth. $^\dagger$GPT-6 Astra runs at its highest reasoning effort, whereas every other run uses the standard effort; it is shown for comparison and not ranked. All scores are multiplied by 100.}
  \label{tab:res-photo}
  \setlength{\tabcolsep}{3pt}
  \begin{fittabular}{@{}rllrrrrrrrl@{}}
    \toprule
    \multicolumn{3}{@{}l}{\colhead{Model}} & \multicolumn{5}{c}{\colhead{Hidden test metrics}} & \colhead{Validation} & \multicolumn{2}{c}{\colhead{Run}} \\
    \cmidrule(r){1-3}\cmidrule(lr){4-8}\cmidrule(lr){9-9}\cmidrule(l){10-11}
    \colhead{\#} & \colhead{Name} & \colhead{Weights} & \phead{Score} & \colhead{Starter} & \colhead{Appearance} & \colhead{Objects} & \colhead{Geometry} & \colhead{Start $\to$ best} & \colhead{Rounds} & \colhead{Ended by} \\
    \midrule
    \weak{1} & \mlogo{claude}Claude Opus 5.5 & \wclosed & \pbest{66.5} & \weak{7.1} & \best{94.8} & 62.5 & \best{42.3} & \weak{7.7 $\to$} 67.8 & 15 & \weak{round limit} \\\rowrule
    \weak{2} & \mlogo{gpt}GPT-6 Astra & \wclosed & \pcol{53.2} & \weak{7.1} & 63.5 & \best{62.5} & 33.8 & \weak{7.7 $\to$} 54.8 & 15 & \weak{round limit} \\\rowrule
    \weak{3} & \mlogo{qwen}Qwen3.8-Flash-Next & \wopen & \pcol{40.0} & \weak{7.1} & 78.7 & 16.5 & 24.7 & \weak{7.7 $\to$} 43.2 & 10 & \weak{time budget} \\\rowrule
    \weak{4} & \mlogo{minimax}MiniMax-M3 & \wopen & \pcol{33.3} & \weak{7.1} & 81.8 & 0.2 & 17.8 & \weak{7.7 $\to$} 32.9 & 15 & \weak{round limit} \\\rowrule
    \weak{5} & \mlogo{qwen}Qwen3.8-27B & \wopen & \pcol{29.0} & \weak{7.1} & 63.0 & 4.4 & 19.7 & \weak{7.7 $\to$} 28.6 & 11 & \weak{time budget} \\\rowrule
    \weak{6} & \mlogo{gemma}Gemma-4-31B-it & \wopen & \pcol{15.4} & \weak{7.1} & 21.4 & 0.4 & 24.4 & \weak{7.7 $\to$} 17.2 & 11 & \weak{time budget} \\\rowrule
    \weak{--} & \mlogo{gpt}GPT-6 Astra$^\dagger$ & \wclosed & \pcol{61.6} & \weak{7.1} & 81.1 & 60.0 & 43.6 & \weak{7.7 $\to$} 63.1 & 12 & \weak{time budget} \\
    \bottomrule
  \end{fittabular}
\end{table}

\begin{table}[p]
  \centering\footnotesize
  \caption{\textbf{Scene from Video.} Video-to-3D: reconstruct an indoor scene from its RGB frames alone as a simulation-ready scene, one textured mesh per object (HoloScene scenes, CoDimRecon setting). Models are ranked by Score on the hidden test split. Test metrics are those of the final solution, except Starter, which is the primary test score of the starter solution. Validation gives the validation score of the starter and the best validation score reached during the run. Score is the mean of three terms: Geometry combines the F1 score and the normal consistency of the reconstructed surfaces, Appearance combines the SSIM and LPIPS of renderings, and Physics is the fraction of objects that remain stable in simulation. $^\dagger$GPT-6 Astra runs at its highest reasoning effort, whereas every other run uses the standard effort; it is shown for comparison and not ranked. All scores are multiplied by 100.}
  \label{tab:res-video}
  \setlength{\tabcolsep}{3pt}
  \begin{fittabular}{@{}rllrrrrrrrl@{}}
    \toprule
    \multicolumn{3}{@{}l}{\colhead{Model}} & \multicolumn{5}{c}{\colhead{Hidden test metrics}} & \colhead{Validation} & \multicolumn{2}{c}{\colhead{Run}} \\
    \cmidrule(r){1-3}\cmidrule(lr){4-8}\cmidrule(lr){9-9}\cmidrule(l){10-11}
    \colhead{\#} & \colhead{Name} & \colhead{Weights} & \phead{Score} & \colhead{Starter} & \colhead{Geometry} & \colhead{Appearance} & \colhead{Physics} & \colhead{Start $\to$ best} & \colhead{Rounds} & \colhead{Ended by} \\
    \midrule
    \weak{1} & \mlogo{claude}Claude Opus 5.5 & \wclosed & \pbest{70.9} & \weak{26.7} & \best{68.8} & \best{73.2} & \best{70.7} & \weak{17.3 $\to$} 27.9 & 15 & \weak{round limit} \\\rowrule
    \weak{2} & \mlogo{gpt}GPT-6 Astra & \wclosed & \pcol{53.4} & \weak{26.7} & 66.3 & 56.3 & 37.7 & \weak{17.3 $\to$} 56.1 & 15 & \weak{round limit} \\\rowrule
    \weak{3} & \mlogo{qwen}Qwen3.8-Flash-Next & \wopen & \pcol{34.5} & \weak{26.7} & 54.7 & 48.7 & 0.0 & \weak{17.3 $\to$} 17.3 & 15 & \weak{round limit} \\\rowrule
    \weak{4} & \mlogo{minimax}MiniMax-M3 & \wopen & \pcol{26.7} & \weak{26.7} & 32.4 & 47.7 & 0.0 & \weak{17.3 $\to$} 17.3 & 15 & \weak{round limit} \\\rowrule
    \weak{4} & \mlogo{qwen}Qwen3.8-27B & \wopen & \pcol{26.7} & \weak{26.7} & 32.4 & 47.7 & 0.0 & \weak{17.3 $\to$} 17.3 & 11 & \weak{no further progress} \\\rowrule
    \weak{6} & \mlogo{gemma}Gemma-4-31B-it & \wopen & \pcol{17.5} & \weak{26.7} & 28.6 & 24.1 & 0.0 & \weak{17.3 $\to$} 19.7 & 12 & \weak{validation budget} \\\rowrule
    \weak{--} & \mlogo{gpt}GPT-6 Astra$^\dagger$ & \wclosed & \pcol{54.8} & \weak{26.7} & 70.5 & 58.6 & 35.4 & \weak{17.3 $\to$} 47.8 & 12 & \weak{time budget} \\
    \bottomrule
  \end{fittabular}
\end{table}

\begin{table}[p]
  \centering\footnotesize
  \caption{\textbf{Visual Search.} High-resolution visual search (V*Bench, HR-Bench 8K). Models are ranked by Accuracy on the hidden test split. Test metrics are those of the final solution, except Starter, which is the primary test score of the starter solution. Validation gives the validation score of the starter and the best validation score reached during the run. Accuracy is measured on the whole test split; V*Bench and HR-Bench 8K give the accuracy on the questions from each source benchmark. All scores are multiplied by 100. $^\dagger$Muse Wonder~1.0 was added after the other runs; it is shown for comparison and not ranked, and its starter solution has no test score.}
  \label{tab:res-vsearch}
  \setlength{\tabcolsep}{3pt}
  \begin{fittabular}{@{}rllrrrrrrl@{}}
    \toprule
    \multicolumn{3}{@{}l}{\colhead{Model}} & \multicolumn{4}{c}{\colhead{Hidden test metrics}} & \colhead{Validation} & \multicolumn{2}{c}{\colhead{Run}} \\
    \cmidrule(r){1-3}\cmidrule(lr){4-7}\cmidrule(lr){8-8}\cmidrule(l){9-10}
    \colhead{\#} & \colhead{Name} & \colhead{Weights} & \phead{Accuracy} & \colhead{Starter} & \colhead{V*Bench} & \colhead{HR-Bench 8K} & \colhead{Start $\to$ best} & \colhead{Rounds} & \colhead{Ended by} \\
    \midrule
    \weak{1} & \mlogo{gemini}Gemini 4 Argon & \wclosed & \pbest{98.6} & \weak{87.9} & \best{100.0} & \best{97.1} & \weak{87.5 $\to$} 98.8 & 10 & \weak{round limit} \\\rowrule
    \weak{2} & \mlogo{claude}Claude Opus 5.5 & \wclosed & \pcol{96.4} & \weak{85.0} & 98.6 & 94.3 & \weak{85.0 $\to$} 95.0 & 10 & \weak{round limit} \\\rowrule
    \weak{2} & \mlogo{gpt}GPT-6.1 Sol & \wclosed & \pcol{96.4} & \weak{89.3} & 95.7 & \best{97.1} & \weak{87.5 $\to$} 93.8 & 10 & \weak{round limit} \\\rowrule
    \weak{4} & \mlogo{gemini}Gemini 3.8 Flash & \wclosed & \pcol{95.0} & \weak{90.0} & 97.1 & 92.9 & \weak{88.8 $\to$} 95.0 & 10 & \weak{round limit} \\\rowrule
    \weak{5} & \mlogo{gpt}GPT-6 Astra & \wclosed & \pcol{93.6} & \weak{84.3} & 94.3 & 92.9 & \weak{86.2 $\to$} 91.2 & 10 & \weak{round limit} \\\rowrule
    \weak{6} & \mlogo{qwen}Qwen3.8-Flash-Next & \wopen & \pcol{88.6} & \weak{78.6} & 97.1 & 80.0 & \weak{78.8 $\to$} 87.5 & 10 & \weak{round limit} \\\rowrule
    \weak{7} & \mlogo{qwen}Qwen3.8-27B & \wopen & \pcol{85.7} & \weak{78.6} & 90.0 & 81.4 & \weak{67.5 $\to$} 80.0 & 10 & \weak{round limit} \\\rowrule
    \weak{7} & \mlogo{glm}GLM-5.3 Flash & \wopen & \pcol{85.7} & \weak{76.4} & 90.0 & 81.4 & \weak{78.8 $\to$} 83.8 & 10 & \weak{round limit} \\\rowrule
    \weak{9} & \mlogo{claude}Claude Sonnet 5.5 & \wclosed & \pcol{80.7} & \weak{74.3} & 85.7 & 75.7 & \weak{81.2 $\to$} 86.2 & 10 & \weak{round limit} \\\rowrule
    \weak{10} & \mlogo{deepseek}DeepSeek-V4.1 Flash & \wopen & \pcol{77.1} & \weak{75.7} & 80.0 & 74.3 & \weak{70.0 $\to$} 81.2 & 10 & \weak{round limit} \\\rowrule
    \weak{11} & \mlogo{gemma}Gemma-4-31B-it & \wopen & \pcol{72.9} & \weak{57.1} & 81.4 & 64.3 & \weak{57.5 $\to$} 77.5 & 10 & \weak{round limit} \\\rowrule
    \weak{12} & \mlogo{minimax}MiniMax-M3 & \wopen & \pcol{70.7} & \weak{62.1} & 77.1 & 64.3 & \weak{61.3 $\to$} 76.2 & 10 & \weak{round limit} \\\rowrule
    \weak{--} & \mlogo{meta}Muse Wonder 1.0$^\dagger$ & \wclosed & \pcol{56.4} & \weak{--} & 55.7 & 57.1 & \weak{53.8 $\to$} 85.0 & 10 & \weak{round limit} \\
    \bottomrule
  \end{fittabular}
\end{table}

\FloatBarrier
\begin{table}[!t]
  \centering\footnotesize
  \caption{\textbf{Video Quality.} Perceptual video quality: score real user videos (YouTube-UGC MOS) and pick the better of AI-generated pairs (Veo 3.1 human votes). Models are ranked by Score on the hidden test split. Test metrics are those of the final solution, except Starter, which is the primary test score of the starter solution. Validation gives the validation score of the starter and the best validation score reached during the run. Score is the mean of SRCC and pairwise accuracy. SRCC and PLCC are the rank and linear correlations between predicted and human quality ratings of real videos; Pair acc.\ is the accuracy in choosing the better of two generated videos. All scores are multiplied by 100.}
  \label{tab:res-vq}
  \setlength{\tabcolsep}{3pt}
  \begin{fittabular}{@{}rllrrrrrrrl@{}}
    \toprule
    \multicolumn{3}{@{}l}{\colhead{Model}} & \multicolumn{5}{c}{\colhead{Hidden test metrics}} & \colhead{Validation} & \multicolumn{2}{c}{\colhead{Run}} \\
    \cmidrule(r){1-3}\cmidrule(lr){4-8}\cmidrule(lr){9-9}\cmidrule(l){10-11}
    \colhead{\#} & \colhead{Name} & \colhead{Weights} & \phead{Score} & \colhead{Starter} & \colhead{SRCC} & \colhead{PLCC} & \colhead{Pair acc.} & \colhead{Start $\to$ best} & \colhead{Rounds} & \colhead{Ended by} \\
    \midrule
    \weak{1} & \mlogo{gpt}GPT-6 Astra & \wclosed & \pbest{82.1} & \weak{82.6} & 86.7 & 90.4 & \best{77.5} & \weak{84.4 $\to$} 84.4 & 10 & \weak{round limit} \\\rowrule
    \weak{2} & \mlogo{claude}Claude Opus 5.5 & \wclosed & \pcol{82.0} & \weak{80.9} & 86.5 & 89.7 & \best{77.5} & \weak{79.9 $\to$} 85.1 & 10 & \weak{round limit} \\\rowrule
    \weak{3} & \mlogo{deepseek}DeepSeek-V4.1 Flash & \wopen & \pcol{82.0} & \weak{79.1} & 90.5 & 89.7 & 73.5 & \weak{67.3 $\to$} 82.1 & 7 & \weak{validation budget} \\\rowrule
    \weak{4} & \mlogo{gpt}GPT-6.1 Sol & \wclosed & \pcol{80.8} & \weak{81.8} & 92.3 & \best{94.2} & 69.4 & \weak{80.4 $\to$} 89.8 & 10 & \weak{round limit} \\\rowrule
    \weak{5} & \mlogo{gemini}Gemini 4 Argon & \wclosed & \pcol{80.6} & \weak{77.0} & 85.6 & 89.7 & 75.5 & \weak{79.8 $\to$} 86.4 & 10 & \weak{round limit} \\\rowrule
    \weak{6} & \mlogo{qwen}Qwen3.8-27B & \wopen & \pcol{78.0} & \weak{74.5} & 86.6 & 87.9 & 69.4 & \weak{65.4 $\to$} 78.4 & 10 & \weak{round limit} \\\rowrule
    \weak{7} & \mlogo{gemini}Gemini 3.8 Flash & \wclosed & \pcol{77.5} & \weak{77.5} & 87.7 & 88.0 & 67.3 & \weak{76.9 $\to$} 90.0 & 9 & \weak{validation budget} \\\rowrule
    \weak{8} & \mlogo{qwen}Qwen3.8-Flash-Next & \wopen & \pcol{76.6} & \weak{78.5} & \best{94.0} & 93.5 & 59.2 & \weak{78.2 $\to$} 88.6 & 10 & \weak{round limit} \\\rowrule
    \weak{9} & \mlogo{glm}GLM-5.3 Flash & \wopen & \pcol{76.2} & \weak{71.5} & 89.2 & 89.9 & 63.3 & \weak{74.5 $\to$} 88.8 & 10 & \weak{round limit} \\\rowrule
    \weak{10} & \mlogo{gemma}Gemma-4-31B-it & \wopen & \pcol{74.3} & \weak{77.5} & 81.3 & 82.5 & 67.3 & \weak{79.0 $\to$} 83.1 & 5 & \weak{validation budget} \\\rowrule
    \weak{11} & \mlogo{minimax}MiniMax-M3 & \wopen & \pcol{69.8} & \weak{72.5} & 74.2 & 10.0 & 65.3 & \weak{60.9 $\to$} 73.4 & 10 & \weak{round limit} \\
    \bottomrule
  \end{fittabular}
\end{table}

\newlength{\arctile}
\newlength{\vqh}
\newlength{\eeh}
\providecommand{\qframe}[1]{\includegraphics[width=2.2cm,height=1.65cm]{figures/qual/robo/#1}}
\providecommand{\qhead}{\noindent\hspace*{2.45cm}\parbox{6.66cm}{\centering\scriptsize\textbf{Before}}%
  \hspace{2mm}\parbox{6.66cm}{\centering\scriptsize\textbf{After}}\par\vspace{1mm}}
\providecommand{\qrobo}[8]{\noindent\parbox[t]{2.35cm}{\vspace{0pt}\raggedright\scriptsize\textbf{#1}\\#2\\\textcolor{muted}{#3}}%
  \hspace{1mm}\parbox[t]{6.66cm}{\vspace{0pt}\qframe{#4-before-0}\hspace{0.3mm}\qframe{#4-before-1}\hspace{0.3mm}\qframe{#4-before-2}\\[0.6mm]%
    \scriptsize\textbf{#5.} #6}%
  \hspace{2mm}\parbox[t]{6.66cm}{\vspace{0pt}\qframe{#4-after-0}\hspace{0.3mm}\qframe{#4-after-1}\hspace{0.3mm}\qframe{#4-after-2}\\[0.6mm]%
    \scriptsize\textbf{#7.} #8}\par\vspace{2.2mm}}
\providecommand{\qbox}[3]{\parbox[c][#2][c]{#1}{\centering\includegraphics[width=#1,height=#2,keepaspectratio]{#3}}}
\providecommand{\qedit}[6]{\parbox[t]{8.1cm}{\vspace{0pt}\scriptsize\parbox[t][3\baselineskip][t]{8.1cm}{\textbf{#1.} \textit{#2}}\\[0.4mm]%
  \qbox{2.65cm}{2.2cm}{figures/qual/edit/#3-in}\hfill\qbox{2.65cm}{2.2cm}{figures/qual/edit/#3-before}\hfill\qbox{2.65cm}{2.2cm}{figures/qual/edit/#3-after}\\[0.3mm]%
  \makebox[2.65cm]{Input}\hfill\makebox[2.65cm]{Starter: #4}\hfill\makebox[2.65cm]{Final: #5}\ifx&#6&\else\\[0.3mm]\textcolor{muted}{#6}\fi}}
\providecommand{\qsearch}[4]{\noindent\parbox{\linewidth}{\scriptsize\textbf{#1.} #2}\\[0.5mm]%
  \includegraphics[width=0.497\linewidth]{figures/qual/hstar/#3-before}\hfill\includegraphics[width=0.497\linewidth]{figures/qual/hstar/#3-after}\par\vspace{2mm}}

\section{Additional qualitative examples}
\label{app:qualitative}

This section extends \cref{sec:qualitative} with the remaining examples of the project page, in the order of its leaderboard.

\begin{figure}[p]
  \centering
\qhead
\qrobo{Throw Bottles Into Dustbin}{RoboDojo}{Claude Opus 5.5}{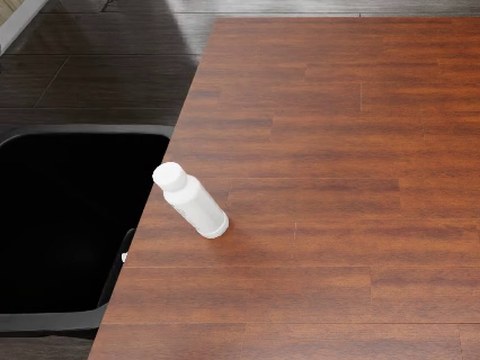}{Round 2, fail, score 40}{The gripper closes beside the upright small bottle and never secures it.}{Round 5, success}{It reaches the same bottle, closes on it and lifts it cleanly.}
\qrobo{Play Xylophone}{RoboDojo}{DeepSeek-V4.1 Flash}{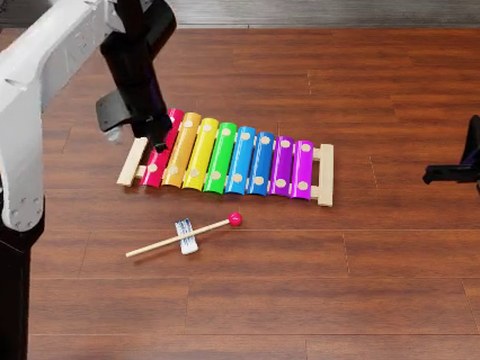}{Round 3, fail, score 0}{The gripper misses the mallet and pokes at the keys with empty fingers.}{Round 6, success}{It picks up the mallet and strikes the keys one after another, from left to right.}
\qrobo{Arrange Largest Number}{RoboDojo}{Claude Opus 5.5}{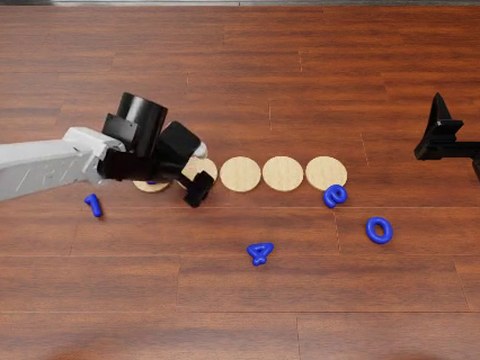}{Round 9, fail}{It picks up the 6 but sets it down beside the row; the second slot stays empty and the row reads 9\_410.}{Round 10, success}{The 6 goes into the second slot and the row reads 96410.}
\qrobo{Push T}{RoboDojo}{Gemini 4 Argon}{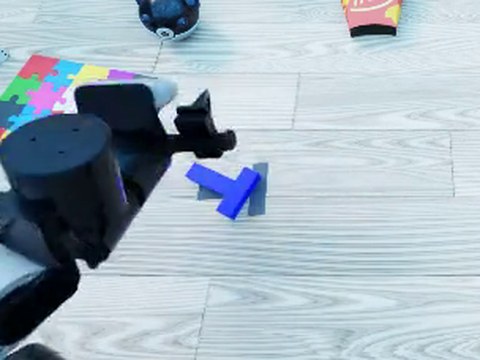}{Round 5, fail, score 0}{After a push near the back of the table the policy stops responding, and the episode ends with the block beside the pad, turned away from it.}{Round 6, success}{It nudges the block in short pushes until it covers the gray pad exactly.}
\qrobo{Sort Nesting Dolls By Size}{RoboDojo}{GPT-6.1 Sol}{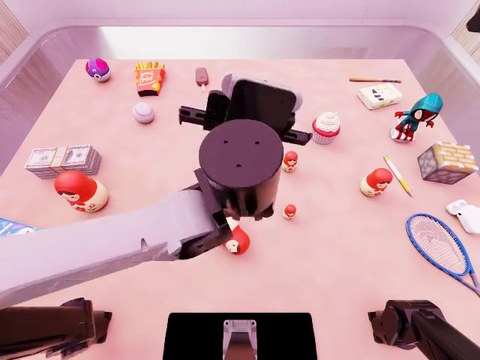}{Round 0 (starter), fail, score 0}{The arms sweep across the table and knock the dolls about; none of them end up in a row.}{Round 9, success}{It picks the dolls up one at a time and stands them in a straight row, smallest on the left.}
\qrobo{Play Tic-Tac-Toe}{RoboDojo}{Muse Wonder 1.0}{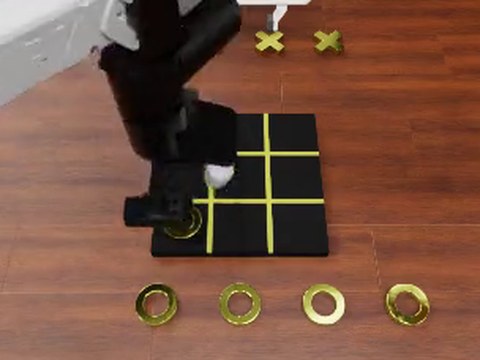}{Round 1, partial, score 75}{It drops its five rings one after another without pausing, so the opponent never places an X and the board is never filled.}{Round 6, success}{After each ring both arms go back to their start pose, the opponent places an X, and play continues until the board is full.}
\qrobo{Insert Tubes}{RoboDojo}{Claude Opus 5.5}{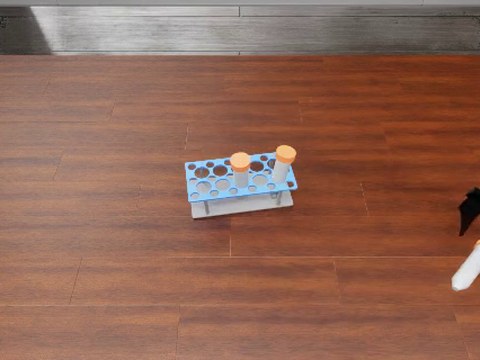}{Round 5, fail}{The third tube enters at an angle and ends lying across the rack.}{Round 7, success}{The third tube goes in upright; all three stand in the rack.}
  \caption{\textbf{RoboDojo, further examples.} Each row shows one episode of a robot task played by the solution of an earlier round (Before) and by that of a later round (After) of the same run, three frames each, cropped to the part of the scene where the two differ. The rows give the benchmark, the model of the run, the round of each solution and the outcome of the episode.}
  \label{fig:qual-robodojo}
\end{figure}

\begin{figure}[p]
  \centering
\qhead
\qrobo{Portion Hot Dogs}{RoboCasa}{GPT-6 Astra}{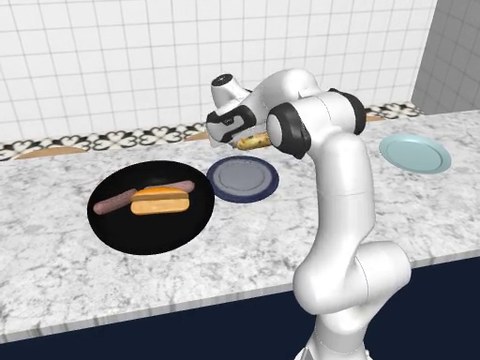}{Round 0 (starter), fail}{The arm pokes at the buns and sausages in the bowl and tips the bowl up; nothing reaches a plate.}{Round 1, success}{With a bun and a sausage already on the first plate, it carries the last sausage over to the second plate.}
\qrobo{Turn On Microwave}{RoboCasa}{Qwen3.8-Flash-Next}{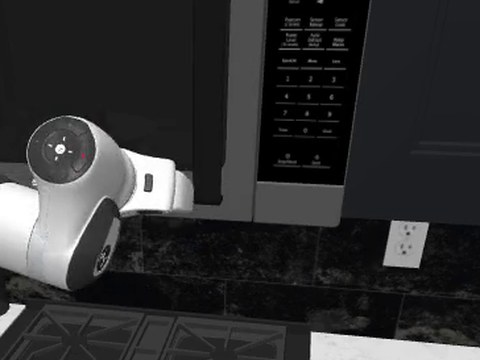}{Round 0 (starter), fail}{The arm swings up past the control panel and pokes at the empty space beside it.}{Round 5, success}{The base turns to face the panel and the fingertip goes straight to the start button at its bottom.}
\qrobo{Turn On Sink Faucet}{RoboCasa}{GLM-5.3 Flash}{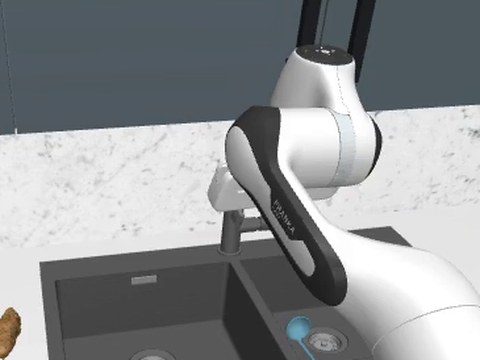}{Round 0 (starter), fail}{The arm never moves: the starting policy is still waiting on the model when the episode's time limit runs out.}{Round 10, success}{It reaches over the basin, takes the lever and lifts it within four seconds.}
\qrobo{Open Cabinet}{RoboCasa}{GPT-6 Astra}{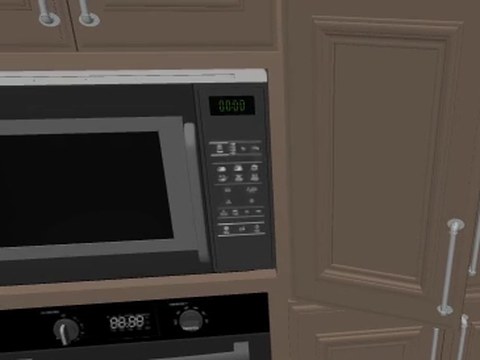}{Round 4, fail}{The gripper keeps reaching for the handles without pulling; the door stays shut.}{Round 7, success}{It hooks the handle and swings the door open.}
\qrobo{Open Drawer}{RoboCasa}{Claude Opus 5.5}{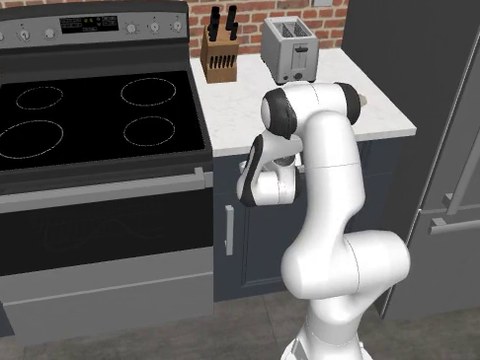}{Round 0 (starter), fail}{The gripper hovers at the side of the drawer front, never takes the handle, and the drawer stays shut.}{Round 1, success}{It drives up to the drawer, closes on the handle and pulls the drawer open.}
\qrobo{Turn Off Stove}{RoboCasa}{GPT-6 Astra}{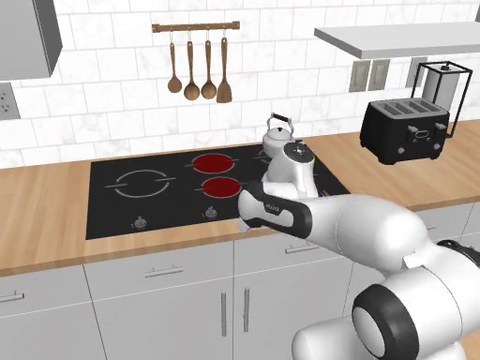}{Round 0 (starter), fail}{Asked to turn the rear center burner off, it turns the wrong knobs: two more burners light up red.}{Round 10, success}{It grips only the rear center burner's knob and turns it; the episode ends as soon as the knob reads off, before the glow has faded.}
  \caption{\textbf{RoboCasa, further examples.} Each row shows one episode of a robot task played by the solution of an earlier round (Before) and by that of a later round (After) of the same run, three frames each, cropped to the part of the scene where the two differ. The rows give the benchmark, the model of the run, the round of each solution and the outcome of the episode.}
  \label{fig:qual-robocasa}
\end{figure}

\begin{figure}[p]
  \centering
\qhead
\qrobo{Stack the right bowl on the left bowl}{RoboLab}{Claude Opus 5.5}{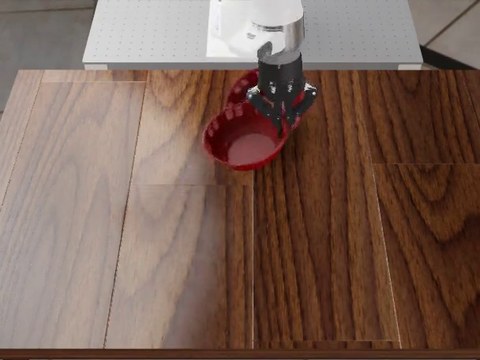}{Round 0 (starter), fail, score 0}{It grabs a bowl by the rim and drags it about, but never sets one inside the other; the bowls end up side by side. It stops when its 40 model calls run out.}{Round 7, success}{It lifts the robot's right bowl (on the image's left: the robot faces the camera) and sets it into the other one.}
\qrobo{Place the smaller object in the grey bin}{RoboLab}{Claude Opus 5.5}{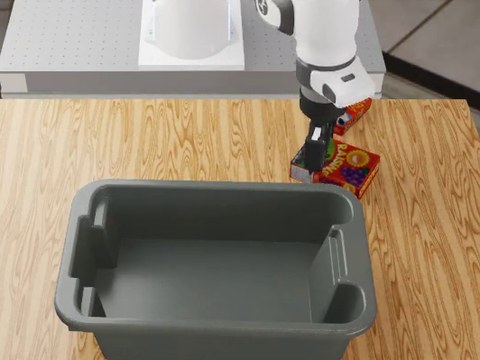}{Round 0 (starter), fail, score 0}{It lifts the butter carton but drops it on the table, then grabs the larger box instead; nothing reaches the bin. It stops when its 40 model calls run out.}{Round 7, success}{It picks the smaller butter carton and lowers it into the grey bin.}
\qrobo{Put the cube and the banana in the bowl}{RoboLab}{Claude Opus 5.5}{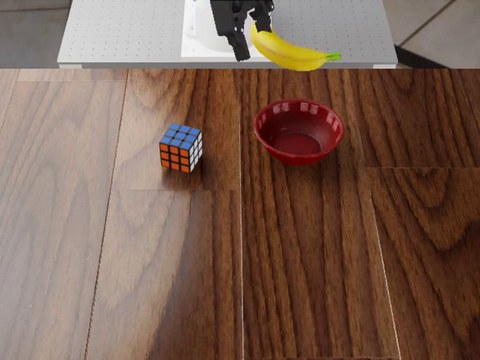}{Round 0 (starter), fail, score 0}{It knocks the cube off the banana, then lifts the banana and carries it away from the bowl; neither ends up inside. It stops when its 40 model calls run out.}{Round 7, success}{It puts the cube in the bowl, then picks up the banana and lowers it in after it.}
\qrobo{Put the cube then the banana in the bowl}{RoboLab}{Claude Opus 5.5}{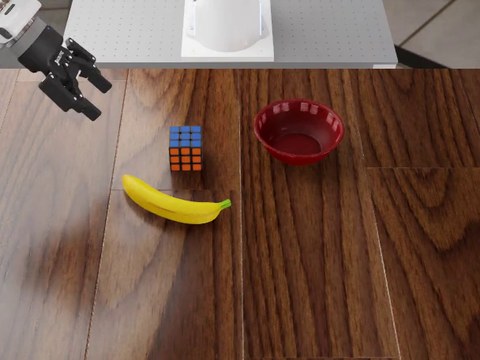}{Round 0 (starter), fail, score 0}{It nudges the cube and the banana, then wanders off to the side; neither goes into the bowl. It stops when its 40 model calls run out.}{Round 7, success}{It puts the cube in the bowl first, then brings the banana and lowers it in on top.}
  \caption{\textbf{RoboLab, further examples.} Each row shows one episode of a robot task played by the solution of an earlier round (Before) and by that of a later round (After) of the same run, three frames each, cropped to the part of the scene where the two differ. The rows give the benchmark, the model of the run, the round of each solution and the outcome of the episode. The RoboLab episodes are development tasks.}
  \label{fig:qual-robolab}
\end{figure}

\begin{figure}[t]
  \centering
\qsearch{Path search}{You want to read a book, where to move?}{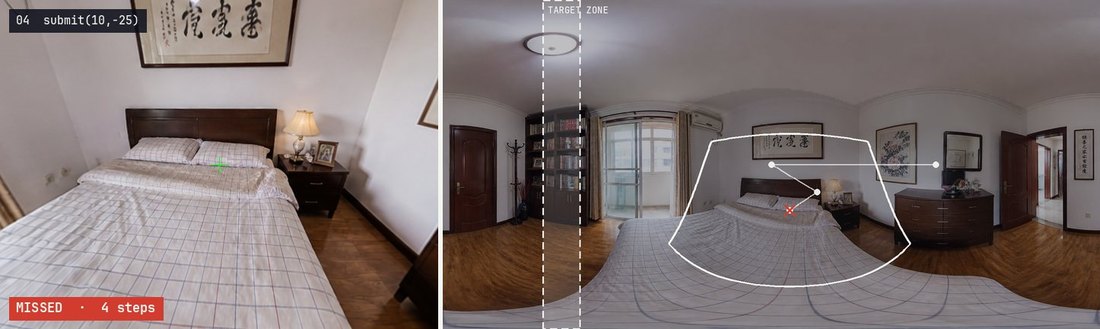}{}
\qsearch{Object search}{Find mannequin.}{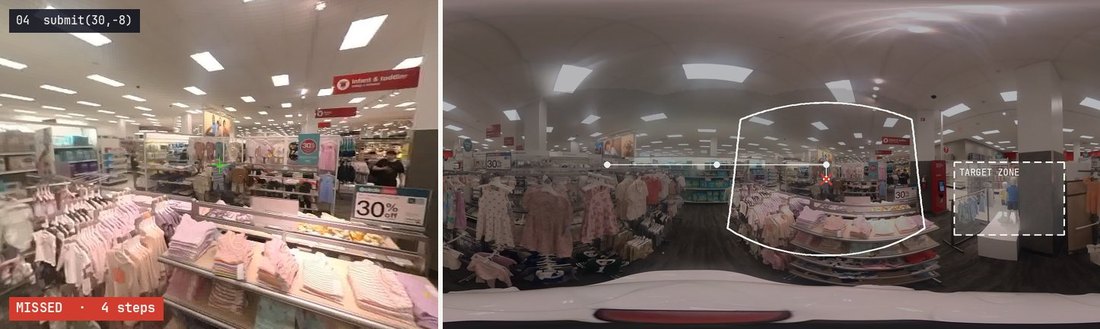}{}
\qsearch{Object search}{Look for the Dog Diapers.}{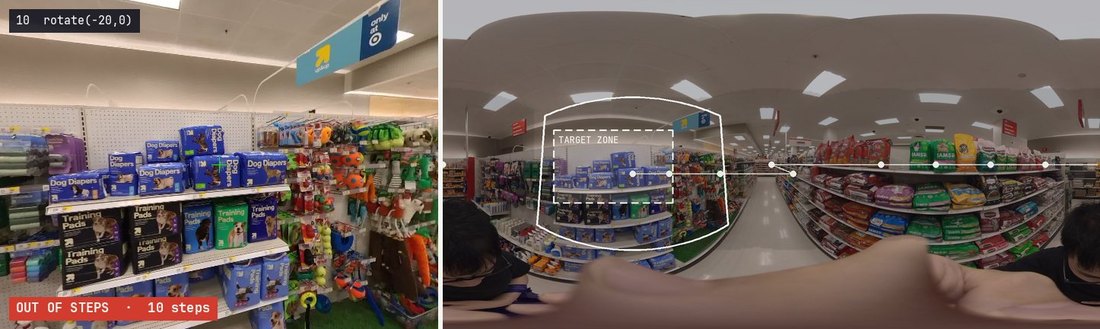}{}
  \caption{\textbf{Active Search, further examples.} Episodes of the starter solution (left of each row) and of the final solution of the run of Claude Opus~5.5 (right). Each image shows the last view of the episode (left part) and the panorama with the path of the view and the target zone (right part); the label gives the outcome and the number of actions, and the top line the last action. Object search asks for a named object, path search for the direction in which to move.}
  \label{fig:qual-search-more}
\end{figure}

\begin{figure}[p]
  \centering
  \setlength{\eeh}{2.05cm}
\noindent\makebox[5.1620\eeh][l]{\scriptsize\textbf{Seen by the starter:} eight evenly spaced times, ego view above and exo view below}\hfill\makebox[1.9203\eeh][l]{\scriptsize\textbf{Chosen by the final solution}}\par\vspace{1mm}
\noindent\parbox{\linewidth}{\scriptsize\raggedright \textbf{Cooking Scrambled Eggs, object state.} While the man in the dark short-sleeve shirt was wiping his hands at the kitchen sink, what was the state of the skillet on the stovetop island? (A) Full of clear water; (B) Covered with chopped vegetables; (C) Filled with mixed egg mixture; (D) Empty and dry\enspace Answer C; starter D, final C.}\\[0.8mm]\includegraphics[height=\eeh]{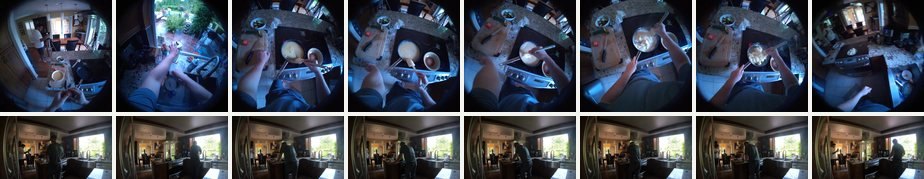}\hfill\includegraphics[height=\eeh]{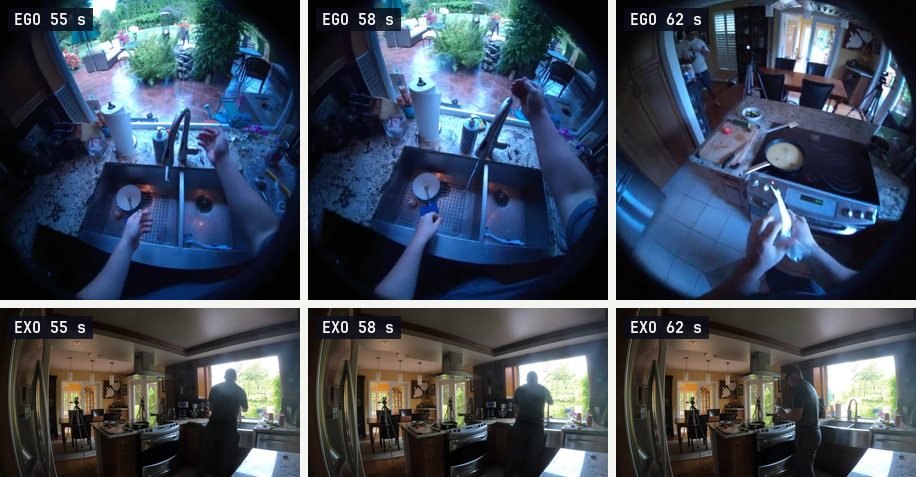}\par\vspace{3mm}
\noindent\parbox{\linewidth}{\scriptsize\raggedright \textbf{Remove a Wheel, egocentric direction.} While the wearer is holding the detached black bicycle wheel over the patterned rug, where is the wheel relative to the wearer? (A) Directly in front of the wearer; (B) To the front-left of the wearer; (C) To the front-right of the wearer; (D) To the left of the wearer\enspace Answer B; starter A, final B.}\\[0.8mm]\includegraphics[height=\eeh]{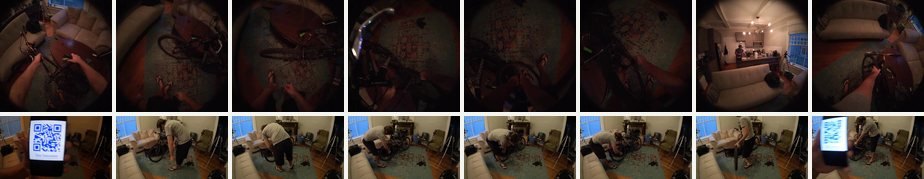}\hfill\includegraphics[height=\eeh]{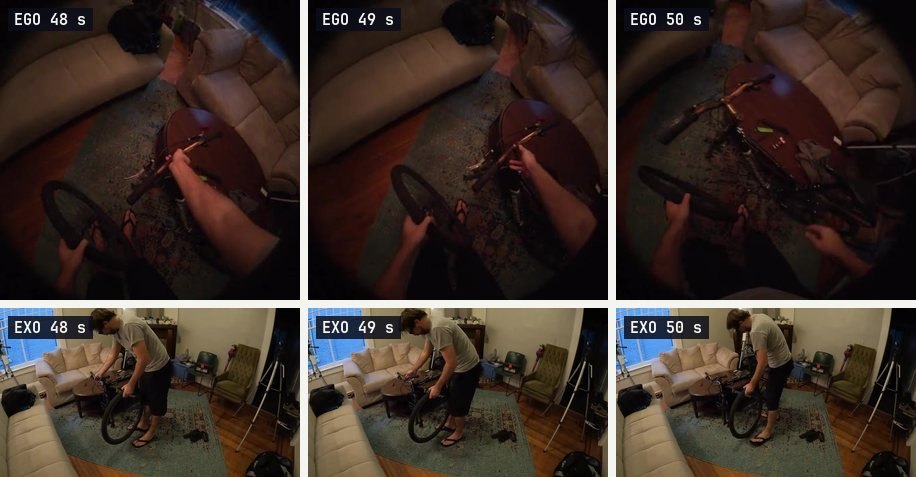}\par\vspace{3mm}
\noindent\parbox{\linewidth}{\scriptsize\raggedright \textbf{Remove a Wheel, habitual location.} Where is the silver combination wrench usually kept in this workshop? (A) Hanging in the lower row of the wooden tool rack filled with other metal wrenches and blue-handled tools; (B) Hanging in the rightmost section of the wooden tool rack filled with other metal wrenches and blue-handled tools; (C) Hanging in the upper row of the wooden tool rack filled with other metal wrenches and blue-handled tools; (D) Hanging in the middle row of the wooden tool rack filled with other metal wrenches and blue-handled tools\enspace Answer D; starter A, final D.}\\[0.8mm]\includegraphics[height=\eeh]{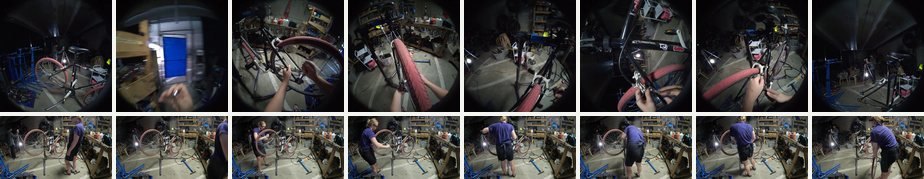}\hfill\includegraphics[height=\eeh]{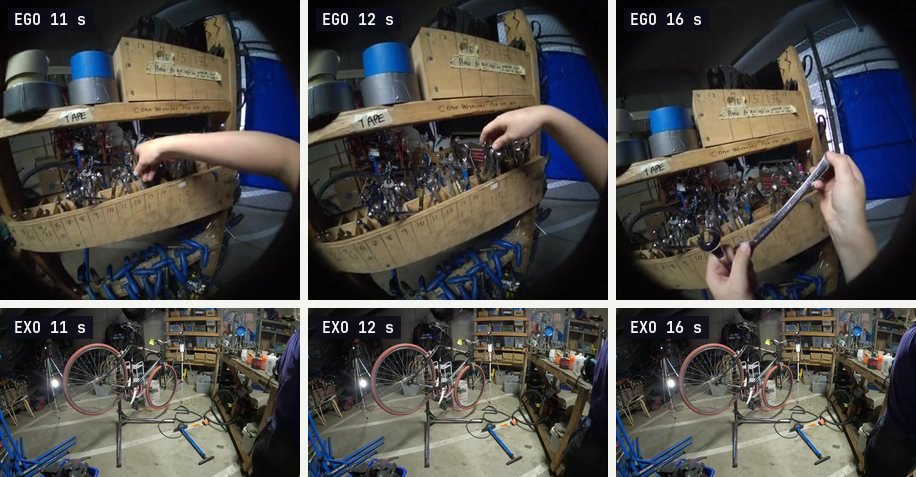}\par\vspace{3mm}
\noindent\parbox{\linewidth}{\scriptsize\raggedright \textbf{Making Sesame-Ginger Asian Salad, object state.} While the man in the black T-shirt and dark glasses was getting the white bowl-like plate from the metal counter, what was the state of the pale dressing in the squeeze bottle? (A) It was still completely full and unopened; (B) It had already been poured out and was no longer full; (C) It had been spilled over the counter; (D) It was sitting empty with the cap off\enspace Answer B; starter A, final B.}\\[0.8mm]\includegraphics[height=\eeh]{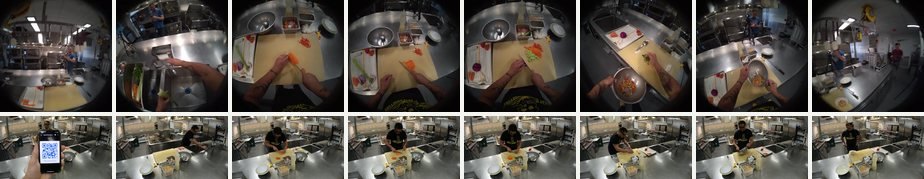}\hfill\includegraphics[height=\eeh]{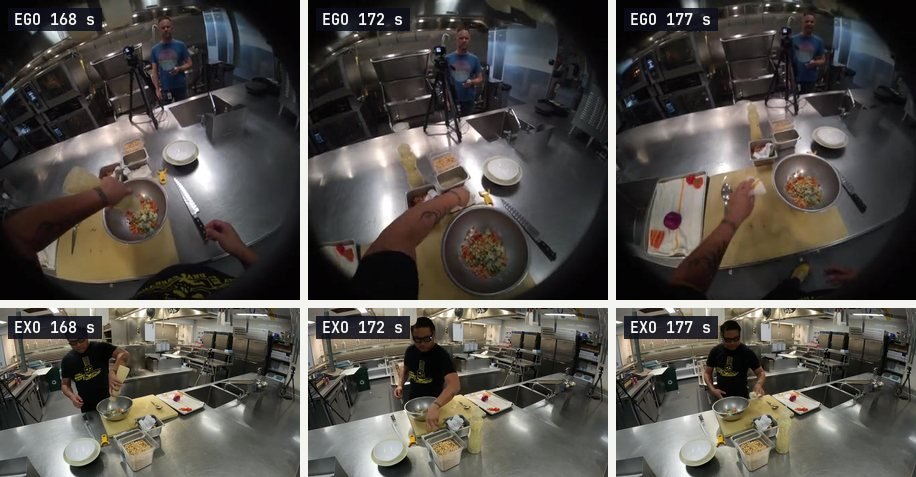}\par\vspace{3mm}
\noindent\parbox{\linewidth}{\scriptsize\raggedright \textbf{Making Cucumber \& Tomato Salad, instantaneous position.} Where was the woman with a ponytail wearing a white floral T-shirt and dark glasses standing the FIRST time getting the cucumber? (A) At the black countertop beside the single-burner silver stove, facing the cabinets; (B) At the black countertop beside the two-burner silver stove, facing the cabinets; (C) At the black countertop beside the two-burner silver stove, facing the sink; (D) At the black countertop beside the two-burner silver stove, facing the refrigerator\enspace Answer B; starter C, final B.}\\[0.8mm]\includegraphics[height=\eeh]{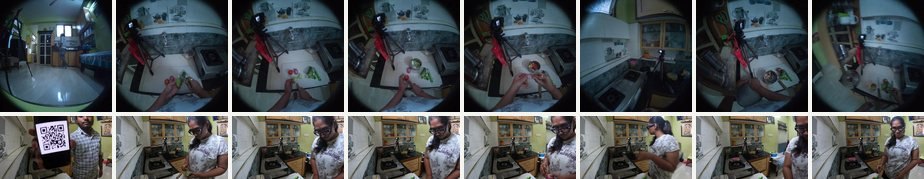}\hfill\includegraphics[height=\eeh]{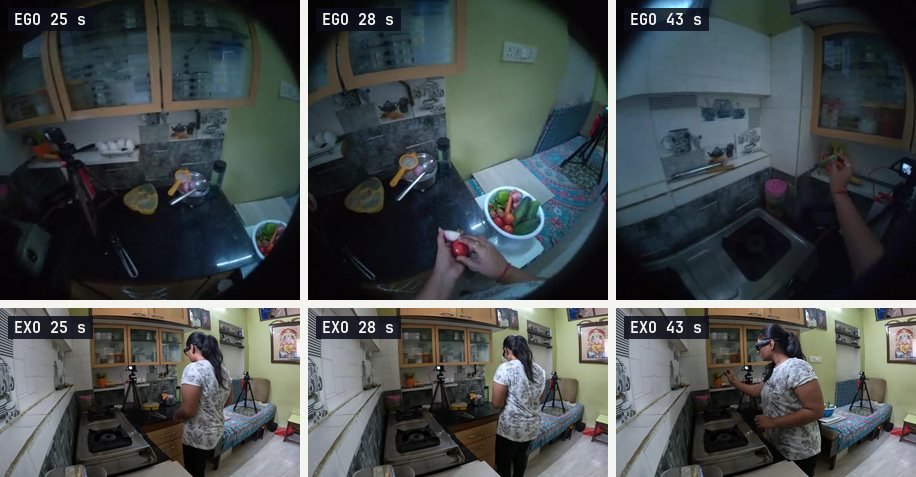}\par\vspace{3mm}
\noindent\parbox{\linewidth}{\scriptsize\raggedright \textbf{Cooking an Omelet, allocentric relation.} While the dark-skinned person in a black shirt pours from the small seasoning container into the pink mixing bowl, where is the white salt container relative to the seasoning container? (A) To the left of the seasoning container; (B) To the right of the seasoning container; (C) Underneath the seasoning container; (D) Directly behind the seasoning container\enspace Answer A; starter B, final A.}\\[0.8mm]\includegraphics[height=\eeh]{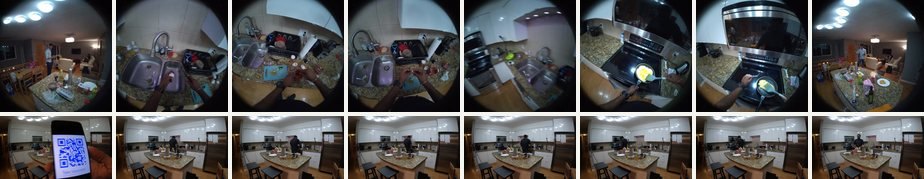}\hfill\includegraphics[height=\eeh]{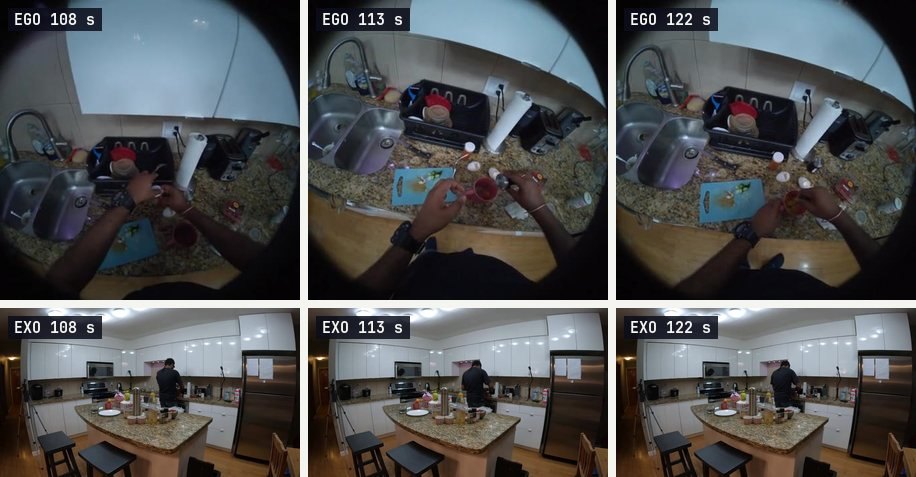}\par\vspace{3mm}
  \caption{\textbf{Ego-Exo Memory.} Test questions, the frames that the starter solution shows the model (evenly spaced over the video), and the frames that the final solution of the run of Claude Opus~5.5 selects before answering, with the answers of both.}
  \label{fig:qual-egoexo}
\end{figure}

\begin{figure}[p]
  \centering
  \setlength{\arctile}{\dimexpr(\linewidth-2.6cm-1.05cm-5mm)/6}
\noindent\parbox[c]{2.45cm}{\scriptsize\raggedright\textbf{re86}, level 5 of 8\\[0.6mm]\textcolor{muted}{Starter: not cleared after 70 actions.\\Final: cleared after 147 actions.}}\hfill\parbox[c]{\dimexpr\linewidth-2.6cm}{\mbox{\makebox[1.05cm][l]{\raisebox{0.47\arctile}{\scriptsize Starter}}\includegraphics[width=\arctile]{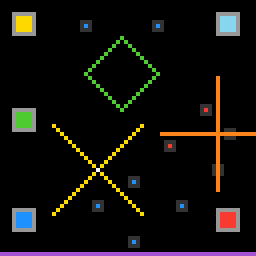}\hspace{0.6mm}\includegraphics[width=\arctile]{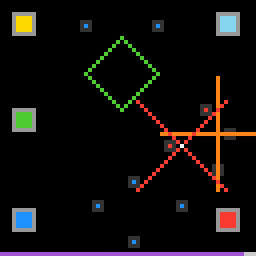}\hspace{0.6mm}\includegraphics[width=\arctile]{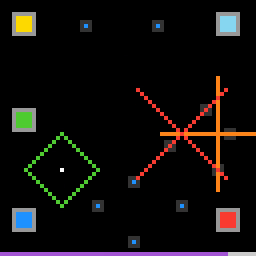}\hspace{0.6mm}\includegraphics[width=\arctile]{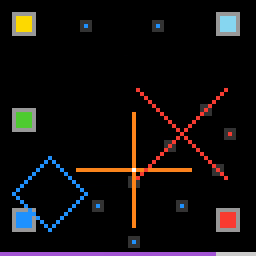}\hspace{0.6mm}\includegraphics[width=\arctile]{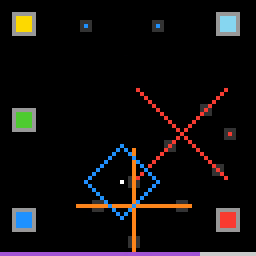}\hspace{0.6mm}\includegraphics[width=\arctile]{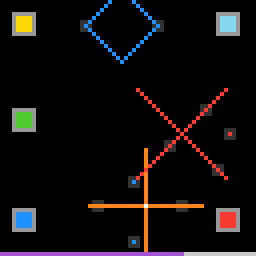}}\\[0.6mm]\mbox{\makebox[1.05cm][l]{\raisebox{0.47\arctile}{\scriptsize Final}}\includegraphics[width=\arctile]{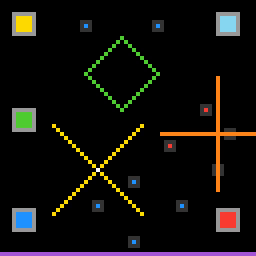}\hspace{0.6mm}\includegraphics[width=\arctile]{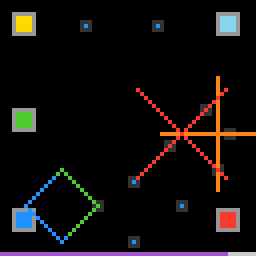}\hspace{0.6mm}\includegraphics[width=\arctile]{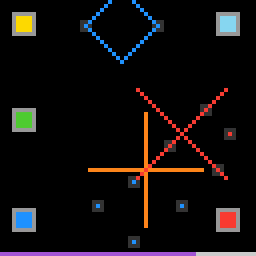}\hspace{0.6mm}\includegraphics[width=\arctile]{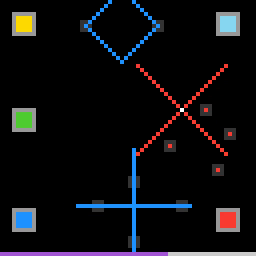}\hspace{0.6mm}\includegraphics[width=\arctile]{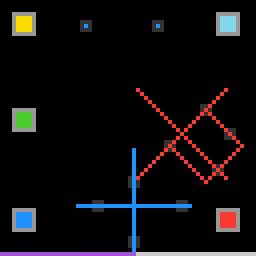}\hspace{0.6mm}\includegraphics[width=\arctile]{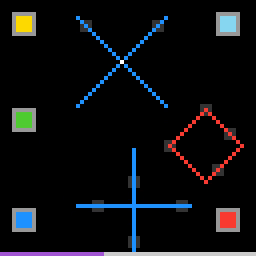}}}\par\vspace{3mm}
\noindent\parbox[c]{2.45cm}{\scriptsize\raggedright\textbf{cn04}, level 3 of 6\\[0.6mm]\textcolor{muted}{Starter: not cleared after 89 actions.\\Final: cleared after 100 actions.}}\hfill\parbox[c]{\dimexpr\linewidth-2.6cm}{\mbox{\makebox[1.05cm][l]{\raisebox{0.47\arctile}{\scriptsize Starter}}\includegraphics[width=\arctile]{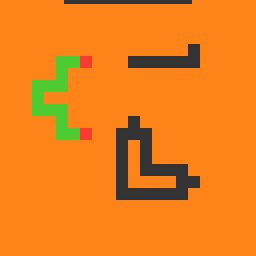}\hspace{0.6mm}\includegraphics[width=\arctile]{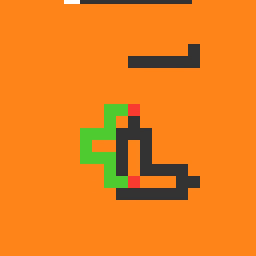}\hspace{0.6mm}\includegraphics[width=\arctile]{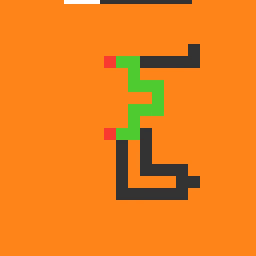}\hspace{0.6mm}\includegraphics[width=\arctile]{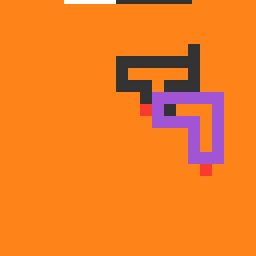}\hspace{0.6mm}\includegraphics[width=\arctile]{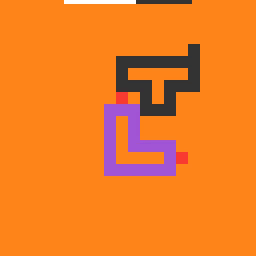}\hspace{0.6mm}\includegraphics[width=\arctile]{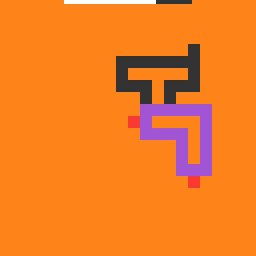}}\\[0.6mm]\mbox{\makebox[1.05cm][l]{\raisebox{0.47\arctile}{\scriptsize Final}}\includegraphics[width=\arctile]{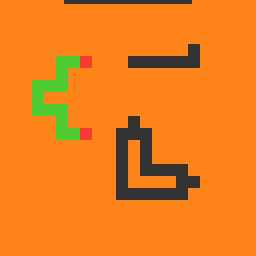}\hspace{0.6mm}\includegraphics[width=\arctile]{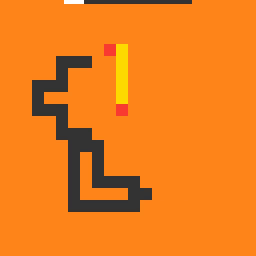}\hspace{0.6mm}\includegraphics[width=\arctile]{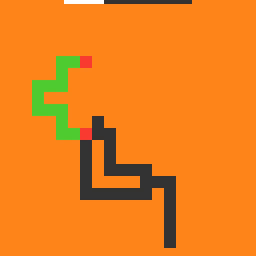}\hspace{0.6mm}\includegraphics[width=\arctile]{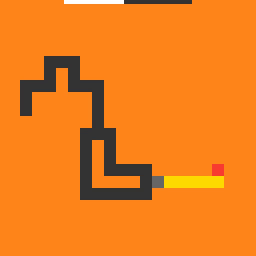}\hspace{0.6mm}\includegraphics[width=\arctile]{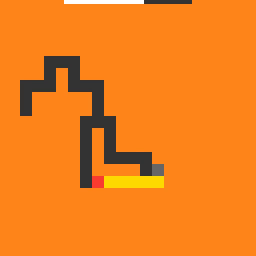}\hspace{0.6mm}\includegraphics[width=\arctile]{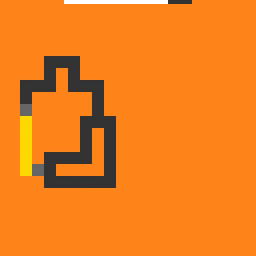}}}\par\vspace{3mm}
\noindent\parbox[c]{2.45cm}{\scriptsize\raggedright\textbf{ls20}, level 3 of 7\\[0.6mm]\textcolor{muted}{Starter: not cleared after 522 actions.\\Final: cleared after 49 actions.}}\hfill\parbox[c]{\dimexpr\linewidth-2.6cm}{\mbox{\makebox[1.05cm][l]{\raisebox{0.47\arctile}{\scriptsize Starter}}\includegraphics[width=\arctile]{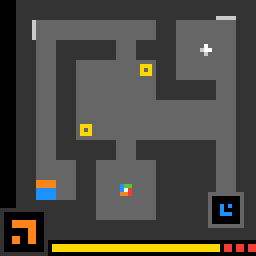}\hspace{0.6mm}\includegraphics[width=\arctile]{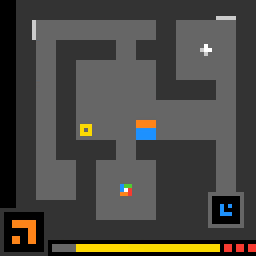}\hspace{0.6mm}\includegraphics[width=\arctile]{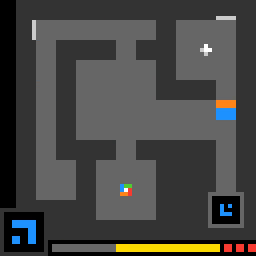}\hspace{0.6mm}\includegraphics[width=\arctile]{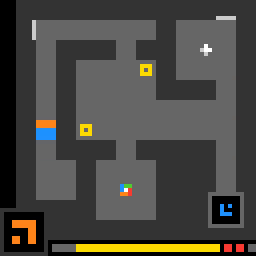}\hspace{0.6mm}\includegraphics[width=\arctile]{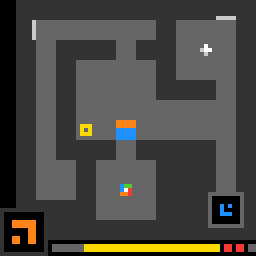}\hspace{0.6mm}\includegraphics[width=\arctile]{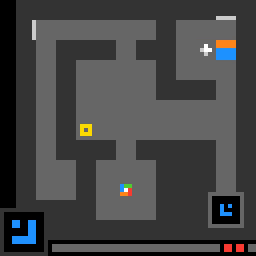}}\\[0.6mm]\mbox{\makebox[1.05cm][l]{\raisebox{0.47\arctile}{\scriptsize Final}}\includegraphics[width=\arctile]{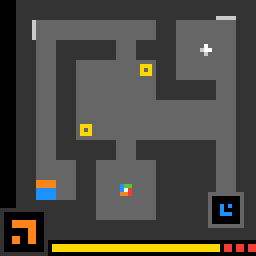}\hspace{0.6mm}\includegraphics[width=\arctile]{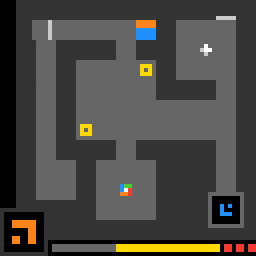}\hspace{0.6mm}\includegraphics[width=\arctile]{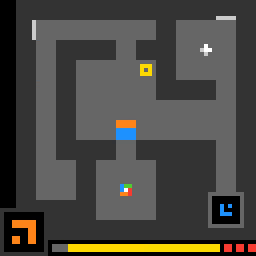}\hspace{0.6mm}\includegraphics[width=\arctile]{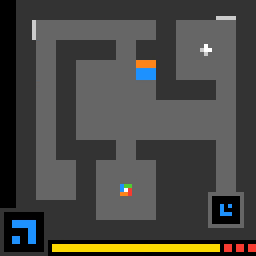}\hspace{0.6mm}\includegraphics[width=\arctile]{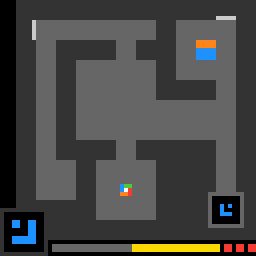}\hspace{0.6mm}\includegraphics[width=\arctile]{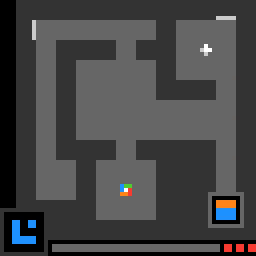}}}\par\vspace{3mm}
\noindent\parbox[c]{2.45cm}{\scriptsize\raggedright\textbf{su15}, level 4 of 9\\[0.6mm]\textcolor{muted}{Starter: not cleared after 43 actions.\\Final: cleared after 35 actions.}}\hfill\parbox[c]{\dimexpr\linewidth-2.6cm}{\mbox{\makebox[1.05cm][l]{\raisebox{0.47\arctile}{\scriptsize Starter}}\includegraphics[width=\arctile]{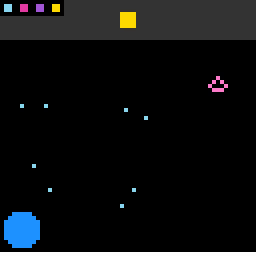}\hspace{0.6mm}\includegraphics[width=\arctile]{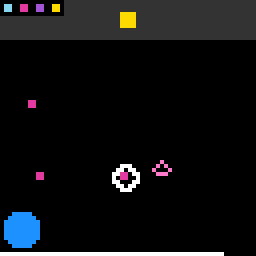}\hspace{0.6mm}\includegraphics[width=\arctile]{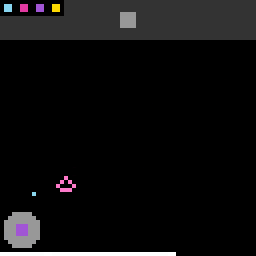}\hspace{0.6mm}\includegraphics[width=\arctile]{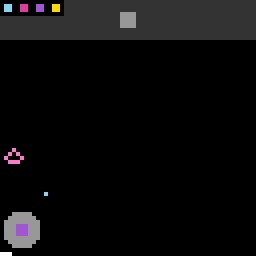}\hspace{0.6mm}\includegraphics[width=\arctile]{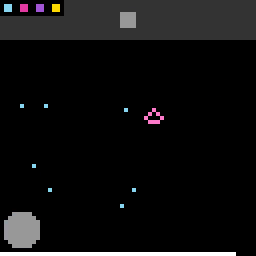}\hspace{0.6mm}\includegraphics[width=\arctile]{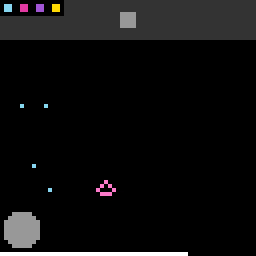}}\\[0.6mm]\mbox{\makebox[1.05cm][l]{\raisebox{0.47\arctile}{\scriptsize Final}}\includegraphics[width=\arctile]{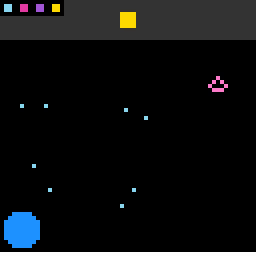}\hspace{0.6mm}\includegraphics[width=\arctile]{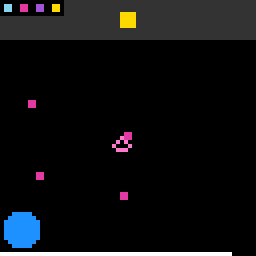}\hspace{0.6mm}\includegraphics[width=\arctile]{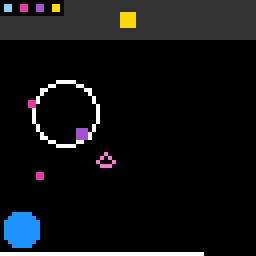}\hspace{0.6mm}\includegraphics[width=\arctile]{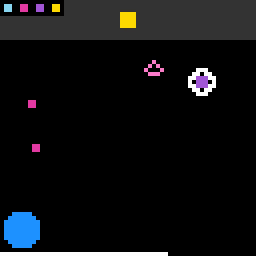}\hspace{0.6mm}\includegraphics[width=\arctile]{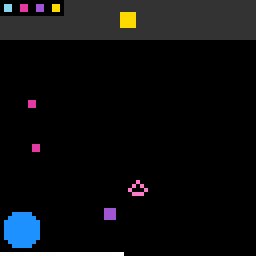}\hspace{0.6mm}\includegraphics[width=\arctile]{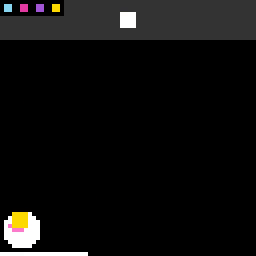}}}\par\vspace{3mm}
  \caption{\textbf{ARC-AGI-3.} Four validation games, each played on one level by the starter solution and by the final solution of the run of Claude Opus~5.5. The games come without instructions: the agent finds out by acting what the controls do and what clears a level. Each row shows six frames evenly spaced from the first to the last frame of the play. The final solution tells the model which pixels each action changed, removes unavailable actions, and restarts after a game over.}
  \label{fig:qual-arc3}
\end{figure}

\begin{figure}[p]
  \centering
\noindent\qedit{Change appearance}{Change the red umbrella canopy to a yellow-and-black plaid pattern}{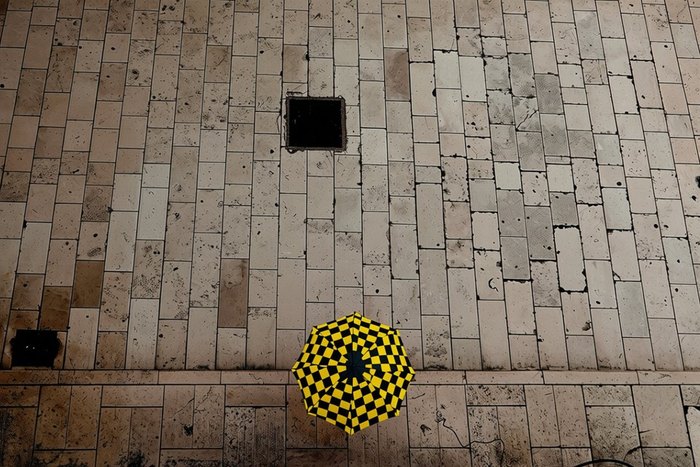}{72}{89}{Starter: A yellow-and-black pattern, but the tile texture around it is smoothed away. Final: Yellow-and-black plaid; the tiles and the light are identical.}\hfill\qedit{Restoration, deblur}{This photo is out of focus; deblur it so the oat stalks and the clouds look sharp}{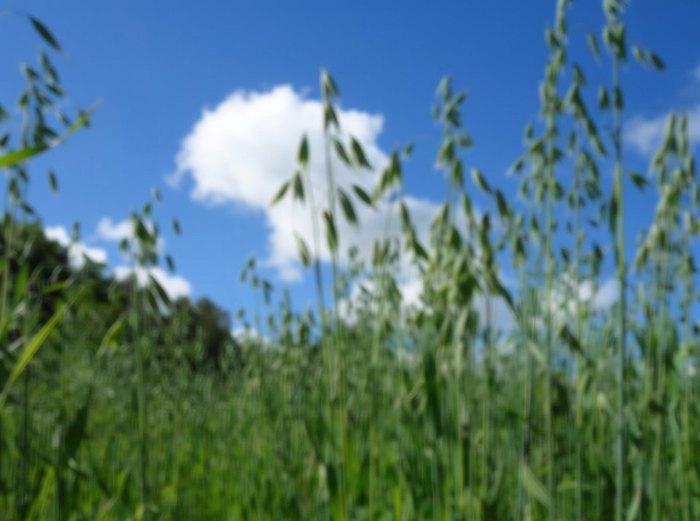}{0}{75}{Starter: Still blurry. Final: Oat stalks and clouds sharp; nothing is redrawn.}\par\vspace{2.5mm}
\noindent\qedit{Soft-effect removal, glare}{Remove the star-burst flares and glare around the street lamps so they look like normal light sources; keep everything else unchanged}{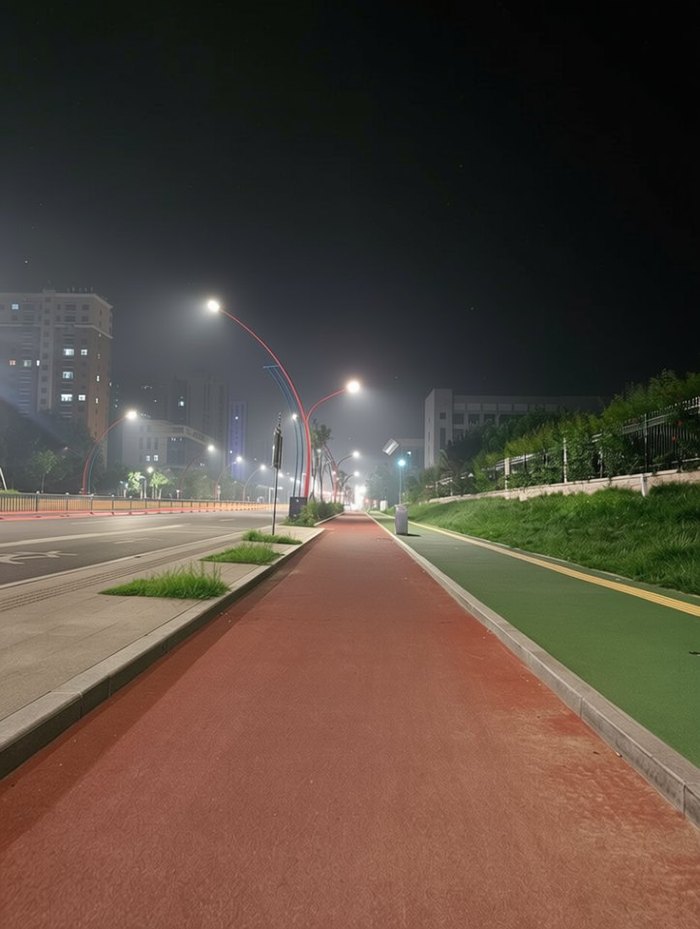}{53}{77}{Starter: The flares soften, but the buildings and the sky change. Final: Flares gone; the lamps, the path and the buildings are unchanged.}\hfill\qedit{Soft-effect removal, shadow}{Remove the large shadow cast on the brick wall; keep the brick pattern and the plants unchanged}{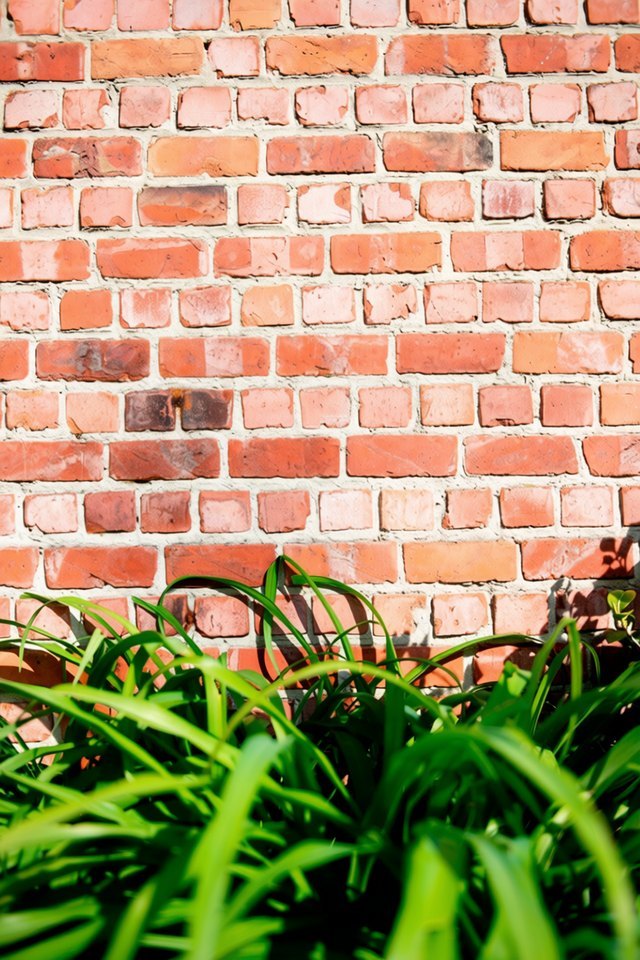}{66}{77}{Starter: The shadow goes, but the wall is redrawn and the bricks shift. Final: Shadow gone; bricks and plants line up with the original.}\par\vspace{2.5mm}
\noindent\qedit{Remove several objects}{Tidy the desk: remove the mug, the second laptop, the green box and the papers, keeping only the laptop in the middle}{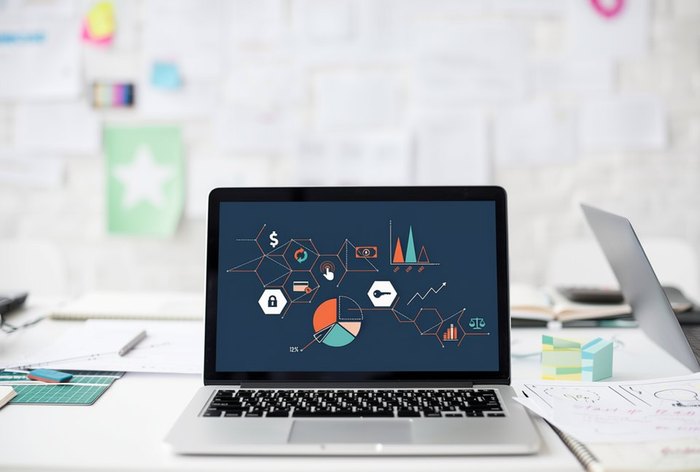}{48}{84}{Starter: The mug is gone, but the second laptop and the papers stay. Final: Only the middle laptop is left, though the desk around it changes a little.}\hfill\qedit{Find the object, then edit}{Make the leftmost bottle cobalt blue, keeping the painterly brushwork}{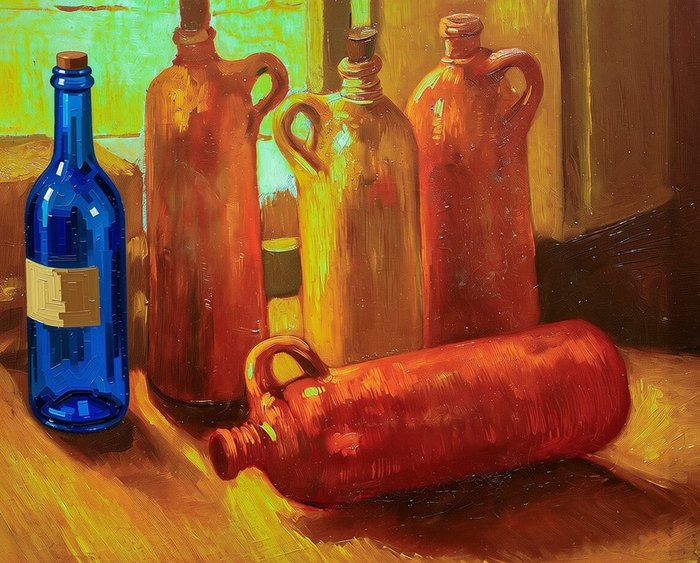}{38}{88}{Starter: The bottle turns blue, but loses its brushwork and the background changes. Final: The leftmost bottle is cobalt blue, still painted; the rest is untouched.}\par\vspace{2.5mm}
  \caption{\textbf{Image Editing, further examples.} For each edit instruction, the input image and the outputs of the starter solution and of the final solution of the run of Claude Opus~5.5, with the score that the task evaluator assigns to each output (multiplied by 100). Below each example, the differences between the two outputs.}
  \label{fig:qual-edit-more}
\end{figure}

\begin{figure}[p]
  \centering
\noindent\makebox[0.25\linewidth]{\scriptsize\textbf{Image}}\makebox[0.25\linewidth]{\scriptsize\textbf{Ground truth}}\makebox[0.25\linewidth]{\scriptsize\textbf{Starter}}\makebox[0.25\linewidth]{\scriptsize\textbf{Final}}\par\vspace{0.5mm}
\noindent\parbox{\linewidth}{\scriptsize\textbf{Lounge} (test split, 6 views, 4 objects): tracking mIoU 17.4 $\to$ 97.1}\\[0.4mm]\includegraphics[width=\linewidth]{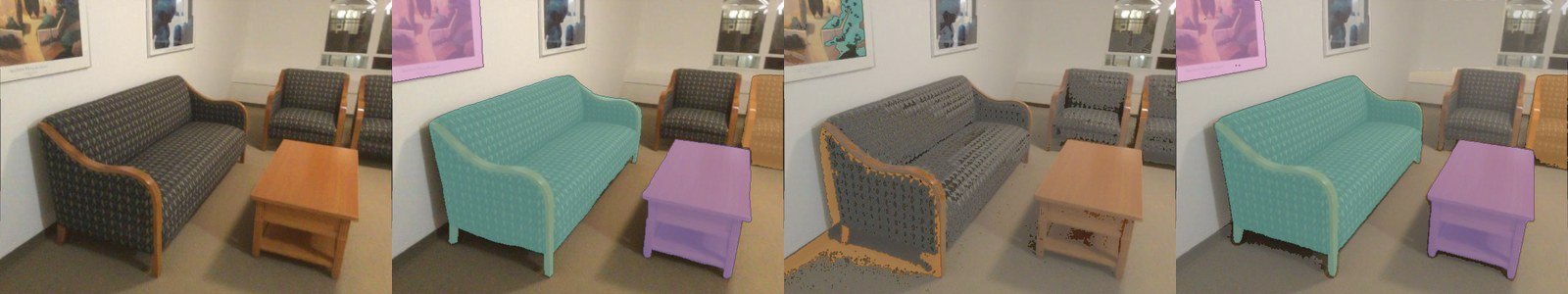}\par\vspace{1.6mm}
\noindent\parbox{\linewidth}{\scriptsize\textbf{Office} (val split, 8 views, 6 objects): tracking mIoU 31.6 $\to$ 86.3}\\[0.4mm]\includegraphics[width=\linewidth]{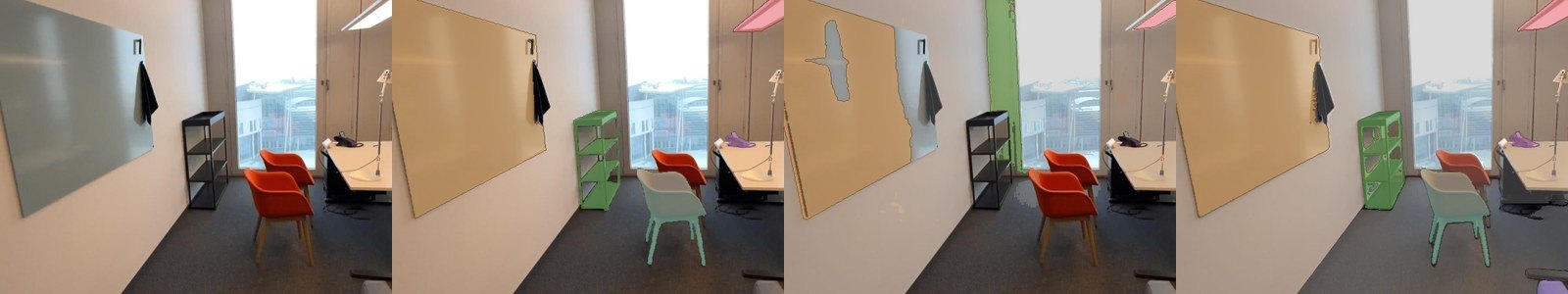}\par\vspace{1.6mm}
\noindent\parbox{\linewidth}{\scriptsize\textbf{Kitchen} (val split, 8 views, 6 objects): tracking mIoU 12.3 $\to$ 86.8}\\[0.4mm]\includegraphics[width=\linewidth]{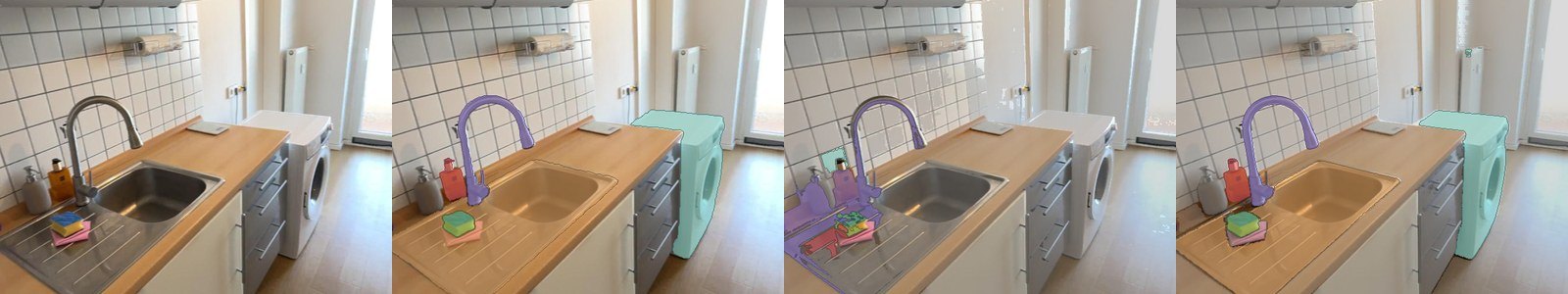}\par\vspace{1.6mm}
\noindent\parbox{\linewidth}{\scriptsize\textbf{Laundry room} (test split, 8 views, 6 objects): tracking mIoU 16.1 $\to$ 75.8}\\[0.4mm]\includegraphics[width=\linewidth]{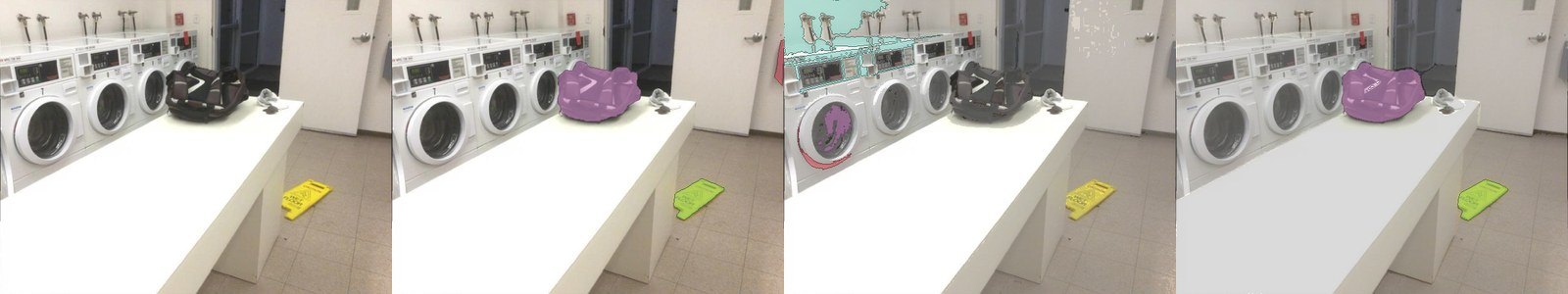}\par\vspace{1.6mm}
\noindent\parbox{\linewidth}{\scriptsize\textbf{Apartment} (test split, 8 views, 8 objects): tracking mIoU 20.6 $\to$ 90.1}\\[0.4mm]\includegraphics[width=\linewidth]{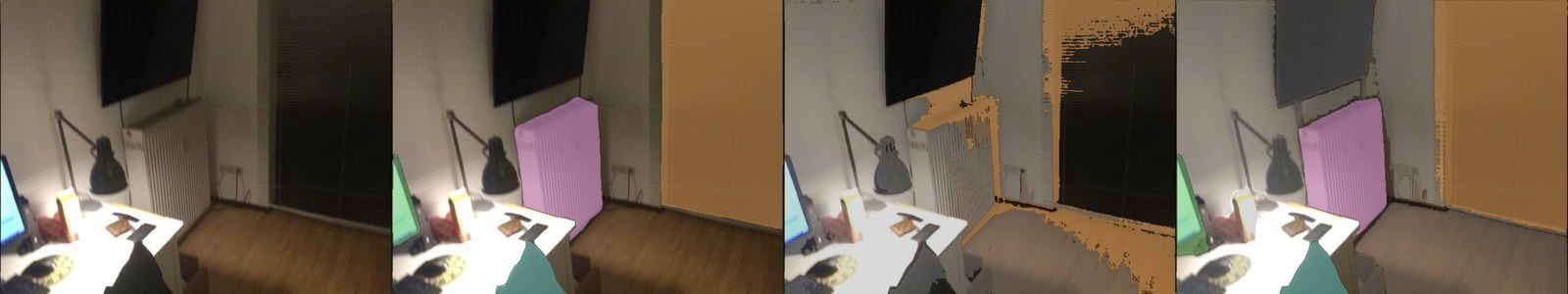}\par\vspace{1.6mm}
\noindent\parbox{\linewidth}{\scriptsize\textbf{Meeting room} (val split, 8 views, 7 objects): tracking mIoU 19.6 $\to$ 92.1}\\[0.4mm]\includegraphics[width=\linewidth]{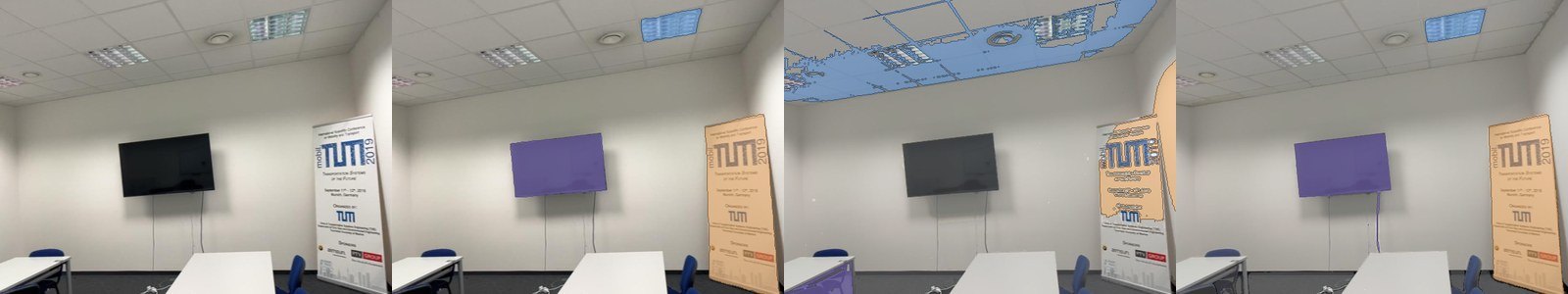}\par\vspace{1.6mm}
  \caption{\textbf{3D Tracking.} One view of each clip with the annotated instances and the instances tracked across views by the starter solution and by the final solution of the run of Claude Opus~5.5, with the tracking mIoU of each (multiplied by 100).}
  \label{fig:qual-iggt}
\end{figure}

\begin{figure}[t]
  \centering
\noindent\makebox[3.1cm]{\scriptsize\textbf{Photo}}\hfill\makebox[3.1cm]{\scriptsize\textbf{Starter}}\hfill\makebox[3.1cm]{\scriptsize\textbf{Round 1}}\hfill\makebox[3.1cm]{\scriptsize\textbf{Final (round 4)}}\hfill\makebox[3.1cm]{\scriptsize\textbf{Ground truth}}\par\vspace{0.8mm}
\noindent{\setlength\fboxsep{0pt}\fcolorbox{hairline}{white}{\includegraphics[width=3.1cm,height=2.32cm]{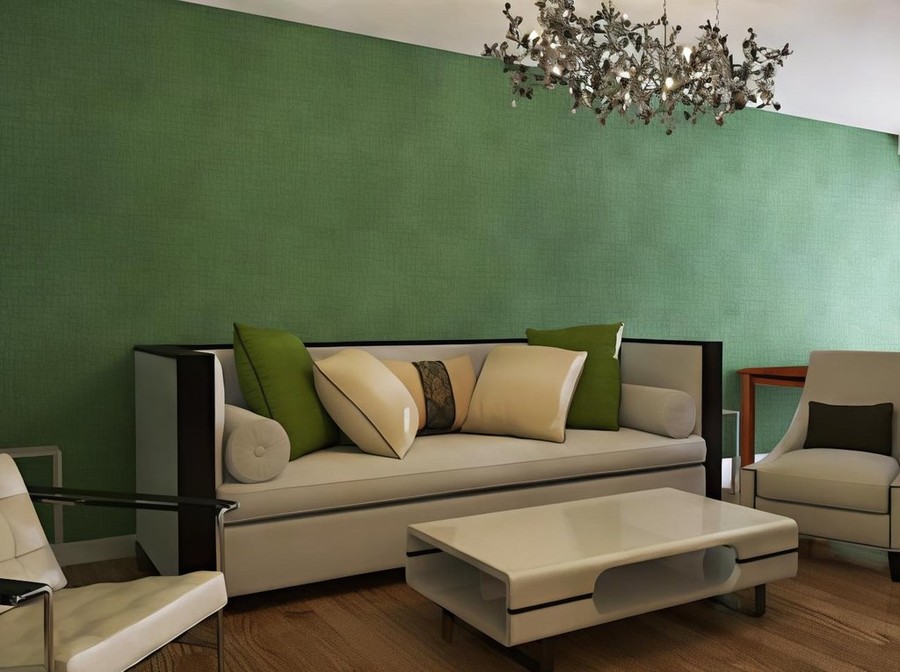}}}\hfill{\setlength\fboxsep{0pt}\fcolorbox{hairline}{white}{\parbox[b][2.32cm][c]{3.1cm}{\centering\scriptsize\textcolor{muted}{empty room,\\no furniture}}}}\hfill{\setlength\fboxsep{0pt}\fcolorbox{hairline}{white}{\includegraphics[width=3.1cm,height=2.32cm]{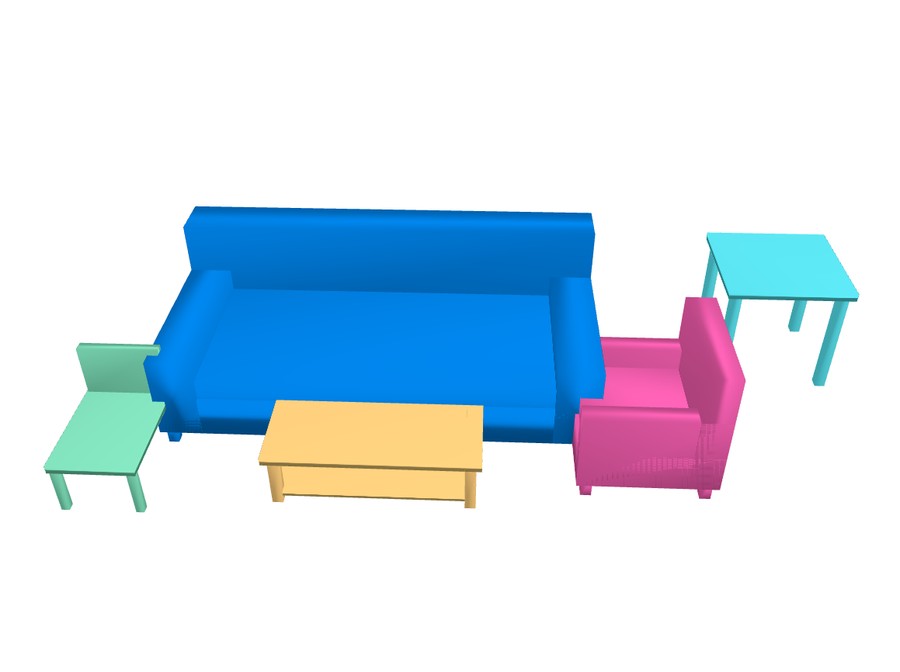}}}\hfill{\setlength\fboxsep{0pt}\fcolorbox{hairline}{white}{\includegraphics[width=3.1cm,height=2.32cm]{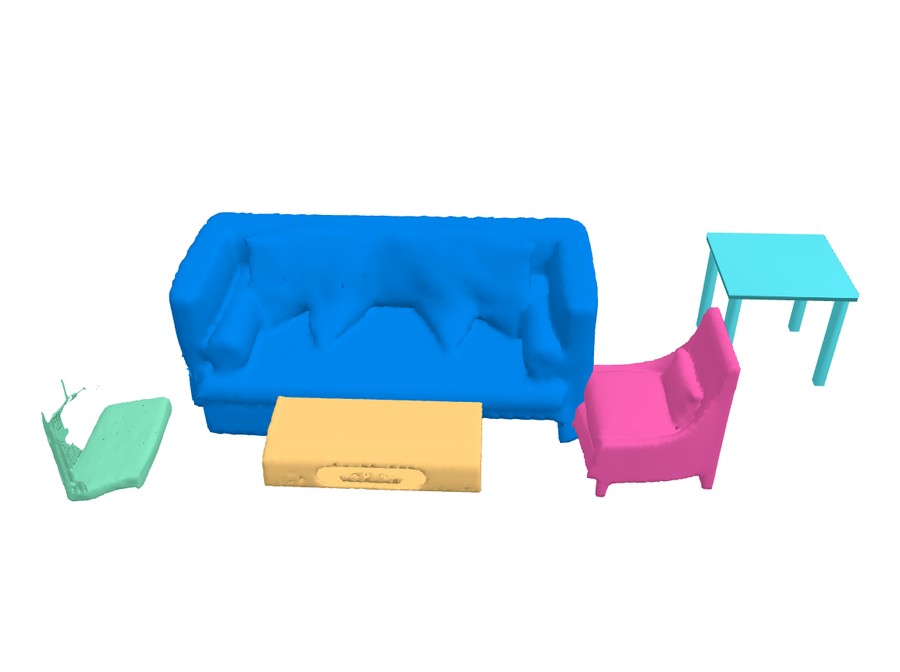}}}\hfill{\setlength\fboxsep{0pt}\fcolorbox{hairline}{white}{\includegraphics[width=3.1cm,height=2.32cm]{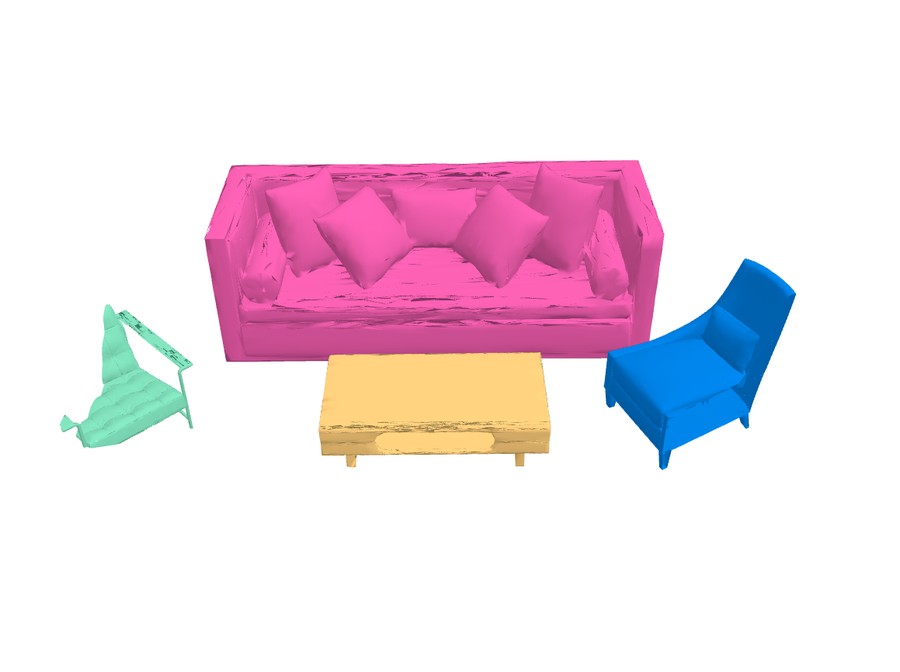}}}\par\vspace{1.2mm}
  \caption{\textbf{Scene from Photo, a further example.} Development photos and the furniture reconstructed by the starter solution, by the solution after round~1, and by the final solution of the run of Claude Opus~5.5, next to the annotated furniture, seen from above the photo's camera; each color is one piece, and the room box is not drawn. The starter returns an empty room box; the solution of round~1 places a box template for each piece; the final solution generates the shape of each piece.}
  \label{fig:qual-scene-more}
\end{figure}

\begin{figure}[t]
  \centering
\noindent\parbox[t]{5.25cm}{\vspace{0pt}\qbox{3.5cm}{2.4cm}{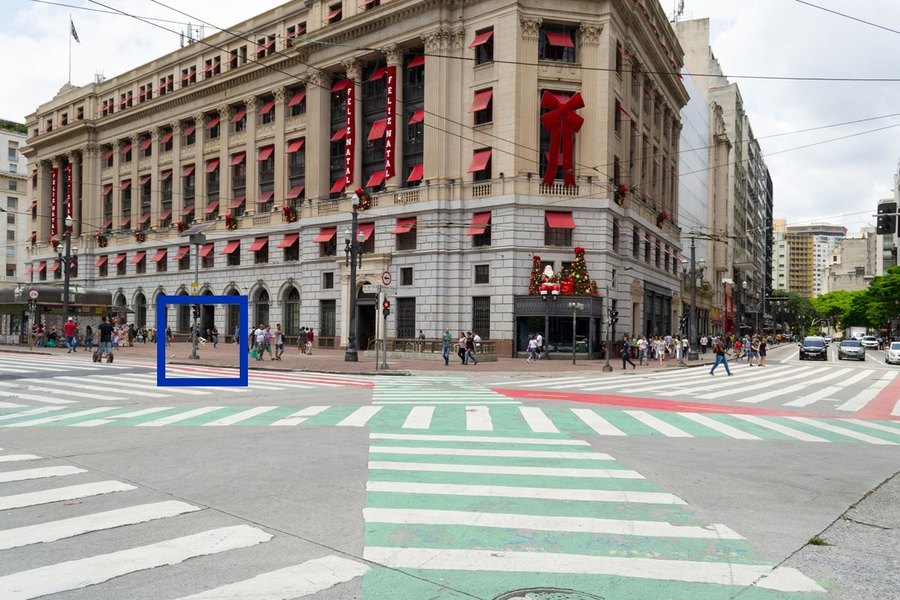}\hfill\qbox{1.65cm}{1.65cm}{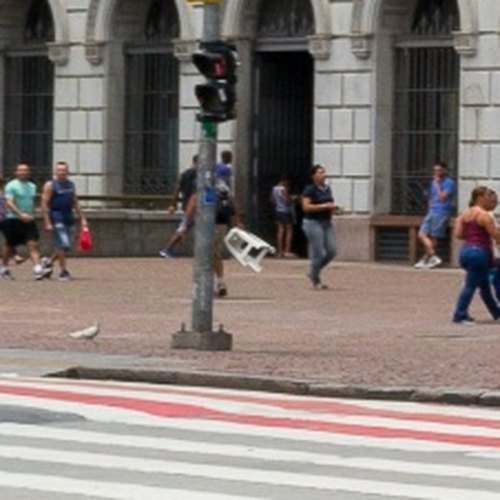}\\[0.4mm]\scriptsize\raggedright\textbf{V* Bench.} What is the color of the plastic stool? Answer C; starter D, final C.}\hfill\parbox[t]{5.25cm}{\vspace{0pt}\qbox{3.5cm}{2.4cm}{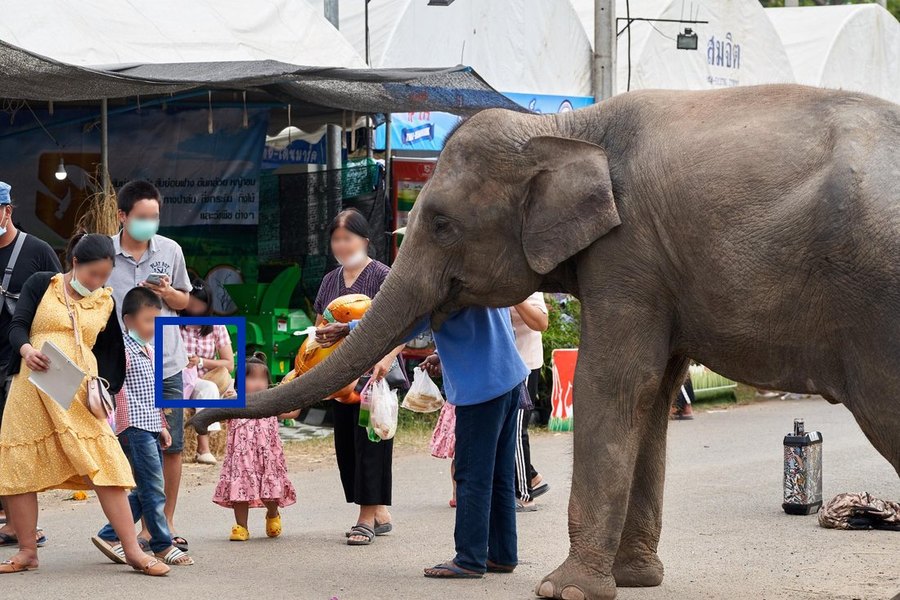}\hfill\qbox{1.65cm}{1.65cm}{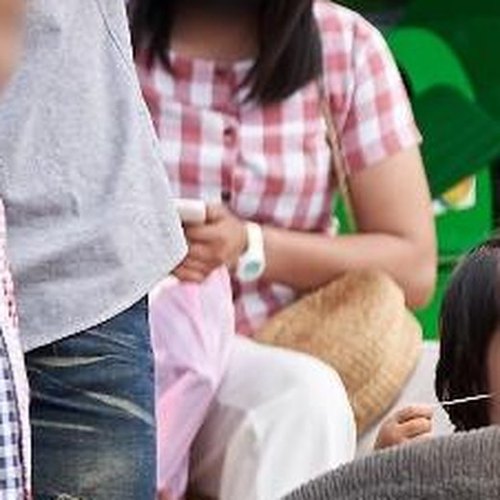}\\[0.4mm]\scriptsize\raggedright\textbf{V* Bench.} What is the color of the watch? Answer C; starter D, final C.}\hfill\parbox[t]{5.25cm}{\vspace{0pt}\qbox{3.5cm}{2.4cm}{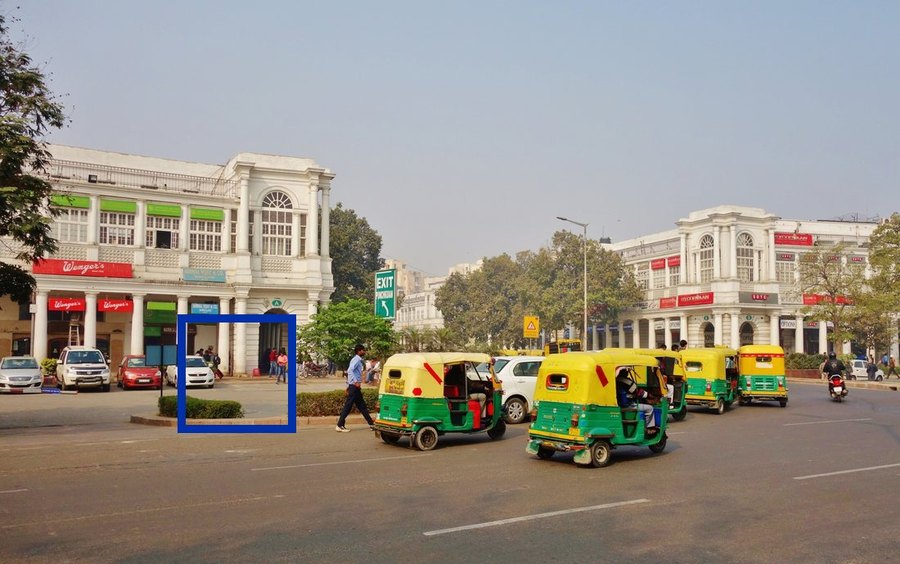}\hfill\qbox{1.65cm}{1.65cm}{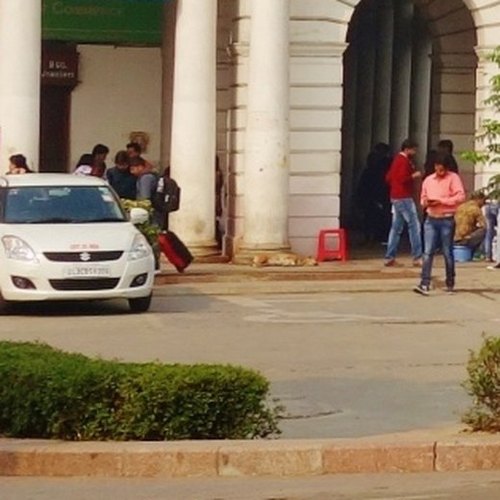}\\[0.4mm]\scriptsize\raggedright\textbf{V* Bench.} Is the red suitcase on the left or right side of the red stool? Answer B; starter A, final B.}\par\vspace{2.5mm}
\noindent\parbox[t]{5.25cm}{\vspace{0pt}\qbox{3.5cm}{2.4cm}{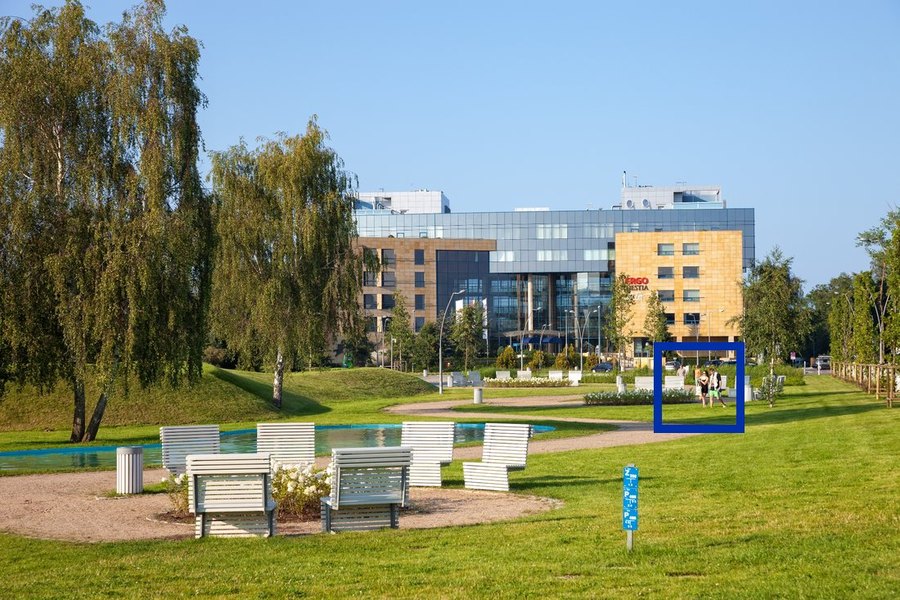}\hfill\qbox{1.65cm}{1.65cm}{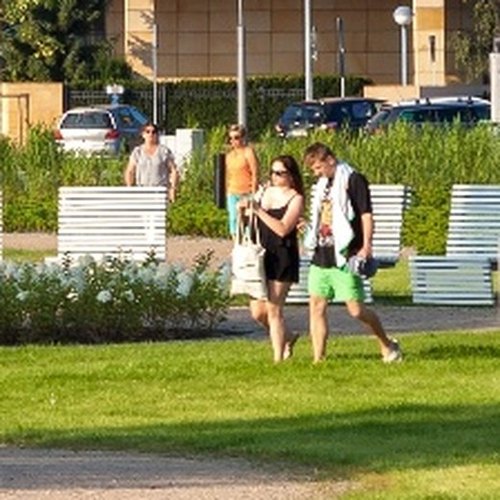}\\[0.4mm]\scriptsize\raggedright\textbf{V* Bench.} What is the color of the handbag? Answer D; starter B, final D.}\hfill\parbox[t]{5.25cm}{\vspace{0pt}\qbox{3.5cm}{2.4cm}{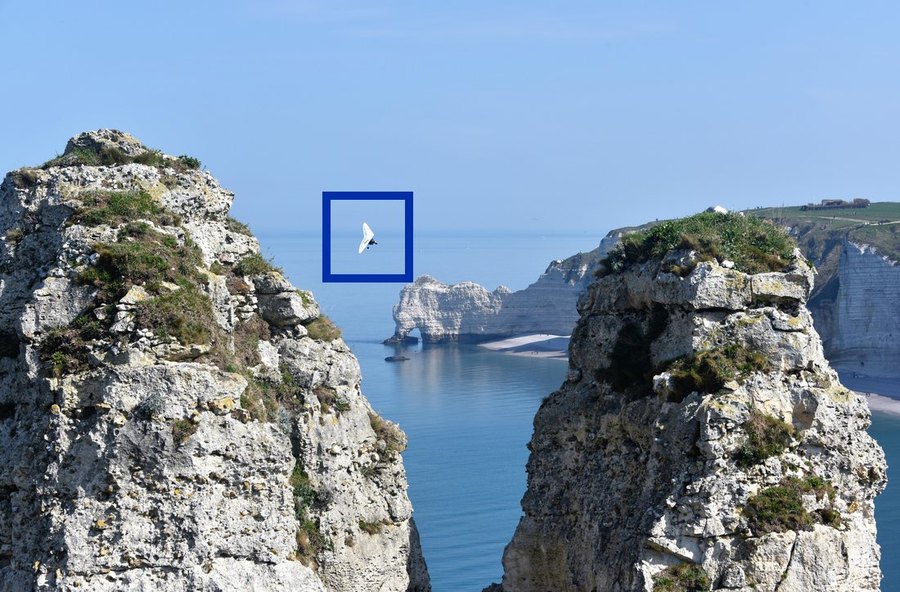}\hfill\qbox{1.65cm}{1.65cm}{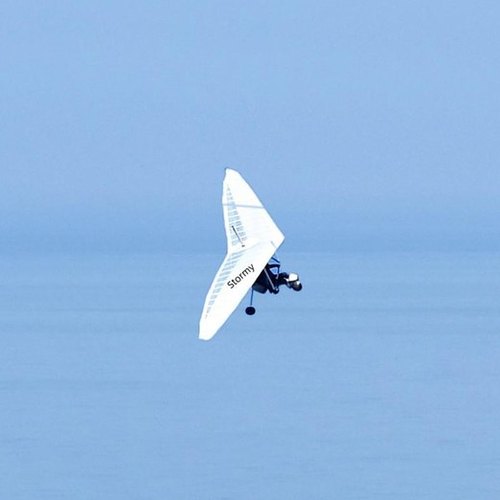}\\[0.4mm]\scriptsize\raggedright\textbf{HR-Bench 8K.} What word is written on the hang glider? Answer D; starter A, final D.}\hfill\parbox[t]{5.25cm}{\vspace{0pt}\qbox{3.5cm}{2.4cm}{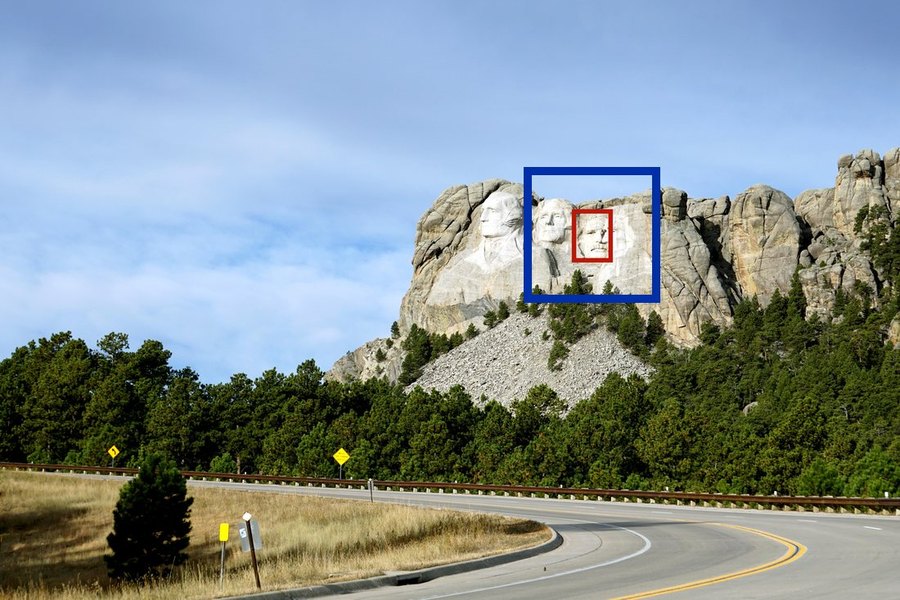}\hfill\qbox{1.65cm}{1.65cm}{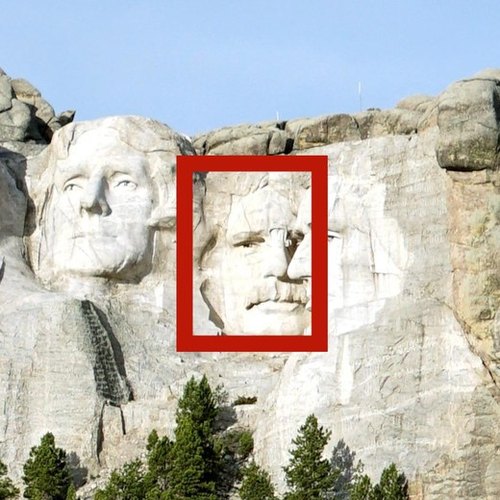}\\[0.4mm]\scriptsize\raggedright\textbf{HR-Bench 8K.} Which president's face is highlighted in the red rectangle on Mount Rushmore? Answer D; starter C, final D.}\par\vspace{2.5mm}
  \caption{\textbf{Visual Search.} Test questions about small objects in high-resolution images: the image, the region that holds the answer (enlarged), and the answers of the starter solution and of the final solution of the run of Claude Opus~5.5.}
  \label{fig:qual-vsearch}
\end{figure}

\begin{figure}[p]
  \centering
  \setlength{\vqh}{1.48cm}
\noindent\parbox[c]{4.3cm}{\scriptsize\raggedright \textbf{Rating, music video.} Human mean opinion score 3.54; starter 3.00, final 3.50 (scale 1--5).}\hfill\parbox[c]{11.7cm}{\raggedleft \includegraphics[height=\vqh]{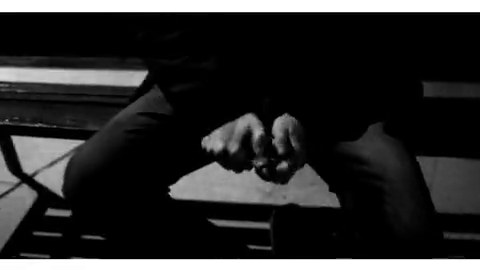}\hspace{0.5mm}\includegraphics[height=\vqh]{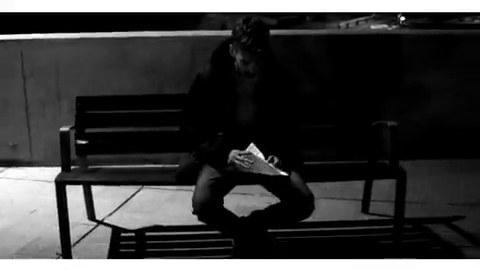}\hspace{0.5mm}\includegraphics[height=\vqh]{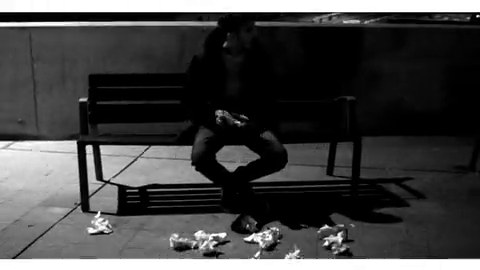}\hspace{0.5mm}\includegraphics[height=\vqh]{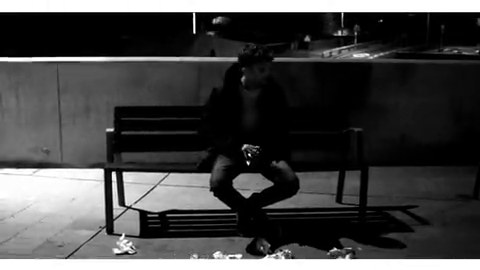}}\par\vspace{2.6mm}
\noindent\parbox[c]{4.3cm}{\scriptsize\raggedright \textbf{Rating, TV clip.} Human mean opinion score 1.99; starter 1.30, final 1.71 (scale 1--5).}\hfill\parbox[c]{11.7cm}{\raggedleft \includegraphics[height=\vqh]{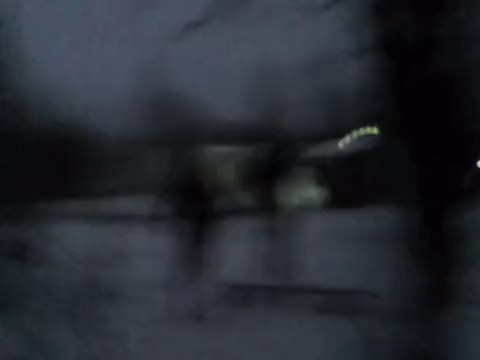}\hspace{0.5mm}\includegraphics[height=\vqh]{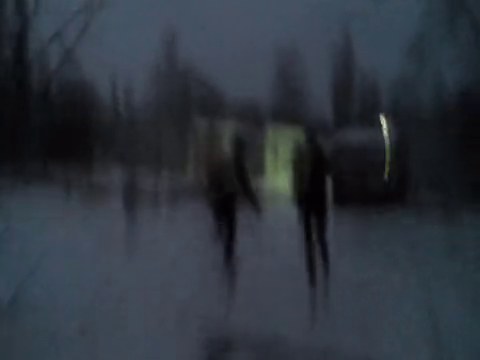}\hspace{0.5mm}\includegraphics[height=\vqh]{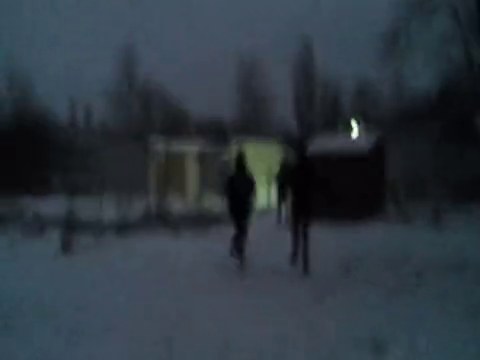}\hspace{0.5mm}\includegraphics[height=\vqh]{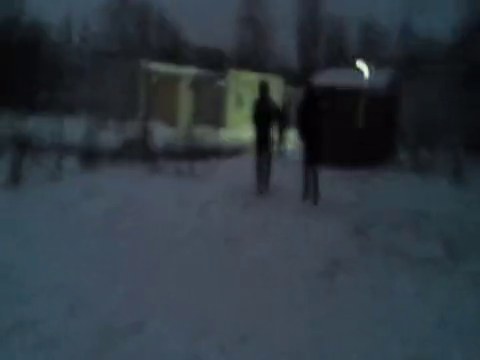}\hspace{0.5mm}\includegraphics[height=\vqh]{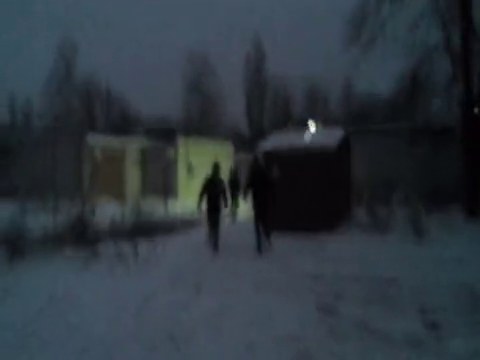}}\par\vspace{2.6mm}
\noindent\parbox[c]{4.3cm}{\scriptsize\raggedright \textbf{Rating, animation.} Human mean opinion score 4.52; starter 4.00, final 4.50 (scale 1--5).}\hfill\parbox[c]{11.7cm}{\raggedleft \includegraphics[height=\vqh]{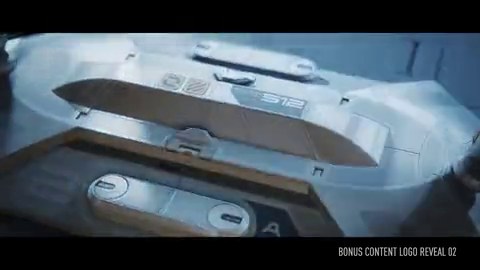}\hspace{0.5mm}\includegraphics[height=\vqh]{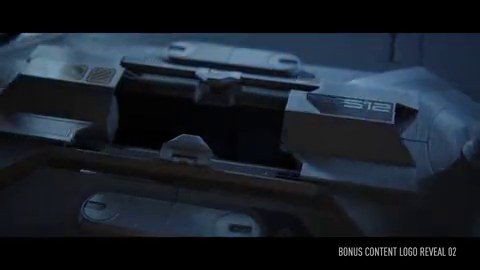}\hspace{0.5mm}\includegraphics[height=\vqh]{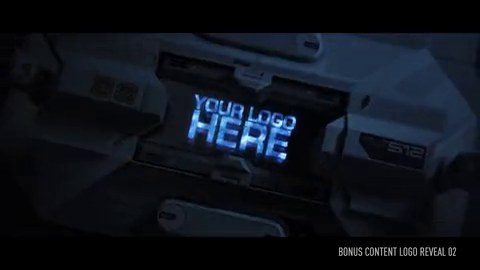}\hspace{0.5mm}\includegraphics[height=\vqh]{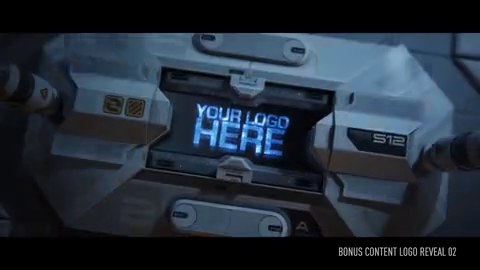}}\par\vspace{2.6mm}
\noindent\parbox[c]{4.3cm}{\scriptsize\raggedright \textbf{Pair.} Which video matches the prompt better? Prompt: A sleek, futuristic hovercar races through a neon-lit metropolis at midnight. The vehicle glides smoothly over elevated tracks, illuminating the skyline with vibrant colors as it weaves between towering skyscrapers, creating a sense of speed and advanced technology. Human choice B; starter A, final B.}\hfill\parbox[c]{11.7cm}{\raggedleft \makebox[3mm][l]{\raisebox{0.45\vqh}{\scriptsize A}}\includegraphics[height=\vqh]{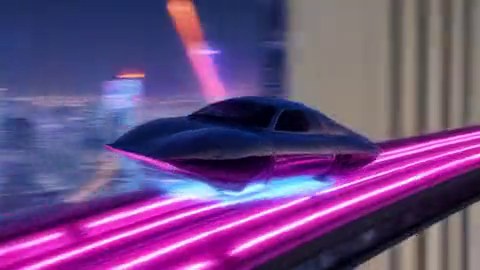}\hspace{0.5mm}\includegraphics[height=\vqh]{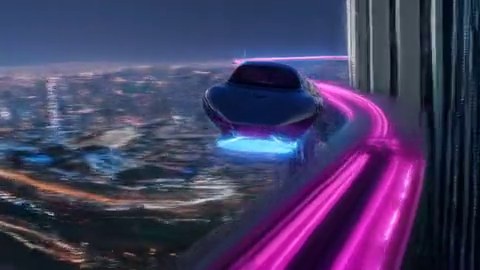}\hspace{0.5mm}\includegraphics[height=\vqh]{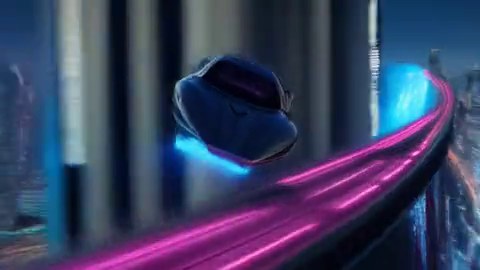}\hspace{0.5mm}\includegraphics[height=\vqh]{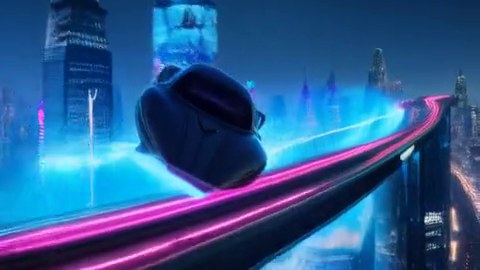}\\[0.6mm]\makebox[3mm][l]{\raisebox{0.45\vqh}{\scriptsize B}}\includegraphics[height=\vqh]{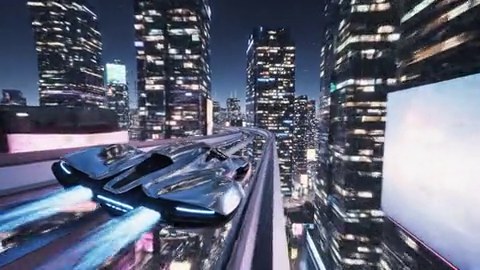}\hspace{0.5mm}\includegraphics[height=\vqh]{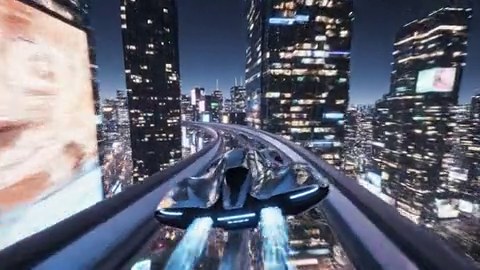}\hspace{0.5mm}\includegraphics[height=\vqh]{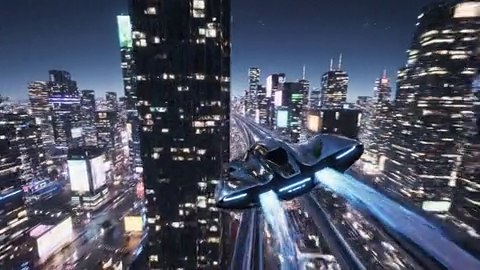}\hspace{0.5mm}\includegraphics[height=\vqh]{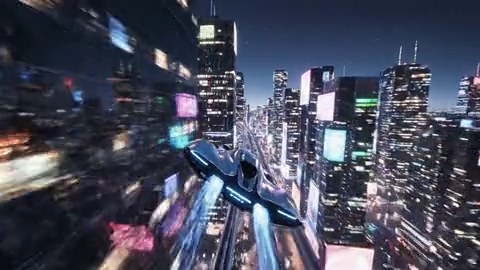}}\par\vspace{2.6mm}
\noindent\parbox[c]{4.3cm}{\scriptsize\raggedright \textbf{Pair.} Which video has fewer visual artifacts? Prompt: A daring escape through a neon-lit cyberpunk city at night. The protagonist weaves between towering skyscrapers as drones hunt, then leaps onto a hovering motorcycle and races into the vibrant skyline. Human choice B; starter A, final B.}\hfill\parbox[c]{11.7cm}{\raggedleft \makebox[3mm][l]{\raisebox{0.45\vqh}{\scriptsize A}}\includegraphics[height=\vqh]{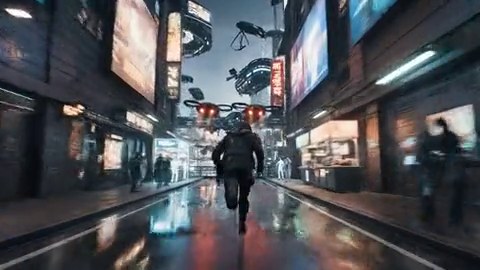}\hspace{0.5mm}\includegraphics[height=\vqh]{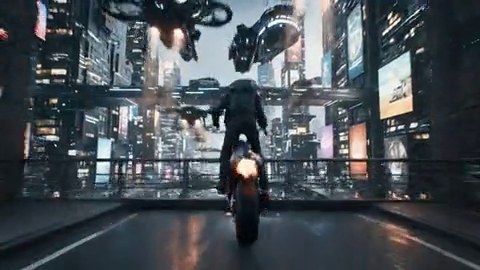}\hspace{0.5mm}\includegraphics[height=\vqh]{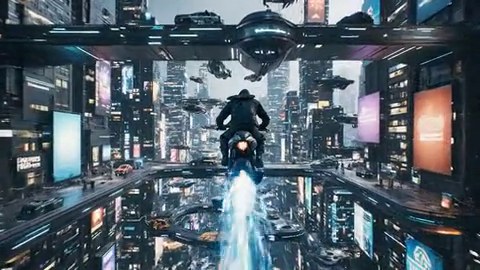}\hspace{0.5mm}\includegraphics[height=\vqh]{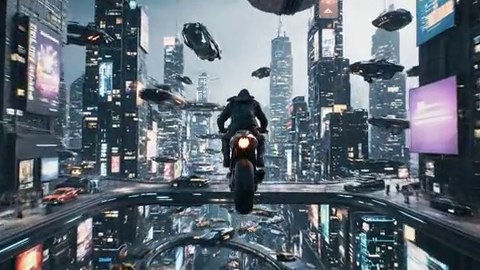}\\[0.6mm]\makebox[3mm][l]{\raisebox{0.45\vqh}{\scriptsize B}}\includegraphics[height=\vqh]{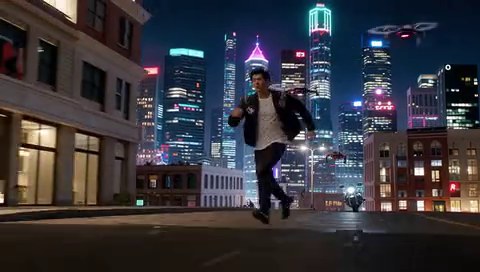}\hspace{0.5mm}\includegraphics[height=\vqh]{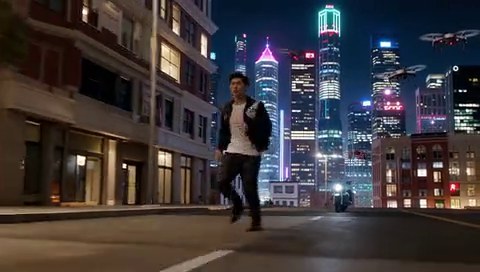}\hspace{0.5mm}\includegraphics[height=\vqh]{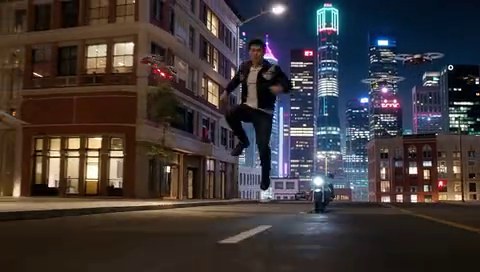}\hspace{0.5mm}\includegraphics[height=\vqh]{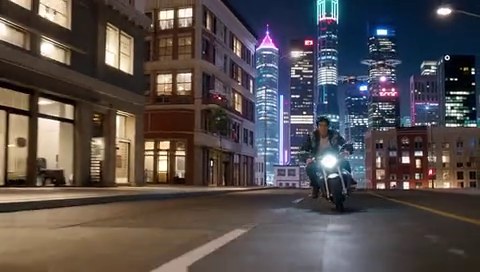}}\par\vspace{2.6mm}
\noindent\parbox[c]{4.3cm}{\scriptsize\raggedright \textbf{Pair.} Which video looks better overall? Prompt: A 3D animated dragon soaring above a medieval castle at sunset, its scales shimmering in the golden light. Flames flicker from its mouth as it navigates through fluffy clouds, casting dynamic shadows below. Human choice B; starter A, final B.}\hfill\parbox[c]{11.7cm}{\raggedleft \makebox[3mm][l]{\raisebox{0.45\vqh}{\scriptsize A}}\includegraphics[height=\vqh]{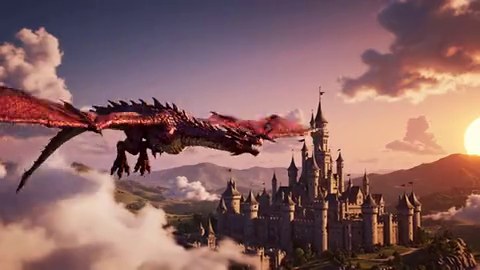}\hspace{0.5mm}\includegraphics[height=\vqh]{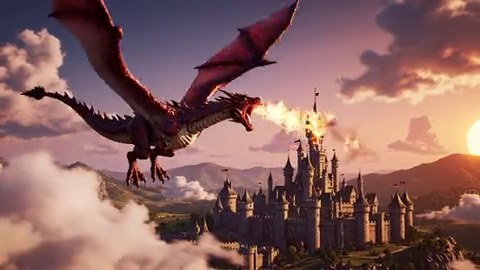}\hspace{0.5mm}\includegraphics[height=\vqh]{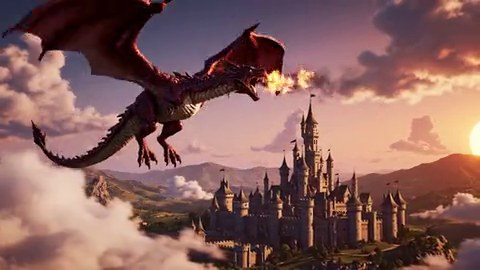}\hspace{0.5mm}\includegraphics[height=\vqh]{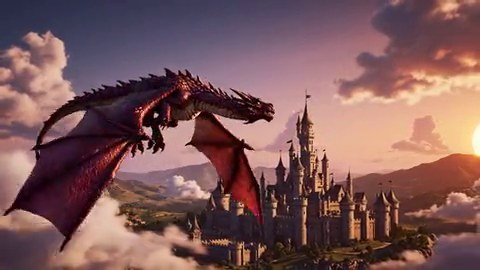}\\[0.6mm]\makebox[3mm][l]{\raisebox{0.45\vqh}{\scriptsize B}}\includegraphics[height=\vqh]{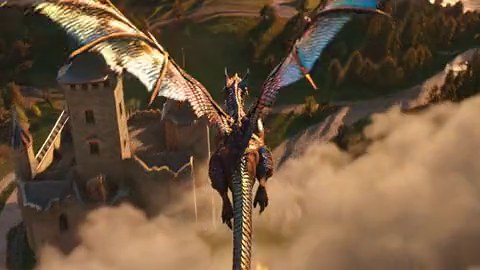}\hspace{0.5mm}\includegraphics[height=\vqh]{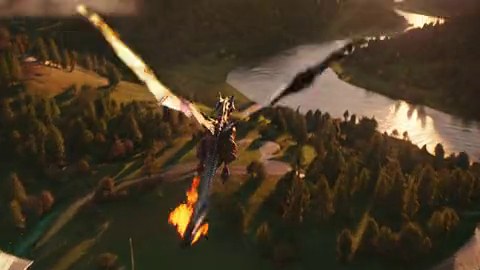}\hspace{0.5mm}\includegraphics[height=\vqh]{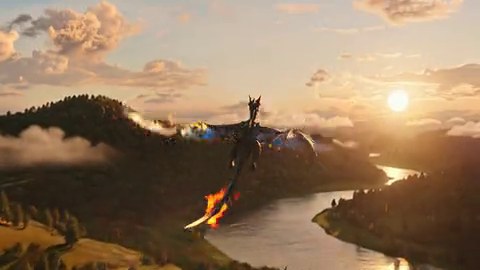}\hspace{0.5mm}\includegraphics[height=\vqh]{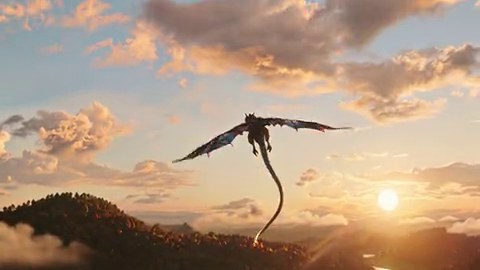}}\par\vspace{2.6mm}
\noindent\parbox[c]{4.3cm}{\scriptsize\raggedright \textbf{Pair.} Which video looks better overall? Prompt: A serene mermaid glides through vibrant coral reefs, sunlight filtering through the water. Fish of every color swim around her gracefully as she dances with the ocean currents. Human choice B; starter A, final B.}\hfill\parbox[c]{11.7cm}{\raggedleft \makebox[3mm][l]{\raisebox{0.45\vqh}{\scriptsize A}}\includegraphics[height=\vqh]{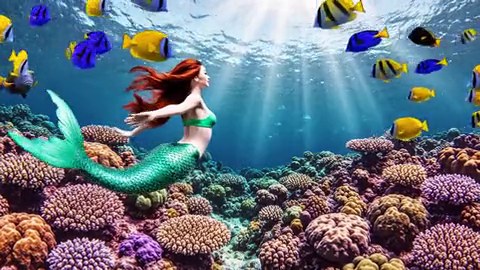}\hspace{0.5mm}\includegraphics[height=\vqh]{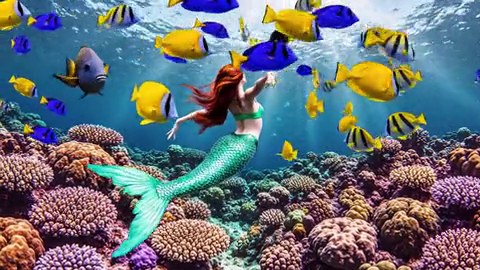}\hspace{0.5mm}\includegraphics[height=\vqh]{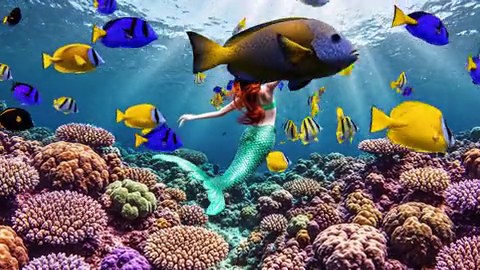}\hspace{0.5mm}\includegraphics[height=\vqh]{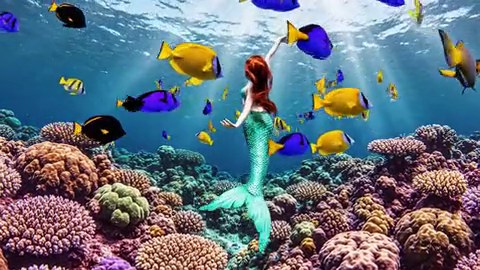}\\[0.6mm]\makebox[3mm][l]{\raisebox{0.45\vqh}{\scriptsize B}}\includegraphics[height=\vqh]{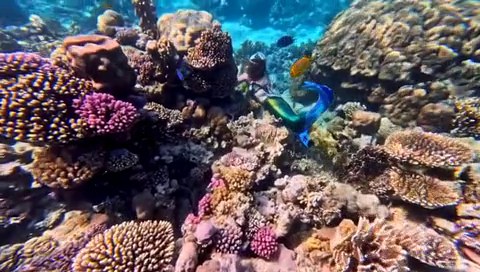}\hspace{0.5mm}\includegraphics[height=\vqh]{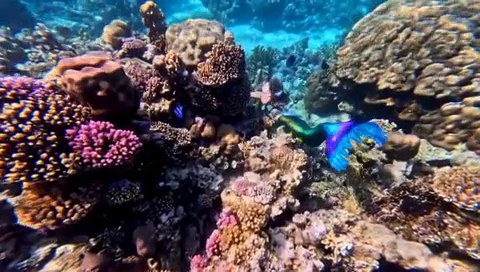}\hspace{0.5mm}\includegraphics[height=\vqh]{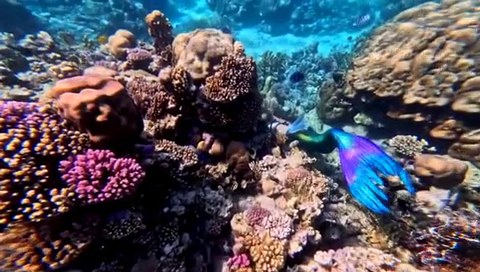}\hspace{0.5mm}\includegraphics[height=\vqh]{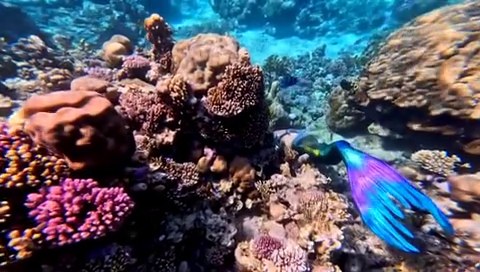}}\par\vspace{2.6mm}
  \caption{\textbf{Video Quality.} Test videos rated on a scale of 1 to 5 and pairs of generated videos compared, with the human judgement and the predictions of the starter solution and of the final solution of the run of Claude Opus~5.5. Each row shows frames evenly spaced over the video.}
  \label{fig:qual-vq}
\end{figure}

\FloatBarrier

\end{document}